\documentclass[12pt,table,xcdraw]{cmuthesis}
\usepackage{helvet}
\usepackage{fullpage}
\usepackage{graphicx}
\usepackage{amssymb}
\usepackage{amsmath}
\usepackage[numbers,sort]{natbib}
\usepackage[backref,pageanchor=true,plainpages=false, pdfpagelabels, bookmarks,bookmarksnumbered,
pdfborder=0 0 0,  
]{hyperref}
\usepackage{subfigure}
\usepackage{wrapfig}
\usepackage{booktabs}
\usepackage{multirow}

\usepackage{lipsum}
\usepackage[skip=4pt,font=small,labelfont=bf]{caption}

\usepackage{appendix}
\usepackage{chngcntr}
\usepackage{etoolbox}
\usepackage{soul}

\AtBeginEnvironment{subappendices}{%
\chapter*{Appendix}
\addcontentsline{toc}{chapter}{Appendices}
\counterwithin{figure}{section}
\counterwithin{table}{section}
}

\newsavebox{\mybox}

\makeatletter
\def\blfootnote{\gdef\@thefnmark{}\@footnotetext}
\makeatother

\usepackage[letterpaper,twoside,vscale=.8,hscale=.75,nomarginpar,hmarginratio=1:1]{geometry}

\DeclareMathOperator*{\argmax}{arg\,max}

\DeclareMathOperator{\EX}{\mathbb{E}}

\begin {document} 
\sbox{\mybox}{\hbadness=10000 \parbox{2cm}{\lipsum[1]}}
\frontmatter

\pagestyle{empty}

\title{ 
{\bf Building Intelligent Autonomous Navigation Agents}}
\author{Devendra Singh Chaplot}
\date{March 2021}
\Year{2021}
\trnumber{CMU-ML-21-101}
\committee{
Ruslan Salakhutdinov, Chair \\
Abhinav Gupta \\
Deva Ramanan\\
Jitendra Malik 
}

\support{This research was supported by: Air Force Research Laboratory award FA875016C0130001, Defense Advanced Research Projects Agency award HR001116C0136, Department of the Interior award D17PC00340, National Science Foundation award IIS1216282, a graduate fellowship from Facebook, and a research grant from Apple.}
\disclaimer{}



\maketitle

\begin{dedication}
To my wife and parents
\end{dedication}

\pagestyle{plain} 


\begin{abstract}
Breakthroughs in machine learning in the last decade have led to `digital intelligence', i.e. machine learning models capable of learning from vast amounts of labeled data to perform several digital tasks such as speech recognition, face recognition, machine translation and so on. The goal of this thesis is to make progress towards designing algorithms capable of `physical intelligence', i.e. building intelligent autonomous navigation agents capable of learning to perform complex navigation tasks in the physical world involving visual perception, natural language understanding, reasoning, planning, and sequential decision making. Despite several advances in classical navigation methods in the last few decades, current navigation agents struggle at long-term semantic navigation tasks. In the first part of the thesis, we discuss our work on short-term navigation using end-to-end reinforcement learning to tackle challenges such as obstacle avoidance, semantic perception, language grounding, and reasoning. In the second part, we present a new class of navigation methods based on modular learning and structured explicit map representations, which leverage the strengths of both classical and end-to-end learning methods, to tackle long-term navigation tasks. We show that these methods are able to effectively tackle challenges such as localization, mapping, long-term planning, exploration and learning semantic priors. These modular learning methods are capable of long-term spatial and semantic understanding and achieve state-of-the-art results on various navigation tasks.
\end{abstract}

\begin{acknowledgments}
I am immensely grateful to my advisor, Ruslan Salakhutdinov. Russ is well-known for his knowledge and research experience. His guidance has been critical in shaping and directing this thesis. Russ also has many other remarkable qualities which made me realize how fortunate I am to be advised by him. He provided me complete intellectual freedom to pursue my research ideas, and yet every time I met with Russ, he was always as excited about my research as I was, if not more. Countless times, I have left his office feeling more inspired and motivated. He has provided me endless guidance and encouragement, and unconditional support for anything I have ever needed for my research.

I am also extremely fortunate to work with Abhinav Gupta during the latter half of my Ph.D. Abhinav has an extraordinary ability to create a fun research environment that makes everyone around him feel comfortable in having honest and candid discussions with him about any topic. His sincere feedback has proved to be invaluable in not only research but also in making career decisions. Abhinav has constantly reminded me to think about the big picture, to think about what really matters and without him, I would certainly not be able to realize the full potential of our research.
    
I would also like to thank Jitendra Malik and Deva Ramanan for being my thesis committee members. Discussions with Deva and Jitendra have enabled me to get a more holistic view of my thesis research and given me more confidence in my research ideas. I also enjoyed my discussions with Jitendra during my internship which have significantly influenced and shaped my thinking and research. 
    
While my thesis committee members have been critical in ideation, I am also lucky to be helped by numerous brilliant students and post-docs in executing these research ideas. In particular, I am immensely grateful to Guillaume Lample and Saurabh Gupta, who have both been close friends and extraordinary mentors to me. They have made me realize the importance of minute details in experimental implementations and often their simple but extremely useful suggestions have turned a method from a failure to success. I have learnt a great deal from them, my Ph.D. experience and this thesis would not be remotely the same without their help. 
    
This dissertation is a result of a series of research collaborations and internships. I am grateful to all my collaborators for helping and guiding me during various research projects, Emilio Parisotto, Lisa Lee, Yao-Hung Hubert Tsai, Yue Wu, Harry Li, Jian Zhang, Ramakumar Pasumarthi, Dheeraj Rajagopal, Helen Jiang, Dhruv Batra, Devi Parikh, Deepak Pathak, Louis-Philippe Morency, and Eric Xing. Many of my collaborators have also been great friends and made long research hours more enjoyable.

\clearpage
The Machine Learning Department at CMU has been a great place to foster research ideas. The first person that comes to mind whenever I think of MLD is Diane Stidle. I cannot count the number of times Diane has answered my questions and helped me when no one else could. I would like to thank Diane for her patience and help and for making my life easier. I am also grateful to all my delightful friends in the department, Jason, Shaojie, Will, Leqi, Sebastian, Kin, Stephania, Siddarth, Amanda, Biswajit, Kartik, Jin, Nicholay, and Maruan for all the thoughtful discussions and our great times together.

I spent close to two years at Facebook AI Research in Pittsburgh during my Ph.D. where I have had the pleasure to interact with several talented students and scientists. In particular, I am grateful to Shubham Tulsiani, Dhiraj Gandhi, Tanmay Shankar, Adithya Murali, Lerrel Pinto, Kenneth Marino, and Senthil Purushwalkam for their friendship and for intellectually stimulating discussions during our coffee breaks. I would also like to especially thank Shubham for his invaluable feedback on my ideas, presentations, and papers and Dhiraj for helping with all the robot experiments.

My Ph.D. life would not so easy and fun if not for my truly amazing friends. I am grateful to Ankush, Bhavya, Bhuwan, Chaitanya, Danish, Navyata, Priyank, Rolly, Sreecharan, Shrimai, Sushil, Vidhisha, and Vikesh for making all these years more enjoyable. 

I am grateful to my parents, Gautam and Sarla, and my brother, Surendra, for their endless love, support, and encouragement throughout my life. And finally, I would like to thank my wife Kanthashree whom I met and got married to during the Ph.D. She has been a collaborator on many research projects and often the first person to bounce off my wild research ideas. More importantly, she made sure I did not have any tough times and because of her, my Ph.D. has been a smooth and happy ride.
\end{acknowledgments}

\tableofcontents
\listoffigures
\listoftables

\mainmatter

\chapter{Introduction}
Advances in machine learning in the last decade have led to `digital intelligence', i.e. machine learning models capable of learning from vast amounts of labeled data to perform several digital tasks such as speech recognition, face recognition, machine translation, and so on. The goal of this thesis is to make progress towards designing algorithms capable of `physical intelligence', i.e. building intelligent embodied autonomous agents capable of learning to perform complex tasks in the physical world involving perception, natural language understanding, reasoning, planning, and sequential decision making. To achieve this goal, this thesis focuses on training embodied navigation agents in 3D environments capable of learning to localize, building semantic maps, path planning, navigating to semantic goals, following language instructions, and answering questions. 

In order to navigate in 3D environments and perform complex tasks, embodied agents require both spatial and semantic understanding of the scene. Spatial understanding involves recognizing obstacles and traversable space from raw RGB images (perception), estimating egomotion (pose estimation), remembering previously seen obstacles as the agent moves (mapping), exploring the environment efficiently, and planning a path to the goal under uncertainty (path planning).

While spatial understanding gives an agent basic obstacle avoidance and geometric navigation capabilities, semantic understanding is essential for performing complex tasks such as finding semantic goals (specific objects, rooms, exit, and so on), following natural language instructions, and answering questions. Semantic understanding not only involves recognizing objects, regions, and their properties from raw RGB observations but also understanding visual cues (like Exit signs) and learning common-sense (beds are more likely to be found in bedrooms). It also involves grounding words in visual objects and their properties (what does 'green' look like?), and complex contextual reasoning (such as relational reasoning: `left of', `not green', or pragmatics: `largest'). Semantic understanding requires the agent to build a semantic map and perform complex reasoning on this map to follow natural language instructions and answer questions.

In addition to spatial and semantic understanding, another aspect which makes embodied intelligence challenging but also powerful at the same time is activeness or the ability to choose actions. Unlike traditional machine learning models which learn from a static dataset passively, embodied agents have the ability to choose actions which affect their future observations. They can learn to decide task-dependent actions to be more efficient and effective. This makes training embodied navigation agents challenging, as they not only need to learn rich and meaningful spatial and semantic representations but also learn a task-dependent policy on top of these representations. 

\begin{figure*}[t]
\centering
\includegraphics[width=0.99\linewidth,height=\textheight,keepaspectratio]{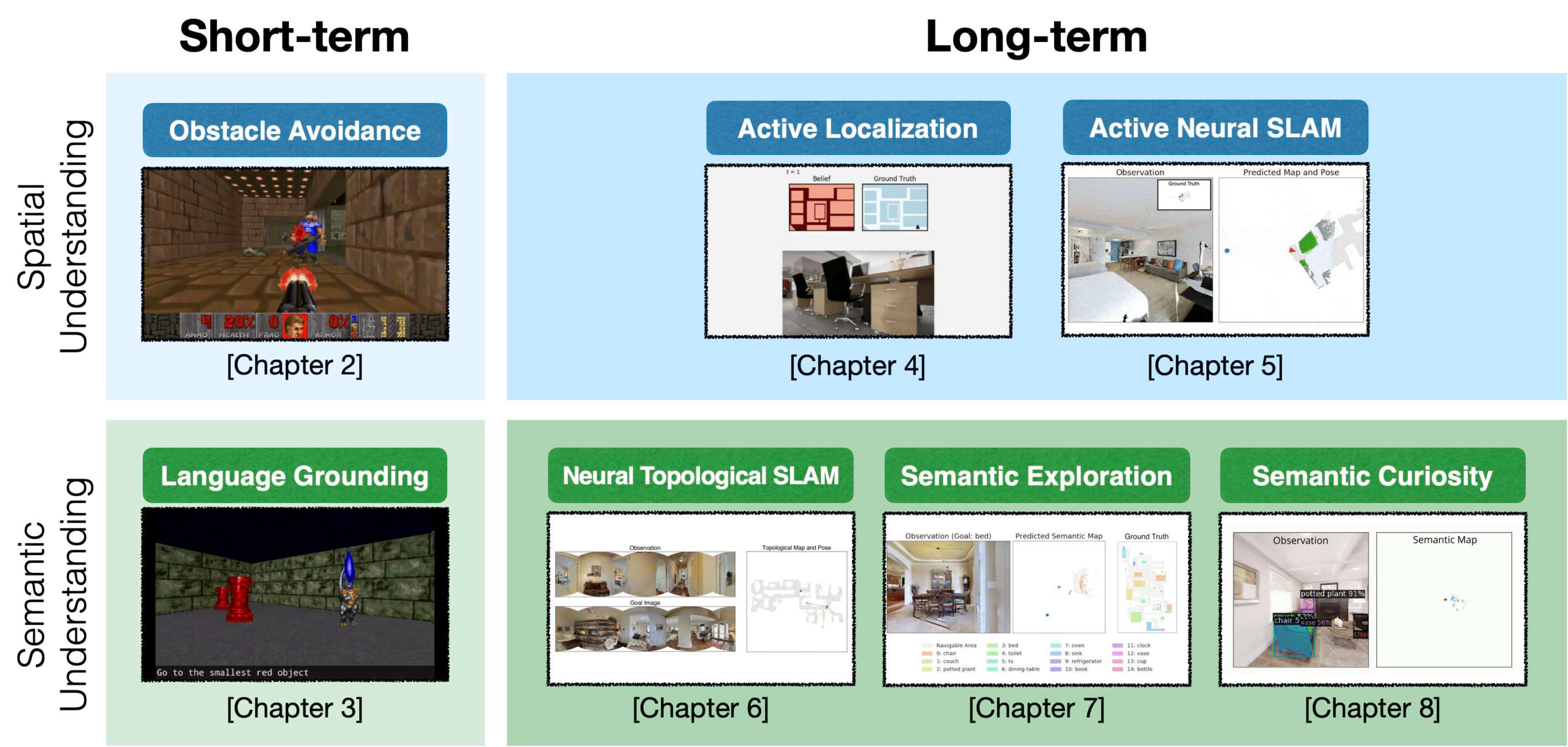}
	\caption{\small \textbf{Short-term and Long-term Navigation.} Figure showing a summary of different methods presented in this thesis tackling short-term and long-term navigation tasks requiring both spatial as well as semantic understanding.}
\label{fig:summary}
\end{figure*} 

The problem of navigation has a very rich history and has been extensively studied in classical robotics for over three decades. Classical robot navigation systems typically use a modular navigation pipeline which requires a map to be pre-computed or pre-specified. The map is then used to localize the robot using specialized laser and sonar sensors. And based on the location of the robot, a path is planned to the goal location using analytical path planning algorithms. The use of explicit maps allows classical methods to be effective at spatial or geometric understanding of the scene. Explicit maps are also an effective form of long-term memory and can be used for planning to distant goals easily. However, they lack semantic understanding and can not tackle semantic navigation tasks as they build only geometric maps and use a rule-based policy. Also, their scalability is limited as the long-term policy or goal locations need to be manually coded, they typically require pre-computed maps, and they often rely on specialized sensors.

In the recent few years, there has been a lot of work on training navigation policies using end-to-end reinforcement learning. Under this framework, the agent is given positive or negative rewards for reaching correct or wrong locations, and these rewards to learn the parameters of a neural network which predict actions directly from input observations. Unlike classical navigation methods, learning-based methods can learn a policy to go to different types of semantic goals, improve with data and generalize to new scenes, allowing them to tackle semantic navigation tasks more effectively. In the first part of this thesis, we present several methods which employ end-to-end reinforcement learning methods to tackle \textit{short-term navigation} tasks involving both spatial and semantic understanding (as shown in the left column in Figure~\ref{fig:summary}).

\section{Short-term Navigation}

\noindent\textbf{Visual Navigation from first-person observations.} Traditional visual navigation systems often require a global map of the environment, human-guided training phase, predefined models of known objects or hand-designed features which often lead to domain-specific systems which are unable to generalize to new environments~\cite{bonin2008visual}. In Chapter~\ref{chap:fps}, we present a visual navigation system which learns fine-grained control to navigate in 3D environments from only raw pixels~\cite{lample2016playing}. Our model was one of the first navigation systems trained end-to-end from pixels to actions using deep reinforcement learning. It is shown to perform well in unseen environments using a technique of domain randomization. This navigation system augmented with auxiliary tasks was also effective in playing First-Person shooter games.\\

\noindent\textbf{Language Grounding and Reasoning.} Embodied navigation agents need to able to understand natural language instructions and communicate with humans in order to be used for large-scale consumer applications. To perform tasks specified by natural language instructions, autonomous agents need to extract semantically meaningful representations of language and map it to visual elements and actions in the environment. 
In Chapter~\ref{chap:grounding}, we present our work on building autonomous agents which learn to navigate to semantic goals described by natural language instructions in partially-observable 3D environments~\cite{chaplot2017gated}. This task poses several challenges: the agent has to learn to recognize objects in raw pixel input, explore the environment as the objects might be occluded or outside the field-of-view of the agent, ground each concept of the instruction in visual elements or actions in the environment, reason about the pragmatics of language based on the objects in the current environment and navigate to the correct object while avoiding incorrect ones. 
In contrast to prior work~\cite{chen2011learning, artzi2013weakly, mei2016listen}, our model assumes no prior linguistic or perceptual knowledge and requires only raw pixels from the environment and the natural language instruction as input. The proposed model combines the image and text representations using a novel Gated-Attention mechanism and learns a policy using reinforcement learning. It performs well on complex instructions involving language grounding and visual reasoning and even generalizes to unseen instructions referring to new objects.\\

\section{Long-term Navigation}
End-to-end learning methods can be effective at short-term spatial and semantic understanding in small game-like environments, but they have several shortcomings when applied to larger and more realistic environments. In learning-based methods, memory and planning are usually implicitly encoded in the parameters of a neural network. Learning about mapping and planning from just rewards becomes exceedingly difficult as the size and fidelity of the environment increases. Furthermore, in more realistic environments, rewards are sparse and provide less visual feedback. This makes exploration difficult for an end-to-end RL policy and leads to high sample complexity, typically requiring tens or hundreds of millions of frames of experience for training. Interestingly, some of the shortcomings of learning-based approaches are the strengths of classical approaches. Specifically, classical approaches are effective at spatial or geometric understanding and long-term planning, require very little or no training data.

We can roughly divide the navigation problem space into 4 quadrants, based on spatial and semantic understanding and short-term and long-term navigation. End-to-end learning-based models can be very good at short-term spatial as well as semantic understanding. On the other hand, the classical methods are good at spatial understanding in the short-term and to some extent even long-term. In the second part of this thesis, we design a new class of methods, which we call `modular learning' methods, to tackle \textit{long-term navigation} tasks requiring both spatial and semantic understanding (see the right column in Figure~\ref{fig:summary}), which are difficult for both prior classes of methods. These modular learning methods leverage the strengths of both classical and end-to-end learning-based methods. They have two key properties: first, we use modularize the navigation system but use learning to train each module so that they can learn semantics, improve with data and generalize to unseen environments. And second, we use explicit structured map representations. We present several modular learning methods to show how modularization and explicit maps help in addressing the drawbacks of end-to-end learning and lead to effective long-term memory and planning with much lower sample complexity. \\

\noindent\textbf{Active Neural Localization.} Effective long-term memory and planning require access to the agent location which is not available in most real-world scenarios. Traditional methods of localization, which filter the belief based on the observations, are sub-optimal in the number of steps required, as they do not decide the actions taken by the agent. In Chapter~\ref{chap:anl}, we present `Active Neural Localizer', a fully differentiable neural network that learns to localize efficiently. The model incorporates ideas of traditional filtering-based localization methods, by using a structured belief of the state with multiplicative interactions to propagate belief, and combines it with a policy model to minimize the number of steps required for localization. 
Our results in photorealistic 3D simulation environments demonstrate the model's capability of learning the policy and perceptual model jointly from raw-pixel based RGB observations and generalizing to unseen environments. \\

\noindent\textbf{Active Neural SLAM.} In addition to localization, embodied agents need to build maps for exploring the environment efficiently. In Chapter~\ref{chap:ans}, we present `Active Neural SLAM' which is a modular model capable of building maps from raw-pixel data, learning exploration policies, and long-term planning. Specifically, the model consists of 3 components: a Mapper, a Global Policy, and a Local Policy. The Mapper builds and updates the map of the environment as the agent receives observations. The Global Policy learns to sample long-term goals based on the task and the map. We then plan a path to the long-term goal using an analytical planner and sample a short-term goal on the planned path. The Local Policy learns navigational actions to reach this short-term goal. 
The modularity breaks down the complex navigation problem into easier tasks leading to an absolute improvement of 21-43\% over prior methods at the exploration and pointgoal navigation tasks, while also improving the sample efficiency over 75 times. We also showed that our model generalizes much better to new domains and harder goals. This model was also the winner of \textbf{CVPR 2019 Habitat Navigation Challenge} in both RGB and RGB-D tracks among over 150 submissions from 16 teams~\cite{habitatchallenge}. We were also able to successfully transfer the trained policy to the real-world on a mobile robot. \\

\noindent\textbf{Neural Topological SLAM.} The above model relies on metric spatial maps for planning. While metric maps are effective at spatial navigation tasks like PointGoal or Exploration, they have two major shortcomings: first, metric maps do not scale well with environment size and amount of experience. Actuation noise on real-robots makes it challenging to build consistent representations over long trajectories, and precise localization may not always be possible. Second, metric spatial maps are unable to capture many relevant semantic properties of the environment. In Chapter~\ref{chap:nts}, we design topological representations for space that effectively leverage semantics and afford approximate geometric reasoning. At the heart of our representations are nodes with associated semantic features, that are interconnected using coarse geometric information. We describe supervised learning-based algorithms that can build, maintain and use such representations under noisy actuation. Experimental study in visually and physically realistic simulation suggests that our method builds effective representations that capture structural regularities and efficiently solve long-horizon navigation problems. We observe a relative improvement of more than 50\% over existing methods at the task of ImageGoal navigation. \\

\noindent\textbf{Goal-Oriented Semantic Exploration.} The Active Neural SLAM (ANS) model builds metric obstacle maps which do not encode any semantics explicitly. Tackling semantic navigations tasks like Object Goal Navigation requires encoding semantics in the episodic memory and learning semantic priors. Furthermore, the exploration policy learnt by Active Neural SLAM is goal-agnostic as it learns to maximize coverage. In Chapter~\ref{chap:semexp}, we propose `Goal-Oriented Semantic Exploration' (SemExp), which makes two improvements over ANS to tackle semantic navigation tasks. First, it builds top-down metric maps similar to ANS but adds extra channels to encode semantic categories explicitly. Instead of predicting the top-down maps directly from the first-person image, it uses first-person predictions followed by differentiable geometric projections. This allows us to leverage existing pretrained object detection and semantic segmentation models to build semantic maps instead of learning from scratch. Second, instead of using a coverage maximizing goal-agnostic exploration policy based only on obstacle maps, we train a goal-oriented semantic exploration policy which learns semantic priors for efficient navigation. Our experiments in visually realistic simulation environments show that SemExp outperforms prior methods by a significant margin. The proposed model also won the CVPR 2020 Habitat ObjectNav Challenge~\cite{batra2020objectnav}~\cite{habitatchallenge2020}. We also demonstrate that SemExp achieves similar performance in the real-world when transferred to a mobile robot platform. \\

\noindent\textbf{Semantic Curiosity.} The semantic mapping learnt by the SemExp model can also be used for active visual learning, where an embodied agent actively explores the environment to collect observations to learn an object detection model. In Chapter~\ref{chap:semcur}, we explore a self-supervised approach for training an exploration policy by introducing a notion of semantic curiosity. Our semantic curiosity policy is based on a simple observation -- the detection outputs should be consistent. Therefore, our semantic curiosity rewards trajectories with inconsistent labeling behavior and encourages the exploration policy to explore such areas. We operationalize the notion of semantic curiosity by calculating the inconsistency in building a semantic map. The exploration policy trained via semantic curiosity generalizes to novel scenes and helps train an object detector that outperforms baselines trained with other possible alternatives such as random exploration, prediction-error curiosity, and coverage-maximizing exploration (such as in Active Neural SLAM). \\

\chapter{Learning visual navigation from first-person observations}
\label{chap:fps}

Deep reinforcement learning has proved to be very successful in mastering human-level control policies in a wide variety of tasks such as object recognition with visual attention \cite{ba2014multiple}, high-dimensional robot control \cite{levine2016end} and solving physics-based control problems \cite{heess2015learning}. In particular, Deep Q-Networks (DQN) are shown to be effective in playing Atari 2600 games \cite{mnih2013playing} and more recently, in defeating world-class Go players \cite{silver2016mastering}.

However, there is a limitation in all of the above applications in their assumption of having the full knowledge of the current state of the environment, which is usually not true in real-world scenarios. In the case of partially observable states, the learning agent needs to remember previous states in order to select optimal actions. Recently, there have been attempts to handle partially observable states in deep reinforcement learning by introducing recurrency in Deep Q-networks. For example, \citet{hausknecht2015deep} use a deep recurrent neural network, particularly a Long-Short-Term-Memory (LSTM) Network, to learn the Q-function to play Atari 2600 games. \citet{foerster2016learning} consider a multi-agent scenario where they use deep distributed recurrent neural networks to communicate between different agent in order to solve riddles. The use of recurrent neural networks is effective in scenarios with partially observable states due to its ability to remember information for an arbitrarily long amount of time.

\begin{wrapfigure}{r}{0.5\textwidth}
\vspace{-12pt}
\centering
\includegraphics[width=1.0\linewidth,height=\textheight,keepaspectratio]{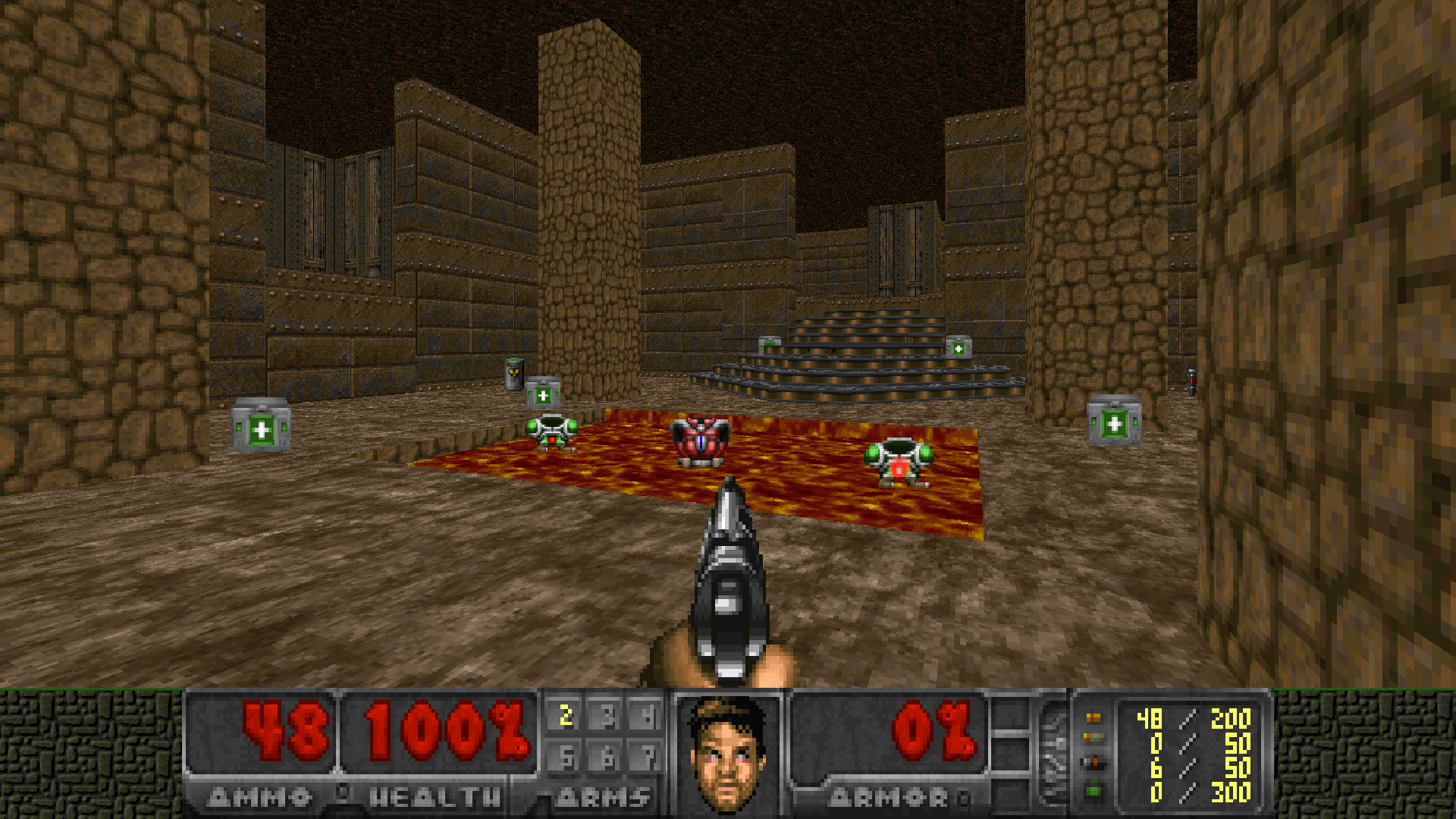}
\caption{A screenshot of Doom.}
\label{fig_fps:doom_scrn}
\vspace{-6pt}
\end{wrapfigure}

Previous methods have usually been applied to 2D environments that hardly resemble the real world. In this chapter, we tackle the task of playing a First-Person-Shooting (FPS) game in a 3D environment. This task is much more challenging than playing most Atari games as it involves a wide variety of skills, such as navigating through a map, collecting items, recognizing and fighting enemies, etc. Furthermore, states are partially observable, and the agent navigates a 3D environment in a first-person perspective, which makes the task more suitable for real-world robotics applications.

In this chapter, we present an AI-agent\footnote{Code: \url{https://github.com/glample/Arnold}} for playing deathmatches\footnote{A deathmatch is a scenario in FPS games where the objective is to maximize the number of kills by a player/agent.} in FPS games using only the pixels on the screen. Our agent divides the problem into two phases: navigation (exploring the map to collect items and find enemies) and action (fighting enemies when they are observed), and uses separate networks for each phase of the game. Furthermore, the agent infers high-level game information, such as the presence of enemies on the screen, to decide its current phase and to improve its performance. We also introduce a method for co-training a DQN with game features, which turned out to be critical in guiding the convolutional layers of the network to detect enemies. We show that co-training significantly improves the training speed and performance of the model.

We evaluate our model on the two different tasks adapted from the Visual Doom AI Competition (ViZDoom)\footnote{ViZDoom Competition at IEEE Computational Intelligence And Games (CIG) Conference, 2016 (http://vizdoom.cs.put.edu.pl/competition-cig-2016)} using the API developed by \citet{kempka2016vizdoom} (Figure~\ref{fig_fps:doom_scrn} shows a screenshot of Doom).
The API gives a direct access to the Doom game engine and allows to synchronously send commands to the game agent and receive inputs of the current state of the game. We show that the proposed architecture substantially outperforms built-in AI agents of the game as well as humans in deathmatch scenarios and we demonstrate the importance of each component of our architecture.

\section{Background}
\label{sec_fps:background}

Below we give a brief summary of the DQN and DRQN models.

\subsection{Deep Q-Networks}

Reinforcement learning deals with learning a policy for an agent interacting in an unknown environment. At each step, an agent observes the current state $s_t$ of the environment, decides of an action $a_t$ according to a policy $\pi$, and observes a reward signal $r_t$. The goal of the agent is to find a policy that maximizes the expected sum of discounted rewards $R_t$

$$ R_t = \sum\limits_{t'=t}^T \gamma^{t'-t} r_{t'}$$

\noindent where T is the time at which the game terminates, and $\gamma \in \left[ 0, 1\right]$ is a discount factor that determines the importance of future rewards. The $Q$-function of a given policy $\pi$ is defined as the expected return from executing an action $a$ in a state $s$:

$$Q^{\pi}(s, a) = \mathbb{E} \left[ R_t | s_t = s, a_t = a \right]$$

\noindent It is common to use a function approximator to estimate the action-value function $Q$. In particular, DQN uses a neural network parametrized by $\theta$, and the idea is to obtain an estimate of the $Q$-function of the current policy which is close to the optimal $Q$-function $Q^{*}$ defined as the highest return we can expect to achieve by following any strategy:

$$Q^{*}(s, a) = \max_{\pi} \mathbb{E} \left[ R_t | s_t = s, a_t = a \right] = \max_{\pi} Q^{\pi}(s, a)$$

\noindent In other words, the goal is to find $\theta$ such that $Q_{\theta}(s, a) \approx Q^{*}(s, a)$. The optimal $Q$-function verifies the Bellman optimality equation

$$Q^{*}(s, a) = \mathbb{E} \big[ r + \gamma \max_{a'} Q^{*}(s', a') | s, a \big]$$

\noindent If $Q_{\theta} \approx Q^{*}$, it is natural to think that $Q_{\theta}$ should be close from also verifying the Bellman equation. This leads to the following loss function:

$$L_t (\theta_t) = \mathbb{E}_{s,a,r,s'} \bigg[ \big( y_t - Q_{\theta_t}(s, a) \big)^{2} \bigg]$$

\noindent where $t$ is the current time step, and $y_t = r + \gamma \max_{a'} Q_{\theta_t}(s', a')$. The value of $y_t$ is fixed, which leads to the following gradient:

$$\nabla_{\theta_t} L_t (\theta_t) = \mathbb{E}_{s,a,r,s'} \bigg[ \big(y_t - Q_{\theta}(s, a) \big) \nabla_{\theta_t} Q_{\theta_t}(s, a) \bigg]$$

\noindent Instead of using an accurate estimate of the above gradient, we compute it using the following approximation:

$$\nabla_{\theta_t} L_t (\theta_t) \approx \big(y_t - Q_{\theta}(s, a) \big) \nabla_{\theta_t} Q_{\theta_t}(s, a)$$

Although being a very rough approximation, these updates have been shown to be stable and to perform well in practice.

Instead of performing the Q-learning updates in an online fashion, it is popular to use experience replay \cite{lin1993reinforcement} to break correlation between successive samples. At each time steps, agent experiences $(s_t, a_t, r_t, s_{t + 1})$ are stored in a replay memory, and the Q-learning updates are done on batches of experiences randomly sampled from the memory.

At every training step, the next action is generated using an $\epsilon$-greedy strategy: with a probability $\epsilon$ the next action is selected randomly, and with probability $1 - \epsilon$ according to the network best action. In practice, it is common to start with $\epsilon = 1$ and to progressively decay $\epsilon$.

\subsection{Deep Recurrent Q-Networks}

The above model assumes that at each step, the agent receives a full observation $s_t$ of the environment - as opposed to games like Go, Atari games actually rarely return a full observation, since they still contain hidden variables, but the current screen buffer is usually enough to infer a very good sequence of actions. But in partially observable environments, the agent only receives an observation $o_t$ of the environment which is usually not enough to infer the full state of the system. A FPS game like DOOM, where the agent field of view is limited to $90^{\circ}$ centered around its position, obviously falls into this category.

To deal with such environments, \citet{hausknecht2015deep} introduced the Deep Recurrent Q-Networks (DRQN), which does not estimate $Q(s_t, a_t)$, but $Q(o_t, h_{t - 1}, a_t)$, where $h_t$ is an extra input returned by the network at the previous step, that represents the hidden state of the agent. A recurrent neural network like a LSTM \cite{hochreiter1997long} can be implemented on top of the normal DQN model to do that. In that case, $h_t = \textrm{LSTM}(h_{t - 1}, o_t)$, and we estimate $Q(h_t, a_t)$. Our model is built on top of the DRQN architecture.

\begin{figure*}
\includegraphics[width=\textwidth,height=\textheight,keepaspectratio]{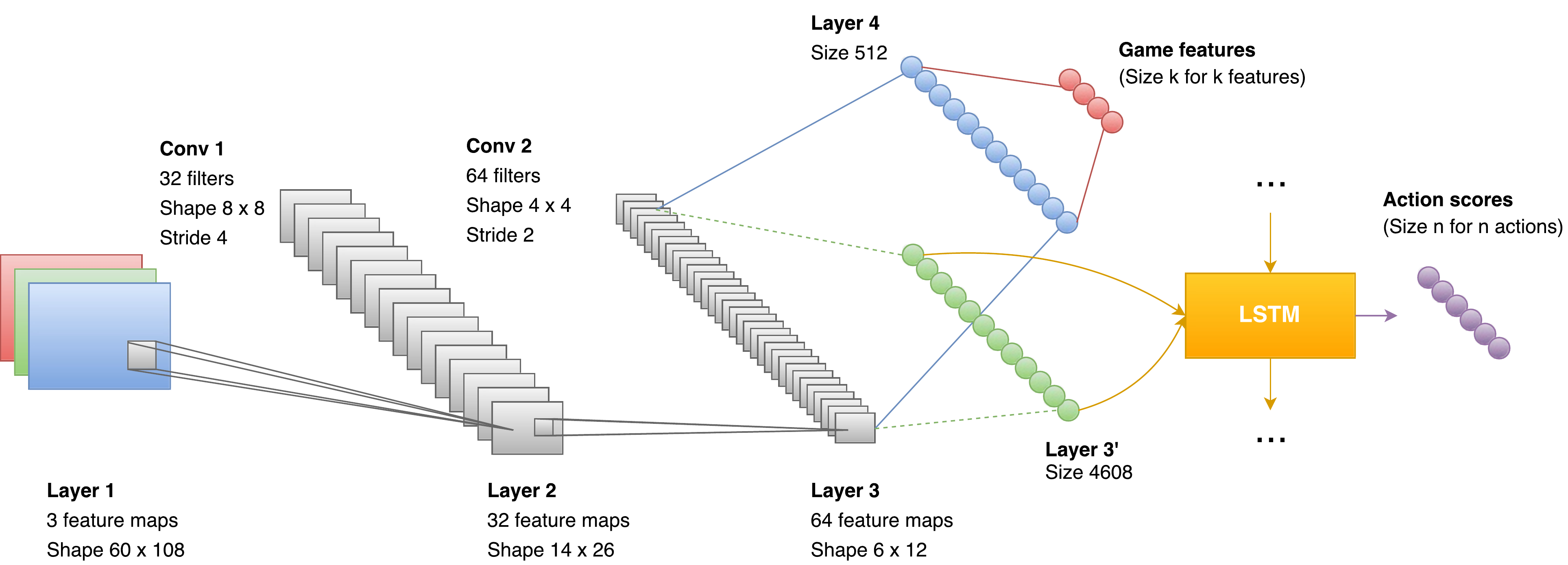}
\caption{An illustration of the architecture of our model. The input image is given to two convolutional layers. The output of the convolutional layers is split into two streams. The first one (bottom) flattens the output (layer 3') and feeds it to a LSTM, as in the DRQN model. The second one (top) projects it to an extra hidden layer (layer 4), then to a final layer representing each game feature. During the training, the game features and the Q-learning objectives are trained jointly.}
\label{fig_fps:model-architecture}
\end{figure*}

\section{Model}
\label{sec_fps:model}

Our first approach to solving the problem was to use a baseline DRQN model. Although this model achieved good performance in relatively simple scenarios (where the only available actions were to turn or attack), it did not perform well on deathmatch tasks. The resulting agents were firing at will, hoping for an enemy to come under their lines of fire. Giving a penalty for using ammo did not help: with a small penalty, agents would keep firing, and with a big one they would just never fire.

\subsection{Game feature augmentation}
\label{sec_fps:game-feature-augmentation}

We reason that the agents were not able to accurately detect enemies. The ViZDoom environment gives access to internal variables generated by the game engine. We modified the game engine so that it returns, with every frame, information about the visible entities. Therefore, at each step, the network receives a frame, as well as a Boolean value for each entity, indicating whether this entity appears in the frame or not (an entity can be an enemy, a health pack, a weapon, ammo, etc). Although this internal information is not available at test time, it can be exploited during training. We modified the DRQN architecture to incorporate this information and to make it sensitive to game features. In the initial model, the output of the convolutional neural network (CNN) is given to a LSTM that predicts a score for each action based on the current frame and its hidden state. We added two fully-connected layers of size $512$ and $k$ connected to the output of the CNN, where $k$ is the number of game features we want to detect. At training time, the cost of the network is a combination of the normal DRQN cost and the cross-entropy loss. Note that the LSTM only takes as input the CNN output, and is never directly provided with the game features. An illustration of the architecture is presented in Figure~\ref{fig_fps:model-architecture}.

\begin{figure*}
\minipage{0.32\textwidth}
\includegraphics[width=\linewidth,height=\textheight,keepaspectratio]{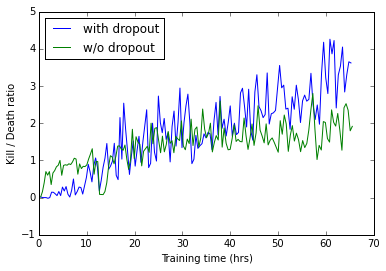}
\endminipage\hfill
\minipage{0.33\textwidth}
\includegraphics[width=\linewidth,height=\textheight,keepaspectratio]{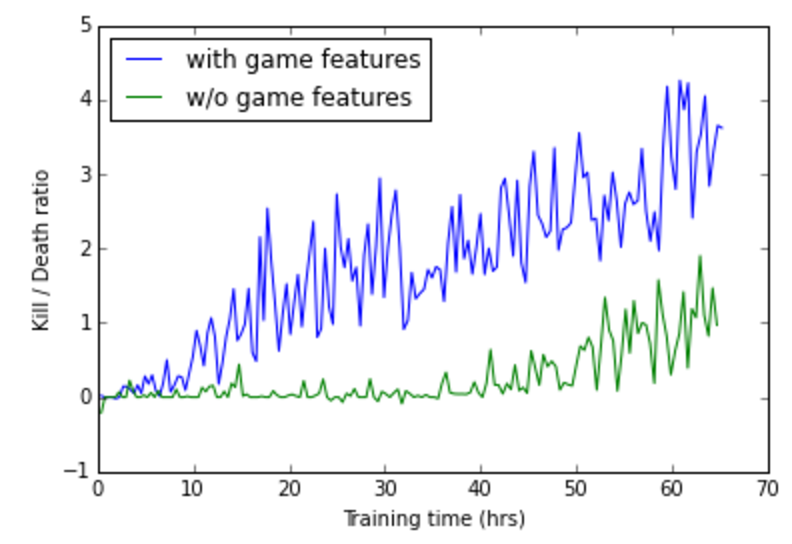}
\endminipage\hfill
\minipage{0.32\textwidth}%
\includegraphics[width=\linewidth,height=\textheight,keepaspectratio]{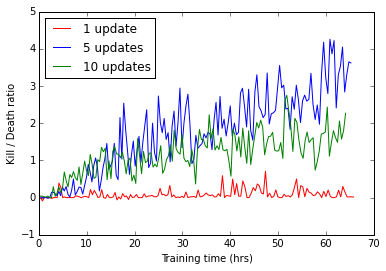}
\endminipage
\caption{Plot of K/D score of action network on limited deathmatch as a function of training time (a) with and without dropout (b) with and without game features, and (c) with different number of updates in the LSTM.}
\label{fig_fps:graphs}
\end{figure*}

Although a lot of game information was available, we only used an indicator about the presence of enemies on the current frame. Adding this game feature dramatically improved the performance of the model on every scenario we tried. Figure~\ref{fig_fps:graphs} shows the performance of the DRQN with and without the game features. We explored other architectures to incorporate game features, such as using a separate network to make predictions and reinjecting the predicted features into the LSTM, but this did not achieve results better than the initial baseline, suggesting that sharing the convolutional layers is decisive in the performance of the model. Jointly training the DRQN model and the game feature detection allows the kernels of the convolutional layers to capture the relevant information about the game. In our experiments, it only takes a few hours for the model to reach an optimal enemy detection accuracy of 90\%. After that, the LSTM is given features that often contain information about the presence of enemy and their positions, resulting in accelerated training.

Augmenting a DRQN model with game features is straightforward. However, the above method can not be applied easily to a DQN model. Indeed, the important aspect of the model is the sharing of the convolution filters between predicting game features and the Q-learning objective. The DRQN is perfectly adapted to this setting since the network takes as input a single frame, and has to predict what is visible in this specific frame. However, in a DQN model, the network receives $k$ frames at each time step, and will have to predict whether some features appear in the last frame only, independently of the content of the $k-1$ previous frames. Convolutional layers do not perform well in this setting, and even with dropout we never obtained an enemy detection accuracy above 70\% using that model.

\subsection{Divide and conquer} 
The deathmatch task is typically divided into two phases, one involves exploring the map to collect items and to find enemies, and the other consists in fighting enemies \cite{mcpartland2008learning,tastan2011learning}. We call these phases the navigation and action phases. Having two networks work together, each trained to act in a specific phase of the game should naturally lead to a better overall performance. Current DQN models do not allow for the combination of different networks optimized on different tasks. However, the current phase of the game can be determined by predicting whether an enemy is visible in the current frame (action phase) or not (navigation phase), which can be inferred directly from the game features present in the proposed model architecture.

There are various advantages of splitting the task into two phases and training a different network for each phase. First, this makes the architecture modular and allows different models to be trained and tested independently for each phase. Both networks can be trained in parallel, which makes the training much faster as compared to training a single network for the whole task. Furthermore, the navigation phase only requires three actions (move forward, turn left and turn right), which dramatically reduces the number of state-action pairs required to learn the Q-function, and makes the training much faster \cite{gaskett1999q}. More importantly, using two networks also mitigates ``camper'' behavior, i.e. tendency to stay in one area of the map and wait for enemies, which was exhibited by the agent when we tried to train a single DQN or DRQN for the deathmatch task. 

We trained two different networks for our agent. We used a DRQN augmented with game features for the action network, and a simple DQN for the navigation network. During the evaluation, the action network is called at each step. If no enemies are detected in the current frame, or if the agent does not have any ammo left, the navigation network is called to decide the next action. Otherwise, the decision is given to the action network. Results in Table~\ref{tab_fps:splitting} demonstrate the effectiveness of the navigation network in improving the performance of our agent.

\begin{figure}
\begin{center}
\includegraphics[width=0.6\linewidth,height=\textheight,keepaspectratio]{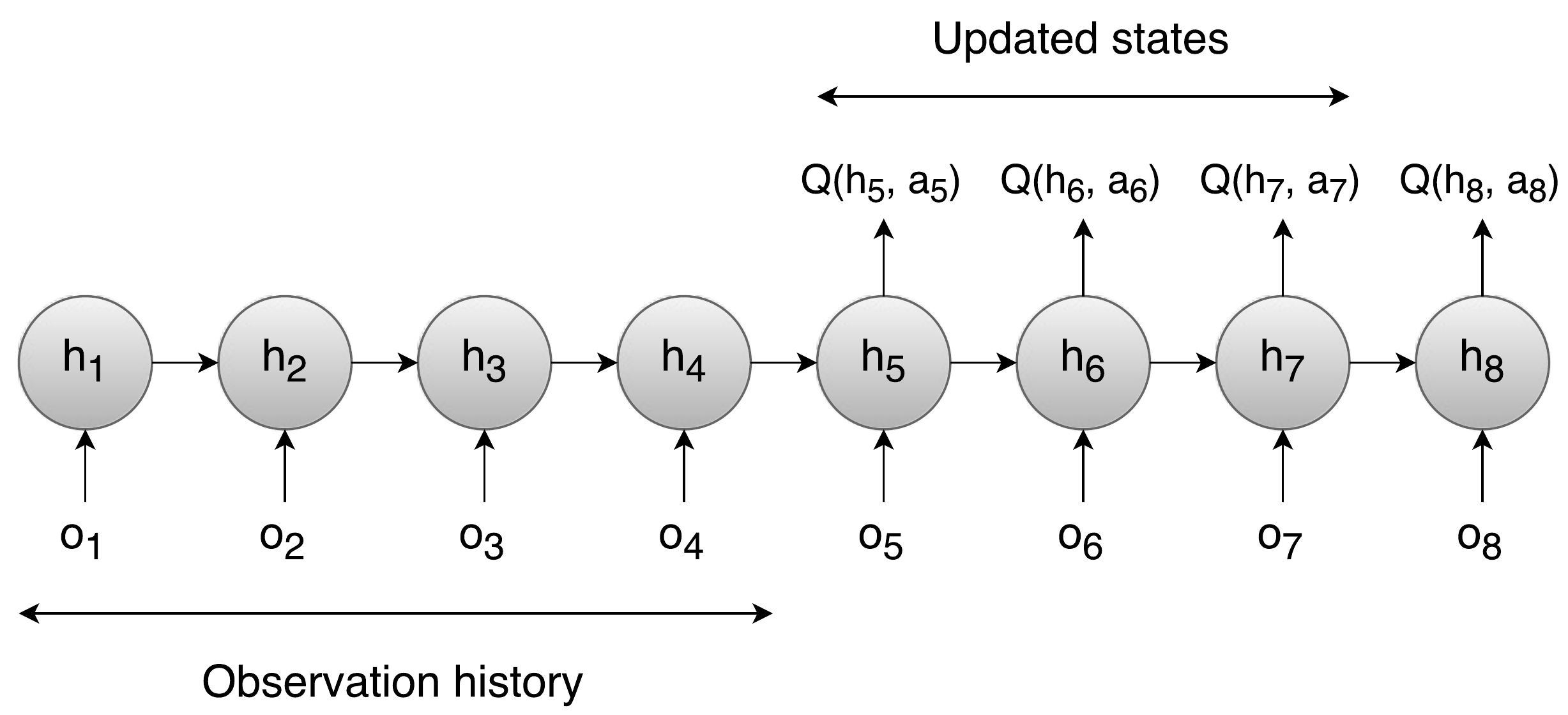}
\caption{DQN updates in the LSTM. Only the scores of the actions taken in states 5, 6 and 7 will be updated. First four states provide a more accurate hidden state to the LSTM, while the last state provide a target for state 7.}
\label{fig_fps:drqnu}
\end{center}
\end{figure}

\section{Training}
\label{sec_fps:training}
\subsection{Reward shaping}
The score in the deathmatch scenario is defined as the number of frags, i.e. number of kills minus number of suicides. If the reward is only based on the score, the replay table is extremely sparse w.r.t state-action pairs having non-zero rewards, which makes it very difficult for the agent to learn favorable actions. Moreover, rewards are extremely delayed and are usually not the result of a specific action: getting a positive reward requires the agent to explore the map to find an enemy and accurately aim and shoot it with a slow projectile rocket. The delay in reward makes it difficult for the agent to learn which set of actions is responsible for what reward.
To tackle the problem of sparse replay table and delayed rewards, we introduce reward shaping, i.e. the modification of reward function to include small intermediate rewards to speed up the learning process \cite{ng2003shaping}. In addition to positive reward for kills and negative rewards for suicides, we introduce the following intermediate rewards for shaping the reward function of the action network:
\begin{itemize}
\item positive reward for object pickup (health, weapons and ammo)
\item negative reward for loosing health (attacked by enemies or walking on lava)
\item negative reward for shooting, or loosing ammo
\end{itemize}

We used different rewards for the navigation network. Since it evolves on a map without enemies and its goal is just to gather items, we simply give it a positive reward when it picks up an item, and a negative reward when it's walking on lava. We also found it very helpful to give the network a small positive reward proportional to the distance it travelled since the last step. That way, the agent is faster to explore the map, and avoids turning in circles.

\subsection{Frame skip}
Like in most previous approaches, we used the frame-skip technique \cite{bellemare2012arcade}. In this approach, the agent only receives a screen input every $k + 1$ frames, where $k$ is the number of frames skipped between each step. The action decided by the network is then repeated over all the skipped frames. A higher frame-skip rate accelerates the training, but can hurt the performance. Typically, aiming at an enemy sometimes requires to rotate by a few degrees, which is impossible when the frame skip rate is too high, even for human players, because the agent will repeat the rotate action many times and ultimately rotate more than it intended to. A frame skip of $k = 4$ turned out to be the best trade-off.

\subsection{Sequential updates}

To perform the DRQN updates, we use a different approach from the one presented by \citet{hausknecht2015deep}. A sequence of $n$ observations $o_1, o_2, ..., o_n$ is randomly sampled from the replay memory, but instead of updating all action-states in the sequence, we only consider the ones that are provided with enough history. Indeed, the first states of the sequence will be estimated from an almost non-existent history (since $h_0$ is reinitialized at the beginning of the updates), and might be inaccurate. As a result, updating them might lead to imprecise updates.

To prevent this problem, errors from states $o_1 ... o_h$, where $h$ is the minimum history size for a state to be updated, are not backpropagated through the network. Errors from states $o_{h + 1} .. o_{n - 1}$ will be backpropagated, $o_n$ only being used to create a target for the $o_{n - 1}$ action-state. An illustration of the updating process is presented in Figure~\ref{fig_fps:drqnu}, where $h=4$ and $n=8$. In all our experiments, we set the minimum history size to $4$, and we perform the updates on $5$ states.
Figure~\ref{fig_fps:graphs} shows the importance of selecting an appropriate number of updates. Increasing the number of updates leads to high correlation in sampled frames, violating the DQN random sampling policy, while decreasing the number of updates makes it very difficult for the network to converge to a good policy.

\setlength\tabcolsep{4.8pt}
\begin{table}[t]
\centering
\label{my-label}

\begin{tabular}{@{}lccccc@{}}
\toprule
                         & \multicolumn{2}{c}{Single Player} & \multicolumn{1}{l}{} & \multicolumn{2}{c}{Multiplayer} \\ \midrule
Evaluation Metric        & Human           & Agent          & \multicolumn{1}{l}{} & Human          & Agent         \\ \midrule
Number of objects        & 5.2             & 9.2             &                      & 6.1            & 10.5           \\
Number of kills          & 12.6            & 27.6            &                      & 5.5            & 8.0            \\
Number of deaths         & 8.3             & 5.0             &                      & 11.2           & 6.0            \\
Number of suicides       & 3.6             & 2.0             &                      & 3.2            & 0.5            \\
K/D Ratio      & 1.52            & 5.12            &                      & 0.49           & 1.33           \\ \bottomrule
\end{tabular}
\caption{Comparison of human players with agent. Single player scenario is both humans and the agent playing against bots in separate games. Multiplayer scenario is agent and human playing against each other in the same game.}
\label{tab_fps:results}
\end{table}

\section{Experiments}
\label{sec_fps:experiment}
\setlength\tabcolsep{8pt}
\begin{table*}[t]
\centering
\resizebox{\textwidth}{!}{%
\begin{tabular}{@{}lcccccccc@{}}
\toprule
                         & \multicolumn{2}{c}{Limited Deathmatch}                                                                                    & \multicolumn{1}{l}{} & \multicolumn{5}{c}{Full Deathmatch}                                                                                                                                                                                                                                        \\ \midrule
                         & \multicolumn{2}{c}{Known Map}                                                                                             & \multicolumn{1}{l}{} & \multicolumn{2}{c}{Train maps}                                                                                 & \multicolumn{1}{l}{} & \multicolumn{2}{c}{Test maps}                                                                                  \\ \midrule
Evaluation Metric        & \begin{tabular}[c]{@{}c@{}}Without\\ navigation\end{tabular} & \begin{tabular}[c]{@{}c@{}}With \\ navigation\end{tabular} & \multicolumn{1}{l}{} & \begin{tabular}[c]{@{}c@{}}Without\\ navigation\end{tabular} & \begin{tabular}[c]{@{}c@{}}With\\ navigation\end{tabular} & \multicolumn{1}{l}{} & \begin{tabular}[c]{@{}c@{}}Without\\ navigation\end{tabular} & \begin{tabular}[c]{@{}c@{}}With\\ navigation\end{tabular} \\ \midrule
Number of objects        & 14        & 46        &           & 52.9               & 92.2                                                        &                      & 62.3                                                           &  94.7                                                        \\
Number of kills          & 167       & 138       &           & 43.0               & 66.8                                                         &                      & 32.0                                                            & 43.0                                                         \\
Number of deaths         & 36        & 25        &           & 15.2               & 14.6                                                         &                      & 10.0                                                            & 6.0                                                         \\
Number of suicides       & 15        & 10        &           & 1.7                & 3.1                                                         &                      & 0.3                                                            & 1.3                                                         \\
Kill to Death Ratio      & 4.64      & 5.52      &           & 2.83               & 4.58                                                         &                      & 3.12                                                            & 6.94                                                         \\ \bottomrule
\end{tabular}}
\caption{Performance of the agent against in-built game bots with and without navigation. The agent was evaluated 15 minutes on each map. The performance on the full deathmatch task was averaged over 10 train maps and 3 test maps.}
\label{tab_fps:splitting}
\end{table*}

\subsection{Hyperparameters}
All networks were trained using the RMSProp algorithm and minibatches of size $32$. Network weights were updated every $4$ steps, so experiences are sampled on average 8 times during the training \cite{van2015deep}. The replay memory contained the one million most recent frames. The discount factor was set to $\gamma = 0.99$. We used an $\epsilon$-greedy policy during the training, where $\epsilon$ was linearly decreased from $1$ to $0.1$ over the first million steps, and then fixed to $0.1$.

Different screen resolutions of the game can lead to a different field of view. In particular, a 4/3 resolution provides a 90 degree field of view, while a 16/9 resolution in Doom has a 108 degree field of view (as shown in Figure~\ref{fig_fps:doom_scrn}). In order to maximize the agent game awareness, we used a 16/9 resolution of 440x225 which we resized to 108x60. Although faster, our model obtained a lower performance using grayscale images, so we decided to use colors in all experiments.

\subsection{Scenario}
We use the ViZDoom platform \cite{kempka2016vizdoom} to conduct all our experiments and evaluate our methods on the deathmatch scenario. In this scenario, the agent plays against built-in Doom bots, and the final score is the number of frags, i.e. number of bots killed by the agent minus the number of suicides committed. We consider two variations of this scenario, adapted from the ViZDoom AI Competition:
\subsubsection{Limited deathmatch on a known map.}
The agent is trained and evaluated on the same map, and the only available weapon is a rocket launcher. Agents can gather health packs and ammo.
\subsubsection{Full deathmatch on unknown maps.}
The agent is trained and tested on different maps. The agent starts with a pistol, but can pick up different weapons around the map, as well as gather health packs and ammo. We use 10 maps for training and 3 maps for testing. We further randomize the textures of the maps during the training, as it improved the generalizability of the model.
\\
The limited deathmatch task is ideal for demonstrating the model design effectiveness and to chose hyperparameters, as the training time is significantly lower than on the full deathmatch task. In order to demonstrate the generalizability of our model, we use the full deathmatch task to show that our model also works effectively on unknown maps.

\subsection{Evaluation Metrics}
For evaluation in deathmatch scenarios, we use Kill to death (K/D) ratio as the scoring metric. Since K/D ratio is susceptible to ``camper'' behavior to minimize deaths, we also report number of kills to determine if the agent is able to explore the map to find enemies. In addition to these, we also report the total number of objects gathered, the total number of deaths and total number of suicides (to analyze the effects of different design choices). Suicides are caused when the agent shoots too close to itself, with a weapon having blast radius like rocket launcher. Since suicides are counted in deaths, they provide a good way for penalizing K/D score when the agent is shooting arbitrarily. 

\subsection{Results \& Analysis}
\label{sec_fps:results}
Demonstrations of navigation and deathmatch on known and unknown maps are available \href{https://www.youtube.com/playlist?list=PLduGZax9wmiHg-XPFSgqGg8PEAV51q1FT}{here}\footnote{\url{https://www.youtube.com/playlist?list=PLduGZax9wmiHg-XPFSgqGg8PEAV51q1FT}}. Arnold, an agent trained using the proposed Action-Navigation architecture placed second in both the tracks Visual Doom AI Competition with the highest K/D Ratio \cite{chaplot2017arnold}.

\subsubsection{Navigation network enhancement.} Scores on both the tasks with and without navigation are presented in Table~\ref{tab_fps:splitting}. The agent was evaluated 15 minutes on all the maps, and the results have been averaged for the full deathmatch maps. In both scenarios, the total number of objects picked up dramatically increases with navigation, as well as the K/D ratio. In the full deathmatch, the agent starts with a pistol, with which it is relatively difficult to kill enemies. Therefore, picking up weapons and ammo is much more important in the full deathmatch, which explains the larger improvement in K/D ratio in this scenario. The improvement in limited deathmatch scenario is limited because the map was relatively small, and since there were many bots, navigating was not crucial to find other agents. However, the agent was able to pick up more than three times as many objects, such as health packs and ammo, with navigation. Being able to heal itself regularly, the agent decreased its number of deaths and improved its K/D ratio. Note that the scores across the two different tasks are not comparable due to difference in map sizes and number of objects between the different maps. The performance on the test maps is better than on the training maps, which is not necessarily surprising given that the maps all look very different. In particular, the test maps contain less stairs and differences in level, that are usually difficult for the network to handle since we did not train it to look up and down.

\subsubsection{Comparison to human players.}
Table~\ref{tab_fps:results} shows that our agent outperforms human players in both the single player and multiplayer scenarios. In the single player scenario, human players and the agent play separately against 10 bots on the limited deathmatch map, for three minutes. In the multiplayer scenario, human players and the agent play against each other on the same map, for five minutes. Human scores are averaged over 20 human players in both scenarios. Note that the suicide rate of humans is particularly high indicating that it is difficult for humans to aim accurately in a limited reaction time.

\subsubsection{Game features.} Detecting enemies is critical to our agent's performance, but it is not a trivial task as enemies can appear at various distances, from different angles and in different environments. Including game features while training resulted in a significant improvement in the performance of the model, as shown in Figure~\ref{fig_fps:graphs}. After 65 hours of training, the best K/D score of the network without game features is less than $2.0$, while the network with game features is able to achieve a maximum score over $4.0$.

Another advantage of using game features is that it gives immediate feedback about the quality of the features given by the convolutional network. If the enemy detection accuracy is very low, the LSTM will not receive relevant information about the presence of enemies in the frame, and Q-learning network will struggle to learn a good policy. The enemy detection accuracy takes few hours to converge while training the whole model takes up to a week. Since the enemy detection accuracy correlates with the final model performance, our architecture allows us to quickly tune our hyperparameters without training the complete model.

For instance, the enemy detection accuracy with and without dropout quickly converged to 90\% and 70\% respectively, which allowed us to infer that dropout is crucial for the effective performance of the model. Figure~\ref{fig_fps:graphs} supports our inference that using a dropout layer significantly improves the performance of the action network on the limited deathmatch. 

As explained in Section~\ref{sec_fps:game-feature-augmentation}, game features surprisingly don't improve the results when used as input to the DQN, but only when used for co-training. This suggests that co-training might be useful in any DQN application even with independent image classification tasks like CIFAR100.

\section{Discussion}
\label{sec_fps:conclusions}
\citet{mcpartland2008learning} and \citet{tastan2011learning} divide the tasks of navigation and combat in FPS Games and present reinforcement learning approaches using game-engine information.
\citet{koutnik2013evolving} previously applied a Recurrent Neural Network to learn TORCS, a racing video game, from raw pixels only. 
\citet{kempka2016vizdoom} previously applied a vanilla DQN to simpler scenarios within Doom and provide an empirical study of the effect of changing number of skipped frames during training and testing on the performance of a DQN.

In this chapter, we have presented a complete architecture for playing deathmatch scenarios in FPS games. We introduced a method to augment a DRQN model with high-level game information, and modularized our architecture to incorporate independent networks responsible for different phases of the game. These methods lead to dramatic improvements over the standard DRQN model when applied to complicated tasks like a deathmatch. We showed that the proposed model is able to outperform built-in bots as well as human players and demonstrated the generalizability of our model to unknown maps. Moreover, our methods are complementary to recent improvements in DQN, and could easily be combined with dueling architectures \cite{wang2015dueling}, and prioritized replay \cite{schaul2015prioritized}.

The navigation policies presented in this chapter are capable of obstacle avoidance and semantic perception. They were effective at navigating to particular objects and performing a particular task such as playing deathmatches. In order to train agents capable of navigating to different types of objects, there has been a lot of interest in training goal-conditioned navigation agents. Among different ways of specifying goals, language offers several advantages. First, the compositionality of language allows generalization to new tasks without additional learning and second, language is also a convenient means for humans to communicate with autonomous agents. In the next chapter, we will present methods to train navigation agents capable of language grounding and reasoning and following unseen language instructions.

\chapter{Language Grounding and Reasoning}
\label{chap:grounding}
\blfootnote{Webpage: \url{https://devendrachaplot.github.io/projects/language-grounding}} 
Artificial Intelligence (AI) systems are expected to perceive the environment and take actions to perform a certain task \citep{russell1995modern}. 
Task-oriented language grounding refers to the process of extracting semantically meaningful representations of language by mapping it to visual elements and actions in the environment in order to perform the task specified by the instruction.

\begin{wrapfigure}{r}{0.5\textwidth}
\vspace{-12pt}
\centering
\includegraphics[width=0.99\linewidth,keepaspectratio]{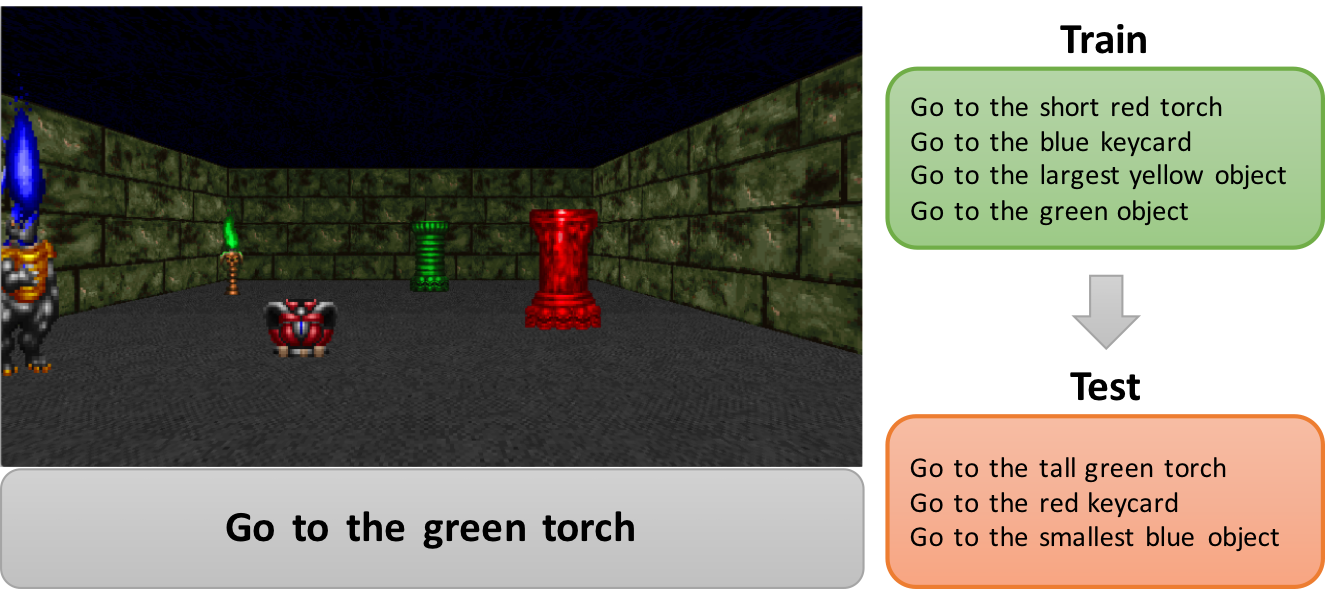}
\caption{\small An example of task-oriented language grounding in the 3D Doom environment with sample instructions. The test set consists of unseen instructions.}
\label{fig_grounding:scrn2}
\vspace{-6pt}
\end{wrapfigure}

Consider the scenario shown in Figure \ref{fig_grounding:scrn2}, where an agent takes natural language instruction and pixel-level visual information as input to carry out the task in the real world. To accomplish this goal, the agent has to draw semantic correspondences between the visual and verbal modalities and learn a policy to perform the task. This problem poses several challenges: the agent has to learn to \emph{recognize} objects in raw pixel input, \emph{explore} the environment as the objects might be occluded or outside the field-of-view of the agent, \emph{ground} each concept of the instruction in visual elements or actions in the environment, \emph{reason} about the pragmatics of language based on the objects in the current environment (for example instructions with superlative tokens, such as `Go to the largest object') and \emph{navigate} to the correct object while avoiding incorrect ones.

\begin{figure*}
\centering
\includegraphics[width=0.99\linewidth,keepaspectratio]{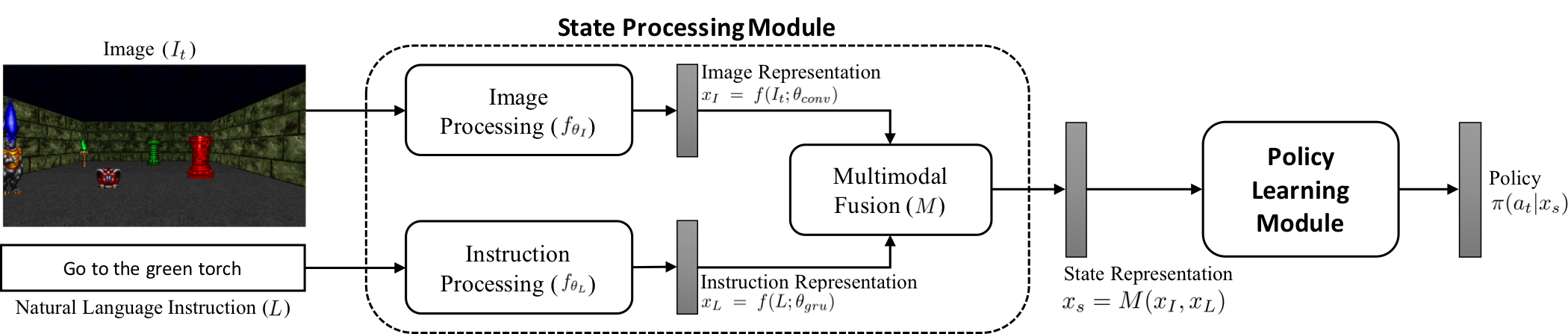}
\caption{\small The proposed model architecture to estimate the policy given the natural language instruction and the image showing the first-person view of the environment.}
\label{fig_grounding:ga_arch}
\end{figure*}

To tackle this problem, we propose an architecture that comprises of a \textit{state processing module} that creates a joint representation of the instruction and the images observed by the agent, and a \textit{policy learner} to predict the optimal action the agent has to take in that timestep.   
The state processing module consists of a novel Gated-Attention multimodal fusion mechanism, 
which is based on multiplicative interactions between both modalities~\citep{dhingra2016gated,Wu2016}. 
 
The contributions of this chapter are summarized as follows:
1) We propose an end-to-end trainable architecture that handles raw pixel-based input for task-oriented language grounding in a 3D environment and assumes no prior linguistic or perceptual knowledge. We show that the proposed model generalizes well to unseen instructions as well as unseen maps.
2) We develop a novel Gated-Attention mechanism for multimodal fusion of representations of verbal and visual modalities. We show that the gated-attention mechanism outperforms the baseline method of concatenating the representations using various policy learning methods. The visualization of the attention weights in the gated-attention unit shows that the model learns to associate attributes of the object mentioned in the instruction with the visual representations learned by the model.
3) We introduce a new environment, built over ViZDoom \citep{kempka2016vizdoom}, for task-oriented language grounding with a rich set of actions, objects and their attributes. The environment provides a first-person view of the world state, and allows for simulating complex scenarios for tasks such as navigation.
\footnote{The code for the environment and the proposed model is available at \url{https://github.com/devendrachaplot/DeepRL-Grounding}}

There has been a lot of related work on language grounding and instruction following in the recent few years. 
In the context of grounding language in objects and their attributes, \citet{guadarrama2014open} present a method to ground open vocabulary to objects in the environment. Several works look at grounding concepts through human-robot interaction \citep{chao2011towards,lemaignan2012grounding}. Other works in grounding include attempts to ground natural language instructions in haptic signals \citep{chu2013using} and teaching robot to ground natural language using active learning \citep{kulick2013active}. 
Some of the work that aims to ground navigational instructions include \citep{guadarrama2013grounding}, \citep{bollini2013interpreting} and \citep{beetz2011robotic}, where the focus was to ground verbs like \emph{go, follow, etc.} and spatial relations of verbs \citep{tellex2011understanding,fasola2013using}. 

Regarding mapping instructions to action sequences,
\citet{chen2011learning} and \citet{artzi2013weakly} present methods based on semantic parsing to map navigational instructions to a sequence of actions. \citet{mei2015listen} look at neural mapping of instructions to sequence of actions, along with input from bag-of-word features extracted from the visual image. While these works focus on grounding navigational instructions to actions in the environment, we aim to ground visual attributes of objects such as shape, size and color.  

Prior work has also explored using Deep Reinforcement learning approaches for playing FPS games \citep{lample2016playing,kempka2016vizdoom,kulkarni2016deep}. The challenge here is to learn optimal policy for a variety of tasks, including navigation using raw visual pixel information. \citet{chaplottransfer} look at transfer learning between different tasks in the Doom Environment. In all these methods, the policy for each task is learned separately using a deep Q-Learning \citep{mnih2013playing}. In contrast, we train a single network for multiple tasks/instructions. 
\citet{zhu2017target} look at target-driven visual navigation, given the image of the target object. We use the natural language instruction and do not have the visual image of the object.
\citet{yu2017deep} look at learning to navigate in a 2D maze-like environment and execute commands, for both seen and \textit{zero-shot} setting, where the combination of words are not seen before. \citet{misra2017mapping} also look at mapping raw visual observations and text input to actions in a 2D Blocks environment. 
While these works also looks at executing a variety of instructions, they tackle only 2D environments. 

Compared to the prior work, this chapter aims to address grounding of natural language instruction in a challenging 3D setting involving raw-pixel input and partially observable envrionment, which poses additional challenges of perception, exploration and reasoning. Unlike many of the previous methods, our model assumes no prior linguistic or perceptual knowledge, and is trainable end-to-end.

\section{Problem Formulation}
\label{sec_grounding:prob}

We tackle the problem of task-oriented language grounding in the context of target-driven visual navigation conditioned on a natural language instruction, where the agent has to navigate to the object described in the instruction. 
Consider an agent interacting with an episodic environment~$\mathcal{E}$. 
In the beginning of each episode, the agent receives a natural language instruction ($L$) which indicates the description of the target, a visual object in the environment. 
At each time step, the agent receives a raw pixel-level image of the first person view of the environment ($I_t$), and performs an action $a_t$. 
The episode terminates whenever the agent reaches any object or the number of time steps exceeds the maximum episode length. 
Let $s_t = \{I_t,L\}$ denote the state at each time step. 
The objective of the agent is to learn an optimal policy $\pi(a_t|s_t)$, which maps the observed states to actions, eventually leading to successful completion of the task. In this case, the task is to reach the correct object before the episode terminates. 
We consider two different learning approaches for language grounding using target-driven visual navigation:
(1) Imitation Learning \citep{bagnell2015invitation}: where the agent has access to an oracle which specifies the optimal action given any state in the environment;
(2) Reinforcement Learning \citep{sutton2018reinforcement}: where the agent receives a positive reward when it reaches the target object and a negative reward when it reaches any other object.

\section{Methods}
\label{sec_grounding:model-arch}

We propose a novel architecture for task-oriented visual language grounding, which assumes no prior linguistic or perceptual knowledge and can be trained end-to-end. The proposed model is divided into two modules, state processing and policy learning, as shown in Figure~\ref{fig_grounding:ga_arch}. 

{\bf State Processing Module}: The state processing module takes the current state $s_t = \{I_t,L\}$ as the input and creates a joint representation for the image and the instruction. 
This joint representation is used by the policy learner to predict the optimal action to take at that timestep. 
It consists of a convolutional network \citep{lecun1995convolutional} to process the image $I_t$, a Gated Recurrent Unit (GRU) \citep{cho2014properties} network to process the instruction~$L$ and
a multimodal fusion unit that combines the representations of the instruction and the image. Let $x_I = f(I_t;\theta_{conv}) \in \mathcal{R}^{d\times H\times W}$ be the representation of the image, where $\theta_{conv}$ denote the parameters of the convolutional network, $d$ denotes number of feature maps (intermediate representations) in the convolutional network output, while $H$ and $W$ denote the height and width of each feature map. Let $x_L = f(L;\theta_{gru})$ be the representation of the instruction, where $\theta_{gru}$ denotes the parameters of the GRU network. The multimodal fusion unit, $M(x_I,x_L)$ combines the image and instruction representations. Many prior methods combine the multimodal representations by concatenation \cite{mei2015listen,misra2017mapping}. We develop a multimodal fusion unit, \textit{Gated-Attention}, based on multiplicative interactions between instruction and image representation.

{\bf Concatenation}:
In this approach, the representations of the image and instruction are simply flattened and concatenated to create a joint state representation:
$$ M_{concat}(x_I,x_L) =  [\textrm{vec}(x_I);\textrm{vec}(x_L)],$$ 

\noindent where $\textrm{vec}(.)$ denotes the flattening operation. The concatenation unit is used as a baseline for the proposed Gated-Attention unit as it is used by prior methods \cite{mei2015listen,misra2017mapping}.

\begin{figure*}
\centering
\begin{minipage}{0.49\textwidth}
\includegraphics[width=0.95\linewidth,height=\textheight,keepaspectratio]{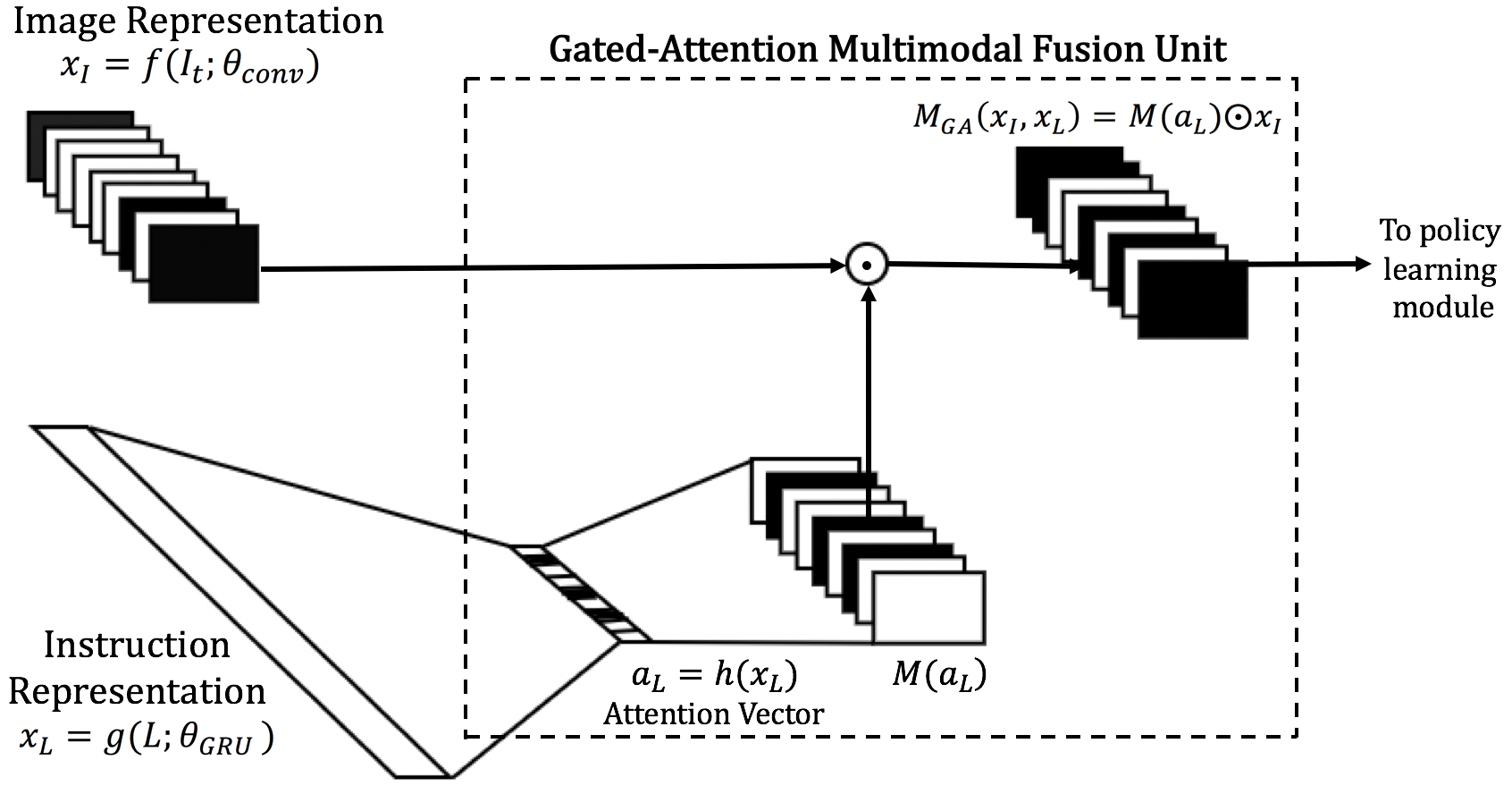}
\caption{Gated-Attention architecture.}
\label{fig_grounding:ga}
\end{minipage}\hfill
\begin{minipage}{0.49\textwidth}
\includegraphics[width=0.95\linewidth,height=\textheight,keepaspectratio]{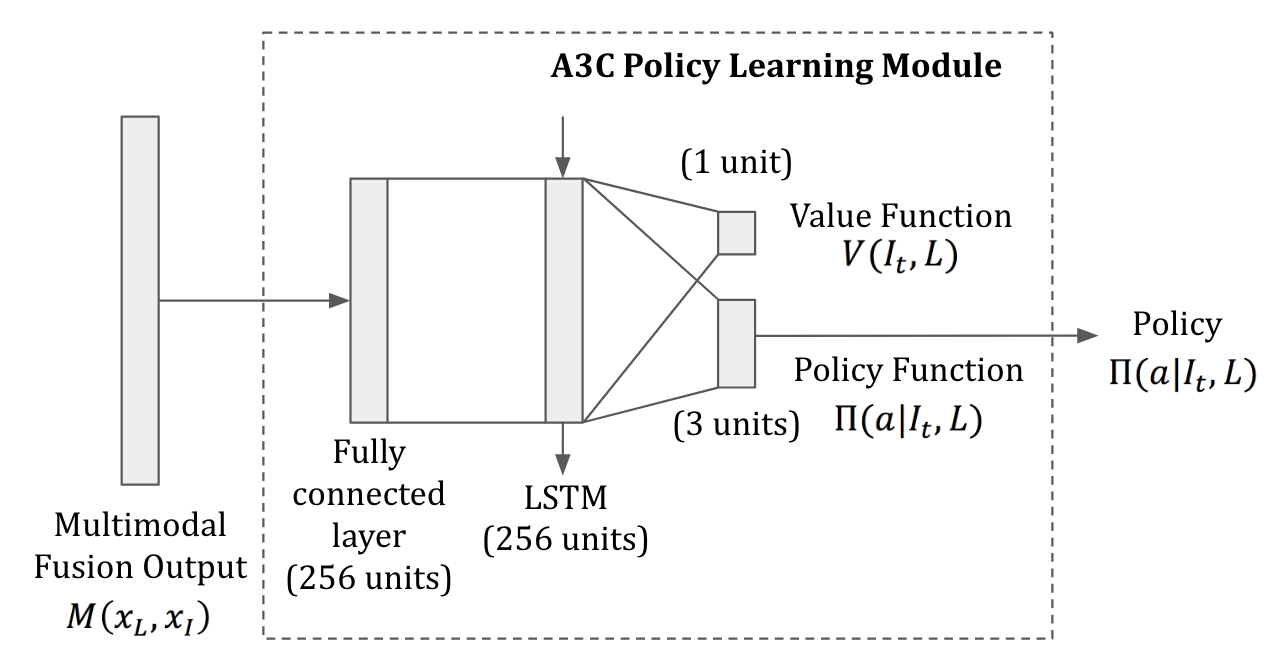}
\caption{A3C policy model architecture.}
\label{fig_grounding:a3c}
\end{minipage}
\end{figure*}

{\bf Gated-Attention}:
In the Gated-Attention unit, the instruction embedding is passed through a fully-connected linear layer with a sigmoid activation. The output dimension of this linear layer, $d$, is equal to the number of feature maps in the output of the convolutional network (first dimension of $x_I$). The output of this linear layer is called the attention vector $a_L = h(x_L) \in \mathcal{R}^d$, where $h(.)$ denotes the fully-connected layer with sigmoid activation. Each element of $a_L$ is expanded to a $H \times W$ matrix. This results in a  3-dimensional matrix,  $M({a_L}) \in \mathcal{R}^{d\times H \times W}$  whose $(i,j,k)^{th}$ element is given by: $M_{a_L}[i,j,k] = a_L[i]$.
This matrix is multiplied element-wise with the output of the convolutional network:
$$M_{GA}(x_I,x_L) = M(h(x_L)) \odot x_I = M(a_L)\odot x_I,$$

\noindent where $\odot$ denotes the Hadamard product \citep{horn1990hadamard}. The architecture of the Gated-Attention unit is shown in Figure~\ref{fig_grounding:ga}. The whole unit is differentiable which makes the architecture end-to-end trainable. 

The proposed Gated-Attention unit is inspired by the Gated-Attention Reader architecture for text comprehension \citep{dhingra2016gated}. They integrate a multi-hop architecture with a Gated-attention mechanism, which is based on multiplicative interactions between the query embedding and the intermediate states of a recurrent neural network document reader. In contrast, we propose a Gated-Attention multimodal fusion unit which is based on multiplicative interactions between the instruction representation and the convolutional feature maps of the image representation. This architecture can be extended to any application of multimodal fusion of verbal and visual modalities.

The intuition behind Gated-Attention unit is that the trained convolutional feature maps detect different attributes of the objects in the frame, such as color and shape. The agent needs to attend to specific attributes of the objects based on the instruction. For example, depending on the whether the instruction is ``Go to the green object'', ``Go to the pillar'' or ``Go to the green pillar'' the agent needs to attend to objects which are `green', objects which look like a `pillar' or both. The Gated-Attention unit is designed to gate specific feature maps based on the attention vector from the instruction, $a_L$. 

\subsection{Policy Learning Module}
The output of the multimodal fusion unit ($M_{concat}$ or $M_{GA}$) is fed to the policy learning module. The architecture of the policy learning module is specific to the learning paradigm: (1) Imitation Learning or (2) Reinforcement Learning.

For imitation learning, we consider two algorithms, Behavioral Cloning \citep{bagnell2015invitation} and DAgger \citep{ross2011reduction}. Both the algorithms require an oracle that can return an optimal action given the current state. The oracle is implemented by extracting agent and target object locations and orientations from the Doom game engine. Given any state, the oracle determines the optimal action as follows: The agent first reorients (using \textit{turn\_left, turn\_right} actions) towards the target object. It moves forward (\textit{move\_forward} action), reorienting towards the target object if deviation of the agent's orientation is greater than the minimum turn angle supported by the environment. 

For reinforcement learning, we use the Asynchronous Advantage Actor-Critic (A3C) algorithm \citep{mnih2016asynchronous} which uses a deep neural network to learn the policy and value functions and runs multiple parallel threads to update the network parameters. We also use the entropy regularization for improved exploration as described by \citep{mnih2016asynchronous}. 
In addition, we use the Generalized Advantage Estimator \citep{schulman2015high} to reduce the variance of the policy gradient updates. 

The policy learning module for imitation learning contains a fully connected layer to estimate the policy function.
The policy learning module for reinforcement learning using A3C (shown in Figure \ref{fig_grounding:a3c}) consists of an LSTM layer, followed by fully connected layers to estimate the policy function as well as the value function. The LSTM layer is introduced so that the agent can have some memory of previous states. This is important as a reinforcement learning agent might explore states where all objects are not visible and need to remember the objects seen previously.

\section{Environment}
\label{sec_grounding:environment}

We create an environment for task-oriented language grounding, where the agent can execute a natural language instruction and obtain a positive reward on successful completion of the task. 
Our environment is built on top of the ViZDoom API \citep{kempka2016vizdoom}, based on Doom, a classic first person shooting game. It provides the raw visual information from a first-person perspective at every timestep. Each scenario in the environment comprises of an agent and a list of objects (a subset of ViZDoom objects) - one correct and rest incorrect in a customized map. The agent can interact with the environment by performing navigational actions such as turn left, turn right, move forward. Given an instruction ``Go to the green torch", the task is considered successful if the agent is able to reach the \emph{green torch} correctly. 
The customizable nature of the environment enables us to create scenarios with varying levels of difficulty which we believe leads to designing sophisticated learning algorithms to address the challenge of multi-task and zero-shot reinforcement learning. 

\begin{wrapfigure}{r}{0.5\textwidth}
\vspace{-12pt}
\centering
\includegraphics[width=0.85\linewidth,height=\textheight,keepaspectratio]{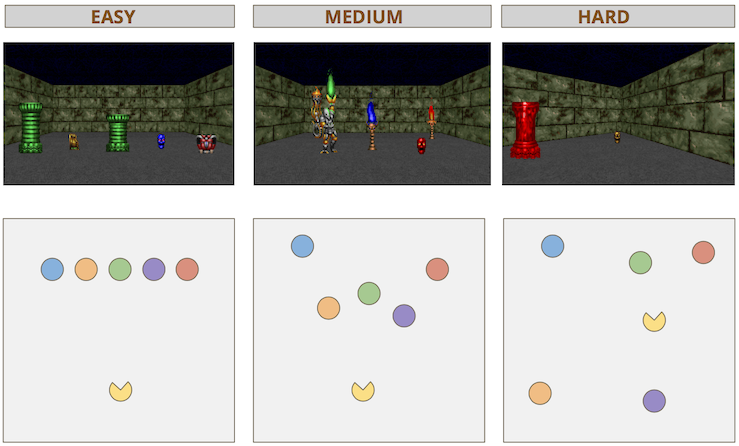}
\caption{Sample starting states and bird's eye view of the map (not visible to the agent) showing agent and object locations in Easy, Medium and Hard modes.}
\label{fig_grounding:env_sample}
\vspace{-6pt}
\end{wrapfigure}

An \emph{instruction} is a combination of (action, attribute(s), object) triple. Each instruction can have more than one attribute but we limit the number of actions and objects to one each.  
The environment  allows a variety of objects to be spawned at different locations in the map. The objects can have various visual attributes such as color, shape and size\footnote{See the list of objects and instructions at \url{https://goo.gl/rPWlMy}}. We provide a set of 70 manually generated instructions\footnotemark[1]. For each of these instructions, the environment allows for automatic creation of multiple episodes, each randomly created with its own set of correct object and incorrect objects.
Although the number of instructions are limited, the combinations of correct and incorrect objects for each instruction allows us to create multiple settings for the same instruction. Each time an instruction is selected, the environment generates a random combination of incorrect objects and the correct object in randomized locations. 
One of the significant challenges posed for a learning algorithm is to understand that the same instruction can refer to different objects in the different episodes. For example, ``Go to the red object'' can refer to a red keycard in one episode, and a red torch in another episode. Similarly, ``Go to the keycard'' can refer to keycards of various colors in different episodes. 
Objects could also occlude each other, or might not even be present in the agent's field of view, or the map could be more complicated, making it difficult for the agent to make a decision based solely on the current input, stressing the need for efficient exploration.

Our environment also provides different modes with respect to spawning of objects each with varying difficulty levels (Figure~\ref{fig_grounding:env_sample}):
\textbf{Easy}: The agent is spawned at a fixed location. The candidate objects are spawned at five fixed locations along a single horizontal line along the field of view of the agent.
\textbf{Medium}: The candidate objects are spawned in random locations, but the environment ensures that they are in the field of view of the agent. The agent is still spawned at a fixed location.
\textbf{Hard}: The candidate objects and the agent are spawned at random locations and the objects may or may not be in the agents field of view in the initial configuration. The agent needs to explore the map to view all the objects.

\begin{figure*}
\centering
\begin{minipage}{0.32\textwidth}
\includegraphics[width=\linewidth,height=\textheight,keepaspectratio]{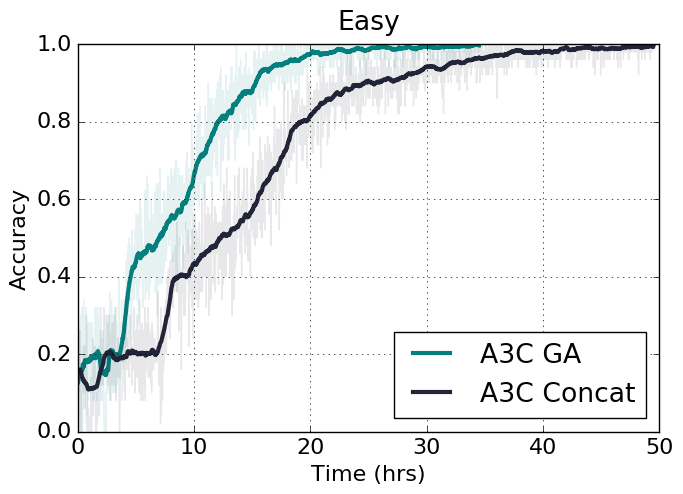}
\end{minipage}
\begin{minipage}{0.32\textwidth}
\centering
\includegraphics[width=\linewidth,height=\textheight,keepaspectratio]{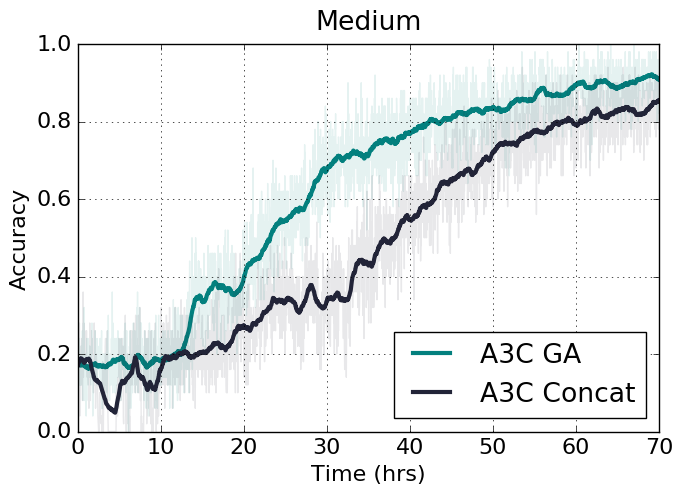}
\end{minipage}
\begin{minipage}{0.32\textwidth}
\centering
\includegraphics[width=\linewidth,height=\textheight,keepaspectratio]{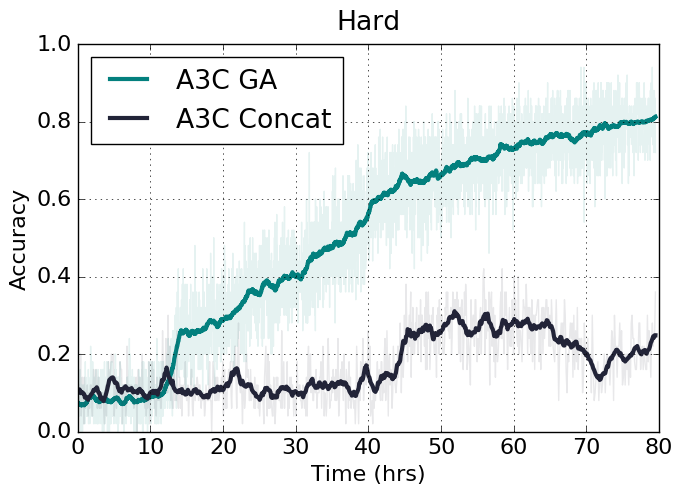}
\end{minipage}
\caption{\small Comparison of the performance of the proposed Gated-Attention (GA) unit to the baseline Concatenation unit using Reinforcement learning algorithm, A3C for (a) easy, (b) medium and (c) hard environments.}
\label{fig_grounding:policy_evaluation}
\end{figure*}

\section{Experimental Setup}
\label{sec_grounding:exp}
We perform our experiments
in all of the three environment difficulty modes, where we restrict the number of objects to 5 for each episode (one correct object, four incorrect objects and the agent are spawned for each episode). 
During training, the objects are spawned from a training set of 55 instructions, while 15 instructions pertaining to unseen \textit{attribute-object} combinations are held out in a test set for zero-shot evaluation.
During training, at the start of each episode, one of the train instructions is selected randomly. A correct target object is selected and 4 incorrect objects are selected at random. These objects are placed at random locations depending on the difficulty level of the environment. The episode terminates if the agent reaches any object or time step exceeds the maximum episode length ($T=30$). The evaluation metric is the \emph{accuracy} of the agent which is success rate of reaching the correct object before the episode terminates. 
We consider two scenarios for evaluation:

\noindent (1) \textbf{Multitask Generalization} (MT), where the agent is evaluated on unseen maps with instructions in the train set. Unseen maps comprise of unseen combination of objects placed at randomized locations. This scenario tests that the agent doesn't overfit to or memorize the training maps and can execute multiple instructions or tasks in unseen maps. 

\noindent (2) \textbf{Zero-shot Task Generalization} (ZSL), where the agent is evaluated on unseen test instructions. This scenario tests whether the agent can generalize to new combinations of attribute-object pairs which are not seen during the training. The maps in this scenario are also unseen.

\subsection{Baseline Approaches}
\textbf{Reinforcement Learning:} We adapt \citep{misra2017mapping} as a reinforcement learning baseline in the proposed environment. \citet{misra2017mapping} looks at jointly reasoning on linguistic and visual inputs for moving blocks in a 2D grid environment to execute an instruction. Their work uses raw features from the 2D grid, processed by a CNN, while the instruction is processed by an LSTM. Text and visual representations are combined using concatenation. The agent is trained using reinforcement learning and enhanced using distance based reward shaping. We do not use reward shaping as we would like the method to generalize to environments where the distance from the target is not available. 

\noindent \textbf{Imitation Learning:} We adapt \citep{mei2015listen} as an imitation learning baseline in the proposed environment. \citet{mei2015listen} map sequence of instructions to actions, treated as a sequence-to-sequence learning problem, with visual state input received by the decoder at each decode timestep. While they use a bag-of-visual words representation for visual state, we adapt the baseline to directly process raw pixels from the 3D environment using CNNs.

To ensure fairness in comparison, we use exact same architecture of CNNs (to process visual input), GRUs (to process textual instruction) and policy learning across baseline and proposed models. This reduces the reinforcement learning baseline to A3C algorithm with concatenation multimodal fusion (\textit{A3C-Concat}), and imitation learning baseline to Behavioral Cloning with Concatenation (\textit{BC-Concat}).

\setlength{\tabcolsep}{1.2em}
\begin{table*}[t]
\centering
\resizebox{\textwidth}{!}{
\begin{tabular}{@{}clcllllll@{}}
\toprule
\multicolumn{2}{c}{\multirow{2}{*}{Model}}                                                        & \multicolumn{1}{c}{\begin{tabular}[c]{@{}c@{}}Parameters\end{tabular}} & \multicolumn{2}{c}{Easy} & \multicolumn{2}{c}{Medium} & \multicolumn{2}{c}{Hard} \\
\multicolumn{2}{c}{}                                                                              &                                                                                     & MT          & ZSL        & MT          & ZSL          & MT          & ZSL        \\ \midrule
\multirow{4}{*}{\begin{tabular}[c]{@{}c@{}}Imitation\\ Learning\end{tabular}}                                            
& BC Concat     & 5.21M & 0.86 & 0.71 & 0.23 & 0.15 & 0.20 & 0.15 \\
& BC GA         & 5.09M & \bf{0.97} & 0.81 & 0.30 & 0.23 & \bf{0.36} & 0.29 \\
& DAgger Concat & 5.21M & 0.92 & 0.73 & 0.45 & 0.23 & 0.19 & 0.13 \\
& DAgger GA     & 5.09M & 0.94 & \bf{0.85} & \bf{0.55} & \bf{0.40} & 0.29 & \bf{0.30} \\ \midrule
\multirow{2}{*}{\begin{tabular}[c]{@{}c@{}}Reinforcement\\ Learning\end{tabular}} 
& A3C Concat    & 3.44M                                                                               & 1.00        & 0.80           & 0.80            & 0.54             & 0.24            &  0.12         \\
& A3C GA        & 3.39M                                                                               & 1.00        & \bf{0.81}       & \bf{0.89}            & \bf{0.75}             & \bf{0.83}            & \bf{0.73}           \\ \bottomrule
\end{tabular}}%
	\caption{The accuracy of all the models with Concatenation and Gated-Attention (GA) units. A3C Concat and BC Concat are the adapted versions of \citet{misra2017mapping}  and \citet{mei2015listen} respectively for the proposed environment. All the accuracy values are averaged over 100 episodes.}
\label{tab_grounding:grounding_results}
\end{table*}

\subsection{Hyper-parameters}
The input to the neural network is the instruction and an RGB image of size 3x300x168. The first layer convolves the image with 128 filters of 8x8 kernel size with stride 4, followed by 64 filters of 4x4 kernel size with stride 2 and another 64 filters of 4x4 kernel size with stride 2. The architecture of the convolutional layers is adapted from previous work on playing deathmatches in Doom \cite{chaplot2017arnold}. The input instruction is encoded through a Gated Recurrent Unit (GRU) \citep{chung2014empirical} of size 256. 

For the imitation learning approach, we run experiments with Behavioral Cloning (BC) and DAgger algorithms in an online fashion, which have data generation and policy update function per outer iteration. The policy learner for imitation learning comprises of a linear layer of size 512 which is fully-connected to 3 neurons to predict the policy function (i.e. probability of each action). In each data generation step, we sample state trajectories based on oracle's policy in BC and based on a mixture of oracle's policy and the currently learned policy in DAgger. The mixing of the policies is governed by an exploration coefficient, which has a linear decay from 1 to 0. For each state, we collect the optimal action given by the policy oracle. Then the policy is updated for 10 epochs over all the state-action pairs collected so far, using the RMSProp optimizer \citep{tieleman2012lecture}. Both methods use Huber loss \citep{huber1964robust} between the estimated policy and the optimal policy given by the policy oracle.

For reinforcement learning, we run experiments with A3C algorithm. 
The policy learning module has a linear layer of size 256 followed by an LSTM layer of size 256 which encodes the history of state observations. The LSTM layer's output is fully-connected to a single neuron to predict the value function as well as three other neurons to predict the policy function. 
All the convolutional layers and fully-connected linear layers have ReLu activations \citep{nair2010rectified}.
The A3C model was trained using Stochastic Gradient Descent (SGD) with a learning rate of 0.001. We used a discount factor of 0.99 for calculating expected rewards and run 16 parallel threads for each experiment. 

\section{Results}
\label{sec_grounding:res}
For all the models described in section \ref{sec_grounding:model-arch}, the performance on both Multitask and Zero-shot Generalization is shown in Table~\ref{tab_grounding:grounding_results}. The performance of A3C models on Multitask Generalization during training is plotted in Figure \ref{fig_grounding:policy_evaluation}. 

\textbf{Performance of GA models:} We observe that models with the Gated-Attention (GA) unit outperform models with the Concatenation unit for Multitask and Zero-Shot Generalization. From Figure~\ref{fig_grounding:policy_evaluation} we observe that A3C models with GA units learn faster than Concat models and converge to higher levels of accuracy. In hard mode, GA achieves $83\%$ accuracy on Multitask Generalization and $73\%$ on Zero-Shot Generalization, whereas Concat achieves $24\%$ and $12\%$ respectively and fails to show any considerable performance. For Imitation Learning, we observe that GA models perform better than Concat, and that as the environment modes get harder, imitation learning does not perform very well as there is a need for exploration in medium and hard settings.
In contrast, the inherent extensive exploration of the reinforcement learning algorithm makes the A3C model more robust to the agent's location and covers more state trajectories. 

\begin{figure*}
\includegraphics[width=\linewidth,height=\textheight,keepaspectratio]{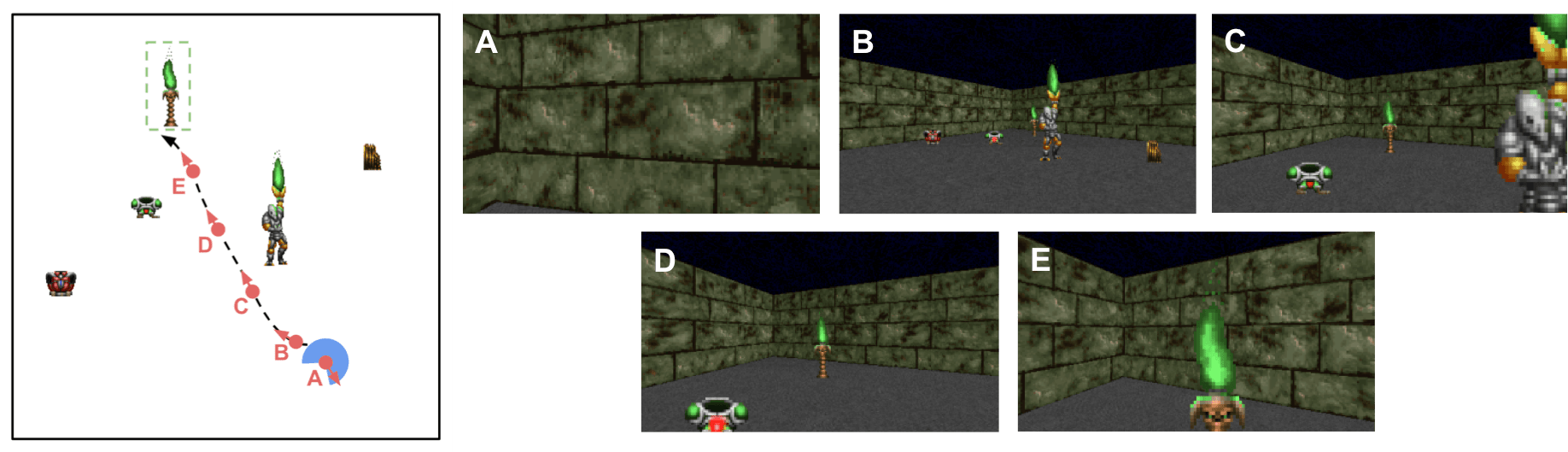}
\caption{\small This figure shows an example of the A3C policy execution at different points for the instruction `Go to the short green torch'. \textit{Left}: Navigation map of the agent, \textit{Right}: frames at each point. \textbf{A:} Initial frame: None of the objects are visible. \textbf{B:} agent has turned so that objects are in the field of view. \textbf{C : } agent successfully avoids the tall green torch. \textbf{D :} agent moves towards the short green torch. \textbf{E :}agent reaches target.}
\label{fig_grounding:policy_example}
\end{figure*}

\textbf{Policy Execution} : Figure \ref{fig_grounding:policy_example} shows a policy execution of the A3C model in the hard mode for the instruction \textit{short green torch}. In this figure, we demonstrate the agent's ability to explore the environment and handle occlusion. In this example, none of the objects are in the field-of-view of the agent in the initial frame.
The agent explores the environment (makes a ~300 degree turn) and eventually navigates towards the target object. It has also learned to distinguish between a \emph{short green torch} and \emph{tall green torch} and to avoid the tall torch before reaching the short torch\footnote{Demo videos: \url{https://goo.gl/rPWlMy}}.

\begin{figure*}
\minipage{0.72\textwidth}
\includegraphics[width=0.99\linewidth,keepaspectratio]{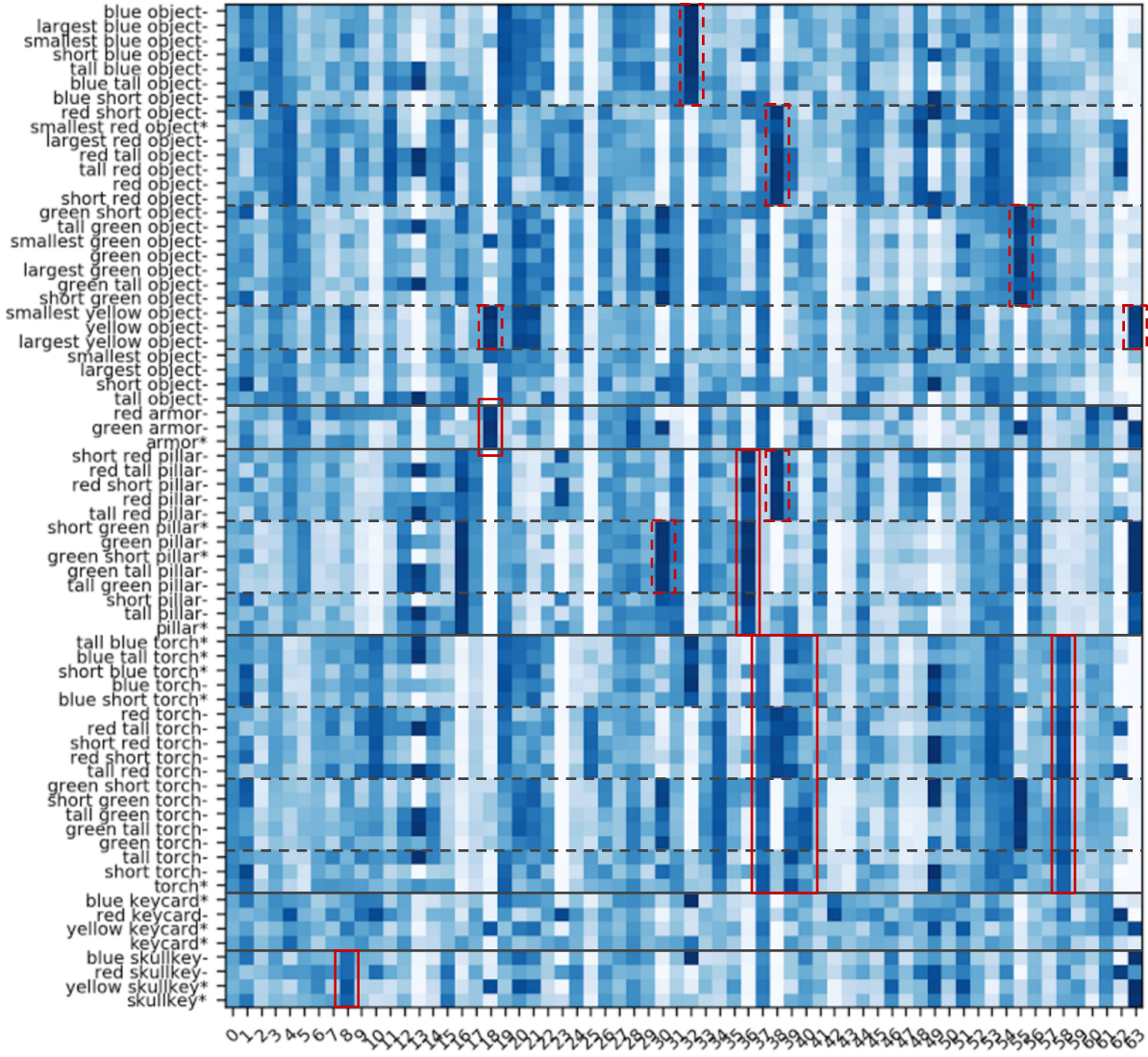}
\caption{\small Heatmap of the values of the 64-dimensional attention vector for different instructions grouped by object type and sub-grouped by object color. The test instructions are marked by *. The red boxes indicate that certain dimensions of the attention vector get activated for particular attributes of the target object referred in the instruction.}
\label{fig_grounding:analysis1}
\endminipage\hfill
\minipage{0.26\textwidth}
\includegraphics[width=0.99\linewidth,height=\textheight,keepaspectratio]{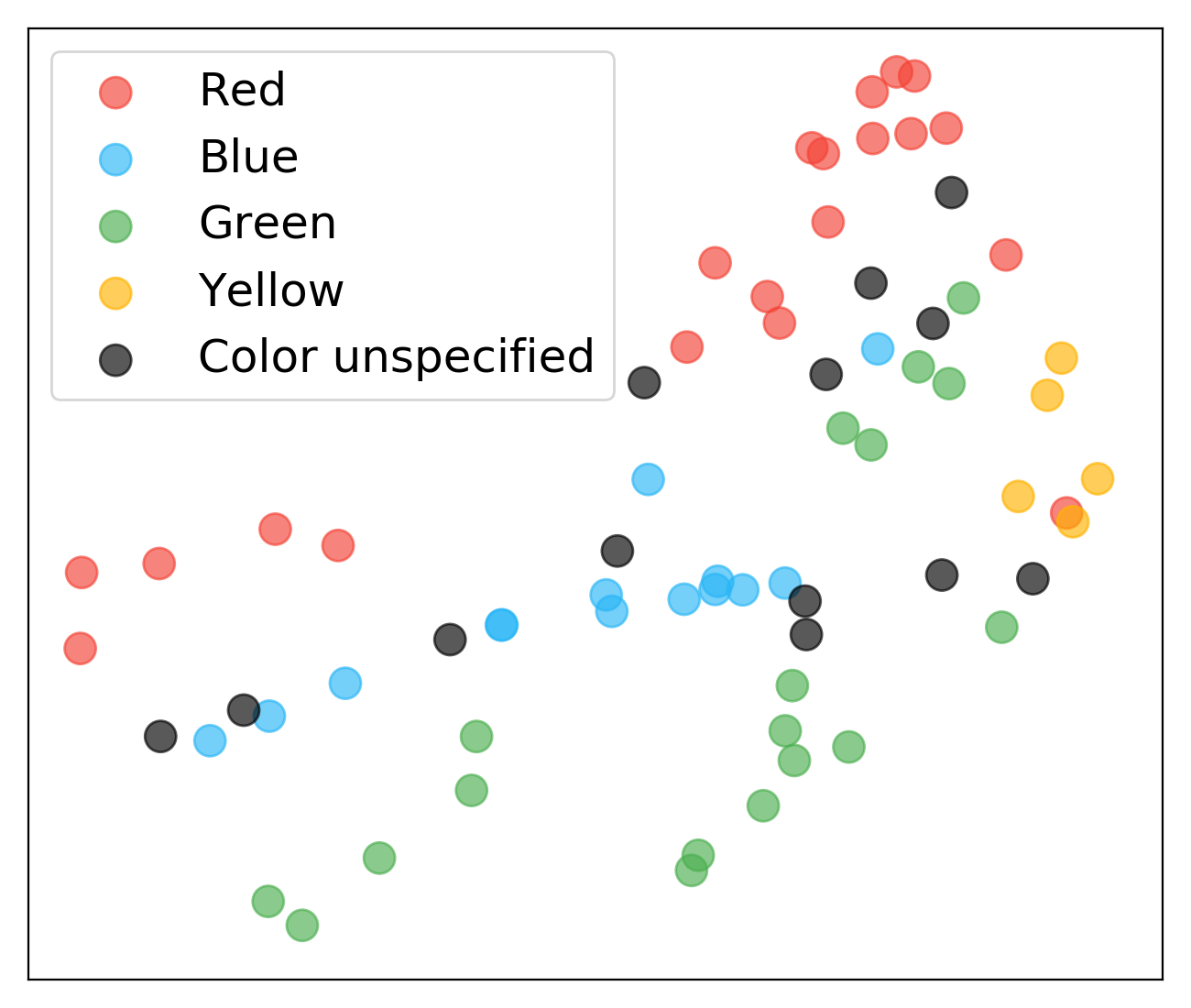}

\includegraphics[width=0.99\linewidth,height=\textheight,keepaspectratio]{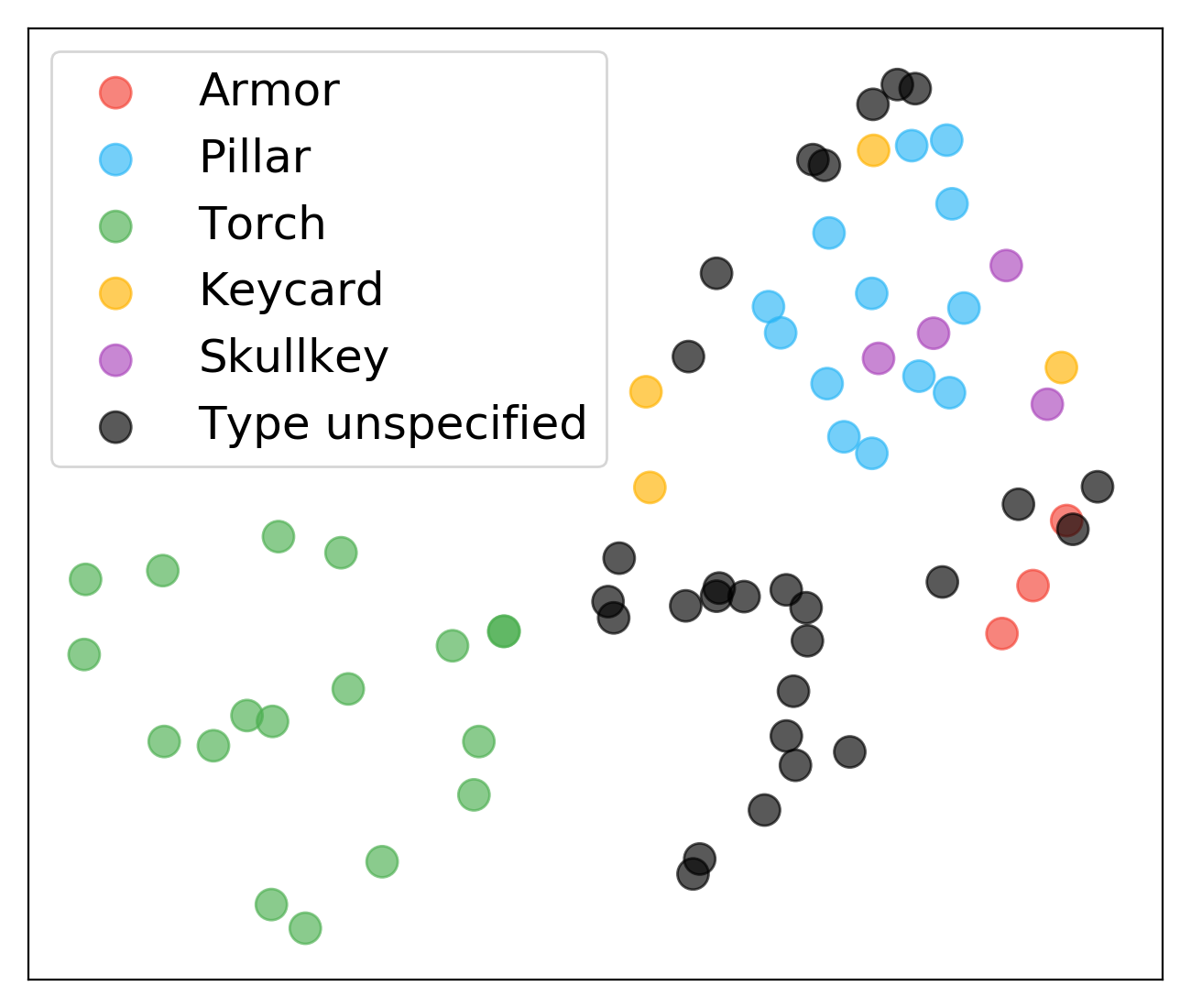}

\includegraphics[width=0.99\linewidth,height=\textheight,keepaspectratio]{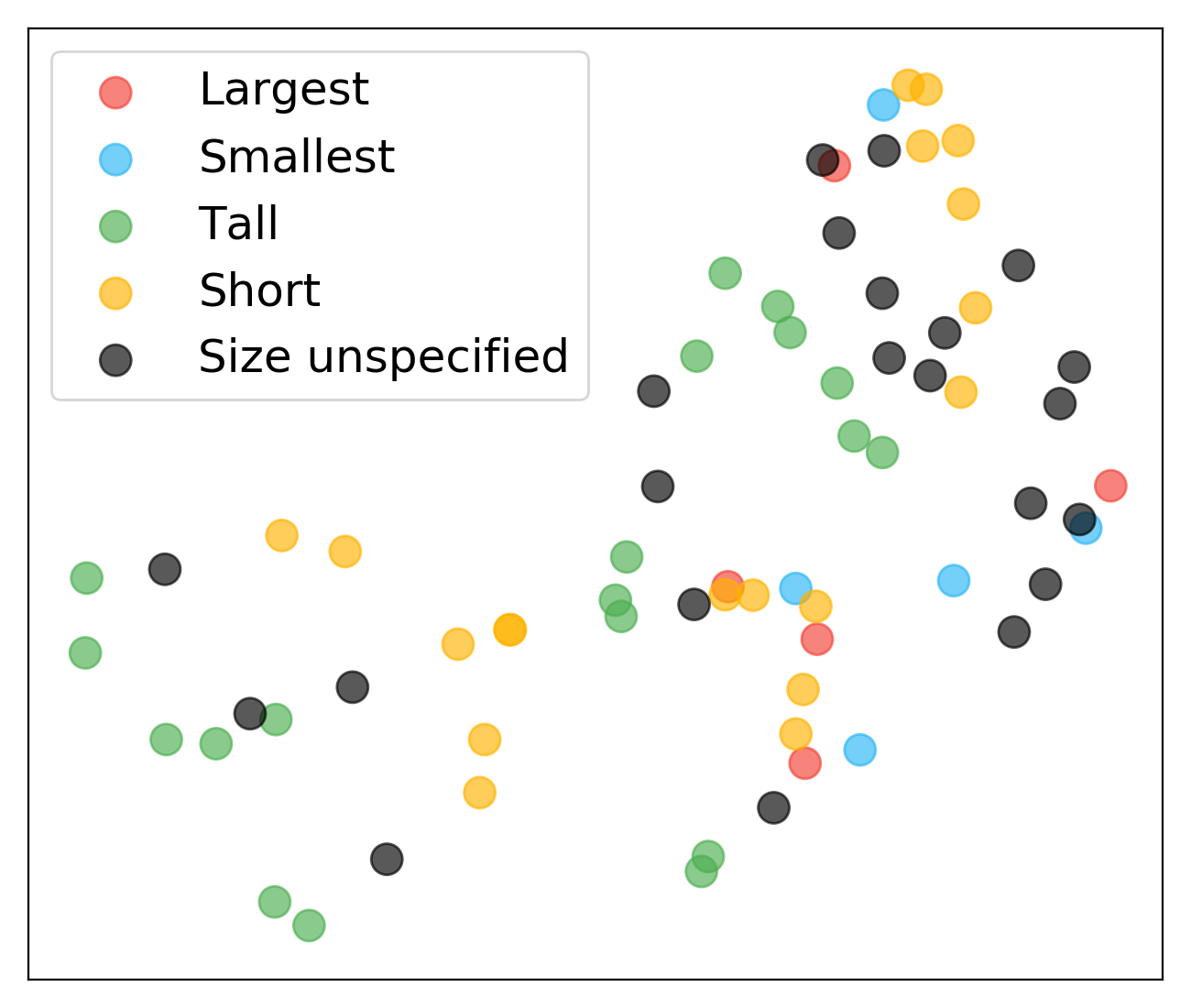}
	\caption{\small The t-SNE visualization of the attention vectors showing clusters based on object color, type and size.}
\label{fig_grounding:tsne}
\endminipage\hfill
\end{figure*}

\textbf{Analysis of Attention Maps:} Figure~\ref{fig_grounding:analysis1} shows the heatmap for values of the attention vector for different instructions grouped by object type of the target object.
As seen in the figure, dimension 18 corresponds to `armor', dimensions 8 corresponds to the `skullkey' and dimension 36 corresponds to the `pillar'. Also, note that there is no dimension which is high for all the instructions in the first group. This indicates that the model also recognizes that the word `object' does not correspond to a particular object type, but rather refers to any object of that color (indicated by dotted red boxes in \ref{fig_grounding:analysis1}). These observations indicate that the model is learning to recognize the attributes of objects such as color and type, and specific feature maps are gated based on these attributes. 
Furthermore, the attention vector weights on the test instructions (marked by * in figure ~\ref{fig_grounding:analysis1}) also indicate that the Gated-Attention unit is also able to recognize attributes of the object in unseen instructions.
We also visualize the t-SNE plots for the attention vectors based on attributes, color and object type as shown in Figure~\ref{fig_grounding:tsne}. 
The attention vectors for objects of red, blue, green, and yellow are present in clusters whereas those for instructions which do not mention the object's color are spread across and belong to the clusters corresponding to the object type. Similarly, objects of a particular type present themselves in clusters. 
The clusters indicate that the model is able to recognize object attributes as it learns similar attention vectors for objects with similar attributes.

\section{Discussion}
\label{sec_grounding:conc}
In this chapter, we proposed an end-to-end architecture for task-oriented language grounding from raw pixels in a 3D environment, for both reinforcement learning and imitation learning.  
The architecture uses a novel multimodal fusion mechanism, \textit{Gated-Attention}, which learns a joint state representation based on multiplicative interactions between instruction and image representation. 
We observe that the models 
(A3C for reinforcement learning and Behavioral Cloning/DAgger for imitation learning) 
which use the Gated-Attention unit outperform the models with concatenation units for both Multitask and Zero-Shot task generalization, across three modes of difficulty. 
The visualization of the attention weights for the Gated-Attention unit indicates that the agent learns to recognize objects, color attributes and size attributes.

\chapter{Active Neural Localization}
\label{chap:anl}

\blfootnote{Webpage: \url{https://devendrachaplot.github.io/projects/Neural-Localization}}

Localization is the problem of estimating the position of an autonomous agent given a map of the environment and agent observations. The ability to localize under uncertainity is required by autonomous agents to perform various downstream tasks such as planning, exploration and target-navigation. Localization is considered as one of the most fundamental problems in mobile robotics \citep{Cox:1990:ARV:93002,Borenstein:1996:NMR:546829}. Localization is useful in many real-world applications such as autonomous vehicles, factory robots and delivery drones.

In this chapter, we tackle the global localization problem where the initial position of the agent is unknown. Despite the long history of research, global localization is still an open problem, and there are not many methods developed which can be learnt from data in an end-to-end manner, instead typically requiring significant hand-tuning and feature selection by domain experts. Another limitation of majority of localization approaches till date is that they are \textit{passive}, meaning that they passively estimate the position of the agent from the stream of incoming observations, and do not have the ability to decide the actions taken by the agent. The ability to decide the actions can result in faster as well as more accurate localization as the agent can learn to navigate quickly to unambiguous locations in the environment. 

We propose ``Active Neural Localizer'', a neural network model capable of \textit{active} localization using raw pixel-based observations and a map of the environment. Based on the Bayesian filtering algorithm for localization \citep{fox2003bayesian}, the proposed model contains a perceptual model to estimate the likelihood of the agent's observations, a structured component for representing the belief, multiplicative interactions to propagate the belief based on observations and a policy model over the current belief to localize accurately while minimizing the number of steps required for localization. The entire model is fully differentiable and trained using reinforcement learning, allowing the perceptual model and the policy model to be learnt simultaneously in an end-to-end fashion. A variety of 2D and 3D simulation environments are used for testing the proposed model. We show that the Active Neural Localizer is capable of generalizing to not only unseen maps in the same domain but also across domains. We also contribute novel simulation scenarios for future active localization research.

Localization has been an active field of research since more than two decades. In the context of mobile autonomous agents, Localization can be refer to two broad classes of problems: Local localization and Global localization. Local localization methods assume that the initial position of the agent is known and they aim to track the position as it moves. A large number of localization methods tackle only the problem of local localization. These include classical methods based on Kalman Filters \citep{kalman1960new, smith1990estimating} geometry-based visual odometry methods \citep{nister2006visual} and most recently, learning-based visual odometry methods which learn to predict motion between consecutive frames using recurrent convolutional neural networks \citep{clark2017vinet, wang2017deepvo}. Local localization techniques often make restrictive assumptions about the agent's location. Kalman filters assume Gaussian distributed initial uncertainty, while the visual odometry-based methods only predict the relative motion between consecutive frames or with respect to the initial frame using camera images. Consequently, they are unable to tackle the global localization problem where the initial position of the agent is unknown. This also results in their inability to handle localization failures, which consequently leads to the requirement of constant human monitoring and interventation \citep{burgard1998integrating}.

Global localization is more challenging than the local localization problem and is also considered as the basic precondition for truly autonomous agents by \cite{burgard1998integrating}. Among the methods for global localization, the proposed method is closest to Markov Localization \citep{fox1998markov}. In contrast to local localization approaches, Markov Localization computes a probability distribution over all the possible locations in the environment. The probability distribution also known as the \textit{belief} is represented using piecewise constant functions (or histograms) over the state space and propagated using the Markov assumption. Other methods for global localization include Multi-hypothesis Kalman filters \citep{cox1994modeling, Roumeliotis00bayesianestimation} which use a mixture of Gaussians to represent the belief and Monte Carlo Localization \citep{thrun2001robust} which use a set of samples (or \textit{particles}) to represent the belief. 

All the above localization methods are \textit{passive}, meaning that they aren't capable of deciding the actions to localize more accurately and efficiently. There has been very little research on \textit{active} localization approaches. Active Markov Localization \citep{fox1998active} is the active variant of Markov Localization where the agent chooses actions greedily to maximize the reduction in the entropy of the belief. \cite{jensfelt2001active} presented the active variant of Multi-hypothesis Kalman filters where actions are chosen to optimise the information gathering for localization. Both of these methods do not learn from data and have very high computational complexity. In contrast, we demonstrate that the proposed method is several order of magnitudes faster while being more accurate and is capable of learning from data and generalizing well to unseen environments.

Recent work has also made progress towards end-to-end localization using deep learning models. \cite{mirowski2016learning} showed that a stacked LSTM can do reasonably well at self-localization. The model consisted of a deep convolutional network which took in at each time step state observations, reward, agent velocity and previous actions. To improve performance, the model also used several auxiliary objectives such as depth prediction and loop closure detection. The agent was successful at navigation tasks within complex 3D mazes. Additionally, the hidden states learned by the models were shown to be quite accurate at predicting agent position, even though the LSTM was not explicitly trained to do so. Other works have looked at doing end-to-end relocalization more explicitly. One such method, called PoseNet~\citep{kendall2015posenet}, used a deep convolutional network to implicitly represent the scene, mapping a single monocular image to a 3D pose (position and orientation). This method is limited by the fact that it requires a new PoseNet trained on each scene since the map is represented implicitly by the convnet weights, and is unable to transfer to scenes not observed during training. An extension to PoseNet, called VidLoc~\citep{clark2017vidloc}, utilized temporal information to make more accurate estimates of the poses by passing a Bidirectional LSTM over each monocular image in a sequence, enabling a trainable smoothing filter over the pose estimates. Both these methods lack a straightforward method to utilize past map data to do localization in a new environment. In contrast, we demonstrate our method is capable of generalizing to new maps that were not previously seen during training time.

\section{Background: Bayesian Filtering}
\vspace{-0.8em}
Bayesian filters \citep{fox2003bayesian} are used to probabilistically estimate a dynamic system's state using observations from the environment and actions taken by the agent. Let $y_t$ be the random variable representing the state at time $t$. Let $s_t$ be the observation received by the agent and $a_t$ be the action taken by the agent at time step $t$. At any point in time, the probability distribution over $y_t$ conditioned over past observations $s_{1:t-1}$ and actions $a_{1:t-1}$ is called the \emph{belief}, $Bel(y_t) = p(y_t|s_{1:t-1},a_{1:t-1}) $
The goal of Bayesian filtering is to estimate the belief sequentially. For the task of localization, $y_t$ represents the location of the agent, although in general it can represent the state of the any object(s) in the environment. Under the Markov assumption, the belief can be recursively computed using the following equations:
\vspace{-0.5em}
$$	\overline{Bel}(y_t)  = \sum_{y_{t-1}}p(y_t|y_{t-1},a_{t-1})Bel(y_{t-1}), \hspace{0.1in}
	Bel(y_t)  =  \frac{1}{Z} Lik(s_t)\overline{Bel}(y_t), $$
\vspace{-1.0em}

where $Lik(s_t) = p(s_t|y_t)$ is the likelihood of observing $s_t$ given the location of the agent is $y_t$, and $Z = \Sigma_{y_t} Lik(s_t)\overline{Bel}(y_t) $ is the normalization constant.
The likelihood of the observation, $Lik(s_t)$ is given by the \emph{perceptual model} and $p(y_t|y_{t-1},a_{t-1})$, i.e. the probability of landing in a state $y_t$ from $y_{t-1}$ based on the action, $a_{t-1}$, taken by the agent is specified by a state transition function, $f_t$. The belief at time $t=0$, $Bel(y_0)$, also known as the \emph{prior}, can be specified based on prior knowledge about the location of the agent. For global localization, prior belief is typically uniform over all possible locations of the agent as the agent position is completely unknown.

\vspace{-0.5em}
\section{Methods}
\label{sec:methods}
\vspace{-0.8em}
\subsection{Problem Formulation}
\vspace{-0.5em}
Let $s_t$ be the observation received by the agent and $a_t$ be the action taken by the agent at time step $t$. Let $y_t$ be a random variable denoting the state of the agent, that includes its x-coordinate, y-coordinate and orientation. In addition to agent's past observations and actions, a localization algorithm requires some information about the map, such as the map design. Let the information about the map be denoted by $M$. In the problem of active localization, we have two goals:
(1) Similar to the standard state estimation problem in the Bayesian filter framework, the goal is to estimate the \emph{belief}, $Bel(y_t)$, or the probability distribution of $y_t$ conditioned over past observations and actions and the information about the map,
$Bel(y_t) = p(y_t|s_{1:t},a_{1:t-1},M)$, 
(2) We also aim to learn a policy $\pi(a_t|Bel(y_t))$ for localizing accurately and efficiently.

\subsection{Proposed Model}
\vspace{-0.5em}
 
\paragraph{Representation of Belief and Likelihood}
Let $y_t$ be a tuple ${A_o,A_x,A_y}$ where $A_x$,$A_y$ and $A_o$ denote agent's x-coordinate, y-coordinate and orientation respectively. Let $M \times N$ be the map size, and $O$ be the number of possible orientations of the agent. Then, $A_x \in [1,M]$, $A_y \in [1,N]$ and $A_o \in [1,O]$. Belief is represented as an $O \times M \times N$ tensor, where $(i,j,k)^{th}$ element denotes the belief of agent being in the corresponding state, $Bel(y_t={i,j,k})$. This kind of grid-based representation of belief is popular among localization methods as if offers several advantages over topological representations \citep{burgard1996estimating, fox2003bayesian}. Let $Lik(s_t) = p(s_t|y_t)$ be the likelihood of observing $s_t$ given the location of the agent is $y_t$, The likelihood of an observation in a certain state is also represented by an $O \times M \times N$ tensor, where $(i,j,k)^{th}$ element denotes the likelihood of the current observation, $s_t$ given that the agent's state is $y_t = {i,j,k}$. We refer to these tensors as Belief Map and Likelihood Map in the rest of the chapter.

\begin{figure*}
\centering
\includegraphics[width=0.99\linewidth,height=\textheight,keepaspectratio]{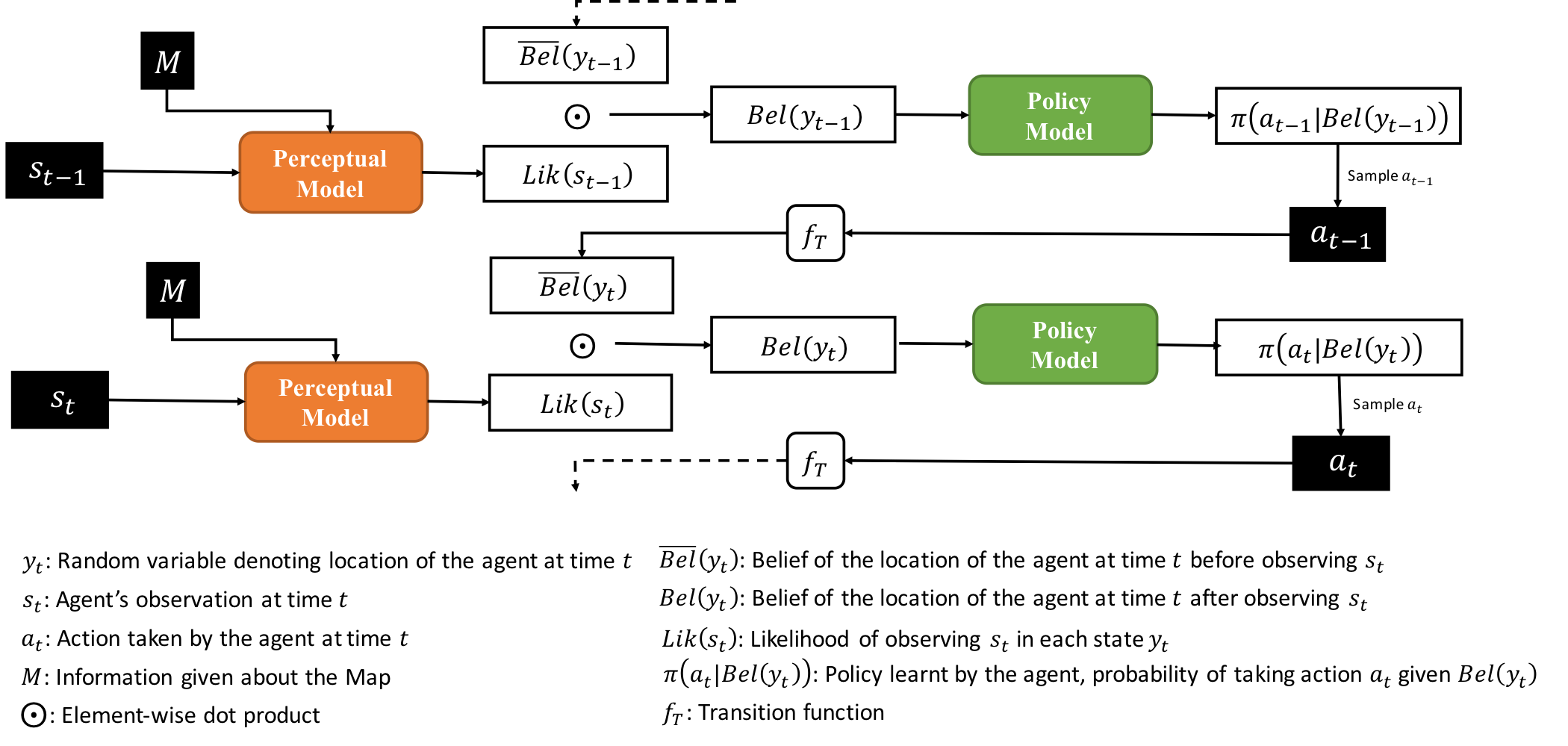}
\caption{\small The architecture of the proposed model. The perceptual model computes the likelihood of the current observation in all possible locations. The belief of agent's location is propagated through time by taking element-wise dot-product with the likelihood. The policy model learns a policy to localize accurately while minimizing the number of steps required for localization. See text for more details.}
\label{fig:block_arch}
\end{figure*}

\paragraph{Model Architecture}
The overall architecture of the proposed model, Active Neural Localizer (ANL), is shown in Figure~\ref{fig:block_arch}. It has two main components: the perceptual model and the policy model. At each timestep $t$, the \emph{perceptual model} takes in the agent's observation, $s_t$ and outputs the Likelihood Map $Lik(s_t)$. The belief is propagated through time by taking an element-wise dot product with the Likelihood Map at each timestep. Let $\overline{Bel}(y_t)$ be the Belief Map at time $t$ before observing $s_t$. Then the belief, after observing $s_t$, denoted by $Bel(y_t)$, is calculated as follows:
\vspace{-0.3em}
$$ Bel(y_t) = \frac{1}{Z}\overline{Bel}(y_t) \odot Lik(s_t) $$
\vspace{-0.4em}
where $\odot$ denotes the Hadamard product, $Z = \sum_{y_t} Lik(s_t)\overline{Bel}(y_t)$ is the normalization constant. 

The Belief Map, after observing $s_t$, is passed through the \emph{policy model} to obtain the probability of taking any action, $\pi(a_t|Bel(y_t))$. The agent takes an action $a_t$ sampled from this policy. The Belief Map at time $t+1$ is calculated using the transition function ($f_T$), which updates the belief at each location according to the action taken by the agent, i.e. $p(y_{t+1}|y_{t},a_{t})$. The transition function is similar to the egomotion model used by \cite{gupta2017cognitive} for mapping and planning. For `turn left' and `turn right' actions, the transition function just swaps the belief in each orientation. For the the `move forward' action, the belief values are shifted one unit according to the orientation. If the next unit is an obstacle, then the value doesn't shift, indicating a collison.

\subsection{Model Components}
\paragraph{Perceptual Model}
The perceptual model computes the feature representation from the agent's observation and the states given in the map information. The likelihood of each state in the map information is calculated by taking the cosine similarity of the feature representation of the agent's observation with the feature representation of the state. Cosine similarity is commonly used for computing the similarity of representations \citep{nair2010rectified,huang2013learning} and has also been used in the context on localization \citep{chaplottransfer}. The benefits of using cosine similarity over dot-product have been highlighted by \cite{chunjie2017cosine}.

\begin{figure*}
\centering
\includegraphics[width=0.99\linewidth,keepaspectratio]{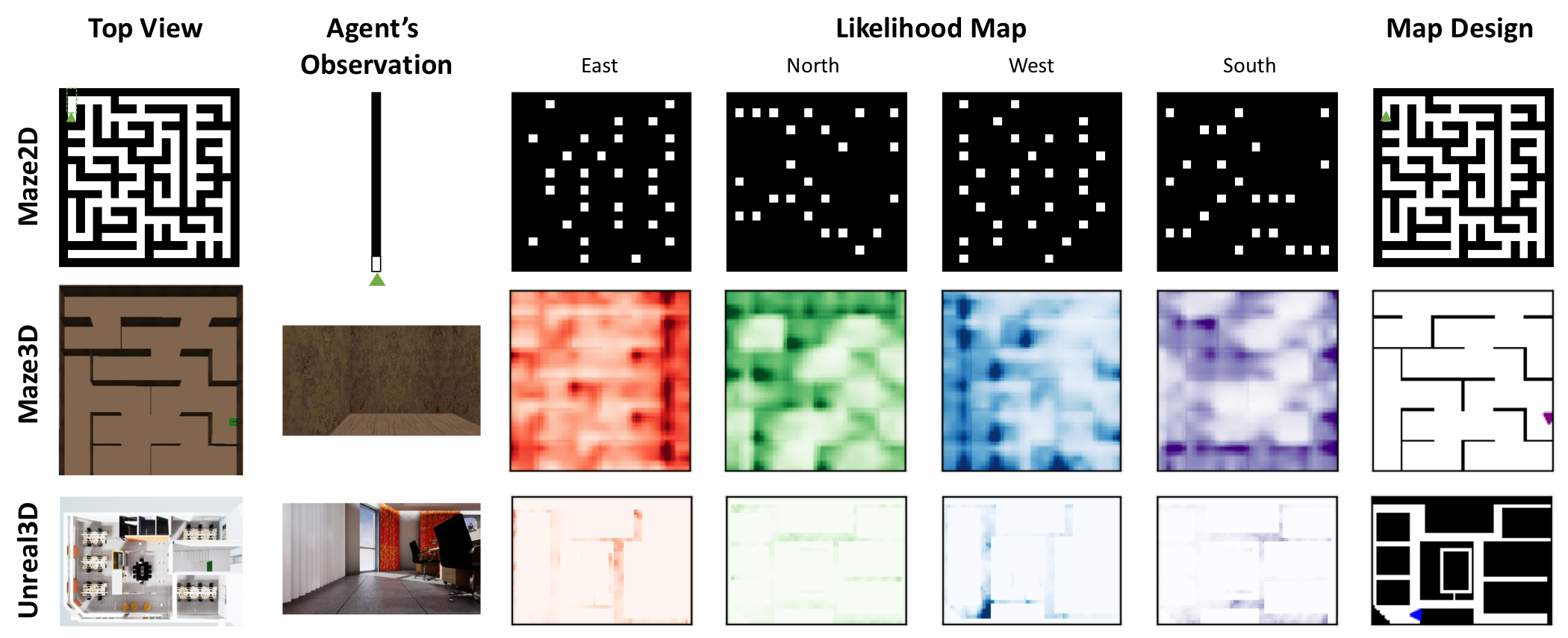}
\vspace{-0.5em}
\caption{\small The map design, agent's observation and the corresponding likelihood maps in different domains. In 2D domains, agent's observation is the pixels in front of the agent until the first obstacle. In the 3D domain, the agent's observation is the image showing the first-person view of the world as seen by the agent.}
\label{fig:domains}
\vspace{-0.5em}
\end{figure*}

In the 2D environments, the observation is used to compute a one-hot vector of the same dimension representing the depth which is used as the feature representation directly. This resultant Likelihood map has uniform non-zero probabilities for all locations having the observed depth and zero probabilities everywhere else. For the 3D environments, the feature representation of each observation is obtained using a trainable deep convolutional network \citep{lecun1995convolutional}. Figure~\ref{fig:domains} shows examples of the agent observation and the corresponding Likelihood Map computed in both 2D and 3D environments. The simulation environments are described in detail in Section \ref{sec:env}.

\paragraph{Policy Model}
The policy model gives the probablity of the next action given the current belief of the agent. It is trained using reinforcement learning, specifically Asynchronous Advantage Actor-Critic (A3C) \citep{mnih2016asynchronous} algorithm. The belief map is stacked with the map design matrix and passed through 2 convolutional layers followed by a fully-connected layer to predict the policy as well as the value function. The policy and value losses are computed using the rewards observed by the agent and backpropagated through the entire model.

\section{Experiments}
\label{sec:experiments}
\vspace{-0.5em}

As described in Section~\ref{sec:methods}, agent's state, $y_t$ is a tuple ${A_o,A_x,A_y}$ where $A_x$,$A_y$ and $A_o$ denote agent's x-coordinate, y-coordinate and orientation respectively. $A_x \in [1,M]$, $A_y \in [1,N]$ and $A_o \in [1,O]$, where $M\times N$ is the map size, and $O$ be the number of possible orientations of the agent. We use a variety of domains for our experiments. The values of $M$ and $N$ vary accross domains but $O=4$ is fixed. The possible actions in all domains are `move forward', `turn left' and `turn right'. The turn angle is fixed at $(360/O = 90)$. This ensures that the agent is always in one of the 4 orientations, North, South, East and West. Note that although we use, $O = 4$ in all our experiments, our method is defined for any value of $O$. 
At each time step, the agent receives an intermediate reward equal to the maximum probability of being in any state, $ r_t = max_{y_t}(Bel(y_t))$. This encourages the agent the reduce the entropy of the Belief Map in order to localize as fast as possible. We observed that the agent converges to similar performance without introducing the intermediate reward, but it helps in speeding up training. 
At the end of the episode, the prediction is the state with the highest probability in the Belief Map. If the prediction is correct, i.e. $y^* = \argmax_{y_t} Bel(y_t)$ where $y*$ is the true state of the agent, then the agent receives a positive reward of $1$. 
The metric for evaluation is accuracy (Acc) which refers to the ratio of the episodes where the agent's prediction was correct over 1000 episodes. We also report the total runtime of the method in seconds taken to evaluate 1000 episodes.  

\subsection{Simulation Environments}
\label{sec:env}

\vspace{-0.5em}
\paragraph{Maze 2D} In the Maze2D environment, maps are represented by a binary matrix, where 0 denotes an obstacle and 1 denotes free space. The map designs are generated randomly using Kruskal's algorithm \cite{kruskal1956shortest}. The agent's observation in 2D environments is the series of pixels in front of the agent. For a map size of $M\times N$, the agent's observation is an array of size $max(M,N)$ containing pixels values in front of the agent. The view of the agent is obscured by obstacles, so all pixel values behind the first obstacle are treated as 0. The information about the map, $M$, received by the agent is the matrix representing the map design. Note that the observation at any state in the map can be computed using the map design. The top row in Figure~\ref{fig:domains} shows examples of map design and agent's observation in this environment.

The 2D environments provide ideal conditions for Bayesian filtering due to lack of observation or motion noise. The experiments in the 2D environments are designed to evaluate and quantify the effectiveness of the policy learning model in ideal conditions. The size of the 2D environments can also be varied to test the scalability of the policy model. This design is similar to previous experimental settings such as by \cite{tamar2016value} and \cite{qmdpnet17} for learning a target-driven navigation policy in grid-based 2D environments.

\vspace{-0.5em}
\paragraph{3D Environments}
In the 3D environments, the observation is an RGB image of the first-person view of the world as seen by the agent. The x-y coordinates of the agent are continuous variables, unlike the discrete grid-based coordinates in the 2D environments. The matrix denoting the belief of the agent is discretized meaning each pixel in the Belief map corresponds to a range of states in the environment. At the start of every epsiode, the agent is spawned at a random location in this continuous range as opposed to a discrete location corresponding to a pixel in the belief map for 2D environments. This makes localization much more challenging than the 2D envrionments. Apart from the map design, the agent also receives a set of images of the visuals seen by the agent at a few locations uniformly placed around the map in all 4 orientations. These images, called memory images, are required by the agent for global localization. They are not very expensive to obtain in real-world environments. We use two types of 3D environments: 

\textbf{Maze3D:} Maze3D consists of virtual 3D maps built using the Doom Game Engine. We use the ViZDoom API \citep{kempka2016vizdoom} to interact with the gane engine. The map designs are generated using Kruskal's algorithm and Doom maps are created based on these map designs using Action Code Scripts\footnote{\url{https://en.wikipedia.org/wiki/Action_Code_Script}}. The design of the map is identical to the Maze2D map designs with the difference that the paths are much thicker than the walls as shown in Figure~\ref{fig:domains}. The texture of each wall, floor and ceiling can be controlled which allows us to create maps with different number of `landmarks'. Landmarks are defined to be walls with a unique texture.

\textbf{Unreal3D:} Unreal3D is a photo-realistic simulation environment built using the Unreal Game Engine. We use the AirSim API \citep{airsim2017fsr} to interact with the game engine. The environment consists of a modern office space as shown in Figure~\ref{fig:domains} obtained from the Unreal Engine Marketplace\footnote{\url{https://www.unrealengine.com/marketplace/small-office-prop-pack}}.

The 3D environments are designed to test the ability of the proposed model to jointly learn the perceptual model along with the policy model as the agent needs to handle raw pixel based input while learning a policy. The Doom environment provides a way to test the model in challenging ambiguous environments by controlling the number of landmarks in the environment, while the Unreal Environment allows us to evaluate the effectiveness of the model in comparatively more realistic settings.

\setlength{\tabcolsep}{6.5pt}
\begin{table}[t]
\footnotesize
\centering
\caption{\small Results on the 2D Environments. `Time' refers to the number of seconds required to evaluate 1000 episodes with the corresponding method and `Acc' stands for accuracy over 1000 episodes.}
\label{tab:maze2d}
\resizebox{\textwidth}{!}{
\begin{tabular}{@{}lllrrrrrrrrrr@{}}
\toprule
Env                                                                                           &      &  & \multicolumn{8}{c}{Maze2D}                                                                                                                                                                        & \multicolumn{1}{c}{} & \multicolumn{1}{c}{All} \\ \midrule
Map Size                                                                                      &      &  & \multicolumn{2}{c}{7x7}                         & \multicolumn{1}{c}{} & \multicolumn{2}{c}{15x15}                       & \multicolumn{1}{c}{} & \multicolumn{2}{c}{21x21}                       & \multicolumn{1}{c}{} & \multicolumn{1}{c}{}    \\ \midrule
Episode Length                                                                                &      &  & \multicolumn{1}{c}{15} & \multicolumn{1}{c}{30} & \multicolumn{1}{c}{} & \multicolumn{1}{c}{20} & \multicolumn{1}{c}{40} & \multicolumn{1}{c}{} & \multicolumn{1}{c}{30} & \multicolumn{1}{c}{60} & \multicolumn{1}{c}{} & \multicolumn{1}{c}{}    \\ \midrule
\multirow{2}{*}{\begin{tabular}[c]{@{}l@{}}Markov \\ Localization\end{tabular}}               & Time &  & 12                     & 15                     &                      & 29                     & 31                     &                      & 49                     & 51                     &                      & 31.2                    \\
                                                                                              & Acc  &  & 0.334                  & 0.529                  &                      & 0.351                  & 0.606                  &                      & 0.414                  & 0.661                  &                      & 0.483                   \\ \midrule
\multirow{2}{*}{\begin{tabular}[c]{@{}l@{}}Active Markov \\ Localization (Fast)\end{tabular}} & Time &  & 29                     & 53                     &                      & 72                     & 165                    &                      & 159                    & 303                    &                      & 130.2                   \\
                                                                                              & Acc  &  & 0.436                  & 0.619                  &                      & 0.468                  & 0.657                  &                      & 0.512                  & 0.735                  &                      & 0.571                   \\ \midrule
\multirow{2}{*}{\begin{tabular}[c]{@{}l@{}}Active Markov \\ Localization (Slow)\end{tabular}} & Time &  & 1698                   & 3066                   &                      & 3791                   & 8649                   &                      & 8409                   & 13554                  &                      & 6527.8                  \\
                                                                                              & Acc  &  & 0.854                  & 0.938                  &                      & 0.846                  & \textbf{0.984}         &                      & 0.845                  & 0.958                  &                      & 0.904                   \\ \midrule
\multirow{2}{*}{\begin{tabular}[c]{@{}l@{}}Active Neural \\ Localization\end{tabular}}        & Time &  & 22                     & 34                     &                      & 44                     & 66                     &                      & 82                     & 124                    &                      & 62.0                    \\
                                                                                              & Acc  &  & \textbf{0.936}         & \textbf{0.939}         &                      & \textbf{0.905}         & 0.939                  &                      & \textbf{0.899}         & \textbf{0.984}         &                      & \textbf{0.934}          \\ \bottomrule
\end{tabular}}
\end{table}

\subsection{Baselines}
{\bf Markov Localization} \citep{fox1998markov} is a passive probabilistic approach based on Bayesian filtering. We use a geometric variant of Markov localization where the state space is represented by fine-grained, regularly spaced grid, called position probability grids \citep{burgard1996estimating}, similar to the state space in the proposed model. Grid-based state space representations is known to offer several advantages over topological representations \citep{burgard1996estimating, fox2003bayesian}. In the passive localization approaches actions taken by the agent are random.

{\bf Active Markov Localization} (AML) \citep{fox1998active} is the active variant of Markov Localization where the actions taken by the agent are chosen to maximize the ratio of the `utility' of the action to the `cost' of the action. The `utility' of an action $a$ at time $t$ is defined as the expected reduction in the uncertainity of the agent state after taking the action $a$ at time $t$ and making the next observation at time $t+1$: $U_t(a) = H(y_t) - \EX_a[H(y_{t+1})]$, where $H(y)$ denotes the entropy of the belief: $H(y) = - \sum_yBel(y)\log Bel(y)$, and $\EX_a[H(y_{t+1})]$ denotes the expected entropy of the agent after taking the action $a$ and observing $y_{t+1}$. The `cost' of an action refers to the time needed to perform the action. In our environment, each action takes a single time step, thus the cost is constant.

We define a generalized version of the AML algorithm. The utility can be maximized over a sequence of actions rather than just a single action. Let $a^* \in \mathbb{A}^{n_l}$ be the action sequence of length $n_l$ that maximizes the utility at time $t$, $a^* = \argmax_a U_t(a)$ (where $\mathbb{A}$ denotes the action space). After computing $a^*$, the agent need not take all the actions in $a^*$ before maximizing the utility again. This is because new observations made while taking the actions in $a^*$ might affect the utility of remaining actions. Let $n_g \in \{1,2,...,n_l\}$ be the number of actions taken by the agent, denoting the greediness of the algorithm. Due to the high computational complexity of calculating utility, the agent performs random action until belief is concentrated on $n_m$ states (ignoring beliefs under a certain threshold). The complexity of the generalized AML is $\mathcal{O}(n_m(n_l-n_g)|\mathbb{A}|^{n_l})$. Given sufficient computational power, the optimal sequence of actions can be calculated with $n_l$ equal to the length of the episode, $n_g = 1$, and $n_m$ equal to the size of the state space.    

In the original AML algorithm, the utility was maximized over single actions, i.e. $n_l=1$ which also makes $n_g=1$. The value of $n_m$ used in their experiments is not reported, however they show an example with $n_m=6$. We run AML with all possible combination of values of $n_l \in \{1,5,10,15\}$, $n_g \in \{1, n_l\}$ and $n_m = \{5,10\}$ and define two versions: (1) Active Markov Localization (Fast): Generalized AML algorithm using the values of $n_l,n_g,n_m$ that maximize the performance while keeping the runtime comparable to ANL, and (2) Active Markov Localization (Slow): Generalized AML algorithm using the values of $n_l,n_g,n_m$ which maximize the performance while keeping the runtime for 1000 episodes below 24hrs (which is the training time of the proposed model) in each environment.

The perceptual model for both Markov Localization and Active Markov Localization needs to be specified separately. For the 2D environments, the perceptual model uses 1-hot vector representation of depth. For the 3D Environments, the perceptual model uses a pretrained Resnet-18 \citep{he2016deep} model to calculate the feature representations for the agent observations and the memory images.

\begin{figure*}
\centering
\includegraphics[width=0.99\linewidth,height=\textheight,keepaspectratio]{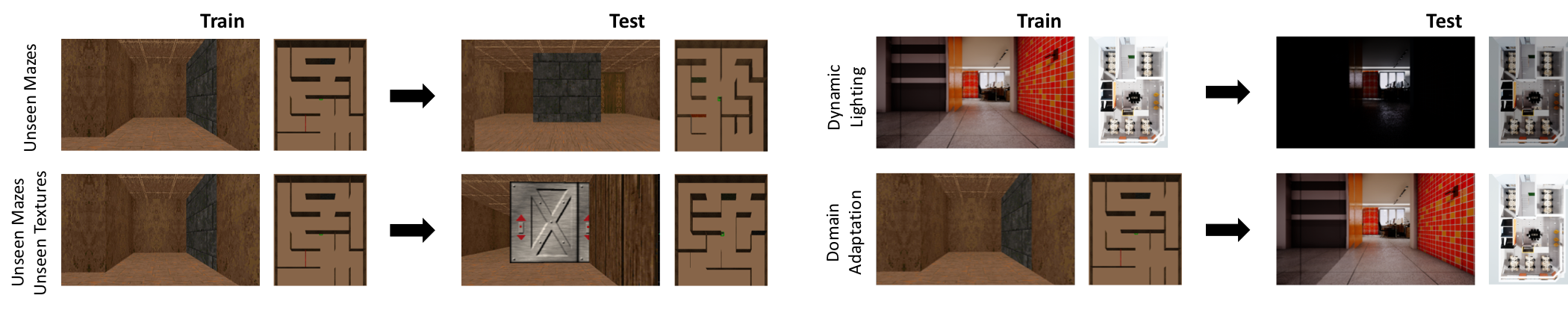}
\caption{Different experiments in the 3D Environments. Refer to the text for more details.}
\label{fig:setup}
\end{figure*}

\subsection{Results}
\paragraph{2D Environments}
For the Maze2D environment, we run all models on mazes having size $7\times7$, $15\times15$ and $21\times21$ with varying episode lengths. We train all the models on randomly generated mazes and test on a fixed set of 1000 mazes (different from the mazes used in training). The results on the Maze2D environment are shown in Table~\ref{tab:maze2d}. As seen in the table, the proposed model, Active Neural Localization, outperforms all the baselines on an average. The proposed method achieves a higher overall accuracy than AML (Slow) while being 100 times faster. Note that the runtime of AML for 1000 episodes is comparable to the total training time of the proposed model. The long runtime of AML (Slow) makes it infeasible to be used in real-time in many cases. When AML has comparable runtime, its performance drops by about 37\% (AML (Fast)). We also observe that the difference in the performance of ANL and baselines is higher for smaller episode lengths. This indicates that ANL is more efficient (meaning it requires fewer actions to localize) in addition to being more accurate.

\setlength{\tabcolsep}{4.0pt}
\begin{table}[t]
\scriptsize
\centering
\caption{\small Results on the 3D environments. `Time' refers to the number of seconds required to evaluate 1000 episodes with the corresponding method and `Acc' stands for accuracy over 1000 episodes.}
\label{tab:maze3d}
\resizebox{\textwidth}{!}{
\begin{tabular}{@{}lllrrrrrrrrrrrrr@{}}
\toprule
Env                                                                                           &      &  & \multicolumn{6}{c}{Maze3D}                                                                                                                                                              & \multicolumn{1}{c}{} & \multicolumn{2}{c}{\begin{tabular}[c]{@{}c@{}}Unreal3D \\ with lights\end{tabular}}                                                                        & \multicolumn{1}{c}{} & \multicolumn{1}{c}{All} & \multicolumn{1}{c}{} & \multicolumn{1}{c}{\begin{tabular}[c]{@{}c@{}}Domain \\ adaptation\end{tabular}}  \\ \midrule
Evaluation Setting                                                                            &      &  & \multicolumn{3}{c}{\begin{tabular}[c]{@{}c@{}}Unseen Mazes \\ Seen Textures\end{tabular}} & \multicolumn{3}{c}{\begin{tabular}[c]{@{}c@{}}Unseen mazes \\ Unseen textures\end{tabular}} & \multicolumn{1}{c}{} & \multicolumn{1}{c}{\begin{tabular}[c]{@{}c@{}}With \\ lights\end{tabular}} & \multicolumn{1}{c}{\begin{tabular}[c]{@{}c@{}}Without \\ lights\end{tabular}} & \multicolumn{1}{c}{} & \multicolumn{1}{c}{}    & \multicolumn{1}{c}{} & \multicolumn{1}{c}{\begin{tabular}[c]{@{}c@{}}Maze3D to \\ Unreal3D\end{tabular}} \\ \midrule
No. of landmarks                                                                              &      &  & \multicolumn{1}{c}{10}        & \multicolumn{1}{c}{5}       & \multicolumn{1}{c}{0}       & \multicolumn{1}{c}{10}        & \multicolumn{1}{c}{5}        & \multicolumn{1}{c}{0}        & \multicolumn{1}{c}{} & \multicolumn{1}{c}{}                                                       & \multicolumn{1}{c}{}                                                          & \multicolumn{1}{c}{} & \multicolumn{1}{c}{}    & \multicolumn{1}{c}{} & \multicolumn{1}{c}{}                                                              \\ \midrule
\multirow{2}{*}{\begin{tabular}[c]{@{}l@{}}Markov Localization \\ (Resnet)\end{tabular}}      & Time &  & 2415                          & 2470                        & 2358                        & 2580                          & 2509                         & 2489                         &                      & 2513                                                                       & 2541                                                                          &                      & 2484.4                  &                      & -                                                                                 \\
                                                                                              & Acc  &  & 0.716                         & 0.657                       & 0.641                       & 0.702                         & 0.669                        & 0.652                        &                      & 0.517                                                                      & 0.249                                                                         &                      & 0.600                   &                      & -                                                                                 \\ \midrule
\multirow{2}{*}{\begin{tabular}[c]{@{}l@{}}Active Markov \\ Localization (Fast)\end{tabular}} & Time &  & 14231                         & 12409                       & 11662                       & 15738                         & 12098                        & 11761                        &                      & 11878                                                                      & 5511                                                                          &                      & 11911.0                 &                      & -                                                                                 \\
                                                                                              & Acc  &  & 0.741                         & 0.701                       & 0.669                       & 0.745                         & 0.687                        & 0.689                        &                      & 0.546                                                                      & 0.279                                                                         &                      & 0.632                   &                      & -                                                                                 \\ \midrule
\multirow{2}{*}{\begin{tabular}[c]{@{}l@{}}Active Markov \\ Localization (Slow)\end{tabular}} & Time &  & 48291                         & 47424                       & 43096                       & 48910                         & 44500                        & 44234                        &                      & 47962                                                                      & 11205                                                                         &                      & 41952.8                 &                      & -                                                                                 \\
                                                                                              & Acc  &  & 0.759                         & 0.749                       & 0.694                       & 0.787                         & 0.730                        & 0.720                        &                      & 0.577                                                                      & 0.302                                                                         &                      & 0.665                   &                      & -                                                                                 \\ \midrule
\multirow{2}{*}{\begin{tabular}[c]{@{}l@{}}Active Neural \\ Localization\end{tabular}}        & Time &  & 297                           & 300                         & 300                         & 300                           & 300                          & 301                          &                      & 2750                                                                       & 2699                                                                          &                      & 905.9                   &                      & 2756                                                                              \\
                                                                                              & Acc  &  & \textbf{0.889}                & \textbf{0.859}              & \textbf{0.852}              & \textbf{0.858}                & \textbf{0.839}               & \textbf{0.871}               &                      & \textbf{0.934}                                                             & \textbf{0.505}                                                                & \textbf{}            & \textbf{0.826}          & \textbf{}            & \textbf{0.921}                                                                    \\ \bottomrule
\end{tabular}}
\end{table}

\paragraph{3D Environments}
All the mazes in the Maze3D environment are of size 70$\times$70 while the office environment environment is of size 70$\times$50. The agent location is a continuous value in this range. Each cell roughly corresponds to an area of 40cm$\times$40cm in the real world. The set of memory images correspond to only about 6\% of the total states. Likelihood of rest of the states are obtained by bilinear smoothing. All episodes have a fixed length of 30 actions. Although the size of the Office Map is 70$\times$50, we represent Likelihood and Belief by a 70$\times$70 in order to transfer the model between both the 3D environments for domain adaptation. We also add a Gaussian noise of 5\% standard deviation to all translations in 3D environments. 

In the \textbf{Maze3D} environment, we vary the difficulty of the environment by varying the number of \emph{landmarks} in the environment. Landmarks are defined to be walls with a unique texture. Each landmark is present only on a single wall in a single cell in the maze grid. All the other walls have a common texture making the map very ambiguous. We expect landmarks to make localization easier as the likelihood maps should have a lower entropy when the agent visits a landmark, which consequently should reduce the entropy of the Belief Map. We run experiments with 10, 5 and 0 landmarks. The textures of the landmarks are randomized during training. This technique of domain randomization has shown to be effective in generalizing to unknown maps within the simulation environment \citep{lample2016playing} and transferring from simulation to real-world \citep{tobin2017domain}. In each experiment, the agent is trained on a set of 40 mazes and evaluated in two settings: (1) Unseen mazes with seen textures: the textures of each wall in the test set mazes have been seen in the training set, however the map design of the test set mazes are unseen and (2) Unseen mazes with unseen textures: both the textures and the map design are unseen. We test on a set of 20 mazes for each evaluation setting. Figure~\ref{fig:setup} shows examples for both the settings.

In the \textbf{Unreal3D} environment, we test the effectiveness of the model in adapting to dynamic lightning changes. We modified the the Office environment using the Unreal Game Engine Editor to create two scenarios: (1) Lights: where all the office lights are switched on; (2) NoLights: where all the office lights are switched off. Figure~\ref{fig:setup} shows sample agent observations with and without lights at the same locations. To test the model's ability to adapt to dynamic lighting changes, we train the model on the Office map with lights and test it on same map without lights. The memory images provided to the agent are taken while lights are switched on. Note that this is different as compared to the experiments on unseen mazes in Maze3D environment, where the agent is given memory images of the unseen environments. 

\begin{figure*}
\centering
\includegraphics[width=0.99\linewidth,height=\textheight,keepaspectratio]{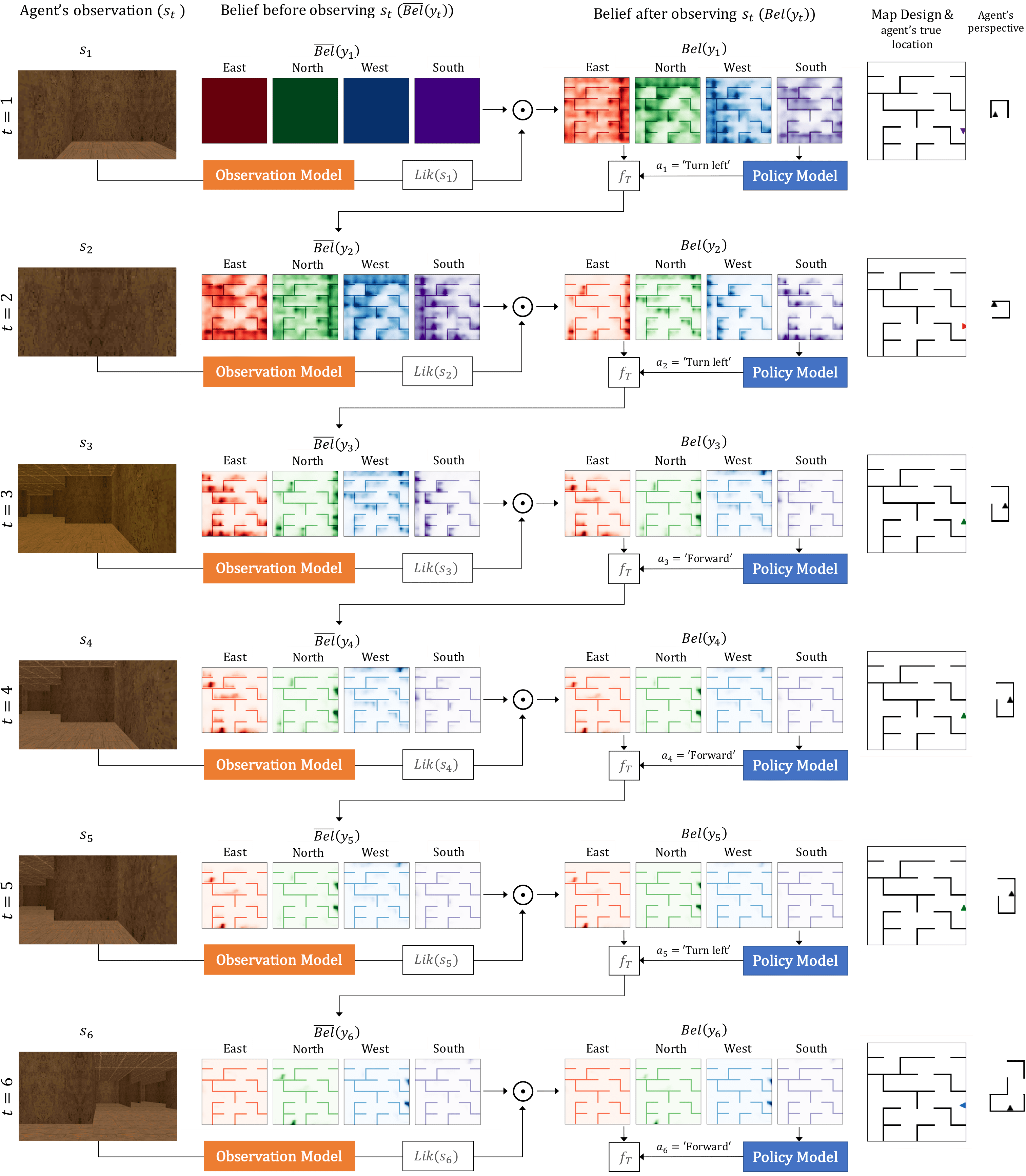}
\caption{\small An example of the policy execution and belief propagation in the Maze3D Environment. The rows shows consecutive time steps in a episode. The columns show Agent's observation, the belief of its location before and after making the observation, the map design and agent's perspective of the world. Agent's true location is also marked in the map design (not visible to the agent). Belief maps show the probability of being at a particular location. Darker shades imply higher probability. The belief of its orientation and agent's true orientation are also highlighted by colors.}
\label{fig:example}
\end{figure*}

The results for the 3D environments are shown in Table~\ref{tab:maze3d} and an example of the policy execution is shown in Figure~\ref{fig:example}\footnote{Policy execution videos can be seen at \url{https://tinyurl.com/y8gp3mkh}}. The proposed model significantly outperforms all baseline models in all evaluation settings with the lowest runtime. We see similar trends of runtime and accuracy trade-off between the two version of AML as seen in the 2D results. The absolute performance of AML (Slow) is rather poor in the 3D environments as compared to Maze2D. This is likely due to the decrease in value of look-ahead parameter, $n_l$, to 3 and the increase in value of the greediness hyper-parameter, $n_g$ to 3, as compared to $n_l=5, n_g=1$ in Maze 2D, in order to ensure runtimes under 24hrs. 

The ANL model performs better on the realistic Unreal environment as compared to Maze3D environment, as most scenes in the Unreal environment consists of unique landmarks while Maze3D environment consists of random mazes with same texture except those of the landmarks. In the Maze3D environment, the model is able to generalize well to not only unseen map design but also to unseen textures. However, the model doesn't generalize well to dynamic lighting changes in the Unreal3D environment. This highlights a current limitation of RGB image-based localization approaches as compared to depth-based approaches, as depth sensors are invariant to lighting changes.

\paragraph{Domain Adaptation} We also test the ability of the proposed model to adapt between different simulation environments. The model trained on the Maze3D is directly tested on the Unreal3D Office Map without any fine-tuning. The results in Table~\ref{tab:maze3d} show that the model is able to generalize well to Unreal environment from the Doom Environment. We believe that the policy model generalizes well because the representation of belief and map design is common in all environments and policy model is based only on the belief and the map design, while the perceptual model generalizes well because it has learned to measure the similarity of the current image with the memory images as it was trained on environments with random textures. This property is similar to siamese networks used for one-shot image recognition \citep{koch2015siamese}.

\section{Discussion}
\vspace{-1em}
In this chapter, we proposed a fully-differentiable model for active global localization which uses structured components for Bayes filter-like belief propagation and learns a policy based on the belief to localize accurately and efficiently. This allows the policy and observation models to be trained jointly using reinforcement learning. We showed the effectiveness of the proposed model on a variety of challenging 2D and 3D environments including a realistic map in the Unreal environment. The results show that our model consistently outperforms the baseline models while being order of magnitudes faster. We also show that a model trained on random textures in the Doom simulation environment is able to generalize to photo-realistic Office map in the Unreal simulation environment. While this gives us hope that model can potentially be transferred to real-world environments, we leave that for future work. The limitation of the model to adapt to dynamic lightning can potentially be tackled by training the model with dynamic lightning in random mazes in the Doom environment. There can be several extensions to the proposed model too. The model can be combined with Neural Map~\citep{parisotto2017neural} to train an end-to-end model for a SLAM-type system and the architecture can also be utilized for end-to-end planning under uncertainity.

In addition to localization, embodied agents need to build maps for exploring the environment efficiently. In the next chapter, we add mapping capabilities to navigation agents in addition to localization, to design a Simulataneous Localization and Mapping (SLAM) system called `Active Neural SLAM', which is a capable of building maps from raw-pixel data, learning exploration policies and long-term planning.

\chapter{Active Neural SLAM}
\label{chap:ans}

\blfootnote{Webpage: \url{https://devendrachaplot.github.io/projects/neural-slam}}

Navigation is a critical task in building intelligent agents. Navigation tasks can be expressed in many forms, for example, point goal tasks involve navigating to a specific coordinates and semantic navigation involves finding path to a specific scene or object. Irrespective of the task, a core problem for navigation in unknown environments is exploration, \textit{i.e.}, how to efficiently visit as much of the environment. This is useful for maximizing the coverage to give the best chance of finding the target in unknown environments or for efficiently pre-mapping environments on a limited time-budget.

Recent work from \citet{chen2018learning} has used end-to-end 
learning to tackle this problem.
Their motivation is three fold: \textit{a)} learning
provides flexibility to the choice of input modalities
(classical systems rely on observing geometry through use
of specialized sensors, while learning systems can infer
geometry directly from RGB images), \textit{b)} use of learning
can improve robustness to errors in explicit state estimation, 
and \textit{c)} learning can effectively leverage structural
regularities of the real world, leading to more efficient behavior in
previously unseen environments. This lead to their design
of an end-to-end trained neural network based policy that
processed raw sensory observations to directly output
actions that the agent should execute.

While use of learning for exploration is well motivated, casting
the exploration problem as an end-to-end learning problem has its
own drawbacks. Learning about mapping, state-estimation and path-planning
purely from data in an end-to-end manner can be prohibitively
expensive. Consequently, past end-to-end learning work for exploration
from \citet{chen2018learning} relies on use of imitation learning
and many millions of frames of experience, but still performs worse than 
classical methods that don't require any training at all.

This motivates our work. In this chapter, we investigate alternate
formulations of employing learning for exploration that retains 
the advantages that learning has to offer, but doesn't suffer
from the drawbacks of full-blown end-to-end learning. 
Our key conceptual insight is that use of learning for 
leveraging structural regularities of indoor environments,
robustness to state-estimation errors, 
and flexibility with respect to input modalities, happens at different time scales
and can thus be factored out.
This motivates use of learning in a modular and hierarchical fashion 
inside of what one may call a `classical navigation pipeline'. 
This results in navigation policies that can work with raw sensory
inputs such as RGB images, are robust to state estimation errors,
and leverage regularities of real world layout. This results in
extremely competitive performance over both geometry-based methods and
recent learning-based methods; at the same time requiring a fraction of
the number of samples.

More specifically, our proposed exploration architecture comprises
of a learned Neural SLAM module, a global policy, and a local policy, 
that are interfaced via the map and an analytical path planner. The learned
Neural SLAM module produces free space maps and estimates agent pose from input RGB images and motion sensors. 
The global policy consumes this free-space map with agent pose and employs learning 
to exploit structural regularities in layout of real world environments
to produce long-term goals. 
These long-term goals are used to generate short-term
goals for the local policy (using a geometric path-planner).
This local policy uses learning to directly map 
raw RGB images to actions that the agent should execute.
Use of learning in the SLAM module provides flexibility with respect to
input modality, learned global policy can exploit regularities
in layout of real world layout of environments, while learned local
policies can use visual feedback to exhibit more robust behaviour.
At the same time, hierarchical and modular design and use of analytical 
planning, significantly cuts down the search space during training,
leading to better performance as well as sample efficiency.

We demonstrate our proposed approach in \textit{visually} and 
\textit{physically} realistic simulators for the task of geometric 
exploration (visit as much area as possible). We work with the
Habitat simulator from \citet{savva2019habitat}. While Habitat is already visually 
realistic (it uses real world scans from 
\citet{Matterport3D} and \citet{xiazamirhe2018gibsonenv} as environments),
we improve its physical realism by
using actuation and odometry sensor noise models, that we collected
by conducting physical experiments on a real mobile robot.
Our experiments and ablations in this realistic simulation reveal
the effectiveness of our proposed approach for the task of exploration.
A straight-forward modification of our method also tackles 
point-goal navigation tasks, and won the AI Habitat challenge
at CVPR2019 across all tracks.

Navigation has been well studied in classical robotics. There has been a renewed
interest in the use of learning to arrive at navigation policies, for a variety of
tasks. Our work builds upon concepts in classical robotics and learning 
for navigation. We survey related works below.

\textbf{Navigation Approaches.} Classical approaches to navigation break the problem into two parts: mapping and path planning. Mapping is done via simultaneous localization and mapping~\citep{thrun2005probabilistic, hartley2003multiple, slam-survey:2015}, by fusing information from multiple views of the environment. 
While sparse reconstruction can be done well with monocular RGB images~\citep{mur2017orb}, dense mapping is inefficient~\citep{newcombe2011dtam} or requires specialized scanners such as Kinect~\citep{izadiUIST11}. 
Maps are used to compute paths to goal locations via path planning~\citep{kavraki1996probabilistic, lavalle2000rapidly, canny1988complexity}. These classical methods have inspired recent learning-based techniques. 
Researchers have designed neural network policies that 
reason via spatial representations~\citep{gupta2017cognitive, parisotto2017neural, zhang2017neural, henriques2018mapnet, gordon2018iqa}, 
topological representations~\citep{savinov2018semi, savinov2018episodic}, or use differentiable and trainable planners~\citep{tamar2016value, lee2018gated, gupta2017cognitive, khan2017end}.
Our work furthers this research, and we study a hierarchical and modular  
decomposition of the problem, and employ learning inside these components instead of
end-to-end learning. Research also focuses on incorporating semantics in SLAM~\citep{pronobis2012large, walter2013learning}.

\textbf{Exploration in Navigation.}
While a number of works focus on passive map-building, path planning and 
goal-driven policy learning, a much smaller body of work tackles the
problem of active SLAM, i.e., how to actively control the camera for 
map building. We point readers to \citet{slam-survey:2015} for a detailed survey,
and summarize major themes below.
Most such works frame this problem as a Partially
Observable Markov Decision Process (POMDP) that are approximately 
solved~\citep{martinez2009bayesian, kollar2008trajectory}, and or seek to find a sequence of 
actions that minimizes uncertainty of maps~\citep{stachniss2005information, carlone2014active}.
Another line of work, explores by picking vantage points (such
as on the frontier between explored and unexplored 
regions~\citep{dornhege2013frontier, holz2010evaluating, yamauchi1997frontier, xu2017autonomous}).
Recent works from \citet{chen2018learning, savinov2018episodic, fang2019scene}
attack this problem via learning. Our proposed modular policies
unify the last two lines of research, and we show improvements over
representative methods from both these lines of work. Exploration
has also been studied more generally in RL in the context of exploration-exploitation trade-off~\citep{sutton2018reinforcement, kearns2002near, auer2002using, jaksch2010near}. 

\textbf{Hierarchical and Modular Policies.} Hierarchical RL~\citep{dayan1993feudal, sutton1999between, barto2003recent} is an active area of research, aimed at automatically discovering
hierarchies to speed up learning. However, this has proven to be 
challenging, and thus most work has resorted to using hand-defining
hierarchies. For example in context of navigation, \citet{bansal2019combining} 
and \citet{kaufmann2019beauty} design modular policies for navigation, 
that interface learned policies with low-level feedback controllers.
Hierarchical and modular policies have also been used for Embodied Question Answering \citep{das2017embodied, gordon2018iqa, das2018neural}.

\section{Task Setup}
We follow the exploration task setup proposed by~\citet{chen2018learning} where the objective is to maximize the coverage in a fixed time budget. The coverage is defined as the total area in the map known to be traversable. Our objective is train a policy which takes in an observation $s_t$ at each time step $t$ and outputs a navigational action $a_t$ to maximize the coverage.

We try to make our experimental setup in simulation as realistic as possible with the goal of transferring trained policies to the real world. We use the Habitat simulator~\citep{savva2019habitat} with the Gibson~\citep{xiazamirhe2018gibsonenv} and Matterport (MP3D)~\citep{Matterport3D} datasets for our experiments. Both Gibson and Matterport datasets are based on real-world scene reconstructions are thus significantly more realistic than synthetic SUNCG dataset~\citep{song2016ssc} used for past research on exploration~\citep{chen2018learning, fang2019scene}.

In addition to synthetic scenes, prior works on learning-based navigation have also assumed simplistic agent motion. Some works limit agent motion on a grid with 90 degree rotations~\citep{zhu2017target, gupta2017cognitive, parisotto2017neural, chaplot2018active}. Other works which implement fine-grained control, typically assume unrealistic agent motion with no noise~\citep{savva2019habitat} or perfect knowledge of agent pose~\citep{chaplottransfer}. Since the motion is simplistic, it becomes trivial to estimate the agent pose in most cases even if it is not assumed to be known. The reason behind these assumptions on agent motion and pose is that motion and sensor noise models are not known. In order to relax both these assumptions, we collect motion and sensor data in the real-world and implement more realistic agent motion and sensor noise models in the simulator as described in the following subsection. 

\subsection{Actuation and Sensor Noise Model}
\label{sec_ans:noise}
We represent the agent pose by $(x, y, o)$ where $x$ and $y$ represent the $xy$ co-ordinate of the agent measured in metres and $o$ represents the orientation of the agent in radians (measured counter-clockwise from $x$-axis). Without loss of generality, assume agents starts at $p_0 = (0, 0, 0)$. Now, suppose the agent takes an action $a_t$. Each action is implemented as a control command on a robot. Let the corresponding control command be $\Delta u_a = (x_a, y_a, o_a)$. Let the agent pose after the action be $p_{1} = (x^\star, y^\star, o^\star)$. The actuation noise ($\epsilon_{act}$) is the difference between the actual agent pose ($p_{1}$) after the action and the intended agent pose ($p_0 + \Delta u$):
\begin{equation*}
	\vspace{-5pt}
	\epsilon_{act} = p_1 - (p_0 + \Delta u) = (x^\star - x_a, y^\star - y_a, o^\star - o_a)
\end{equation*}

Mobile robots typically have sensors which estimate the robot pose as it moves. Let the sensor estimate of the agent pose after the action be $p_{1}' = (x', y', o')$. The sensor noise ($\epsilon_{sen}$) is given by the difference between the sensor pose estimate ($p_{1}'$) and the actual agent pose($p_{1}$):
\begin{equation*}
	\vspace{-5pt}
	\epsilon_{sen} = p_1' - p_1 = (x' - x^\star, y'- y^\star, o' - o^\star)
\end{equation*}

In order to implement the actuation and sensor noise models, we would like to collect data for navigational actions in the Habitat simulator. We use three default navigational actions: Forward: move forward by 25cm, Turn Right: on the spot rotation clockwise by 10 degrees, and Turn Left: on the spot rotation counter-clockwise by 10 degrees. The control commands are implemented as $u_{Forward} = (0.25, 0, 0)$, $u_{Right}: (0, 0, -10*\pi/180)$ and $u_{Left}: (0, 0, 10*\pi/180)$. In practice, a robot can also rotate slightly while moving forward and translate a bit while rotating on-the-spot, creating rotational actuation noise in forward action and similarly, a translation actuation noise in on-the-spot rotation actions.

We use a LoCoBot\footnote{http://locobot.org} to collect data for building the actuation and sensor noise models. We use the pyrobot API~\citep{pyrobot2019} along with ROS~\citep{Quigley09} to implement the control commands and get sensor readings. For each action $a$, we fit a separate Gaussian Mixture Model for the actuation noise and sensor noise, making a total of 6 models.  Each component in these Gaussian mixture models is a multi-variate Gaussian in 3 variables, $x$, $y$ and $o$. For each model, we collect 600 datapoints. The number of components in each Gaussian mixture model are chosen using cross-validation. 
We implement these actuation and sensor noise models in the Habitat simulator for our experiments. We have released the noise models, along with their implementation in the Habitat simulator in the open-source \href{https://github.com/devendrachaplot/Neural-SLAM}{code}.

\section{Methods}

\begin{figure*}
\centering
\includegraphics[width=0.99\linewidth,keepaspectratio]{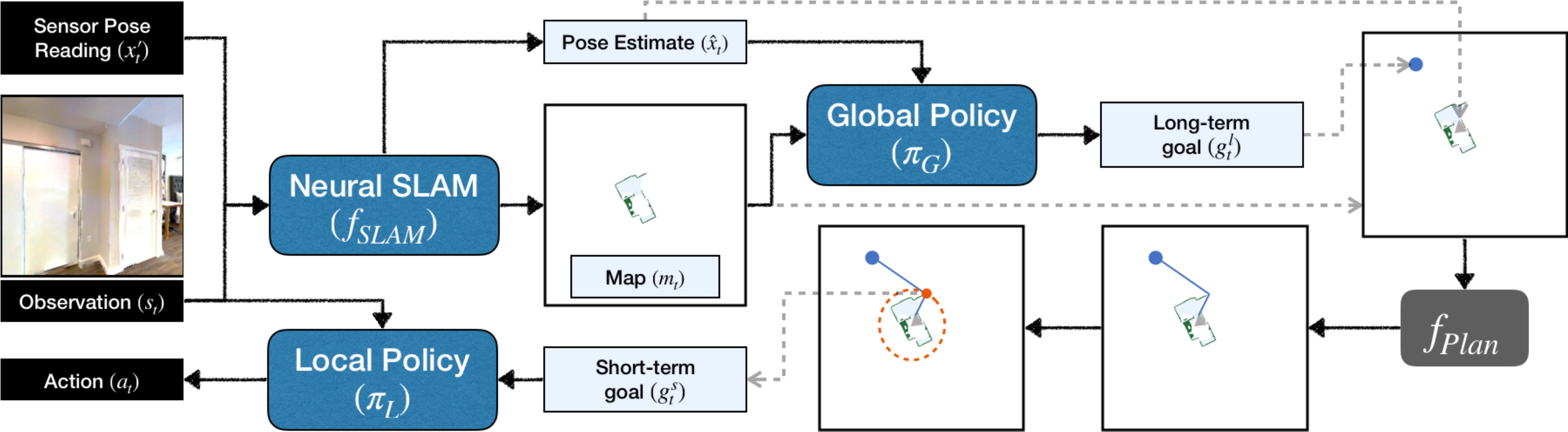}
\caption{\small Overview of our approach. We use a neural network based Mapper
that predicts a map and agent pose estimate from incoming RGB observations and sensor readings. This map is used by a Global policy to output a long-term goal, which is converted to a short-term goal using an analytic path planner. A Local Policy is trained to navigate to this short-term goal.}
\label{fig_ans:anm_arch}
\end{figure*}

We propose a modular navigation model, `Active Neural SLAM'. It consists of three components: a \textit{Neural SLAM module}, a \textit{Global policy} and a \textit{Local policy} as shown in Figure~\ref{fig_ans:anm_arch}. The Neural SLAM module predicts the map of the environment and the agent pose based on the current observations and previous predictions. The Global policy uses the predicted map and agent pose to produce a long-term goal. The long-term goal is converted into a short-term goal using path planning. The Local policy takes navigational actions based on the current observation to reach the short-term goal. \\

\noindent\textbf{Map Representation.} The Active Neural SLAM model internally maintains a spatial map, $m_t$ and pose of the agent $x_t$. The spatial map, $m_t$, is a ${2 \times M \times M}$ matrix where $M \times M$ denotes the map size and each element in this spatial map corresponds to a cell of size $25cm^2$ ($5cm \times 5cm$) in the physical world. Each element in the first channel denotes the probability of an obstacle at the corresponding location and each element in the second channel denotes the probability of that location being explored. A cell is considered to be explored when it is known to be free space or an obstacle. The spatial map is initialized with all zeros at the beginning of an episode, $m_0 = [0]^{2 \times M \times M}$. The pose $x_t \in \mathbb{R}^3$ denotes the $x$ and $y$ coordinates of the agent and the orientation of the agent at time $t$. The agent always starts at the center of the map facing east at the beginning of the episode, $x_0 = (M/2, M/2, 0.0)$.\\ 

\begin{figure*}
\centering
\begin{minipage}{\textwidth}
\includegraphics[width=\linewidth,height=\textheight,keepaspectratio]{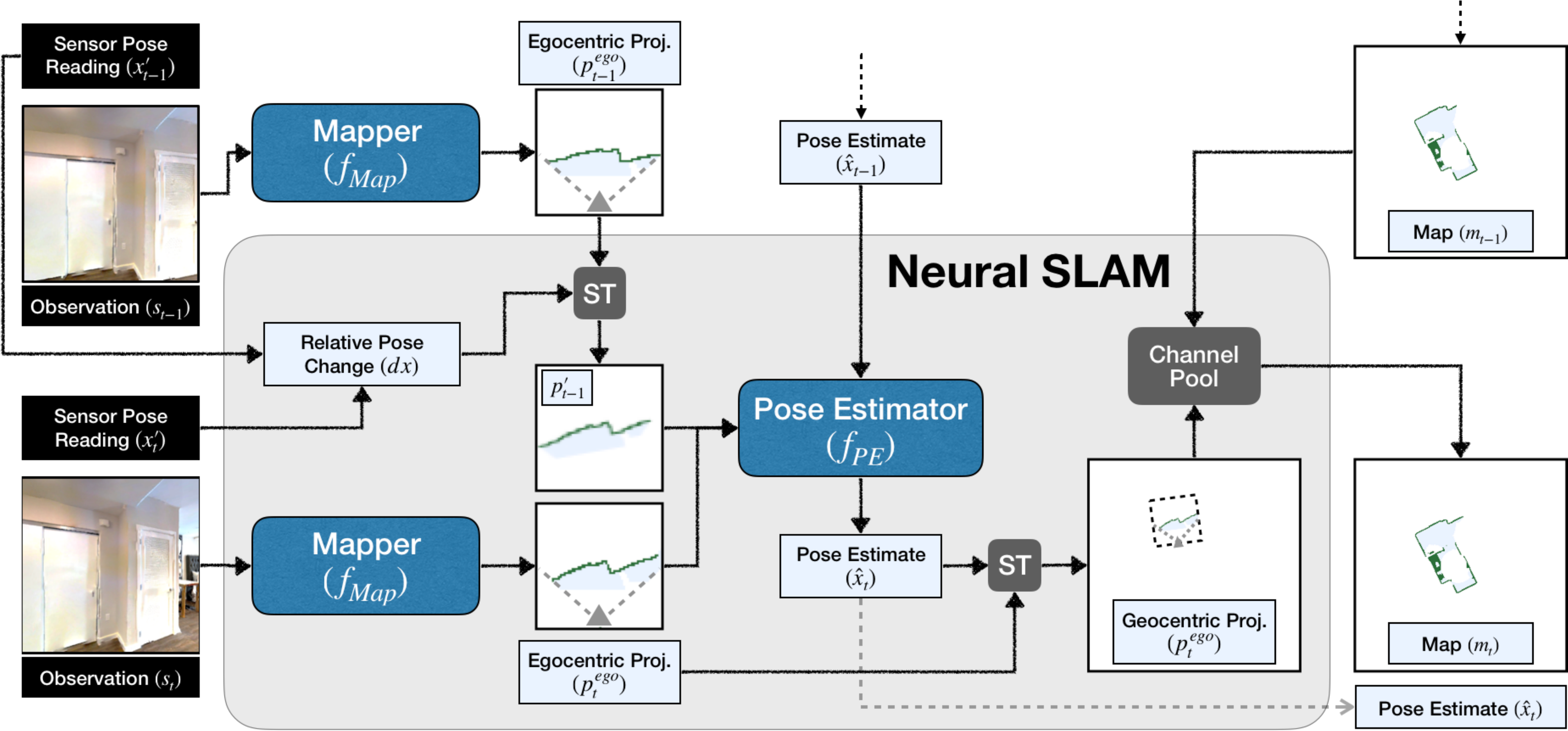}
\caption{\textbf{Architecture of the Neural SLAM module:} The Neural SLAM module ($f_{Map}$) takes in the current RGB observation, $s_t$, the current and last sensor reading of the agent pose $x_{t-1:t}'$, last agent pose estimate, $\hat{x}_{t-1}$ and the map at the previous time step $m_{t-1}$ and outputs an updated map, $m_{t}$ and the current agent pose estimate, $\hat{x}_t$. `ST' denotes spatial transformation. }
\label{fig_ans:map}
\end{minipage}\hfill
\end{figure*}

\noindent\textbf{Neural SLAM Module.}
The Neural SLAM Module ($f_{Map}$) takes in the current RGB observation, $s_t$, the current and last sensor reading of the agent pose $x_{t-1:t}'$, last agent pose and map estimates, $\hat{x}_{t-1}, m_{t-1}$ and outputs an updated map, $m_{t}$, and the current agent pose estimate, $\hat{x}_t$, (see Figure \ref{fig_ans:map}): $m_t, \hat{x}_t = f_{Map}(s_t, x_{t-1:t}', \hat{x}_{t-1}, m_{t-1}| \theta_M)$, where $\theta_M$ denote the trainable parameters of the Neural SLAM module. It consists of two learned components, a Mapper and a Pose Estimator. The Mapper outputs a egocentric top-down 2D spatial map, $p_t^{ego} \in [0, 1]^{2 \times V \times V}$ (where $V$ is the vision range), predicting the obstacles and the explored area in the current observation. The Pose Estimator predicts the agent pose ($\hat{x_t}$) based on past pose estimate ($\hat{x}_{t-1}$) and last two egocentric map predictions ($p_{t-1:t}^{ego}$). It essentially compares the current egocentric map prediction to the last egocentric map prediction transformed to the current frame to predict the pose change between the two maps. The egocentric map from the Mapper is transformed to a geocentric map based on the pose estimate given by the Pose Estimator and then aggregated with the previous spatial map ($m_{t-1}$) to get the current map($m_t$). \\

\noindent\textbf{Global Policy.}
The Global Policy takes $h_t \in [0,1]^{4 \times M \times M}$ as input, where the first two channels of $h_t$ are the spatial map $m_t$ given by the SLAM module, the third channel represents the current agent position estimated by the SLAM module, the fourth channel represents the visited locations, i.e. $\forall i, j \in \{1, 2, \ldots, m\}$:
\vspace{-5pt}
\begin{equation*}
\begin{split}
    h_t[c, i, j] & = m_t[c, i, j] \quad \forall c \in \{0,1\}\\
    h_t[2, i, j] & = 1 \qquad \qquad \text{ if } i = \hat{x}_t[0] \text{ and } j = \hat{x}_t[1]\\
    h_t[3, i, j] & = 1 \qquad \qquad \text{ if } (i,j) \in [(\hat{x}_k[0], \hat{x}_k[1])]_{k \in \{0, 1, \ldots, t\}}
\end{split}
\end{equation*}
\vspace{-5pt}

We perform two transformations before passing $h_t$ to the Global Policy model. The first transformation subsamples a window of size $4 \times G \times G$ around the agent from $h_t$. The second transformation performs max pooling operations to get an output of size $4 \times G \times G$ from $h_t$. Both the transformations are stacked to form a tensor of size $8 \times G \times G$ and passed as input to the Global Policy model. The Global Policy uses a 5-layer convolutional neural network to predict a long-term goal, $g_t^l$ in $G \times G$ space: $g_t^l = \pi_{G}(h_t|\theta_{G})$, where $\theta_{G}$ are the parameters of the Global Policy.\\

\noindent\textbf{Planner.}
The Planner takes the long-term goal ($g_t^l$), the spatial obstacle map ($m_t$) and the agnet pose estimate ($\hat{x}_t$) as input and computes the short-term goal $g_t^s$, i.e. $g_t^s = f_{Plan}(g_t^l, m_t, \hat{x}_t)$. It computes the shortest path from the current agent location to the long-term goal ($g_t^l$) using the Fast Marching Method~\citep{sethian1996fast} based on the current spatial map $m_t$. The unexplored area is considered as free space for planning. We compute a short-term goal coordinate (farthest point within $d_s(=1.25m)$ from the agent) on the planned path. \\

\noindent\textbf{Local Policy.}
The Local Policy takes as input the current RGB observation ($s_t$) and the short-term goal ($g_t^s$) and outputs a navigational action, $a_t = \pi_{L}(s_t, g_t^s|\theta_L)$, where $\theta_L$ are the parameters of the Local Policy. The short-term goal coordinate is transformed into relative distance and angle from the agent's location before being passed to the Local Policy. It consists of a 3-layer convolutional neural network followed by a GRU layer.

\subsection{Neural SLAM Module Implementation}
The Neural SLAM module ($f_{Map}$) takes in the current RGB observation, $s_t \in \mathbb{R}^{3 \times H \times W}$, the current and last sensor reading of the agent pose $x_{t-1:t}'$ and the map at the previous time step $m_{t-1} \in \mathbb{R}^{2 \times M \times M}$ and outputs an updated map, $m_{t} \in \mathbb{R}^{2 \times M \times M}$ (see Figure \ref{fig_ans:map}):
$$m_t, \hat{x}_t = f_{Map}(s_t, x_{t-1:t}', \hat{x}_{t-1}, m_{t-1}| \theta_M, b_{t-1})$$

where $\theta_M$ denote the trainable parameters and $p_{t-1}$ denotes internal representations of the Neural SLAM module. The Neural SLAM module can be broken down into two parts, a Mapper ($f_{Pr}$) and a Pose Estimator Unit ($f_{PE}$,). The Mapper outputs a egocentric top-down 2D spatial map, $p_t^{ego} \in [0, 1]^{2 \times V \times V}$ (where $V$ is the vision range), predicting the obstacles and the explored area in the current observation: $p_t^{ego} = f_{Pr}(s_t|\theta_{Pr})$, where $\theta_{Pr}$ are the parameters of the Mapper. It consists of Resnet18 convolutional layers to produce an embedding of the observation. This embedding is passed through two fully-connected layers followed by 3 deconvolutional layers to get the first-person top-down 2D spatial map prediction.

Now, we would like to add the egocentric map prediction ($p_t^{ego}$) to the geocentric map from the previous time step ($m_{t-1}$). In order to transform the egocentric map to the geocentric frame, we need the pose of the agent in the geocentric frame. The sensor reading $x_t'$ is typically noisy. Thus, we have a Pose Estimator to correct the sensor reading and give an accurate estimate of the agent's geocentric pose. 

In order to estimate the pose of the agent, we first calculate the relative pose change ($dx$) from the last time step using the sensor readings at the current and last time step ($x_{t-1}', x_t'$). Then we use a Spatial Transformation~\citep{jaderberg2015spatial} on the egocentric map prediction at the last frame ($p_{t-1}^{ego}$) based on the relative pose change ($dx$),  $p_{t-1}' = f_{ST}(p_{t-1}^{ego}|dx)$. Note that the parameters of this Spatial Transformation are not learnt, but calculated using the pose change ($dx$). This transforms the projection at the last step to the current egocentric frame of reference. If the sensor was accurate, $p_{t-1}'$ would highly overlap with $p_t^{ego}$. The Pose Estimator Unit takes in $p_{t-1}'$ and $p_t^{ego}$ as input and predicts the relative pose change: $\hat{dx_t} = f_{PE}(p_{t-1}',p_t^{ego}|\theta_{PE})$ The intuition is that by looking at the egocentric predictions of the last two frames, the pose estimator can learn to predict the small translation and/or rotation that would align them better. The predicted relative pose change is then added to the last pose estimate to get the final pose estimate $\hat{x_t} = \hat{x}_{t-1} + \hat{dx_t}$.

Finally, the egocentric spatial map prediction is transformed to the geocentric frame using the current pose prediction of the agent ($\hat{x_t}$) using another Spatial Transformation and aggregated with the previous spatial map ($m_{t-1}$) using Channel-wise Pooling operation: $m_t = m_{t-1} + f_{ST}(p_t | \hat{x_t})$. Combing all the functions and transformations:
\begin{align*}
m_t & = f_{Map}(s_t, x_{t-1:t}', m_{t-1}| \theta_M, b_{t-1})\\
     & = m_{t-1} + f_{ST}(p_t | x_t' + f_{PE}(f_{ST}(p_{t-1}^{ego}|x_{t-1:t}) , f_{Pr}(s_t|\theta_{Pr})|\theta_{PE})) \\
     & \quad\quad \text{where } {\theta_{Pr}, \theta_{PE}} \in \theta_M, \quad\text{and}\quad p_{t-1}^{ego} \in b_{t-1}
\end{align*}

\section{Experimental setup}
We use the Habitat simulator~\citep{savva2019habitat} with the Gibson~\citep{xiazamirhe2018gibsonenv} and Matterport (MP3D)~\citep{Matterport3D} datasets for our experiments. Both Gibson and MP3D consist of scenes which are 3D reconstructions of real-world environments, however Gibson is collected using a different set of cameras, consists mostly of office spaces while MP3D consists of mostly homes with a larger average scene area. We will use Gibson as our training domain, and use MP3D for domain generalization experiments. 
The observation space consists of RGB images of size $3 \times 128 \times 128$ and base odometry sensor readings of size $3 \times 1$ denoting the change in agent's x-y coordinates and orientation. The actions space consists of three actions: \texttt{move\_forward, turn\_left, turn\_right}. Both the base odometry sensor readings and the agent motion based on the actions are noisy. They are implemented using the sensor and actuation noise models based on real-world data as discussed in Section~\ref{sec_ans:noise}.

We follow the Exploration task setup proposed by~\citet{chen2018learning} where the objective to maximize the coverage in a fixed time budget. Coverage is the total area in the map known to be traversable. We define a traversable point to be known if it is in the field-of-view of the agent and is less than $3m$ away. We use two evaluation metrics, the absolute coverage area in $m^2$ (\textbf{Cov}) and percentage of area explored in the scene (\textbf{\% Cov}), i.e. ratio of coverage to maximum possible coverage in the corresponding scene. During training, each episode lasts for a fixed length of 1000 steps. 

We use train/val/test splits provided by~\citet{savva2019habitat} for both the datasets. Note that the set of scenes used in each split is disjoint, which means the agent is tested on new scenes never seen during training. Gibson test set is not public but rather held out on an online evaluation server for the Pointgoal task. We use the validation as the test set for comparison and analysis for the Gibson domain. We do not use the validation set for hyper-parameter tuning. To analyze the performance of all the models with respect to the size of the scene, we split the Gibson validation set into two parts, a small set of 10 scenes with explorable area ranging from $16m^2$ to $36m^2$, and a large set of 4 scenes with traversable area ranging from $55m^2$ to $100m^2$. Note that the size of the map is usually much larger than the traversable area, with the largest map being about $23m$ long and $11m$ wide. \\

\noindent\textbf{Training Details.}
We train our model in the Gibson domain and transfer it to the Matterport domain. The Mapper is trained to predict egocentric projections, and the Pose Estimator is trained to predict agent pose using supervised learning. The ground truth egocentric projection is computed using geometric projections from ground truth depth. 
The Global and Local policies are both trained using Reinforcement Learning. The reward for the Global policy is the increase in coverage and the reward for the Local policy is the reduction in Euclidean distance to the short-term goal. All the modules are trained simultaneously. Their parameters are independent, but the data distribution is inter-dependent. Based on the actions taken by the Local policy, the future input to Neural SLAM module changes, which in turn changes the map and agent pose input to the Global policy and consequently affects the short-term goal given to the Local Policy. \\

\noindent\textbf{Baselines.} We use a range of end-to-end Reinforcement Learning (RL) methods as baselines:\\
\textbf{RL + 3LConv:} An RL Policy with 3 layer convolutional network followed by a GRU~\citep{cho2014properties} as described by~\citet{savva2019habitat} which is also identical to our Local Policy architecture.\\
\textbf{RL + Res18:} A RL Policy initialized with ResNet18~\citep{he2016deep} pre-trained on ImageNet followed by a GRU.\\
\textbf{RL + Res18 + AuxDepth:} This baseline is adapted from~\citet{mirowski2016learning} who use depth prediction as an auxiliary task. We use the same architecture as our Neural SLAM module (conv layers from ResNet18) with one additional deconvolutional layer for Depth prediction followed by 3 layer convolution and GRU for the policy.\\
\textbf{RL + Res18 + ProjDepth:} This baseline is adapted form~\citet{chen2018learning} who project the depth image in an egocentric top-down in addition to the RGB image as input to the RL policy. Since we do not have depth as input, we use the architecture from RL + Res18 + AuxDepth for depth prediction and project the predicted depth before passing to 3Layer Conv and GRU policy.

For all the baselines, we also feed a 32-dimensional embedding of the sensor pose reading to the GRU along with the image-based representation. This embedding is also learnt end-to-end using RL. All baselines are trained using PPO~\citep{schulman2017proximal} with increase in coverage as the reward.

\setlength{\tabcolsep}{10pt}
\begin{table}[t]
\centering
\caption{\small Exploration performance of the proposed model, Active Neural SLAM (ANS) and baselines. The baselines are adated from [1]~\citet{savva2019habitat}, [2]~\citet{mirowski2016learning} and [3]~\citet{chen2018learning}.}
\label{tab:exp_results}
\resizebox{\textwidth}{!}{
\begin{tabular}{@{}llccl
>{\columncolor[HTML]{DEDEDE}}c 
>{\columncolor[HTML]{DEDEDE}}c l@{}}
\toprule
                       &  &               &                &  & \multicolumn{2}{c}{\cellcolor[HTML]{DEDEDE}Domain Generalization} &  \\
                       &  & \multicolumn{2}{c}{Gibson Val} &  & \multicolumn{2}{c}{\cellcolor[HTML]{DEDEDE}MP3D Test}             &  \\ \midrule
Method                 &  & \% Cov.       & Cov. (m2)      &  & \% Cov.                        & Cov. (m2)                        &  \\ \midrule
RL + 3LConv [1]           &  & 0.72          & 22.382         &  & 0.297                          & 41.277                           &  \\
RL + Res18             &  & 0.725         & 22.446         &  & 0.295                          & 40.916                           &  \\
RL + Res18 + AuxDepth [2] &  & 0.744         & 23.896         &  & 0.286                          & 39.944                           &  \\
RL + Res18 + ProjDepth [3] &  & 0.766         & 24.958         &  & 0.294                          & 41.549                           &  \\
	\textbf{Active Neural SLAM (ANS)}	       &  & \textbf{0.924}         & \textbf{31.379}         &  & \textbf{0.405}                          & \textbf{57.228}                           &  \\ \bottomrule
\end{tabular}}
\end{table}

\begin{figure*}[t!]
\centering
\includegraphics[width=0.99\linewidth,height=\textheight,keepaspectratio]{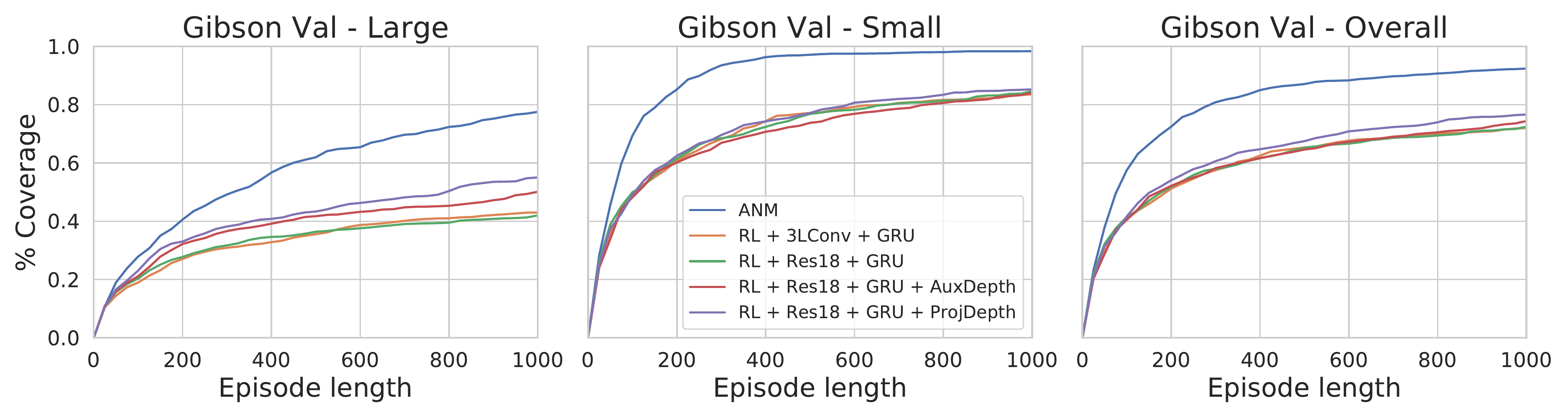}
\caption{\small Plot showing the $\%$ Coverage as the episode progresses for ANS and the baselines on the large and small scenes in the Gibson Val set as well as the overall Gibson Val set.}
\label{fig_ans:plot_ratio}
\end{figure*}

\section{Results}
\begin{figure*}[ht!]
\includegraphics[width=0.99\linewidth,height=\textheight,keepaspectratio]{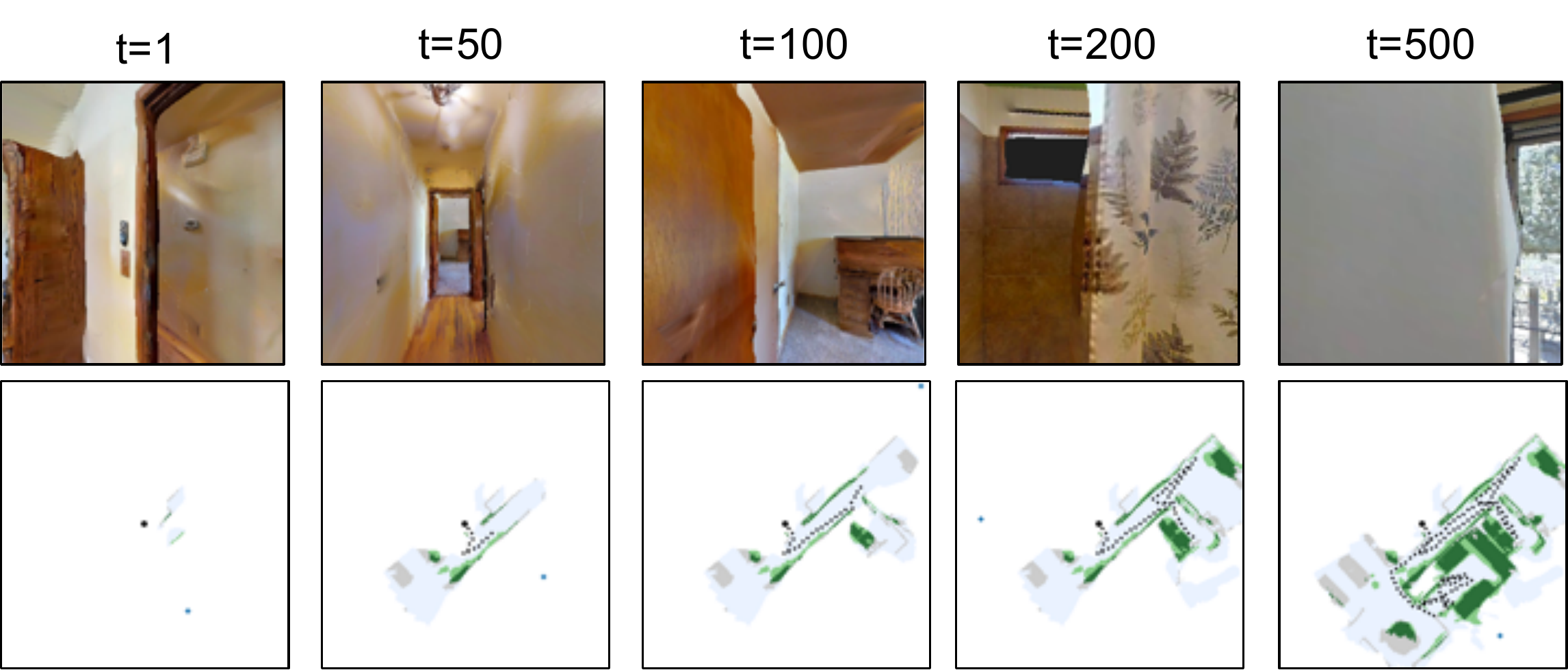}
\caption{\small \textbf{Exploration visualization}. Figure showing sample trajectories of the proposed model along with predicted map in the Exploration task as the episode progresses. The starting location is shown in black square. Long-term goals selected by the Global policy are shown by blue circles. Ground truth map is under-laid in grey. Map prediction is overlaid in green, with dark green denoting correct predictions and light green denoting false positives. Blue shaded region shows the explored area prediction.}
\label{fig_ans:exp_traj}
\end{figure*}

We train the proposed ANS model and all the baselines for the Exploration task with 10 million frames on the Gibson training set. The results are shown in Table~\ref{tab:exp_results}. The results on the Gibson Val set are averaged over a total of 994 episodes in 14 different unseen scenes. The proposed model achieves an average absolute and relative coverage of $31.379m^2 / 0.924$ as compared to $24.958m^2 / 0.766$ for the best baseline. This indicates that the proposed model is more efficient and effective at exhaustive exploration as compared to the baselines. This is because our hierarchical policy architecture reduces the horizon of the long-term exploration problem as instead of taking tens of low-level navigational actions, the Global policy only takes few long-term goal actions. We also report the domain generalization performance on the Exploration task in Table~\ref{tab:exp_results}~(see shaded region), where all models trained on Gibson are evaluated on the Matterport domain. ANS leads to higher domain generalization performance ($57.228m^2 / 0.405$ vs $41.549m^2 / 0.297$). The abosulte coverage is higher for the Matterport domain as it consists of larger scenes on average. Some visualizations of policy execution are provided in Figure~\ref{fig_ans:exp_traj}\footnote{See {\scriptsize \url{https://devendrachaplot.github.io/projects/Neural-SLAM}} for visualization videos.}.

In Fig.~\ref{fig_ans:plot_ratio}, we plot the relative coverage (\% Cov) of all the models as the episode progresses on the large and small scene sets, as well as the overall Gibson Val set. The plot on the small scene set shows that ANS is able to almost completely explore the small scenes in around 500 steps, however the baselines are only able to explore 85\% of the small scenes in 1000 steps (see Fig.~\ref{fig_ans:plot_ratio} center). This indicates that ANS explores more efficiently in small scenes. The plot on the large scenes set shows that the performance gap between ANS and baselines widens as the episode progresses (see Fig.~\ref{fig_ans:plot_ratio} left). Looking at the behaviour of the baselines, we saw that they often got stuck in local areas. This behaviour indicates that they are unable to remember explored areas over long-time horizons and are ineffective at long-term planning. On the other hand, ANS uses a Global policy on the map which allows it to have memory of explored areas over long-time horizons, and plan effectively to reach distant long-term goals by leveraging analytical planners. As a result, it is able to explore effectively in large scenes with long episode lengths.

\setlength{\tabcolsep}{5.9pt}
\begin{table}[t]
	\centering
\caption{\small Results of the ablation experiments on the Gibson environment.}
\label{tab:ablation}
\resizebox{\textwidth}{!}{
\begin{tabular}{@{}llcccccccc@{}}
\toprule
                        &           & \multicolumn{2}{c}{Gibson Val Overall} &           & \multicolumn{2}{c}{Gibson Val Large} &           & \multicolumn{2}{c}{Gibson Val Small} \\
Method                  &           & \% Cov.           & Cov. (m2)          &           & \% Cov.          & Cov. (m2)         &           & \% Cov.          & Cov. (m2)         \\ \midrule
ANS w/o Local Policy    &           & 0.882             & 29.673             &           & 0.713            & 45.810            &           & 0.950            & 23.219            \\
ANS w/o Global Policy   &           & 0.879             & 29.242             &           & 0.678            & 44.443            &           & 0.960            & 23.161            \\
ANS w/o Pose Estimation &           & 0.889             & 29.964             &           & 0.724            & 47.585            &           & 0.955            & 22.916            \\
\textbf{ANS}            & \textbf{} & \textbf{0.924}    & \textbf{31.379}    & \textbf{} & \textbf{0.776}   & \textbf{50.082}   & \textbf{} & \textbf{0.984}   & \textbf{23.898}   \\ \bottomrule
\end{tabular}	}
\end{table}

\subsection{Ablations}

\textbf{Local Policy.} An alternative to learning a Local Policy is to have a deterministic policy which follows the plan given by the Planner. As shown in Table~\ref{tab:ablation}, the ANS model performs much worse without the Local Policy. The Local Policy is designed to adapt to small errors in Mapping. We observed Local policy overcoming both false positives and false negatives encountered in mapping. For example, the Neural SLAM module could sometime wrongly predict a carpet as an obstacle. In this case, the planner would plan to go around the carpet. However, if the short-term goal is beyond the carpet, the Local policy can understand that the carpet is not an obstacle based on the RGB observation and learn to walk over it. Similarly, we also observed cases where the Neural SLAM module didn't predict small obstacles very close to the agent as they were not in the field-of-view due to the height of the camera. In this case, the planner would plan a path through the obstacle where the deterministic policy would get stuck. Since the local policy is recurrent, it learns to navigate around these obstacles by getting feedback from the environment. When the policy tries to move forward but it can not, it gets feedback that there must be an obstacle.

\textbf{Global Policy.} An alternative to learning a Global Policy for sampling long-term goals is to use a classical algorithm called Frontier-based exploration~\citep{yamauchi1997frontier}. A frontier is defined as the boundary between the explored free space and the unexplored space. Frontier-based exploration essentially sample points on this frontier as goals to explore the space. There are different variants of Frontier-based exploration based on the sampling strategy. \citet{holz2010evaluating} compare different sampling strategies and find that sampling the point on the frontier closest to the agent gives the best results empirically. We implement this variant and replace it with our learned Global Policy. As shown in Table~\ref{tab:ablation}, Frontier-based exploration policy perform worse than the Global Policy. We observed that Frontier-based exploration spent a lot of time exploring corners or small area behind furniture. In contrast, the trained Global policy ignored small spaces and chose distant long-term goals which led to exploring more area.

\textbf{Pose Estimation.} A difference between ANS and the baselines is that ANS uses additional supervision to train the Pose Estimator. In order to understand whether the performance gain is coming from this additional supervision, we remove the Pose Estimator from ANS and just use the input sensor reading as our pose estimate. Results in Table~\ref{tab:ablation} show that the ANS still outperforms the baselines even without the Pose Estimator. Furthermore, passing the ground truth pose as input the baselines instead of the sensor reading did not improve the performance of the baselines.

\begin{figure*}[t!]
\centering
\includegraphics[width=0.99\linewidth,height=\textheight,keepaspectratio]{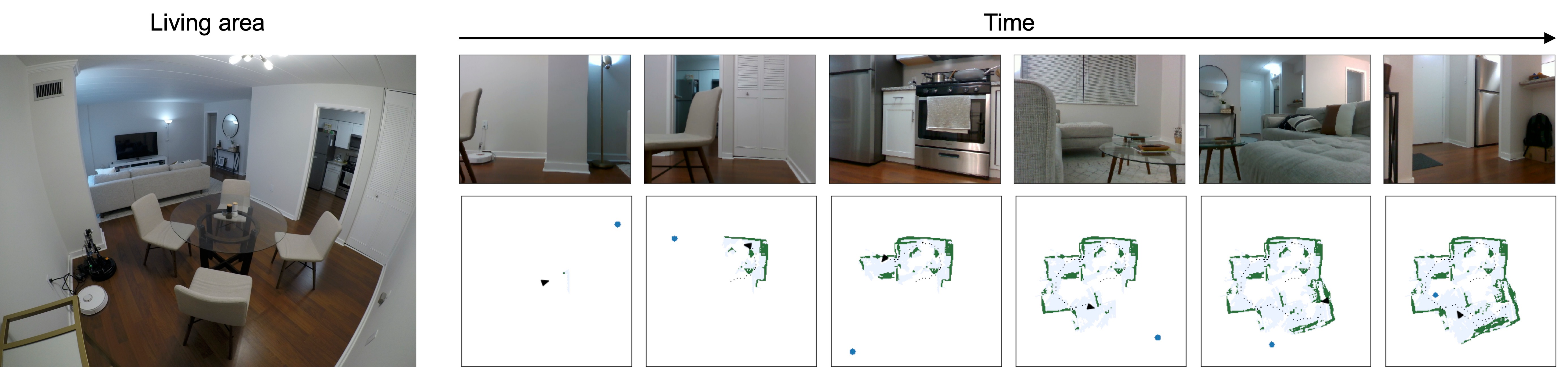}\\
\vspace{5pt}
\includegraphics[width=0.99\linewidth,height=\textheight,keepaspectratio]{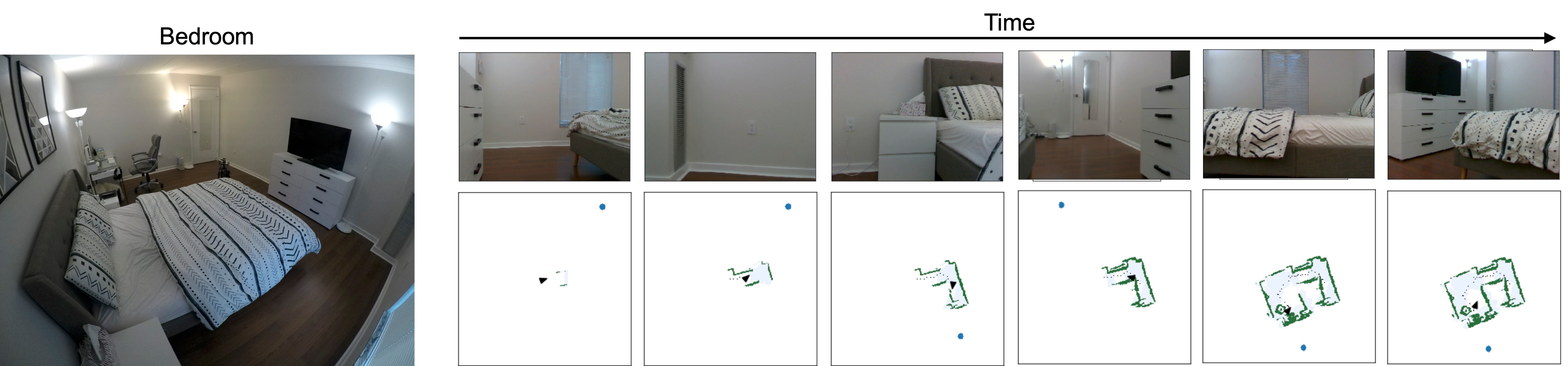}

\caption{\small \textbf{Real-world Transfer.} \textbf{Left:} Image showing the living area in an apartment used for the real-world experiments. \textbf{Right:} Sample images seen by the robot and the predicted map. The long-term goal selected by the Global Policy is shown by a blue circle on the map.}
\label{fig_ans:real_world_living}
\vspace{-10pt}
\end{figure*}

\subsection{Real-world transfer}
We deploy the trained ANS policy on a Locobot in the real-world. In order to match the real-world observations to the simulator observations as closely as possible, we change the simulator input configuration to match the camera intrinsics on the Locobot. This includes the camera height and horizontal and vertical field-of-views. In Figure~\ref{fig_ans:real_world_living}, we show an episode of ANS exploring the living area in an apartment. The figure shows that the policy transfers well to the real-world and is able to effectively explore the environment. The long-term goals sampled by the Global policy (shown by blue circles on the map) are often towards open spaces in the explored map, which indicates that it is learning to exploit the structure in the map. Please refer to the project \href{https://devendrachaplot.github.io/projects/Neural-SLAM}{webpage} for real-world videos.

\setlength{\tabcolsep}{3.8pt}
\begin{table}[t]
\centering
	\caption{\small Performance of the proposed model, Active Neural SLAM (ANS) and all the baselines on the Exploration task. `ANS - Task Transfer' refers to the ANS model transferred to the PointGoal task after training on the Exploration task.}
\label{tab:pg_results}
\resizebox{\textwidth}{!}{
\begin{tabular}{@{}lllccc
>{\columncolor[HTML]{DEDEDE}}c 
>{\columncolor[HTML]{DEDEDE}}c c
>{\columncolor[HTML]{DEDEDE}}c 
>{\columncolor[HTML]{DEDEDE}}c 
>{\columncolor[HTML]{DEDEDE}}c 
>{\columncolor[HTML]{DEDEDE}}c 
>{\columncolor[HTML]{DEDEDE}}c l@{}}
\toprule
            &                                                                        &  &                           &                           &                      & \multicolumn{2}{c}{\cellcolor[HTML]{DEDEDE}\begin{tabular}[c]{@{}c@{}}Domain \\ Generalization\end{tabular}} &                      & \multicolumn{5}{c}{\cellcolor[HTML]{DEDEDE}\begin{tabular}[c]{@{}c@{}}Goal \\ Generalization\end{tabular}}                                                                                                                                                   &  \\ \midrule
            & Test Setting $\rightarrow$                                           &  & \multicolumn{2}{c}{Gibson Val}                        &                      & \multicolumn{2}{c}{\cellcolor[HTML]{DEDEDE}MP3D Test}                                                        &                      & \multicolumn{2}{c}{\cellcolor[HTML]{DEDEDE}Hard-GEDR}                                                 &                                              & \multicolumn{2}{c}{\cellcolor[HTML]{DEDEDE}Hard-Dist}                                                 &  \\ \midrule
Train Task  & Method                                                                 &  & Succ                      & SPL                       &                      & Succ                                                  & SPL                                                  &                      & Succ                                              & SPL                                               &                                              & Succ                                              & SPL                                               &  \\ \midrule
PointGoal   & Random                                                                 &  & 0.027                     & 0.021                     &                      & 0.010                                                 & 0.010                                                &                      & 0.000                                             & 0.000                                             &                                              & 0.000                                             & 0.000                                             &  \\
            & RL + Blind                                                             &  & 0.625                     & 0.421                     &                      & 0.136                                                 & 0.087                                                &                      & 0.052                                             & 0.020                                             &                                              & 0.008                                             & 0.006                                             &  \\
            & RL + 3LConv + GRU                                                      &  & 0.550                     & 0.406                     &                      & 0.102                                                 & 0.080                                                &                      & 0.072                                             & 0.046                                             &                                              & 0.006                                             & 0.006                                             &  \\
            & RL + Res18 + GRU                                                       &  & 0.561                     & 0.422                     &                      & 0.160                                                 & 0.125                                                &                      & 0.176                                             & 0.109                                             &                                              & 0.004                                             & 0.003                                             &  \\
            & \begin{tabular}[c]{@{}l@{}}RL + Res18 + GRU + AuxDepth\end{tabular} &  & 0.640                     & 0.461                     &                      & 0.189                                                 & 0.143                                                &                      & 0.277                                             & 0.197                                             &                                              & 0.013                                             & 0.011                                             &  \\
            & \begin{tabular}[c]{@{}l@{}}RL + Res18 + GRU + ProjDepth\end{tabular} &  & 0.614                     & 0.436                     &                      & 0.134                                                 & 0.111                                                &                      & 0.180                                             & 0.129                                             &                                              & 0.008                                             & 0.004                                             &  \\
            & IL + Res18 + GRU                                                       &  & 0.823                     & 0.725                     &                      & 0.365                                                 & 0.318                                                &                      & 0.682                                             & 0.558                                             &                                              & 0.359                                             & 0.310                                             &  \\
            & CMP                                                                    &  & 0.827                     & 0.730                     &                      & 0.320                                                 & 0.270                                                &                      & 0.670                                             & 0.553                                             &                                              & 0.369                                             & 0.318                                             &  \\
            & ANS                                                                    &  & \multicolumn{1}{c}{\textbf{0.951}} & \multicolumn{1}{c}{\textbf{0.848}} & \multicolumn{1}{c}{} & \multicolumn{1}{c}{\cellcolor[HTML]{DEDEDE}\textbf{0.593}}     & \multicolumn{1}{c}{\cellcolor[HTML]{DEDEDE}\textbf{0.496}}    & \multicolumn{1}{c}{} & \multicolumn{1}{c}{\cellcolor[HTML]{DEDEDE}\textbf{0.824}} & \multicolumn{1}{c}{\cellcolor[HTML]{DEDEDE}\textbf{0.710}} & \multicolumn{1}{c}{\cellcolor[HTML]{DEDEDE}} & \multicolumn{1}{c}{\cellcolor[HTML]{DEDEDE}0.662} & \multicolumn{1}{c}{\cellcolor[HTML]{DEDEDE}\textbf{0.534}} &  \\ \midrule
Exploration & \begin{tabular}[c]{@{}l@{}}ANS - Task Transfer\end{tabular}                                                    &  & \multicolumn{1}{c}{0.950} & \multicolumn{1}{c}{0.846} & \multicolumn{1}{c}{} & \multicolumn{1}{c}{\cellcolor[HTML]{DEDEDE}0.588}     & \multicolumn{1}{c}{\cellcolor[HTML]{DEDEDE}0.490}    & \multicolumn{1}{c}{} & \multicolumn{1}{c}{\cellcolor[HTML]{DEDEDE}0.821} & \multicolumn{1}{c}{\cellcolor[HTML]{DEDEDE}0.703} & \multicolumn{1}{c}{\cellcolor[HTML]{DEDEDE}} & \multicolumn{1}{c}{\cellcolor[HTML]{DEDEDE}\textbf{0.665}} & \multicolumn{1}{c}{\cellcolor[HTML]{DEDEDE}0.532} &  \\ \bottomrule
\end{tabular}}   

\end{table}

\subsection{Pointgoal Task Transfer}
PointGoal has been the most studied task in recent literature on navigation where the objective is to navigate to a goal location whose relative coordinates are given as input in a limited time budget. We follow the PointGoal task setup from \citet{savva2019habitat}, using train/val/test splits for both Gibson and Matterport datasets. Note that the set of scenes used in each split is disjoint, which means the agent is tested on new scenes never seen during training. Gibson test set is not public but rather held out on an online evaluation server\footnote{\url{https://evalai.cloudcv.org/web/challenges/challenge-page/254}}. We report the performance of our model on the Gibson test set when submitted to the online server but also use the validation set as another test set for extensive comparison and analysis. We do not use the validation set for hyper-parameter tuning. 

\citet{savva2019habitat} identify two measures to quantify the difficulty of a PointGoal dataset. The first is the average geodesic distance (distance along the shortest path) to the goal location from the starting location of the agent, and the second is the average geodesic to Euclidean distance ratio (GED ratio). The GED ratio is always greater than or equal to 1, with higher ratio resulting in harder episodes. The train/val/test splits in Gibson dataset come from the same distribution of having similar average geodesic distance and GED ratio. In order to analyze the performance of the proposed model on out-of-set goal distribution, we create two harder sets, Hard-Dist and Hard-GEDR. In the Hard-Dist set, the geodesic distance to goal is always more than $10\text{m}$ and the average geodesic distance to the goal is $13.48\text{m}$ as compared to $6.9/6.5/7.0\text{m}$ in train/val/test splits~\citep{savva2019habitat}. Hard-GEDR set consists of episodes with an average GED ratio of 2.52 and a minimum GED ratio of 2.0 as compared to average GED ratio 1.37 in the Gibson val set.

We also follow the episode specification from \citet{savva2019habitat}. Each episode ends when either the agent takes the \texttt{stop} action or at a maximum of 500 timesteps. An episode is considered a success when the final position of the agent is within $0.2\text{m}$ of the goal location. In addition to Success rate (Succ), we also use Success weighted by (normalized inverse) Path Length or SPL as a metric for evaluation for the PointGoal task as proposed by \citet{anderson2018evaluation}.

All the baseline models trained for the task of Exploration either need to be retrained or atleast fine-tuned to be transferred to the Pointgoal task. The modularity of ANS provides it another advantage that it can be transferred to the Pointgoal task without any additional training. For transfer to the Pointgoal task, we just fix the Global policy to always output the PointGoal coordinates as the long-term goal and use the Local and Mapper trained for the Exploration task. 

\subsubsection{Pointgoal Results}
In Table~\ref{tab:pg_results}, we show the performance of the proposed model transferred to the PointGoal task along with the baselines trained on the PointGoal task with the same amount of data (10million frames). The proposed model achieves a success rate/SPL of 0.950/0.846 as compared to 
0.827/0.730 
for the best baseline model on Gibson val set. We also report the performance of the proposed model trained from scratch on the PointGoal task for 10 million frames. The results indicate that the performance of ANS transferred from Exploration is comparable to ANS trained on PointGoal. This highlights a key advantage of our model that it allows us to transfer the knowledge of obstacle avoidance and control in low-level navigation across tasks, as the Local Policy and Mapper are task-invariant.\\

\noindent\textbf{Sample efficiency.} RL models are typically trained for more than 10 million samples. In order to compare the performance and sample-efficiency, we trained the best performing RL model (RL + Res18 + GRU + ProjDepth) for 75 million frames and it achieved a Succ/SPL of 0.678/0.486. 
ANS reaches the performance of 0.789/0.703 SPL/Succ at only 1 million frames. These numbers indicate that ANS achieves $>75\times$ speedup as compared to the best RL baseline. \\

\begin{wrapfigure}{r}{0.45\textwidth}
\centering
\includegraphics[width=0.45\textwidth]{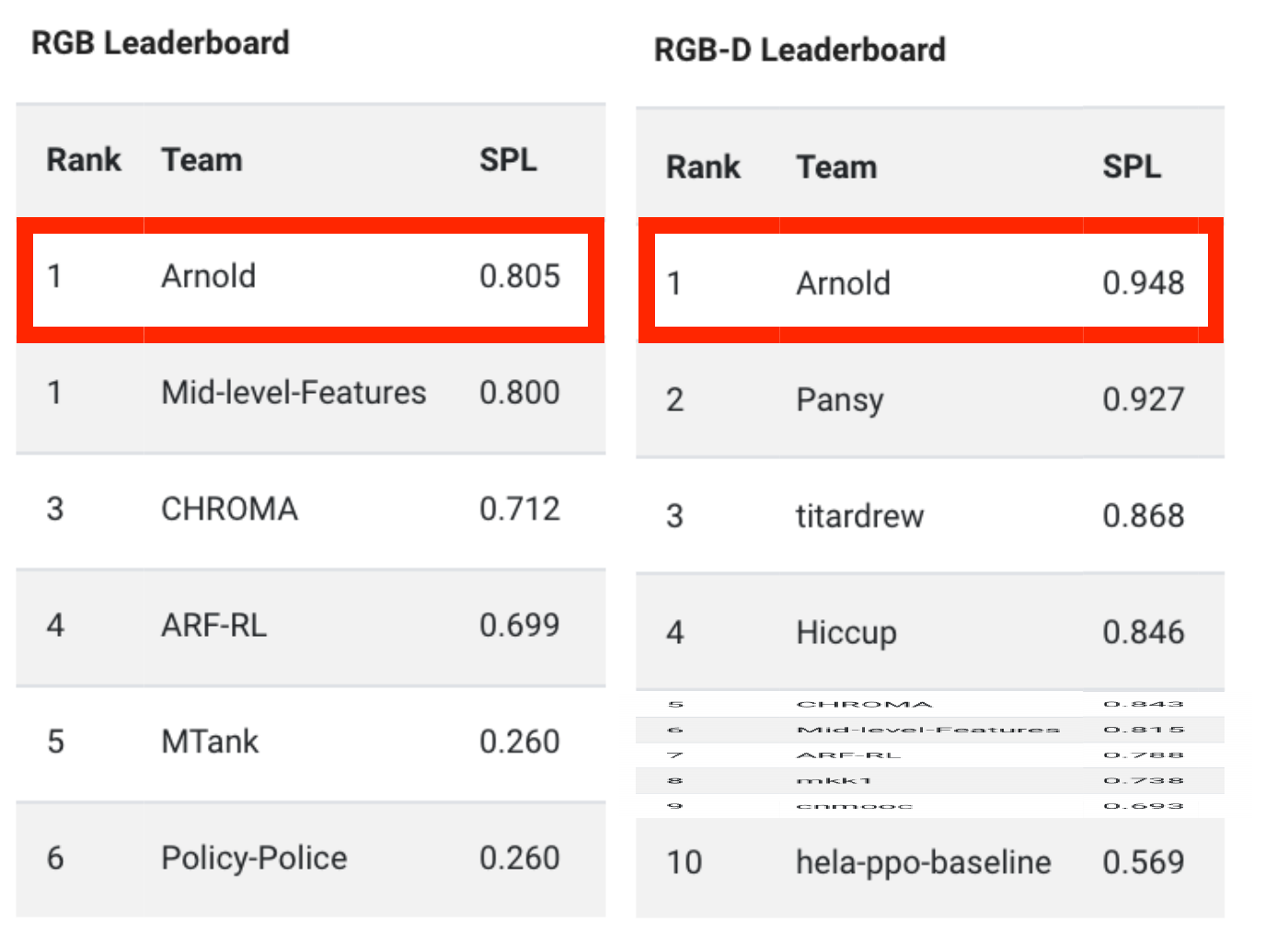}
\caption{\small Screenshot of CVPR 2019 Habitat Challenge Results. The proposed model was submitted under code-name `Arnold'.}
\vspace{-15pt}
\label{fig_ans:challenge}
\end{wrapfigure}

\begin{figure*}
\centering
\begin{minipage}{0.24\textwidth}
\includegraphics[width=\linewidth,height=\textheight,keepaspectratio]{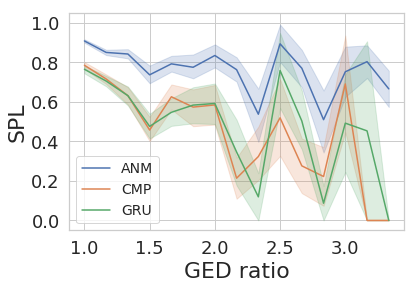}
\end{minipage}
\begin{minipage}{0.24\textwidth}
\centering
\includegraphics[width=\linewidth,height=\textheight,keepaspectratio]{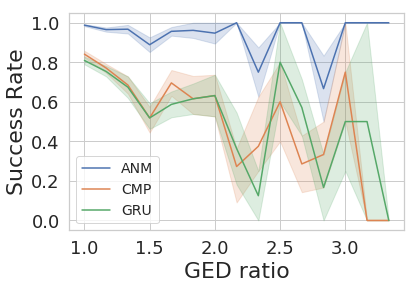}
\end{minipage}
\begin{minipage}{0.24\textwidth}
\centering
\includegraphics[width=\linewidth,height=\textheight,keepaspectratio]{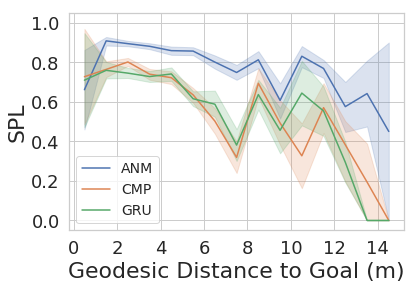}
\end{minipage}
\begin{minipage}{0.24\textwidth}
\centering
\includegraphics[width=\linewidth,height=\textheight,keepaspectratio]{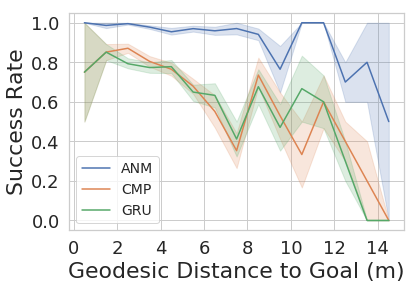}
\end{minipage}
\caption{\small Performance of the proposed ANS model along with CMP and IL + Res18 + GRU (GRU) baselines with increase in geodesic distance to goal and increase in GED Ratio on the Gibson Val set.}
\label{fig_ans:analysis}
\vspace{-5pt}
\end{figure*}

\begin{figure*}
\vspace{-5pt}
\centering
\includegraphics[width=0.99\linewidth,height=\textheight,keepaspectratio]{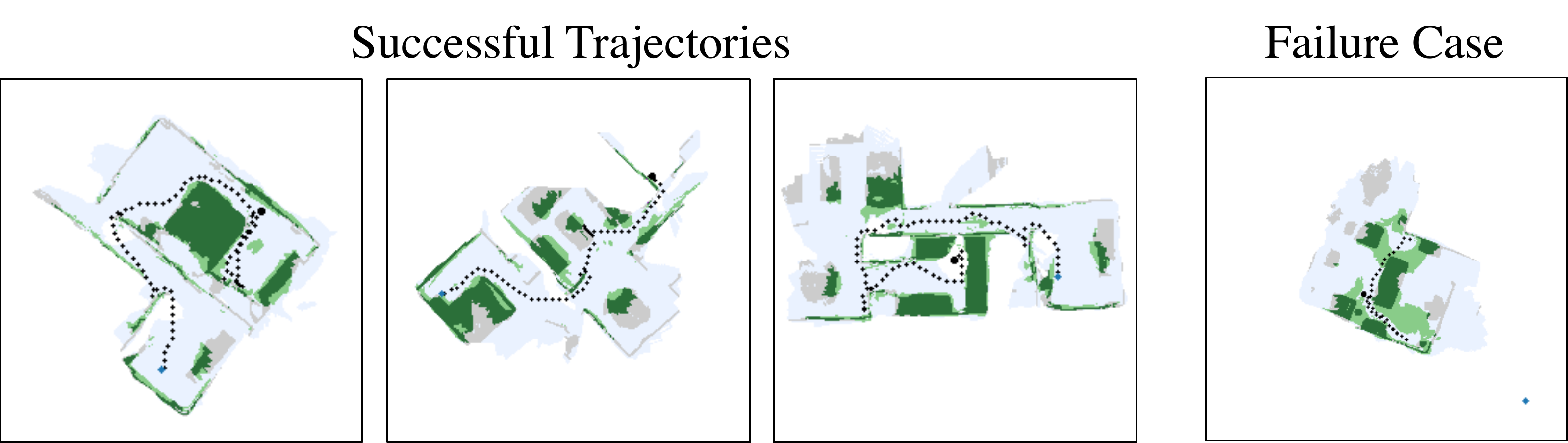}
\caption{\small Figure showing sample trajectories of the proposed model along with predicted map in the PointGoal task. The starting and goal locations are shown by black squares and blue circles, respectively. Ground truth map is under-laid in grey. Map prediction is overlaid in green, with dark green denoting correct predictions and light green denoting false positives. Blue shaded region shows the explored area prediction. On the left, we show some successful trajectories which indicate that the model is effective at long distance goals with high GED ratio. On the right, we show a failure case due to mapping error.}
\label{fig_ans:traj}
\vspace{-10pt}
\end{figure*}

\begin{figure*}[ht!]
\centering
\includegraphics[width=0.99\linewidth,height=\textheight,keepaspectratio]{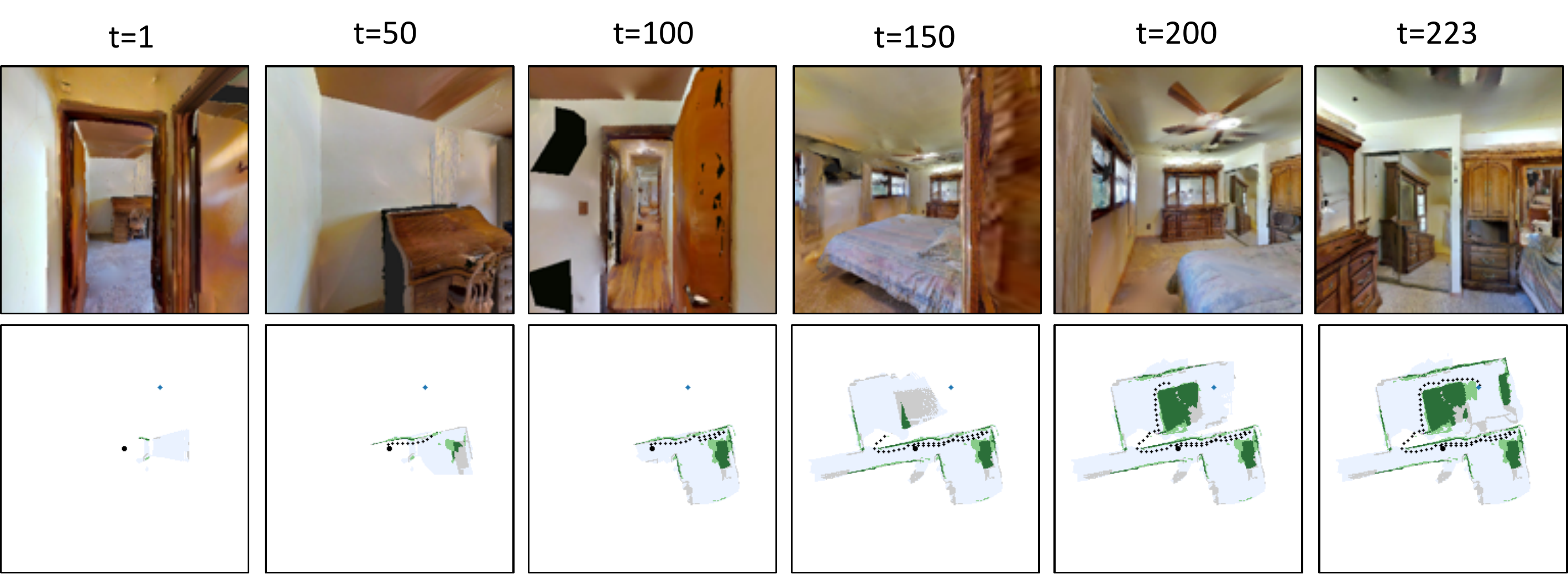}
\caption{\small \textbf{Pointgoal visualization}. Figure showing sample trajectories of the proposed model along with predicted map in the Pointgoal task as the episode progresses. The starting and goal locations are shown by black squares and blue circles, respectively. Ground truth map is under-laid in grey. Map prediction is overlaid in green, with dark green denoting correct predictions and light green denoting false positives. Blue region shows the explored area prediction.}
\label{fig_ans:pg_traj}
\vspace{-10pt}
\end{figure*}

\noindent\textbf{Domain and Goal Generalization:} In Table~\ref{tab:pg_results}~(see shaded region), we evaluate all the baselines and ANS trained on the PointGoal task in the Gibson domain on the test set in Matterport domain as well as the harder goal sets in Gibson. We also transfer ANS trained on Exploration in Gibson on all the 3 sets. The results show that ANS outperforms all the baselines at all generalization sets. Interestingly, RL based methods almost fail completely on the Hard-Dist set. We also analyze the performance of the proposed model as compared to two best baselines CMP and IL + Res18 + GRU as a function of geodesic distance to goal and GED ratio in Figure~\ref{fig_ans:analysis}. The performance of the baselines drops faster as compared to ANS, especially with increase in goal distance. This indicates that end-to-end learning methods are effective at short-term navigation but struggle when long-term planning is required to reach a distant goal. In Figure~\ref{fig_ans:traj}, we show some example trajectories of the ANS model along with the predicted map. The successful trajectories indicate that the model exhibits strong backtracking behavior which makes it effective at distant goals requiring long-term planning. Figure~\ref{fig_ans:pg_traj} visualizes a trajectory in the PointGoal task show first-person observation and corresponding map predictions. Please refer to the project \href{https://devendrachaplot.github.io/projects/Neural-SLAM}{webpage} for visualization videos.\\

\noindent\textbf{Habitat Challenge Results.} We submitted the ANS model to the CVPR 2019 Habitat Pointgoal Navigation Challenge. The results are shown in Figure~\ref{fig_ans:challenge}. ANS was submitted under code-name `Arnold'. ANS was the winning entry for both RGB and RGB-D tracks among over 150 submissions from 16 teams, achieving an SPL of 0.805 (RGB) and 0.948 (RGB-D) on the Test Challenge set.

\section{Discussion}
In this chapter, we proposed a modular navigational model which leverages the strengths of classical and learning-based navigational methods. We show that the proposed model outperforms prior methods on both Exploration and PointGoal tasks and shows strong generalization across domains, goals, and tasks. 

Active Neural SLAM is designed for tasks such as Exploration and PointGoal, which primarily require only spatial understanding. Obstacle-based metric maps used in Active Neural SLAM are not suitable for semantic understanding. In the next chapter, we will discuss methods for tackling navigation tasks requiring semantic understanding such as long-term Image Goal navigation using topological maps.  

\chapter{Neural Topological SLAM}
\label{chap:nts}
\def\methodname{Neural Topological SLAM}
\def\methodabbr{NTS}
\blfootnote{Webpage: \url{https://devendrachaplot.github.io/projects/neural-topological-slam}}

\begin{wrapfigure}{r}{0.5\linewidth}
\centering
\vspace{-12pt}
\includegraphics[width=\linewidth,height=\textheight,keepaspectratio]{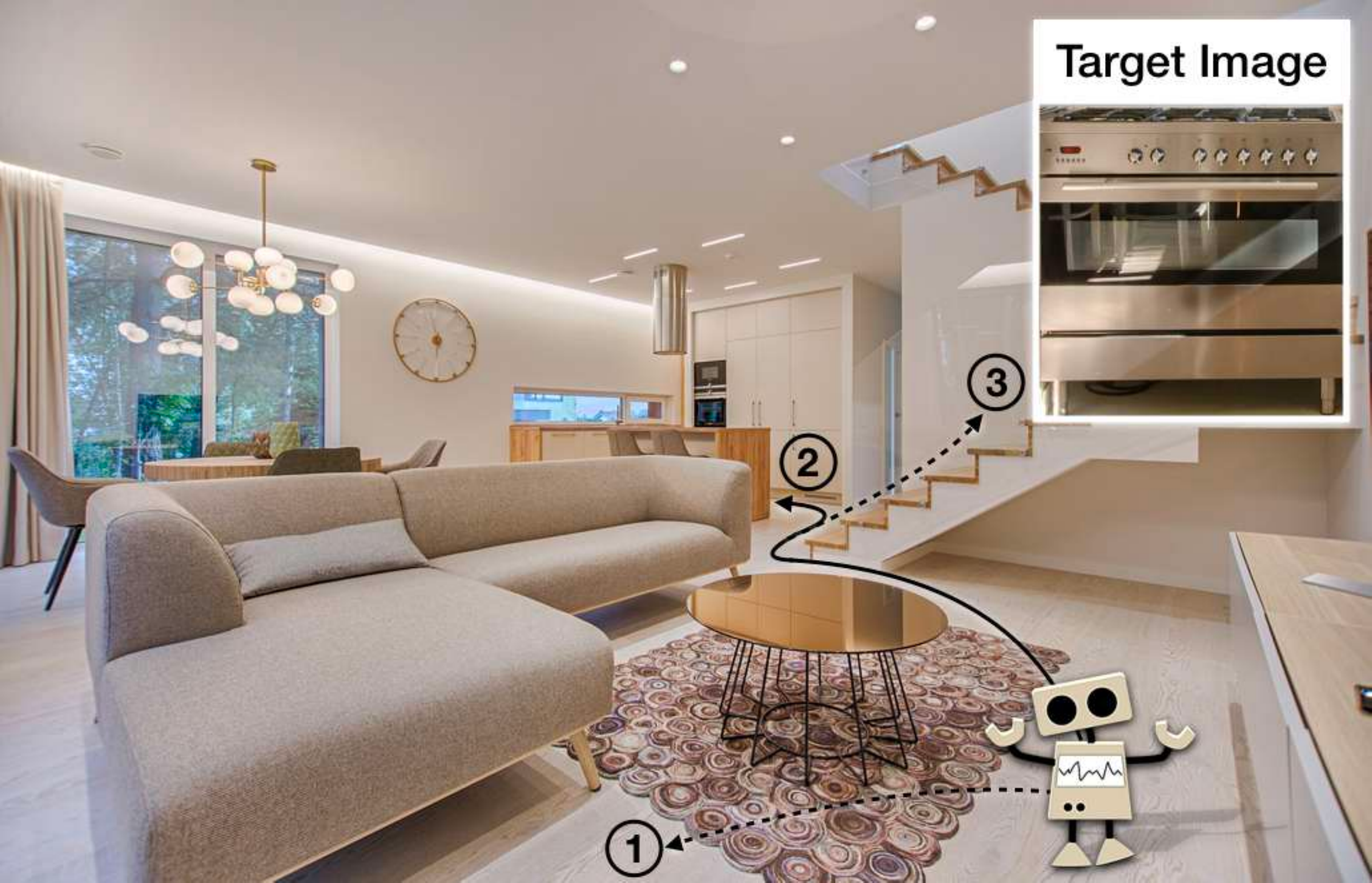}
\vspace{-12pt}
\caption{\small \textbf{Semantic Priors and Landmarks.} When asked to go to target image of an oven most humans would use the path number 2 since it allows access to kitchen. Humans use semantic priors and common-sense to explore and navigate everyday yet most navigation algorithms struggle to do so.}
\label{fig_nts:teaser}
\vspace{-5pt}
\end{wrapfigure}

Imagine you are in a new house as shown in Fig~\ref{fig_nts:teaser} and you are given the task of finding a target object as shown in Fig~\ref{fig_nts:teaser} (top). While there are multiple possible directions to move, most of us would choose the path number 2 to move. This is because we use strong structural priors -- we realize the target is an oven which is more likely to be found in the kitchen which seems accessible via path number 2. Now let us suppose, once you reach the oven, your goal is to reach back to the living room which you saw initially. How would you navigate?  The answer to this question lies in how we humans store maps (or layout) of the house we just traversed. One possible answer would be metric maps, in which case we would know exactly how many steps to take to reach the living room. But this is clearly not how we humans operate~\cite{gillner1998navigation,wang2002human}. Instead, most of us would first get out of the kitchen by moving to the hallway and then navigate to the living room which is visible from the hallway.

It is clear from the above examples, there are two main components of a successful visual navigation algorithm: (a) ability to build spatial representations and store them; (b) ability to exploit structural priors. When it comes to spatial representations, the majority of papers in navigation insist on building metrically precise representations of free space. However, metric maps have two major shortcomings: first, metric maps do not scale well with environment size and amount of experience. But more importantly, actuation noise on real-robots makes it challenging to build consistent representations, and precise localization may not always be possible. When it comes to exploiting structural priors, most learning-based approaches do not model these explicitly. Instead, they hope the learned policy function has these priors encoded implicitly. But it still remains unclear if these policy functions can encode semantic priors when learned via RL.

In this chapter, we propose to tackle both the problems head-on. Instead of using metric-maps which are brittle to localization and noise, we propose a topological representation of the space. Our proposed representation consists of nodes that are connected in the form of a graph, based on local geometry information. Each node is represented visually via a 360-degree panoramic image. Nodes are connected to each other using approximate relative pose between them. But what makes our visual topological maps novel are two directional functions~$\mathcal{F}_g$ and $\mathcal{F}_s$, which extract geometric and semantic properties of nodes. Specifically, $\mathcal{F}_g$ estimates how likely agent will encounter free-space and $\mathcal{F}_s$ estimates how likely target image is to be encountered if the agent moves in a particular direction. By explicitly modeling and learning function $\mathcal{F}_s$, our model ensures that structural priors are encoded and used when exploring and navigating new unseen environments. 

Our representation has few advantages over classical and end-to-end learning-based approaches:  (a) it uses graph-based representation which allows efficient long-term planning; (b) it explicitly encodes structural priors via function $\mathcal{F}_s$; (c) the geometric function $\mathcal{F}_g$ allows efficient exploration and online map building for a new environment; (d) but most importantly, all the functions and policies can be learned in completely supervised manner forgoing the need for unreliable credit assignment via RL.

This work makes contributions across the following multiple aspects of the navigation problem: space representation, training paradigm for navigation policies, and different navigation tasks. We survey works in these areas below.\\\vspace{-6pt}

\noindent \textbf{Navigation Tasks.} 
Navigation tasks can be divided into two main categories. The first category of tasks is ones where the goal location is known, and limited exploration is necessary. This could be in the form of a simply wandering around without colliding~\cite{gandhi2017learning,sadeghi2016cad2rl}, following an object~\cite{lee2017learning}, getting to a goal coordinate~\cite{gupta2017cognitive,anderson2018evaluation}: using sequence of nts/images along the path~\cite{kumar2018visual,bruce2018deployable}, or language-based instructions~\cite{anderson2018vision}. Sometimes, the goal is specified as an image but experience from the environment is available in the form of demonstrations~\cite{eysenbach2019search,savinov2018semi}, or in the form of reward-based training~\cite{mirowski2018learning,zhu2017target}, which again limits the role of exploration. The second category of tasks is when the goal is not known and exploration is necessary. Examples are tasks such as finding an object~\cite{gupta2017cognitive}, or room~\cite{wu2019bayesian}, in a novel environment, or explicit exploration~\cite{chen2018learning,ans}. These task categories involve different challenges. The former tasks focus on effective retrieval and robust execution, while the later tasks involve semantic and common sense reasoning in order to efficiently operate in previously unseen environments. Our focus, in this work, is the task of reaching a target image in a novel environment. No experience is available from the environment, except for the target image. We aren't aware of any works that target this specific problem.\\\vspace{-6pt}

\noindent \textbf{Classical Space Representations.}
Spatial and topological representations have a rich history in robot navigation. Researchers have used explicit metric spatial representations~\cite{elfes1989using}, and have considered how can such representations be built with different sensors~\cite{newcombe2011kinectfusion, thrun2005probabilistic, newcombe2011dtam, mur2015orb, hornung13auro}, and how can agents be localized against such representations~\cite{dellaert1999monte}. Recent work has started associating semantics with such spatial representations~\cite{bowman2017probabilistic}. In a similar vein, non-metric topological representations have also been considered in classical literature~\cite{kuipers1991robot, meng1993mobile, choset2001topological}. Some works combine topological and metric representations~\cite{tomatis2001combining, thrun1998integrating}, and some study topological representations that are semantic~\cite{kuipers1991robot}. While our work builds upon the existing literature on topological maps, the resemblance is only at high-level graph structure. Our work focuses on making visual topological mapping and exploration scalable, robust and efficient. We achieve this via the representation of both semantic and geometric properties in our topological maps; the ability to build topological maps in an online manner and finally, posing the learning problem as a supervised problem.\\\vspace{-6pt}

\noindent \textbf{Learned Space Representations.}
Depending on the problem being considered, different representations have been investigated. For short-range locomotion tasks, purely reactive policies~\cite{sadeghi2016cad2rl, gandhi2017learning, lee2017learning, bansal2019combining} suffice. For more complex problems such as target-driven navigation in a novel environment, such purely reactive strategies do not work well~\cite{zhu2017target}, and memory-based policies have been investigated. This can be in the form of vanilla neural network memories such as LSTMs~\cite{mirowski2016learning, oh2016control}, or transformers~\cite{fang2019smt}. Researchers have also incorporated insights from classical literature into the design of expressive neural memories for navigation. This includes spatial memories~\cite{gupta2017cognitive, parisotto2017neural} and topological approaches~\cite{chen2019behavioral, savinov2018semi, eysenbach2019search, wu2019bayesian, savinov2018episodic, yang2018visual}. 
Learned spatial approaches can acquire expressive spatial representations~\cite{gupta2017cognitive}, they are however bottle-necked by their reliance on metric consistency and thus have mostly been shown to work in discrete state spaces for comparatively short-horizon tasks~\cite{gupta2017cognitive, parisotto2017neural}. Researchers have also tackled the problem of passive and active localization~\cite{mur2015orb,chaplot2018active}, in order to aid building such consistent metric representations. 
Some topological approaches \cite{chen2019behavioral, savinov2018semi, eysenbach2019search} work with human explorations or pre-built topological maps, thus ignoring the problem of exploration. Others build a topological representation with explicit semantics~\cite{wu2019bayesian, zhang2018composable}, which limits tasks and environments that can be tackled. In contrast from past work, we unify spatial and topological representations in a way that is robust to actuation error, show how we can incrementally and autonomously build topological representations, and do semantic reasoning.\\
\vspace{-6pt}

\noindent \textbf{Training Methodology.}
Different tasks have also lead to the design of different training methodologies for training navigation policies. This ranges from reinforcement learning with sparse and shaped rewards~\cite{mirowski2018learning, zhu2017target, pathakICMl17curiosity, sadeghi2016cad2rl, mirowski2016learning}, imitation learning and DAgger~\cite{gupta2017cognitive, ross2011reduction}, self-supervised learning for individual components~\cite{savinov2018semi,gandhi2017learning}. While RL allows learning of rich exploratory behavior, training policies using RL is notoriously hard and sample inefficient. Imitation learning is sample efficient but may not allow learning exploratory behavior. Self-supervised learning is promising but has only been experimented in the context of known goal tasks. We employ a supervised learning approach and show how we can still learn expressive exploration behavior while at the same time not suffering from exuberant sample complexity for training.

\begin{figure*}
\centering
\begin{minipage}{0.99\textwidth}
\centering
\includegraphics[width=\linewidth,height=\textheight,keepaspectratio]{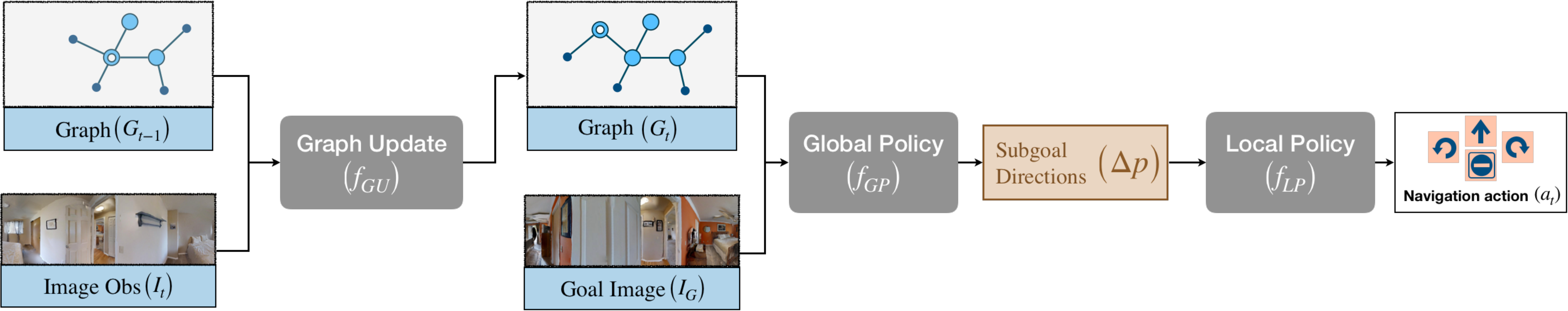}
\end{minipage}
\vspace{3pt}

    \caption{\small \textbf{Model Overview.} Figure showing an overview of the proposed model, \methodname. It consists of 3 components, a Graph Construction module which updates the topological map as it receives observations, a Global Policy which samples subgoals, and a Local Policy which takes navigational actions to reach the subgoal. See text for more details.}
\label{fig_nts:overview}
\end{figure*}

\section{Task Setup}
\noindent We consider an autonomous agent situated in an episodic environment. At the beginning of an episode, the agent receives a target goal image, $I_G$. At each time step $t$, the agent receives observations ($s_t$) from the environment. Each observation consists of the current first-person image observation, $I_t$, from a panoramic camera and a pose estimate from a noisy motion sensor. At each time step, the agent takes a navigational action $a_t$. The objective is to learn a policy $\pi(a_t| s_t, I_G)$ to reach the goal image. In our experimental setup, all nts/images are panoramas, including agent observations and goal image.

\section{Methods}
\noindent We propose a modular model, `\methodname~(\methodabbr)', which builds and maintains a topological map for navigation. The topological map is represented using a graph, denoted by $G_t$ at time $t$. Each node in the graph ($N_i$) is associated with a panoramic image ($I_{N_i}$) and represents the area visible in this image. Two nodes are connected by an edge ($E_{i,j}$) if they represent adjacent areas. Each edge also stores the relative pose between two nodes, $\Delta p_{ij}$.

Our model consists of three components, a \textit{Graph  Update} module, a \textit{Global Policy}, and a \textit{Local Policy}. On a high level, the Graph Update module updates the topological map based on agent observations, the Global Policy selects a node in the graph as the long-term goal and finds a subgoal to reach the goal using path planning, and the Local Policy navigates to the subgoal based on visual observations. Fig.~\ref{fig_nts:overview} provides an overview of the proposed model.

\begin{figure}[t]
\centering
\begin{minipage}{0.69\linewidth}
\centering
\includegraphics[width=\linewidth,height=\textheight,keepaspectratio]{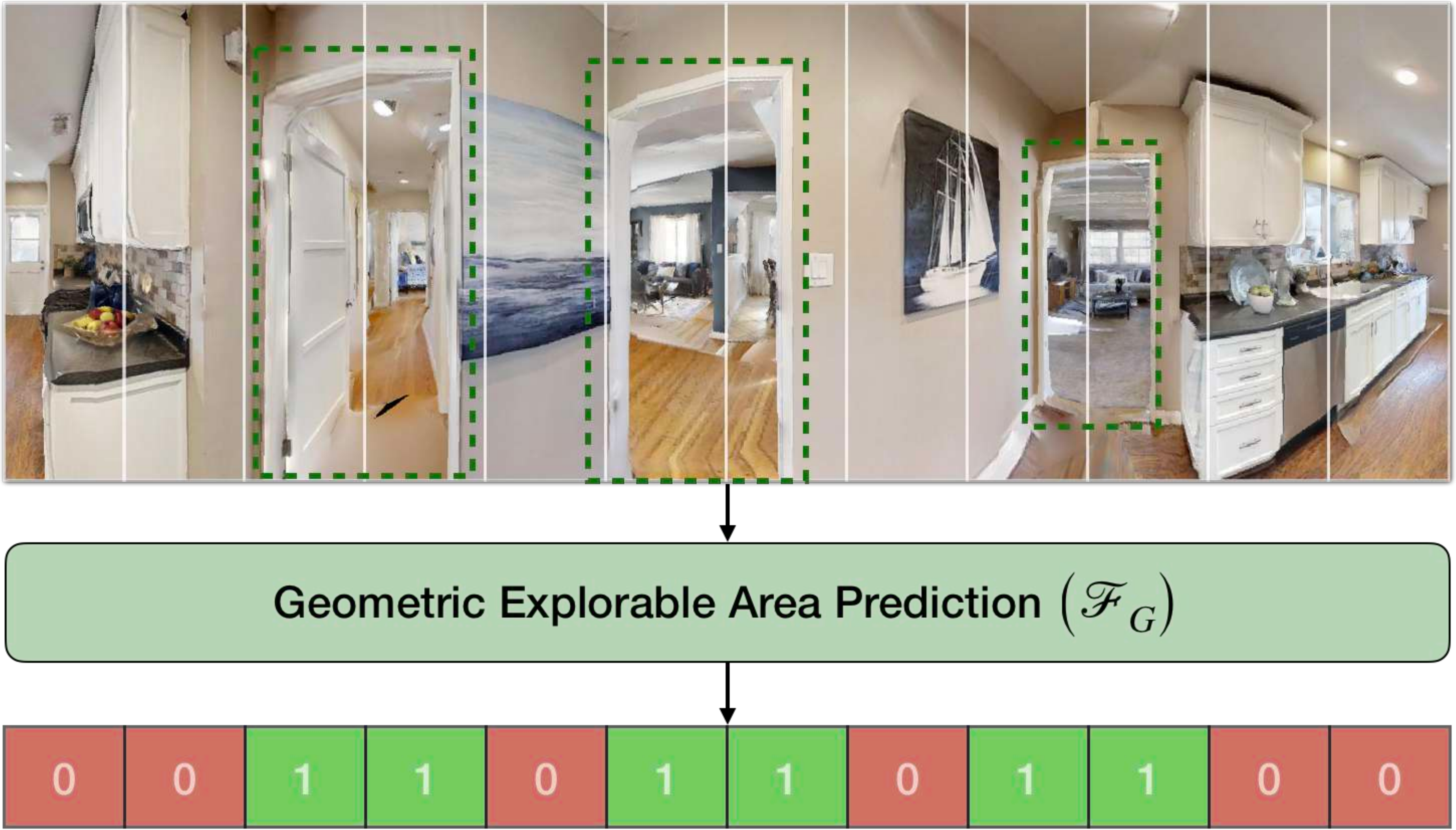}
\end{minipage}

    \caption{\small \textbf{Geometric Explorable Area Prediction ($\mathcal{F}_G$).} Figure showing sample input image ($I$) and output predictions of the Geometric Explorable Area Prediction function ($\mathcal{F}_G$). The green boxes show doorways for reference, and are not available as input.
    }
\label{fig_nts:f_g}
\end{figure}

\begin{figure}[t]
    \centering
\begin{minipage}{0.99\linewidth}
\centering
    \includegraphics[width=\linewidth,height=\textheight,keepaspectratio]{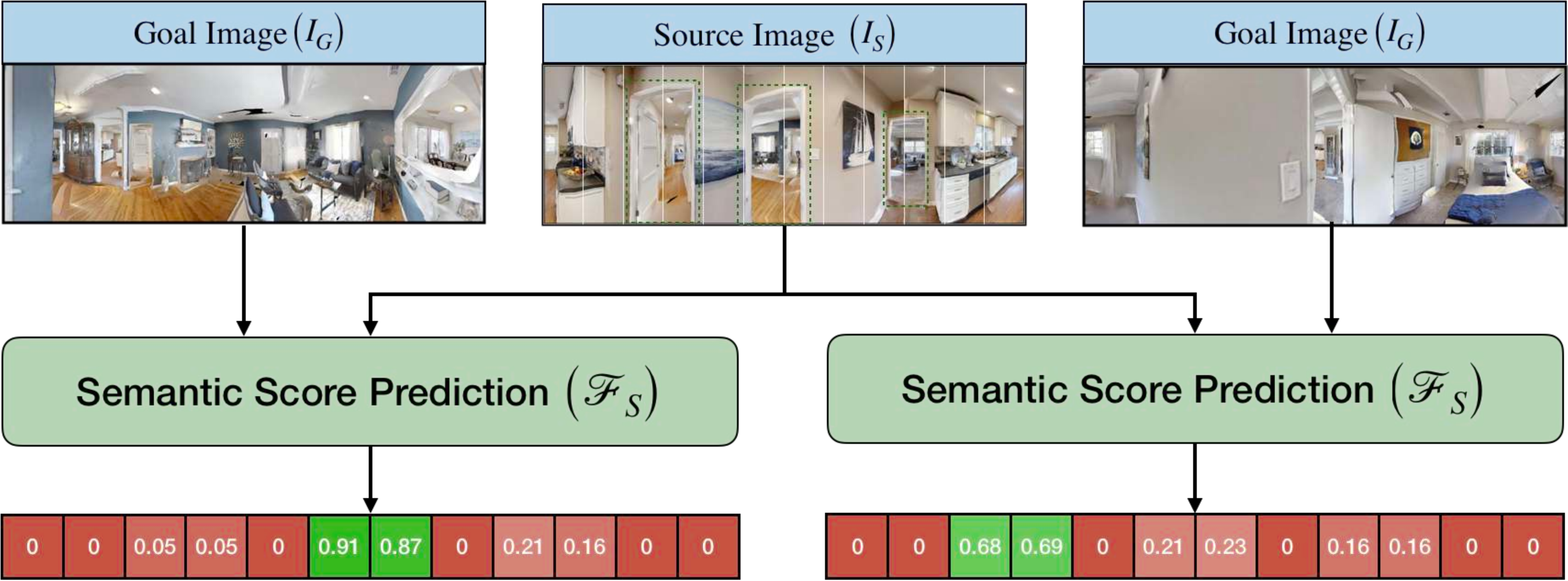}
\end{minipage}
\vspace{3pt}

    \caption{\small \textbf{Semantic Score Prediction ($\mathcal{F}_S$).} Figure showing sample input and output predictions of the Semantic Score Prediction function ($\mathcal{F}_S$). The score predictions change based on the goal image. When the goal image is of the living room (left), the score of the directions in the center are higher as they lead to the living room. When the goal image is of a bedroom (right), the scores corresponding to the pathway on the left are higher as they are more likely to lead to the bedroom.}
    \label{fig_nts:f_s}
\end{figure}

\begin{figure*}[t!]
\centering
\begin{minipage}{0.99\textwidth}
\centering
\includegraphics[width=\linewidth,height=\textheight,keepaspectratio]{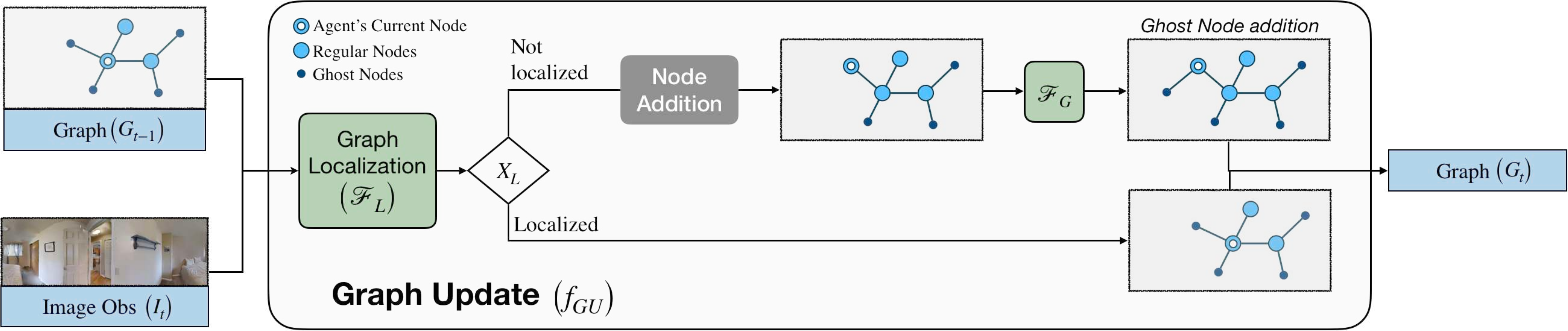}
\end{minipage}
\vspace{3pt}

    \caption{\small \textbf{Graph Update.} Figure showing an overview of the Graph Update Module. It takes the current Graph ($G_t$) and the agent observation ($I_t$) as input. It first tries to localize the agent in the Graph. If the agent is localized in a node different from the last timestep, it changes the location of the agent and adds an edge if required. If the agent is not localized, a new node is added and corresponding ghost nodes are added using the geometric explorable area prediction function ($\mathcal{F}_G$). See the text for more details.}
\label{fig_nts:update}
\end{figure*}

The above components will require access to 4 functions. We first define these 4 functions and then describe how they are used by the components of the model.\\\vspace{-4pt}

\noindent \textbf{Graph Localization $(\mathcal{F}_L)$.} Given a graph $G$ and an image $I$, this function tries to localize $I$ in a node in the graph. An image is localized in a node if its location is visible from the panoramic image associated with the node. Internally, this requires comparing each node image ($I_{N_i}$) with the given image ($I$) to predict whether $I$ belongs to node $N_i$.\\\vspace{-4pt}

\noindent \textbf{Geometric Explorable Area Prediction $(\mathcal{F}_G)$}. Given an image $I$, this function makes $n_\theta = 12$ different predictions of whether there's explorable area in the direction $\theta$ sampled uniformly between $0$ and $2\pi$. Figure~\ref{fig_nts:f_g} shows an example of input and output of this function. Intuitively, it recognizes doors or hallways which lead to other areas.\\\vspace{-4pt}

\noindent \textbf{Semantic Score Prediction $(\mathcal{F}_S)$}. Given a source image $I_S$ and a goal image $I_G$, this function makes $n_\theta = 12$ different predictions of how fast the agent is likely to reach the goal image if it explores in the direction $\theta$ sampled uniformly between $0$ and $2\pi$. Figure~\ref{fig_nts:f_s} shows example input-output pairs for this function. The scores corresponding to the same source image change as the goal image changes. Estimating this score requires the model to learn semantic priors about the environment. \\\vspace{-4pt}

\noindent \textbf{Relative Pose Prediction $(\mathcal{F}_R)$}. Given a source image $I_S$ and a goal image $I_G$ which belong to the same node, this function predicts the relative pose ($\Delta p_{S,G}$) of the goal image from the source image.\\\vspace{-6pt}

\subsection{Model components}
\vspace{-2pt}
\noindent Assuming that we have access to the above functions, we first describe how these functions are used by the three components to perform Image Goal Navigation. We then describe how we train a single model to learn all the above functions using supervised learning.\\\vspace{-4pt}

\noindent \textbf{Graph Update.} The Graph Update module is responsible for updating the topological map given agent observations. At $t=0$, the agent starts with an empty graph. At each time step, the Graph Update module ($f_{GU}$) takes in the current observations $s_t$ and the previous topological map $G_{t-1}$ and outputs the updated topological map, $G_{t} = f_{GU}(s_t, G_{t-1})$. Figure~\ref{fig_nts:update} shows an overview of the Graph Update Module.

In order to update the graph, the module first tries to localize the current image in a node in the current graph using the Graph Localization function ($\mathcal{F}_L$). If the current image is localized in a node different from the last node, we add an edge between the current node and the last node (if it does not exist already). If the current image is not localized, then we create a new node with the current image. We also add an edge between the new node and the last node. Every time we add an edge, we also store the relative pose ($\Delta p$) between the two nodes connected by the edge using the sensor pose estimate. 

The above creates a graph of the explored areas. In order to explore new areas, we also predict and add unexplored areas to the graph. We achieve this by augmenting the graph with `ghost' nodes which are agent's prediction of explorable areas using the Geometric Explorable Area Prediction function ($\mathcal{F}_G$). If there is an explorable area in a direction $\theta$, we add a `ghost' node ($X_k$) and connect it to the new node ($N_i$) using edge $E_{i,k}$. Since we do not have the image at the ghost node location, we associate a patch of the node image in the direction of $\theta$, i.e. ($I_{X_k} = I_{{N_i},\theta}$). The relative pose between the new node and the ghost node is stored as ($r, \theta$), where $\theta$ is the direction and $r = 3m$ is the radius of the node. The ghost nodes are always connected to exactly one regular node and always correspond to unexplored areas. We ensure this by removing ghost nodes when adding regular nodes in the same direction, and not adding ghost nodes in a particular direction if a regular node exists in that direction. Intuitively, ghost nodes correspond to the unexplored areas at the boundary of explored areas denoted by regular nodes. \\\vspace{-4pt}

\begin{figure*}
\centering
\begin{minipage}{0.99\textwidth}
\centering
\includegraphics[width=\linewidth,height=\textheight,keepaspectratio]{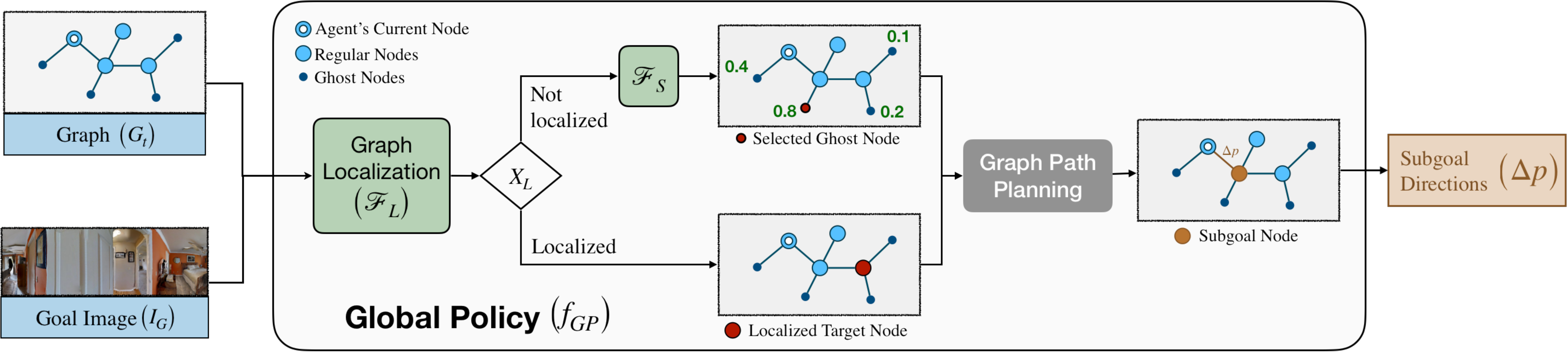}
\end{minipage}
\vspace{3pt}

    \caption{\small \textbf{Global Policy.} Figure showing an overview of the Global Policy. It takes the current Graph ($G_t$) and the Goal Image ($I_G$) as input. It first tries to localize the Goal Image in the Graph. If the Goal Image is localized, the corresponding node is selected as the long-term goal. If the Goal Image is not localized, then the semantic scoring function ($\mathcal{F}_S$) is used to score all the ghost nodes based on how close they are to the goal image. The ghost node with the highest score is selected as the long-term goal. Given a long-term goal, a subgoal node is computed using graph path planning. The relative directions to the subgoal node are the output of the Global Policy which is passed to the Local Policy.}
\label{fig_nts:global}
    \vspace{-8pt}
\end{figure*}

\noindent\textbf{Global Policy.}
The Global Policy is responsible for selecting a node in the above graph as the long-term goal. It first tries to localize the goal image in the current graph using the Graph Localization function ($\mathcal{F}_L$). If the goal image ($I_G$) is localized in a node $N_i$, then $N_i$ is chosen as the long-term goal. If the goal image is not localized, then the Global Policy needs to choose an area to explore, i.e. choose a ghost node for exploration. We use the Semantic Score Prediction ($\mathcal{F}_S$) function to predict the score of all ghost nodes. The Global Policy then just picks the ghost node with the highest score as the long-term goal.

 Once a node is selected as a long-term goal, we plan the path to the selected node from the current node using Djikstra's algorithm on the current graph. The next node on the shortest path is chosen to be the subgoal ($N_{SG}$). The relative pose associated with the edge to the subgoal node is passed to the Local Policy ($\Delta p_{i, SG}$).

If the goal image is localized in the current node (or the agent reaches the node $N_i$ where the goal image ($I_G$) is localized), the Global policy needs to predict the relative pose of $I_G$ with respect to the current agent observation. We use the Relative Pose Prediction ($\mathcal{F}_R$) function to get the relative pose of the goal image, which is then passed to the Local Policy.\\\vspace{-4pt}

\noindent\textbf{Local Policy.}
The Local Policy receives the relative pose as goal directions which comprises of distance and angle to the goal. Given the current image observation and the relative goal directions, the Local Policy takes navigation actions to reach the relative goal. This means the Local Policy is essentially a PointGoal navigation policy. Our Local Policy is adapted from~\cite{ans}. It predicts a local spatial map using a learned mapper model in the case of RGB input or using geometric projections of the depth channel in case of RGBD input. It then plans a path to the relative goal using shortest path planning.

\begin{figure*}
\centering
\begin{minipage}{0.99\textwidth}
\centering
\includegraphics[width=\linewidth,height=\textheight,keepaspectratio]{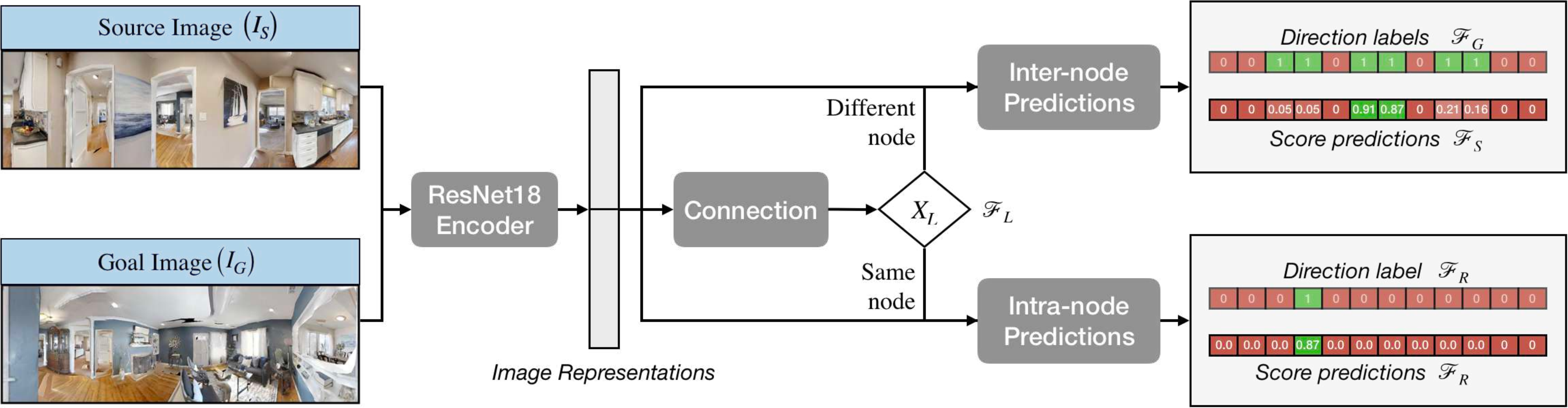}
\end{minipage}
\vspace{3pt}

    \caption{\small \textbf{NTS Multi-task Learning Model.} Figure showing an overview of the NTS Multi-task Learning Model. It takes a Source Image ($I_S$) and a Goal Image ($I_G$) as input and encodes them using a shared ResNet18 encoder. It first predicts whether the two nts/images belong to the same node or not. If they belong to the same node, it makes Intra-Node predictions which include the direction and score (or equivalently) distance of the Goal Image relative to the Source Image. If they belong to different nodes, it makes Inter-Node Predictions which include directions of explorable areas and a semantic score corresponding to each explorable area denoting its proximity of the Goal Image. All the predictions of this model are used at various places in the components of the overall NTS model. See text for more details.}
\label{fig_nts:model}
\end{figure*}

\vspace{2pt}
\subsection{Training NTS Multi-task Learning model}
\vspace{-2pt}
Given access to the four functions described above, we discussed how the different components use these functions for navigation. In this subsection, we describe how we train a single multi-task learning model to learn all the four functions. Figure~\ref{fig_nts:model} shows an overview of this multi-task learning model. It takes a Source Image ($I_S$) and a Goal Image ($I_G$) as input and encodes them using a shared ResNet18~\cite{he2016deep} encoder. It first predicts whether the two nts/images belong to the same node or not. This prediction is used to implement the Graph Localization function ($\mathcal{F}_L$). If they belong to the same node, it makes Intra-Node predictions which include the direction and score (or equivalently) distance of the Goal Image relative to the Source Image. These predictions are used to implement the Relative Pose Prediction function ($\mathcal{F}_R$). If they belong to different nodes, it makes Inter-Node Predictions which include directions of explorable areas (which is used as the Geometric Explorable Area Prediction function ($\mathcal{F}_G$)) and a semantic score corresponding to each explorable area denoting its proximity of the Goal Image (which is used as the Semantic Score Prediction function ($\mathcal{F}_S$)).\\

\begin{figure*}[t!]
\centering
\includegraphics[width=\linewidth,height=\textheight,keepaspectratio]{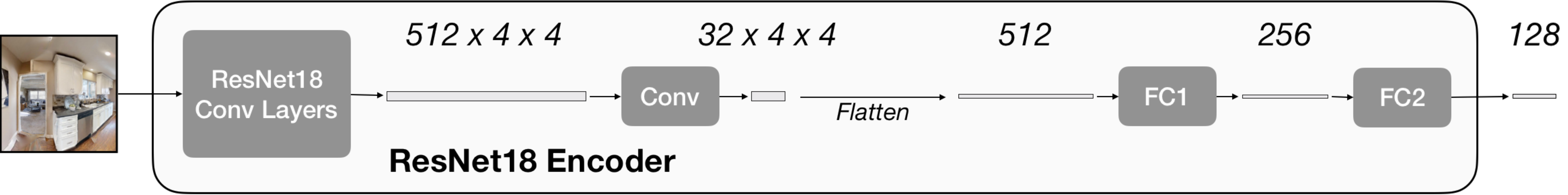}
\caption{\small \textbf{ResNet18 Encoder.}  Figure showing the architecture of the ResNet18 Encoder.}
\label{fig_nts:resnet18_encoder}
	\vspace{-10pt}
\end{figure*}

\noindent\textbf{Architecture and Hyperparameter Details.} Given panoramic source ($I_S$) and goal ($I_G$) nts/images, each of size $128 \times 512$, we first construct $n_\theta = 12$ square patches from each panoramic image. The $i^{th}$ patch consists of $128 \times 128$ image centered at $i / n_{\theta} * 2\pi $ radians. These resulting patches are such that each patch overlaps with 1/3 portion of the adjacent patch on both sides (patches are wrapped around at the edge of the panorama). Each of the $n_\theta$ patches of both source and target nts/images is passed through the ResNet18 encoder to get a patch representation of size 128. The architecture of the ResNet18 encoder is shown in Figure~\ref{fig_nts:resnet18_encoder} for RGB setting. For RGBD, we pass the depth channel through separate non-pretrained ResNet18 Conv layers, followed by 1x1 convolution to get a representation of size 512. This representation is concatenated with the 512 size RGB representation and passed through FC1 and FC2 to finally get the same 128 size patch representation. We use a dropout of 0.5 in FC1 and FC2 and ReLU non-linearity between all layers.

\begin{figure*}[t!]
\includegraphics[height=4.5cm,keepaspectratio]{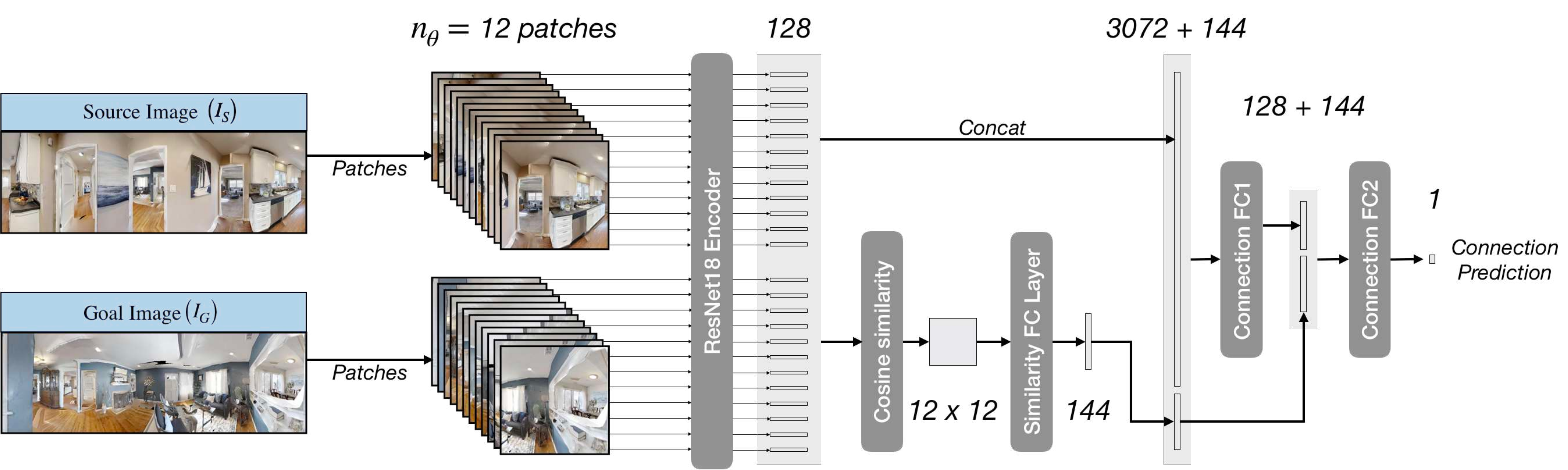}
\vspace{3pt}
\caption{\small \textbf{Connection Model.} Figure showing the architecture of the Connection Model.}
\label{fig_nts:connection}
	\vspace{-10pt}
\end{figure*}

\begin{figure*}[t!]
\includegraphics[height=4.5cm,keepaspectratio]{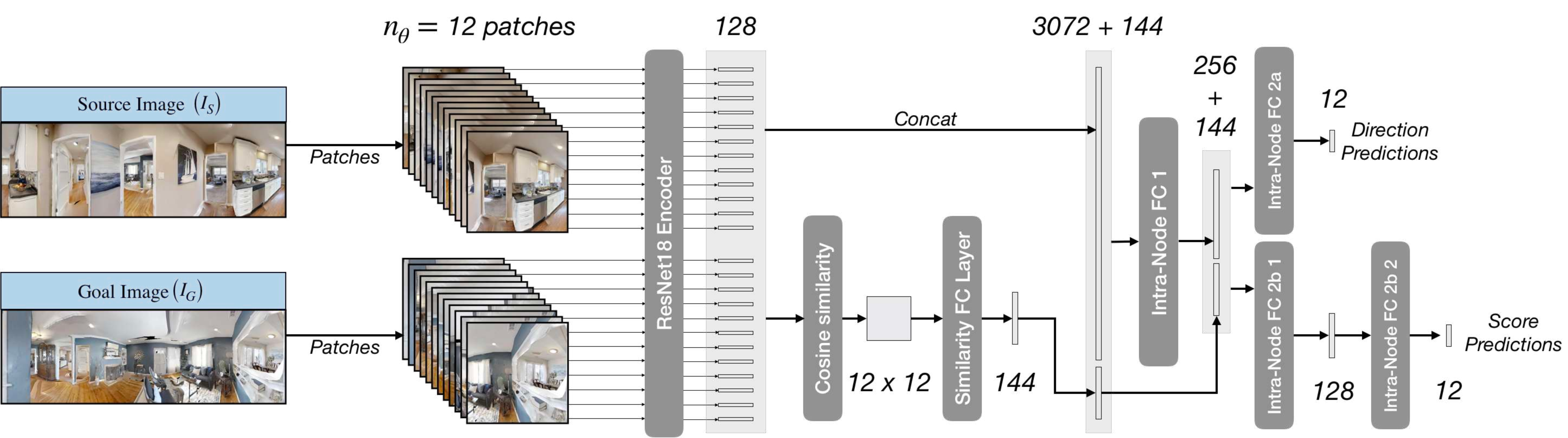}
\vspace{3pt}
\caption{\small \textbf{Intra-Node Predictions Model.} Figure showing the architecture of the Intra-Node Predictions Model.}
\label{fig_nts:intra_node}
	\vspace{-10pt}
\end{figure*}

\begin{figure*}[t!]
\includegraphics[height=4.94cm,keepaspectratio]{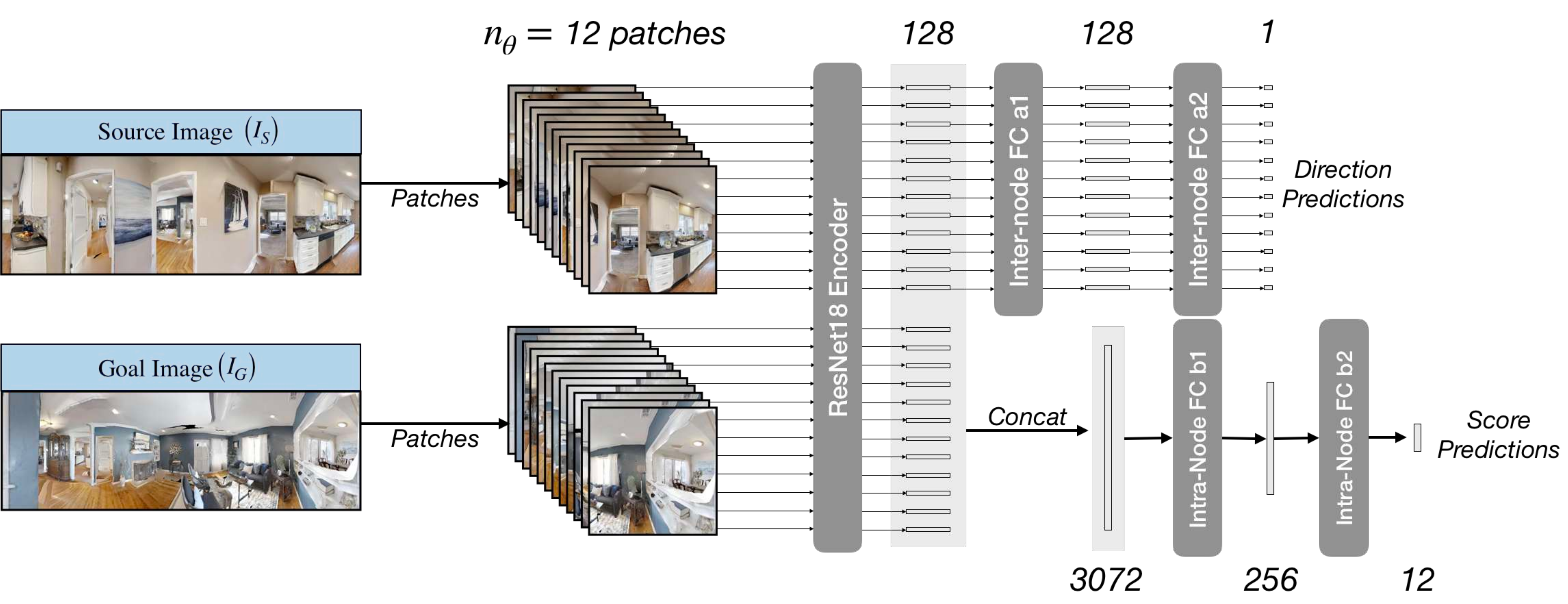}
\caption{\small \textbf{Inter-Node Predictions Model.} Figure showing the architecture of the Inter-Node Predictions Model.}
\label{fig_nts:inter_node}
	\vspace{-5pt}
\end{figure*}

Given 12 patch representations for source and goal nts/images each, the connection model predicts whether the two nts/images belong to the same node or not. The architecture of the Connection Model is shown in Figure~\ref{fig_nts:connection}. If the two nts/images belong to the same node, the Intra-Node Predictions model predicts the direction and score (or equivalently distance) of the goal image relative to the source image. The architecture of the Intra-Node Predictions Model is shown in Figure~\ref{fig_nts:intra_node}. If the two nts/images belong to different nodes, the Inter-Node Predictions model predicts the explorable area directions and the score of each direction. The architecture of the Inter-Node Predictions Model is shown in Figure~\ref{fig_nts:inter_node}. As shown in the figure, the explorable area directions are specific to the source image patches and independent of the goal image. We use ReLU non-linearity between all layers in all the modules.

The entire model is trained using supervised learning. We use Cross-Entropy Loss for the Connection and direction labels and MSE Loss for score labels. We use a loss coefficient of 10 for the score labels and a loss coefficient of 1 for the connection and direction labels. We train the model jointly with all the 5 losses using the Adam optimizer~\cite{kingma2014adam} with a learning rate of 5e-4 and a batch size of 64.

\vspace{-2pt}
\subsection{Dataset collection and automated labeling}
\vspace{-4pt}
\noindent For training the NTS Multi-task Learning model, we need to collect the data and different types of labels. We randomly sample 300 points in each training scene. For each ordered pair of nts/images, source image ($I_S$) and goal image ($I_G$), we gather the different labels as follow:\\\vspace{-9pt}

\noindent\textbf{Connection label:} This label denotes whether the source and goal image belong to the same node or not. We get this label by detecting whether the goal image location is visible from the source image. To compute this, we take a 5-degree patch of depth image centered at the relative direction of the goal image from the source image. If the maximum depth in this patch is greater than the distance to the goal image, then the goal image belongs to the same node as the source image and the label is 1. Intuitively, the above checks whether the maximum depth value in the direction of the goal image is greater than the distance to the goal. If the maximum depth in the direction of goal image is greater than the distance to the goal image or if the distance to the goal image is greater than $r=3m$, then the goal image belongs to a different node and the label is 0.\\\vspace{-9pt}

\noindent\textbf{Intra-node labels:} If the source and goal image belong to the same node, i.e. the above label is 1, then the intra-node direction and score labels are directly obtained from relative position of the goal image from the source image. Let the relative position of the goal image be $\Delta p = (d, \theta)$, where $d$ is the distance and $\theta$ is the angle to the goal image from the source image. Intra-node direction label is just the $360/n_{\theta} = 30$ degree bin in which $\theta$ falls, i.e. $nint(\theta/2\pi \times n_{\theta})$ where $nint(\cdot)$ is the nearest integer function. For the intra-node score label ($s$), we just convert the distance $d$ to a score between $0$ and $1$ using the following function:
$$ s = max((1 - d/r), 0)$$
\noindent where $s$ denotes the score, $r = 3m$ denotes the radius of the node, $d$ is the distance. Note that the direction label is discrete and the score label is continuous. \\\vspace{-9pt}

\noindent\textbf{Inter-node labels:} If the source and goal image do not belong to the same node, i.e. the connection label is 0, we compute the inter-node labels. For computing the inter-node direction labels, we first project the depth channel in the panoramic source image to compute an obstacle map. We ignore obstacles beyond $r=3m$. In this obstacle map, we take $n_{\theta} = 12$ points at angles $i/n_{\theta} \times 2\pi, \forall i \in [1,2,\ldots,12]$ and distance $r=3m$ away. For every, $i \in [1,2,\ldots,12]$, if the shortest path distance to the corresponding is less than $1.05\times r$, then the corresponding inter-node direction label is 1 and otherwise 0.

Let the inter-score labels be denoted by $s_i, \forall i \in [1,2,\ldots,12]$. For the inter-node score label $s_i$, we find the farthest explored and traversable point in direction $i \times (2\pi/n_{\theta})$ in the above obstacle map. Let the shortest path distance (geodesic distance) of the goal image from this point be $d_i$. Then the score label is given by:
$$ s_i = max((1 - d_i/d_{max}), 0)$$
\noindent where $d_{max} = 20m$ is the maximum distance above which the score is 0.

\section{Experimental Setup}
\vspace{-4pt}
\noindent\textbf{Environment.} All our experiments are conducted in the Habitat simulator~\cite{savva2019habitat} with the Gibson~\cite{xiazamirhe2018gibsonenv} dataset. The Gibson dataset is visually realistic as it consists of reconstructions of real-world scenes. We also implement physically realistic motion sensor and actuation noise models as proposed by~\cite{ans}. Actuation motion noise leads to stochastic transitions as the amount translated or rotated by the agent is noisy. This model also adds realistic translational noise in rotation actions and rotational noise in translational actions. The sensor noise model adds realistic noise to the base odometry sensor readings. Both the noise models are based on real-world data and agents trained on these noise models are shown to transfer to the real-world~\cite{ans}. \\\vspace{-6pt}

\noindent\textbf{Task setup.} We use panoramic nts/images of size $128 \times 512$ for both agent image observation and target goal image. We conduct experiments with both RGB and RGBD settings. The base odometry sensor provides a $3 \times 1$ reading denoting the change in agent's x-y coordinates and orientation. The action space consists of four actions: \texttt{move\_forward, turn\_right, turn\_left, stop}. The forward action moves the agent approximately 25cm forward and the turn actions turn the agent approximately 10 degrees. Note that the state space and motion of the agent are continuous. The agent succeeds in an episode if it takes the \texttt{stop} action within a $1m$ radius of the target location. The agent fails if it takes \texttt{stop} action anywhere else or does not take the \texttt{stop} till the episode ends. In addition to the success rate, we also use Success weighted by inverse Path Length (SPL) as an evaluation metric as proposed by \cite{anderson2018evaluation}. It takes into account the efficiency of the agent in reaching the goal (shorter successful trajectories lead to higher SPL). \\\vspace{-6pt}

\noindent\textbf{Training data.} We split the curated set of 86 scenes from ~\cite{savva2019habitat} into sets of 68/4/14 scenes for train/val/test. For training our supervised learning model, we sample 300 nts/images randomly in each of 68 training scenes. We get labels for pairs of source and target nts/images in each scene giving us a total of approximately $ 68 \times 300 \times 300$ = 6.12 million data points. The labeling process is automated and it only requires the ground-truth map already available with the dataset without the need for any additional human annotation. Note that sampling nts/images or the ground-truth map are not required for the test environments.\\\vspace{-6pt}

\noindent\textbf{Test episodes.} For creating test episodes, we sample episodes (given by starting and goal locations) in the test scenes to create 3 different sets of difficulty based on the distance of the goal from starting locations: Easy ($1.5-3m$), Medium ($3-5m$) and Hard ($5-10m$). The maximum episode length is 500 steps for each difficulty level. 

\setlength{\tabcolsep}{4.0pt}
\begin{table*}[]
    \centering
{\footnotesize
\begin{tabular}{@{}llccccccccccc@{}}
\toprule
\multicolumn{1}{c}{}  &                                           & \multicolumn{2}{c}{Easy}      &           & \multicolumn{2}{c}{Medium}    &           & \multicolumn{2}{c}{Hard}      &           & \multicolumn{2}{c}{Overall}   \\
\multicolumn{1}{c}{}  & Model                                     & Succ          & SPL           &           & Succ          & SPL           &           & Succ          & SPL           &           & Succ          & SPL           \\ \midrule
\multirow{4}{*}{RGB}  & ResNet + GRU + IL                        & 0.57          & 0.23          &           & 0.14          & 0.06          &           & 0.04          & 0.02          &           & 0.25          & 0.10          \\
                      & Target-driven RL                  & 0.56          & 0.22          &           & 0.17          & 0.06          &           & 0.06          & 0.02          &           & 0.26          & 0.10          \\
                      & Active Neural SLAM (ANS)       & 0.63          & 0.45          &           & 0.31          & 0.18          &           & 0.12          & 0.07          &           & 0.35          & 0.23          \\
                      & \textbf{Neural Topological SLAM (NTS)} & \textbf{0.80} & \textbf{0.60} & \textbf{} & \textbf{0.47} & \textbf{0.31} & \textbf{} & \textbf{0.37} & \textbf{0.22} & \textbf{} & \textbf{0.55} & \textbf{0.38} \\ \midrule
\multirow{6}{*}{RGBD} & ResNet + GRU + IL                           & 0.72          & 0.32          &           & 0.16          & 0.09          &           & 0.05          & 0.02          &           & 0.31          & 0.14          \\
                      & Target-driven RL        & 0.68          & 0.28          &           & 0.21          & 0.08          &           & 0.09          & 0.03          &           & 0.33          & 0.13          \\
                      & Metric Spatial Map + RL     & 0.69          & 0.27          &           & 0.22          & 0.07          &           & 0.12          & 0.04          &           & 0.34          & 0.13          \\
                      & Metric Spatial Map + FBE + Local                       & 0.77          & 0.56          &           & 0.36          & 0.18          &           & 0.13          & 0.05          &           & 0.42          & 0.26          \\
                      & Active Neural SLAM (ANS)     & 0.76          & 0.55          &           & 0.40          & 0.24          &           & 0.16          & 0.09          &           & 0.44          & 0.29          \\
                      & \textbf{Neural Topological SLAM (NTS)} & \textbf{0.87} & \textbf{0.65} & \textbf{} & \textbf{0.58} & \textbf{0.38} & \textbf{} & \textbf{0.43} & \textbf{0.26} & \textbf{} & \textbf{0.63} & \textbf{0.43} \\ \bottomrule
\end{tabular}
\vspace{-2pt}
\caption{\textbf{Results.} Performance of the proposed model Neural Topological SLAM (NTS) and the baselines in RGB and RGBD settings.}
\vspace{-8pt}
\label{tab_nts:results_main}
	}
\end{table*}

\subsection{Baselines}
\noindent We use the following baselines for our experiments:

\noindent\textbf{ResNet + GRU + IL.} A simple baseline consisting of ResNet18 image encoder and GRU based policy trained with imitation learning (IL).

\noindent\textbf{Target-driven RL.} A siamese style model for encoding the current image and the goal image using shared convolutional networks and trained end-to-end with reinforcement learning, adapted from Zhu et al.~\cite{zhu2017target}.

\noindent\textbf{Metric Spatial Map + RL.} An end-to-end RL model which uses geometric projections of the depth image to create a local map and passes it to the RL policy, adapted from Chen et al.~\cite{chen2018learning}.

\noindent\textbf{Metric Spatial Map + FBE + Local.} This is a hand-designed baseline which creates a map using depth nts/images and then uses a classical exploration heuristic called Frontier-based Exploration (FBE)~\cite{yamauchi1997frontier} which greedily explores nearest unexplored frontiers in the map. We use the Localization model and Local Policy from NTS to detect when a goal is nearby and navigate to it.

\noindent\textbf{Active Neural SLAM.} This is a recent modular model based on Metric Spatial Maps proposed for the task of exploration. We adapt it to the Image Goal task by using the Localization model and Local Policy from NTS to detect when a goal is nearby and navigate to it.

All the baselines are trained for 25 million frames. RL baselines are trained using Proximal Policy Optimization~\cite{schulman2017proximal} with a dense reward function. The reward function includes a high reward for success (=SPL*10.), a shaping reward equal to the decrease in distance to the goal and a per step reward of -0.001 to encourage shorter trajectories. ResNet + GRU + IL is trained using behavioral cloning on the ground-truth trajectory. This means just like the proposed model, all the baselines also use the ground-truth map for training. In terms of the number of training samples, sampling 300 random nts/images in the environment would require 300 episode resets in the RL training setup. 300 episodes in 68 scenes would lead to a maximum of $10.2$ million ($=68 \times 300 \times 500$) samples. Since we use 25 million frames to train our baselines, they use strictly more data than our model. Furthermore, our model does not require any interaction in the environment and can be trained offline with image data.

\setlength{\tabcolsep}{3.2pt}
\begin{table*}[]
    \centering
{\small

\begin{tabular}{@{}lcccccccccc@{}}
\toprule
                                        &  & \multicolumn{4}{c}{RGBD - No stop} &  & \multicolumn{4}{c}{RGBD - No Noise} \\
Model                              &  & Easy   & Med.   & Hard  & Overall  &  & Easy   & Med.   & Hard  & Overall  \\ \midrule
ResNet + GRU + IL                     &  & 0.76   & 0.28   & 0.10  & 0.38     &  & 0.71   & 0.18   & 0.06  & 0.32     \\
Target-driven RL               &  & 0.89   & 0.45   & 0.21  & 0.52     &  & 0.69   & 0.22   & 0.07  & 0.33     \\
Metric Spatial Map + RL~               &  & 0.89   & 0.45   & 0.21  & 0.52     &  & 0.70   & 0.24   & 0.11  & 0.35     \\
Metric Spatial Map + FBE + Local                 &  & 0.92   & 0.46   & 0.29  & 0.56     &  & 0.78   & 0.46   & 0.23  & 0.49     \\
Active Neural SLAM (ANS)  &  & 0.93   & 0.50   & 0.32  & 0.58     &  & 0.79   & 0.53   & 0.30  & 0.54     \\
    \textbf{Neural Topological SLAM (NTS)}    &  & \textbf{0.94}   & \textbf{0.70}   & \textbf{0.60}  & \textbf{0.75}     &  & \textbf{0.87}   & \textbf{0.60}   & \textbf{0.46}  & \textbf{0.64}     \\ \bottomrule
\end{tabular}
\vspace{-2pt}
    \caption{\textbf{No stop and no noise.} Success rate of all the models without stop action (left) and without motion noise (right).}
\vspace{-10pt}
	\label{tab_nts:results_stop}}
\end{table*}

\section{Results}
\begin{figure*}[h!]
\centering
\includegraphics[width=\linewidth,height=\textheight,keepaspectratio]{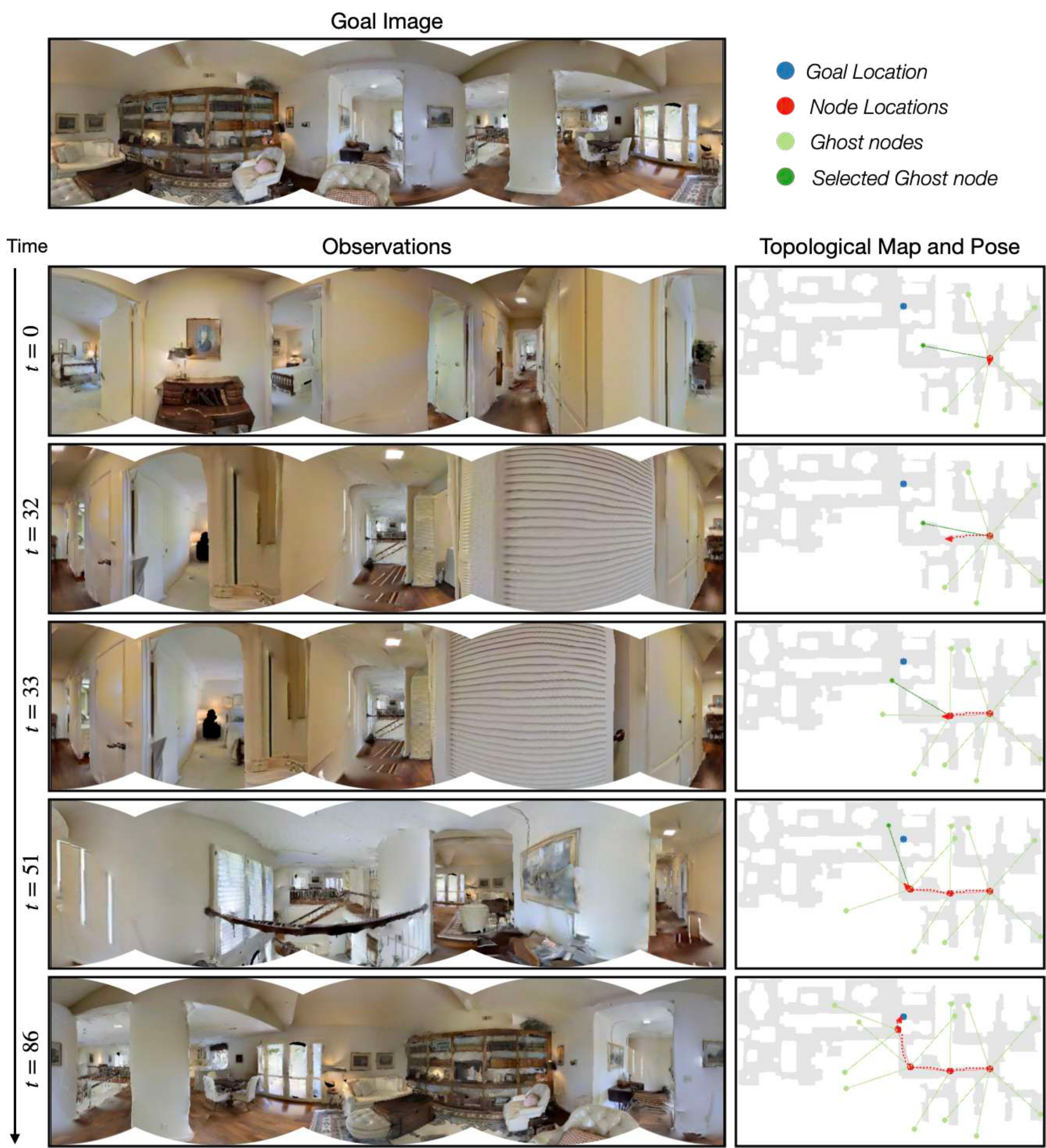}
	\vspace{-6pt}
	\caption{\small \textbf{Example Trajectory Visualization.} Figure showing a visualization of an example trajectory using the NTS model. Agent observations are shown on the left and the topological map and pose are shown on the right. Note that the map and the goal location are shown only for visualization and are not available to the NTS model. \textbf{(t=0)} NTS model creates the first node and adds the ghost nodes. Note that ghost node locations are not predicted, the model predicts only the direction of ghost nodes. The model correctly selects the correct ghost node closest to the goal. \textbf{(t=33)} The NTS model creates the second regular node and adds ghost nodes. Note that the ghost node in the direction of the second node from the first node are removed. \textbf{(t=51)} The NTS model creates another new node. \textbf{(t=86)} The agent reaches the goal after creating 4 nodes as shown in the trajectory on the map and decides to take the stop action.
    }
\label{fig_nts:eg}
\end{figure*}

\vspace{-4pt}
\noindent We evaluate the proposed method and all the baselines on 1000 episodes for each difficulty setting. We compare all the methods across all difficulty levels in both RGB and RGBD settings in Table~\ref{tab_nts:results_main}. The results show that the proposed method outperforms all the baselines by a considerable margin across all difficulty settings with an overall Succ/SPL of 0.55/0.38 vs 0.35/0.23 in RGB and 0.63/0.43 vs 0.44/0.29 in RGBD. The results also indicate the relative improvement of \methodabbr{} over the baselines increases as the difficulty increases leading to a large improvement in the hard setting (0.43/0.26 vs 0.16/0.09 in RGBD). In Fig~\ref{fig_nts:eg}, we visualize an example trajectory using the NTS model.
\\\vspace{-6pt}

\begin{figure*}
\centering
\begin{minipage}{0.49\textwidth}
\includegraphics[width=\linewidth,height=\textheight,keepaspectratio]{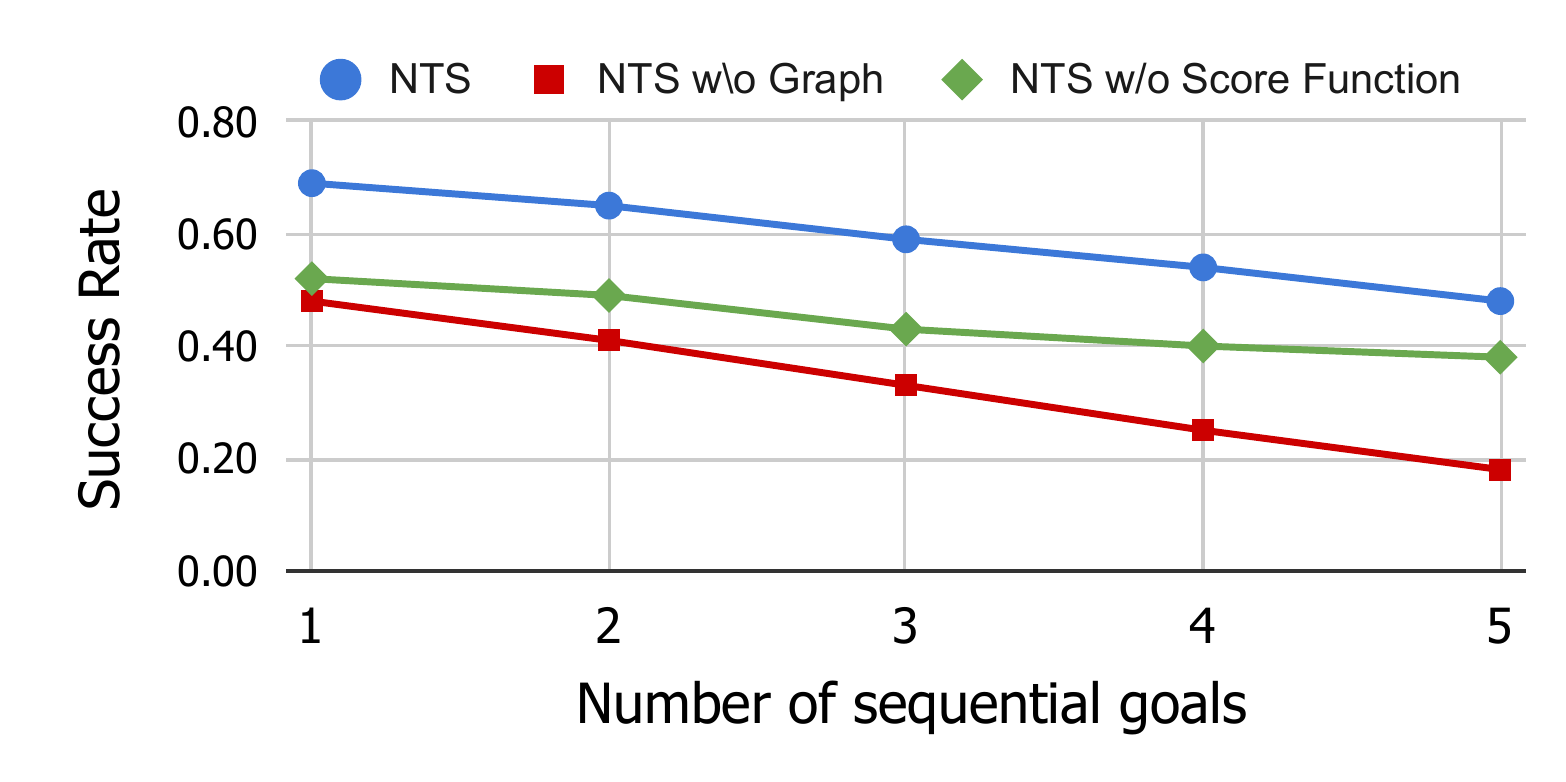}
\end{minipage}\hfill
\begin{minipage}{0.49\textwidth}
\centering
\includegraphics[width=\linewidth,height=\textheight,keepaspectratio]{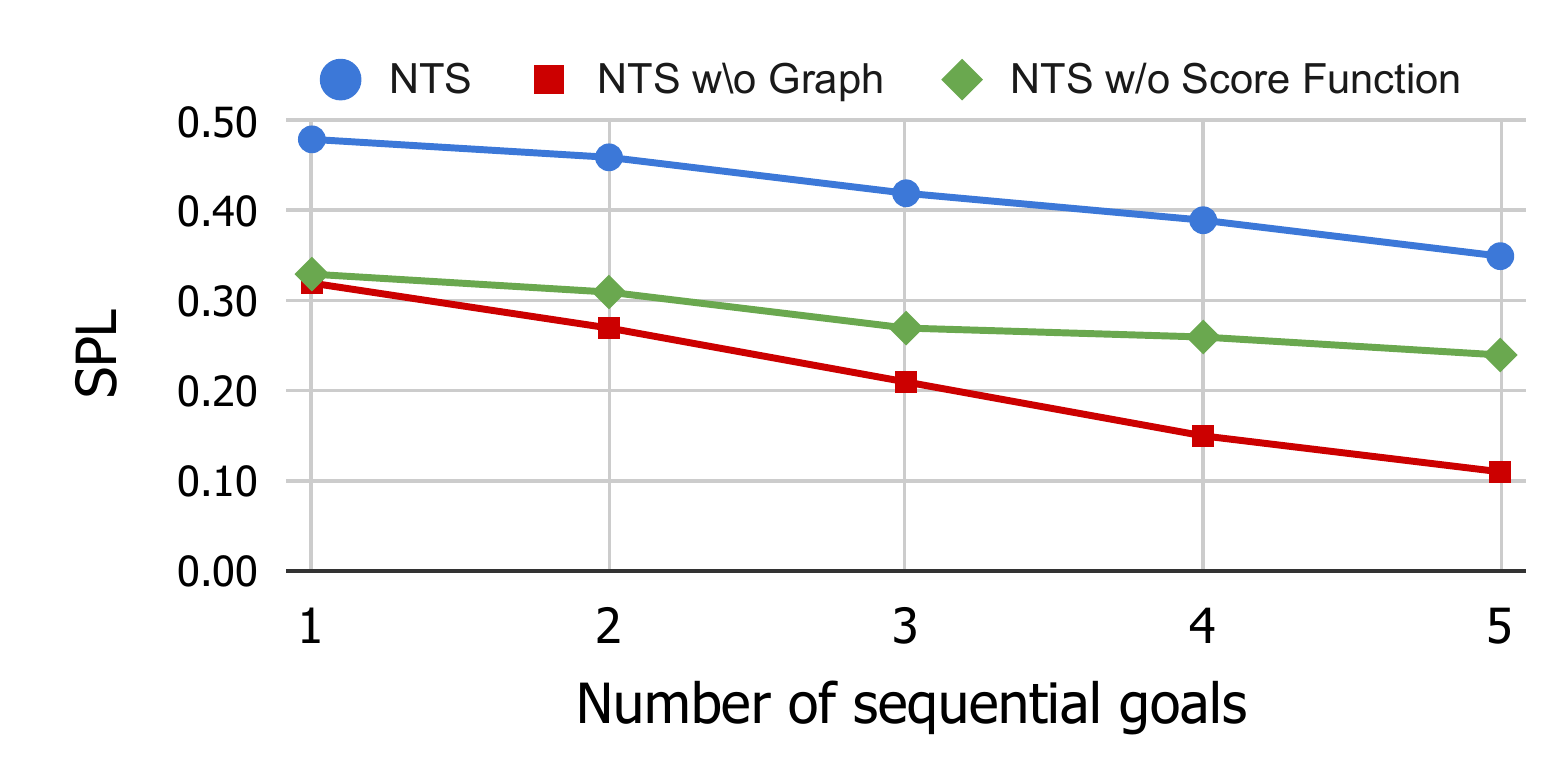}
\end{minipage}
\vspace{-8pt}
\caption{\small Performance of the proposed model \methodabbr{} and two ablations as a function of number of sequential goals. }
\label{fig_nts:ablation}
\vspace{-12pt}
\end{figure*}

\noindent \textbf{Comparison with end-to-end RL and the effect of stop action.} The results indicate that \methodabbr{} performs better than both end-to-end RL based baselines~\cite{zhu2017target, chen2018learning} and methods using metric spatial maps~\cite{chen2018learning, ans}. The performance of the RL-based baselines is much weaker than the proposed model. We believe the reason behind this is the complexity of the exploration search space. Compared to the Pointgoal navigation task where the agent receives updated direction to the goal at each time step, Image Goal navigation is more difficult in terms of exploration as the goal image does not directly provide the direction to explore. Another difficulty is exploring the `stop' action. Prior setups of the Image Goal task where end-to-end RL policies were shown to perform reasonably well assumed that the agent succeeded if it hits the goal state. However, based on the suggestion from~\cite{anderson2018evaluation}, we added the `stop' action as it is more realistic. To quantify the effect of stop action, we report the performance of all the models without the stop action in Table~\ref{tab_nts:results_stop}(left). We see that the performance of RL baselines is much higher. However, the performance of \methodabbr{} also increases as the agent automatically stops when it reaches the goal state instead of using the prediction of the Relative Pose Estimator to stop. Other differences that make our experimental setup more realistic but also makes exploration harder for RL as compared to prior setups include continuous state space as compared to grid-based state space, fine-grained action as compared to 90 degree turns and grid cell forward steps and stochastic transitions due to realistic motion noise.\\\vspace{-6pt}

\noindent \textbf{Comparison with spatial map-based methods and the effect of motion noise.} The performance of metric spatial map-based baselines drops quickly as the distance to the goal increases. This is likely due to the accumulation of pose errors as the trajectory length increases. Errors in pose prediction make the map noisy and eventually lead to incorrect path planning. To quantify this effect, we evaluate all the models without any motion actuation and sensor noise in Table~\ref{tab_nts:results_stop}(right). The results show that the performance of the metric map-based baselines scales much better with distance in the absence of motion noise, however, the performance of NTS does not increase much. This indicates that \methodabbr{} is able to tackle motion noise relatively well. This is because \methodabbr{} only uses pose estimates between consecutive nodes, which do not accumulate much noise as they are a few actions away from each other. The performance of \methodabbr{} is still better than the baselines even without motion noise as it consists of Semantic Score Predictor ($\mathcal{F}_S$) capable of learning structural priors which the metric spatial map-based baselines lack. We quantify the effect of the Semantic Score Predictor in the following subsection.

\subsection{Ablations and Sequential Goals}
\vspace{-4pt}
\noindent In this subsection, we evaluate the proposed model on sequential goals in a single episode and study the importance of the topological map or the graph and the Semantic Score Predictor ($\mathcal{F}_S$). For creating a test episode with sequential goals, we randomly sample a goal between $1.5m$ to $5m$ away from the last goal. The agent gets a time budget of 500 timesteps for each goal. We consider two ablations:\\\vspace{-9pt}

\noindent \textbf{\methodabbr{} w/o Graph.} We pick the direction with the highest score in the current image greedily, not updating or using the graph over time. Intuitively, the performance of this ablation should deteriorate as the number of sequential goals increases as it has no memory of past observations.\\\vspace{-9pt}

\noindent \textbf{\methodname{} w/o Score Function.} In this ablation, we do not use the Semantic Score Predictor ($\mathcal{F}_S$) and pick a ghost node randomly as the long-term goal when the Goal Image is not localized in the current graph. Intuitively, the performance of this ablation should improve with the increase in the number of sequential goals, as random exploration would build the graph over time and increase the likelihood of the Goal Image being localized.\\\vspace{-9pt}

\begin{figure*}[h!]
\centering
\includegraphics[width=\linewidth,height=\textheight,keepaspectratio]{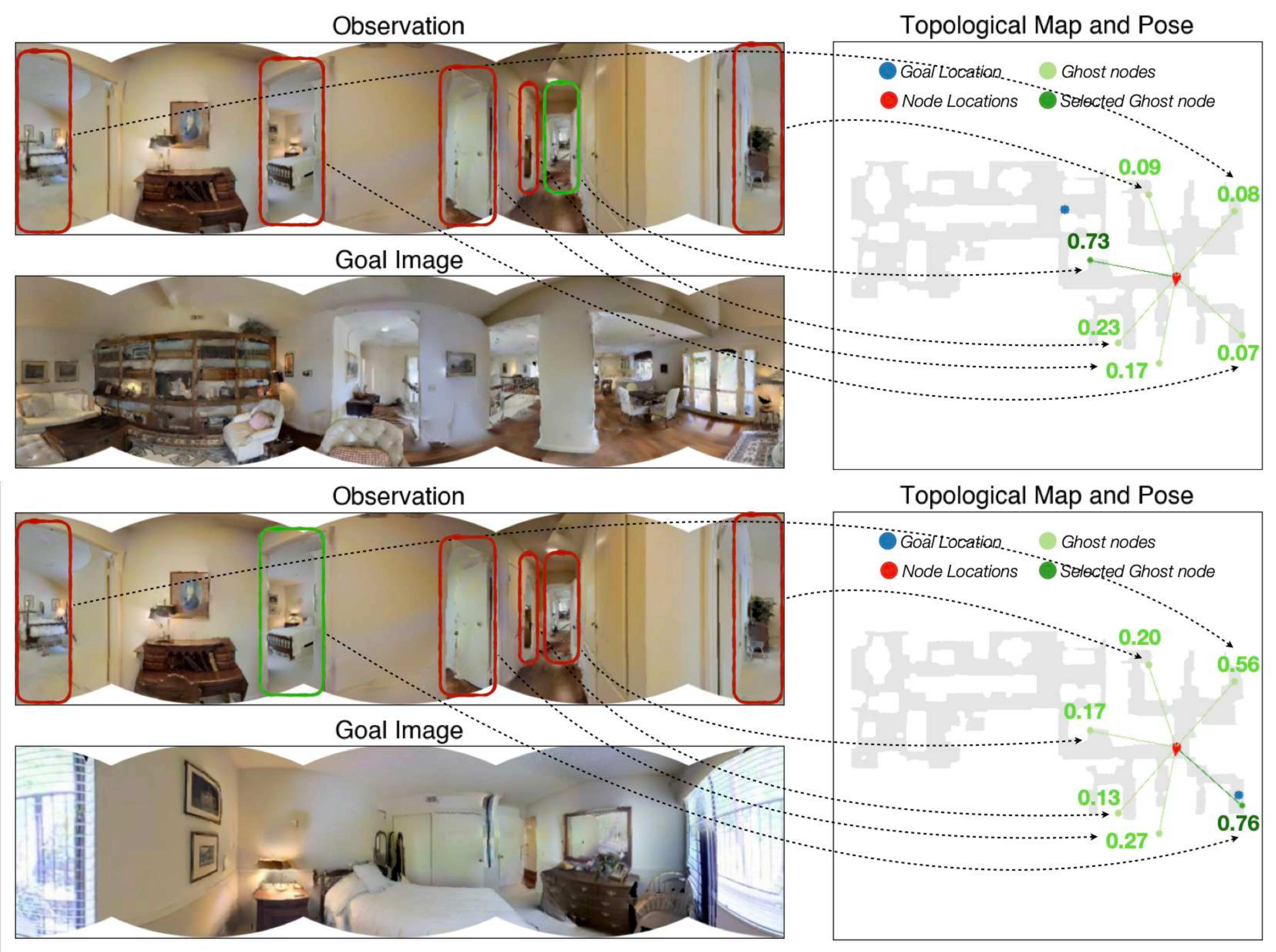}
	\vspace{-12pt}
	\caption{\small \textbf{Learning Semantic Priors.} Figure showing an example indicating that NTS model learns semantic priors. The figure shows the first time step for two different episodes starting at the same location but with different goal nts/images. \textbf{Top:} The goal image is of a living room. The NTS model successfully selects the ghost node leading down the hallway as other ghost nodes lead to different bedrooms. \textbf{Bottom:} The goal nts/images is of a bedroom. With the same starting location and the same ghost nodes, the NTS models select a different ghost node which leads to the correct bedroom.
    }
\label{fig_nts:semantic_priors}
\end{figure*}

We report the success rate and SPL of \methodabbr{} and the two ablations as a function of the number of sequential goals in Figure~\ref{fig_nts:ablation}. Success, in this case, is defined as the ratio of goals reached by the agent across a test set of 1000 episodes. The performance of \methodabbr{} is considerably higher than both the ablations, indicating the importance of both the components. The drop in performance when the semantic score function is removed indicates that the model learns some semantic priors (Figure~\ref{fig_nts:semantic_priors} shows an example). The performance of all the models decreases with an increase in the number of sequential goals because if the agent fails to reach an intermediate goal, there is a high chance that the subsequent goals are farther away. The performance gap between \methodabbr{} and \methodabbr{} w/o Score Function decreases and the performance gap between \methodabbr{} and \methodabbr{} w/o Graph increases with increase in the number of sequential goals as expected. This indicates that the topological map becomes more important over time as the agent explores a new environment, and while the Semantic Score Predictor is the most important at the beginning to explore efficiently.

\vspace{-2pt}
\section{Discussion}
\vspace{-4pt}
\noindent We designed topological representations for space that leverage semantics and afford coarse geometric reasoning. We showed how we can build such representation autonomously and use them for the task of image-goal navigation. Topological representations provided robustness to actuation noise, while semantic features stored at nodes allowed the use of statistical regularities for efficient exploration in novel environments. We showed how advances made in this paper make it possible to study this task in settings where no prior experience from the environment is available, resulting in a relative improvement of over $50\%$. 

Neural Topological SLAM encodes the semantics in the map implicitly using localization and semantic score functions and tackles the Image Goal Navigation task. In the next chapter, we will present methods which encode semantics in the map explicitly and study them in the context of the Object Goal Navigation task.

\chapter{Goal-Oriented Semantic Exploration}
\label{chap:semexp}
\def\methodname{Goal-Oriented Semantic Exploration}
\def\methodabbr{SemExp}
\newcommand\Mycite[1]{%
    \citeauthor{#1}~\cite{#1}}

\blfootnote{Webpage: \url{https://devendrachaplot.github.io/projects/semantic-exploration}}

\begin{figure*}[t!]
\centering
\includegraphics[width=0.99\linewidth,height=\textheight,keepaspectratio]{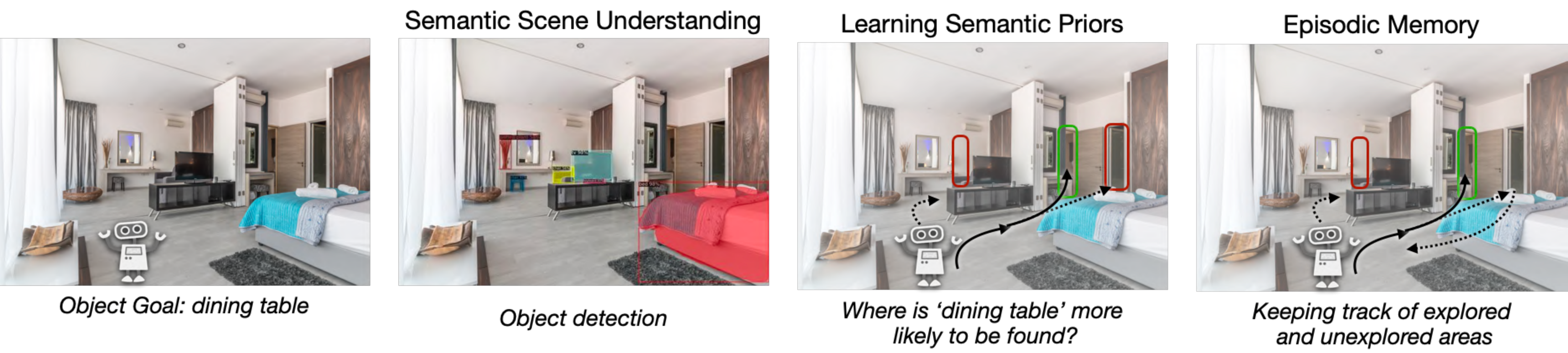}
\caption{\small \textbf{Semantic Skills required for Object Goal navigation.} Efficient Object Goal navigation not only requires passive skills such as object detection, but also active skills such as an building an episodic memory and using it effective to learn semantic priors abour relative arrangements of objects in a scene.}
\label{fig_sem_exp:teaser}
\end{figure*}

Consider an autonomous agent being asked to navigate to a `dining table' in an unseen environment as shown in Figure~\ref{fig_sem_exp:teaser}. In terms of semantic understanding, this task not only involves object detection, i.e. what does a `dining table' look like, but also scene understanding of where `dining tables' are more likely to be found. The latter requires a long-term episodic memory as well as learning semantic priors on the relative arrangement of objects in a scene. Long-term episodic memory allows the agent to keep track of explored and unexplored areas. Learning semantic priors allows the agent to also use the episodic memory to decide which region to explore next in order to find the target object in the least amount of time.

How do we design a computational model for building an episodic memory and using it effectively based on semantic priors for efficient navigation in unseen environments? One popular approach is to use end-to-end reinforcement or imitation learning with recurrent neural networks to build episodic memory and learn semantic priors implicitly~\cite{mirowski2016learning, gupta2017cognitive, zhu2017target, mousavian2019visual}. However, end-to-end learning-based methods suffer from large sample complexity and poor generalization as they memorize object locations and appearance in training environments.

In Chapter~\ref{chap:ans}, we introduced a modular learning-based system called `Active Neural SLAM' which builds explicit obstacle maps to maintain episodic memory. Explicit maps also allow analytical path planning and thus lead to significantly better exploration and sample complexity. However, Active Neural SLAM, designed for maximizing exploration coverage, does not encode semantics in the episodic memory and thus does not learn semantic priors. In this chapter, we extend the Active Neural SLAM system to build explicit semantic maps and learn semantic priors using a semantically-aware long-term policy.

The proposed method, called `\methodname{}' (\methodabbr{}), makes two improvements over Active Neural SLAM~\cite{ans} to tackle semantic navigation tasks. First, it builds top-down metric maps similar to~\cite{ans} but adds extra channels to encode semantic categories explicitly. Instead of predicting the top-down maps directly from the first-person image as in~\cite{ans}, we use first-person predictions followed by differentiable geometric projections. This allows us to leverage existing pretrained object detection and semantic segmentation models to build semantic maps instead of learning from scratch. Second, instead of using a coverage maximizing goal-agnostic exploration policy based only on obstacle maps, we train a goal-oriented semantic exploration policy which learns semantic priors for efficient navigation. These improvements allow us to tackle a challenging object goal navigation task. Our experiments in visually realistic simulation environments show that \methodabbr{} outperforms prior methods by a significant margin. The proposed model also won the CVPR 2020 Habitat ObjectNav Challenge~\cite{batra2020objectnav}\footnote{ \url{https://aihabitat.org/challenge/2020/}}. We also demonstrate that \methodabbr{} achieves similar performance in the real-world when transferred to a mobile robot platform.

We briefly discuss related work on semantic mapping and navigation below.

\textbf{Semantic Mapping.} There's a large body of work on building obstacle maps both in 2D and 3D using structure from motion and Simultaneous Localization and Mapping (SLAM)~\cite{henry2014rgb, izadiUIST11, snavely2008modeling}. We defer the interested readers to the survey by~\Mycite{slam-survey:2015} on SLAM. Some of the more relevant works incorporate semantics in the map using probabilistic graphical models~\cite{bowman2017probabilistic} or using recent learning-based computer vision models~\cite{zhang2018semantic, ma2017multi}. In contrast to these works, we use differentiable projection operations to learn semantic mapping with supervision in the map space. This limits large errors in the map due to small errors in first-person semantic predictions.

\textbf{Navigation.} Classical navigation approaches use explicit geometric maps to compute paths to goal locations via path planning~\cite{kavraki1996probabilistic, lavalle2000rapidly, canny1988complexity, sethian1996fast}. The goals are selected base on heuristics such as the Frontier-based Exploration algorithm~\cite{yamauchi1997frontier}. In contrast, we use a learning-based policy to use semantic priors for selecting exploration goals based on the object goal category.

Recent learning-based approaches use end-to-end reinforcement or imitation learning for training navigation policies. These include methods which use recurrent neural networks~\cite{mirowski2016learning, lample2016playing, chaplot2017arnold, savva2017minos, hermann2017grounded, chaplot2017gated, savva2019habitat, wijmans2019decentralized}, structured spatial representations~\cite{gupta2017cognitive, parisotto2017neural, chaplot2018active, henriques2018mapnet, gordon2018iqa} and topological representations~\cite{savinov2018semi, savinov2018episodic}. Recent works tackling object goal navigation include~\cite{wu2018learning, yang2018visual, wortsman2019learning, mousavian2019visual}. \Mycite{wu2018learning} try to explore structural similarities between the environment by building a probabilistic graphical model over the semantic information like room types. Similarly, ~\Mycite{yang2018visual} propose to incorporate semantic priors into a deep reinforcement learning framework by using Graph Convolutional Networks.~\Mycite{wortsman2019learning} propose a meta-reinforcement learning approach where an agent learns a self-supervised interaction loss that encourages effective navigation to even keep learning in a test environment.~\Mycite{mousavian2019visual} use semantic segmentation and detection masks obtained by running state-of-the-art computer vision algorithms on the input observation and used a deep network to learn the navigation policy based on it. In all the above methods, the learnt representations are implicit and the models need to learn obstacle avoidance, episodic memory, planning as well as semantic priors implicitly from the goal-driven reward. Explicit map representation has been shown to improve performance as well as sample efficiency over end-to-end learning-based methods for different navigation tasks~\cite{ans, chaplot2020neural}, however they learn semantics implicitly. In this work, we use explicit structured semantic map representation, which allows us to learn semantically-aware exploration policies and tackle the object-goal navigation task. Concurrent work studies the use of similar semantic maps in learning exploration policies for improving object detection systems~\cite{chaplot2020semantic}.

\begin{figure*}[t!]
\centering
\includegraphics[width=0.99\linewidth,height=\textheight,keepaspectratio]{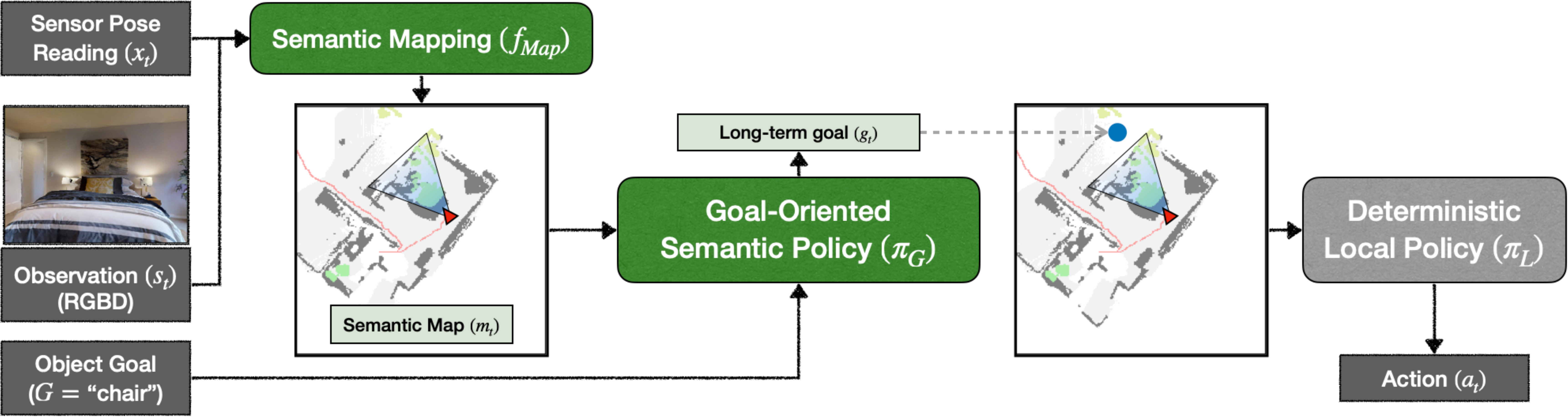}
\caption{\small \textbf{\methodname{} Model Overview.} The proposed model consists of two modules, Semantic Mapping and Goal-Oriented Semantic Policy. The Semantic Mapping model builds a semantic map over time and the Goal-Oriented Semantic Policy selects a long-term goal based on the semantic map to reach the given object goal efficiently. A deterministic local policy based on analytical planners is used to take low-level navigation actions to reach the long-term goal.}
\label{fig_sem_exp:overview}
\end{figure*}

\section{Problem Definition}
In the Object Goal task~\cite{savva2017minos, anderson2018evaluation}, the objective is to navigate to an instance of the given object category such as `chair' or `bed'. The agent is initialized at a random location in the environment and receives the goal object category ($G$) as input. At each time step $t$, the agent receives visual observations ($s_t$) and sensor pose readings $x_t$ and takes navigational actions $a_t$. The visual observations consist of first-person RGB and depth images. The action space $\mathcal{A}$ consists of four actions: \texttt{move\_forward, turn\_left, turn\_right, stop}. The agent needs to take the `stop' action when it believes it has reached close to the goal object. If the distance to the goal object is less than some threshold, $d_s (=1m)$, when the agent takes the stop action, the episode is considered successful. The episode terminates at after a fixed maximum number of timesteps $(=500)$.\\

\vspace{-3pt}
\section{Methods}
\vspace{-7pt}
We propose a modular model called `\methodname{}' (\methodabbr{}) to tackle the Object Goal navigation task (see Figure~\ref{fig_sem_exp:overview} for an overview). It consists of two learnable modules, `Semantic Mapping' and `Goal-Oriented Semantic Policy'. The Semantic Mapping module builds a semantic map over time and the Goal-Oriented Semantic Policy selects a long-term goal based on the semantic map to reach the given object goal efficiently. A deterministic local policy based on analytical planners is used to take low-level navigation actions to reach the long-term goal. We first describe the semantic map representation used by our model and then describe the modules.\\

\noindent\textbf{Semantic Map Representation.} The \methodabbr{} model internally maintains a semantic metric map, $m_t$ and pose of the agent $x_t$. The spatial map, $m_t$, is a ${K \times M \times M}$ matrix where $M \times M$ denotes the map size and each element in this spatial map corresponds to a cell of size $25cm^2$ ($5cm \times 5cm$) in the physical world. $K = C + 2$ is the number of channels in the semantic map, where $C$ is the total number of semantic categories. The first two channels represent obstacles and explored area and the rest of the channels each represent an object category. Each element in a channel represents whether the corresponding location is an obstacle, explored, or contains an object of the corresponding category. The map is initialized with all zeros at the beginning of an episode, $m_0 = [0]^{K \times M \times M}$. The pose $x_t \in \mathbb{R}^3$ denotes the $x$ and $y$ coordinates of the agent and the orientation of the agent at time $t$. The agent always starts at the center of the map facing east at the beginning of the episode, $x_0 = (M/2, M/2, 0.0)$.\\

\noindent\textbf{Semantic Mapping.} In order to a build semantic map, we need to predict semantic categories and segmentation of the objects seen in visual observations. It is desirable to use existing object detection and semantic segmentation models instead of learning from scratch. The Active Neural SLAM model predicts the top-down map directly from RGB observations and thus, does not have any mechanism for incorporating pretrained object detection or semantic segmentation systems. Instead, we predict semantic segmentation in the first-person view and use differentiable projection to transform first-person predictions to top-down maps. This allows us to use existing pretrained models for first-person semantic segmentation. However small errors in first-person semantic segmentation can lead to large errors in the map after projection. We overcome this limitation by imposing a loss in the map space in addition to the first-person space.

\begin{figure*}[t!]
\centering
\includegraphics[width=0.99\linewidth,height=\textheight,keepaspectratio]{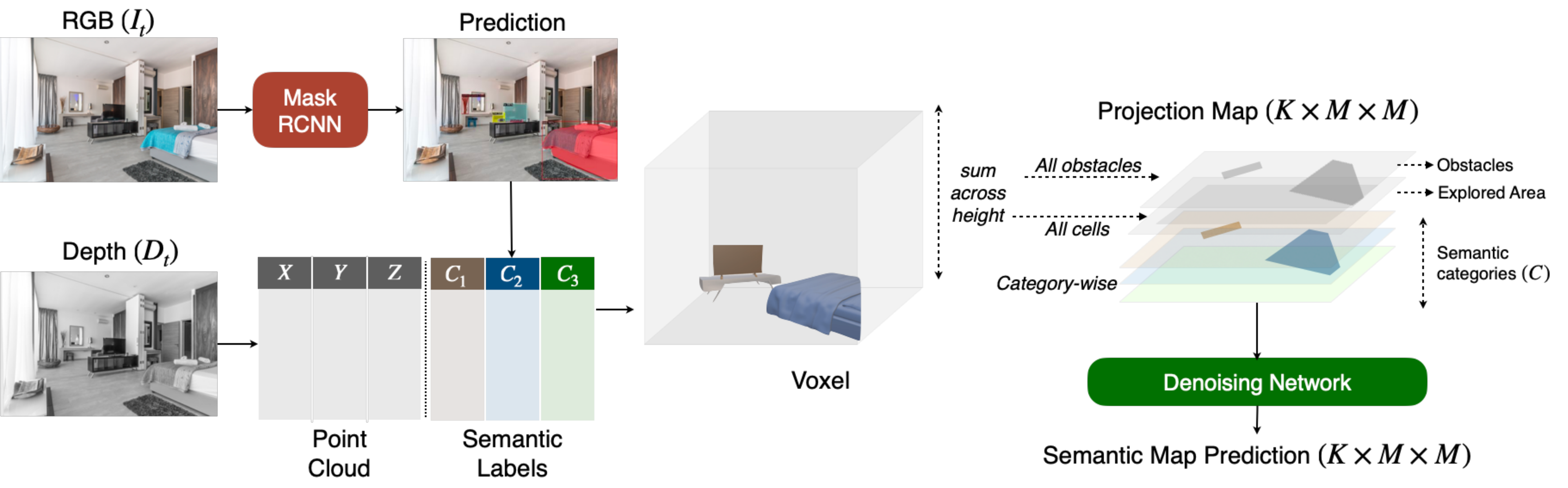}
\caption{\small \textbf{Semantic Mapping.} The Semantic Mapping module takes in a sequence of RGB ($I_t$) and Depth ($D_t$) images and produces a top-down Semantic Map.}
\label{fig_sem_exp:semantic_mapping}
\vspace{-10pt}
\end{figure*}

Figure~\ref{fig_sem_exp:semantic_mapping} shows an overview of the Semantic Mapping module. The
depth observation is used to compute a point cloud. Each point in the point cloud is associated with the predicted semantic categories. The semantic categories are predicted using a pretrained Mask RCNN~\cite{mask_rcnn} on the RGB observation. Each point in the point cloud is then projected in 3D space using differentiable geometric computations to get the voxel representation. The voxel representation is then converted to the semantic map. Summing over the height dimension of the voxel representation for all obstacles, all cells, and each category gives different channels of the projected semantic map. The projected semantic map is then passed through a denoising neural network to get the final semantic map prediction. The map is aggregated over time using spatial transformations and channel-wise pooling as described in~\cite{ans}. The Semantic Mapping module is trained using supervised learning with cross-entropy loss on the semantic segmentation as well as semantic map prediction. The geometric projection is implemented using differentiable operations such that the loss on the semantic map prediction can be backpropagated through the entire module if desired.\\

\noindent\textbf{Goal-Oriented Semantic Policy.} The Goal-Oriented Semantic Policy decides a long-term goal based on the current semantic map to reach the given object goal ($G$). If the channel corresponding to category $G$ has a non-zero element, meaning that the object goal is observed, it simply selects all non-zero elements as the long-term goal. If the object goal is not observed, the Goal-Oriented Semantic Policy needs to select a long-term goal where a goal category object is most likely to be found. This requires learning semantic priors on the relative arrangement of objects and areas. We use a neural network to learn these semantic priors. It takes the semantic map, the agent's current and past locations, and the object goal as input and predicts a long-term goal in the top-down map space.

The Goal-Oriented Semantic Policy is trained using reinforcement learning with distance reduced to the nearest goal object as the reward. We sample the long-term goal at a coarse time-scale, once every $u = 25$ steps, similar to the goal-agnostic Global Policy in~\cite{ans}. This reduces the time-horizon for exploration in RL exponentially and consequently, reduces the sample complexity.\\

\noindent\textbf{Deterministic Local Policy.} The local policy uses Fast Marching Method~\cite{sethian1996fast} to plan a path to the long-term goal from the current location based on the obstacle channel of the semantic map. It simply takes deterministic actions along the path to reach the long-term goal. We use a deterministic local policy as compared to a trained local policy in~\cite{ans} as they led to a similar performance in our experiments. Note that although the above Semantic Policy acts at a coarse time scale, the Local Policy acts at a fine time scale. At each time step, we update the map and replan the path to the long-term goal.

\vspace{-3pt}
\section{Experimental Setup}
\vspace{-7pt}
We use the Gibson~\citep{xiazamirhe2018gibsonenv} and Matterport3D (MP3D)~\citep{Matterport3D} datasets in the Habitat simulator~\citep{savva2019habitat} for our experiments. Both Gibson and MP3D consist of scenes which are 3D reconstructions of real-world environments. For the Gibson dataset,wWe use the train and val splits of Gibson tiny set for training and testing respectively as the test set is held-out for the online evaluation server. We do not use the validation set for hyper-parameter tuning. The semantic annotations for the Gibson tiny set are available from~\Mycite{armeni20193d}. For the MP3D dataset, we use the standard train and test splits.
Our training and test set consists of a total of 86 scenes (25 Gibson tiny and 61 MP3D) and 16 scenes (5 Gibson tiny and 11 MP3D), respectively.

The observation space consists of RGBD images of size $4 \times 640 \times 480$ , base odometry sensor readings of size $3 \times 1$ denoting the change in agent's x-y coordinates and orientation, and goal object category represented as an integer. The actions space consists of four actions: \texttt{move\_forward, turn\_left, turn\_right, stop}. The success threshold $d_s$ is set to $1m$. The maximum episode length is 500 steps. We note that the depth and pose are perfect in simulation, but these challenges are orthogonal to the focus of this chapter and prior works have shown that both can be estimated effectively from RGB images and noisy sensor pose readings~\cite{godard2017unsupervised, ans}. These design choices are identical to the CVPR 2020 Object Goal Navigation Challenge. 

\begin{figure*}[t!]
\centering
\includegraphics[width=0.99\linewidth,height=\textheight,keepaspectratio]{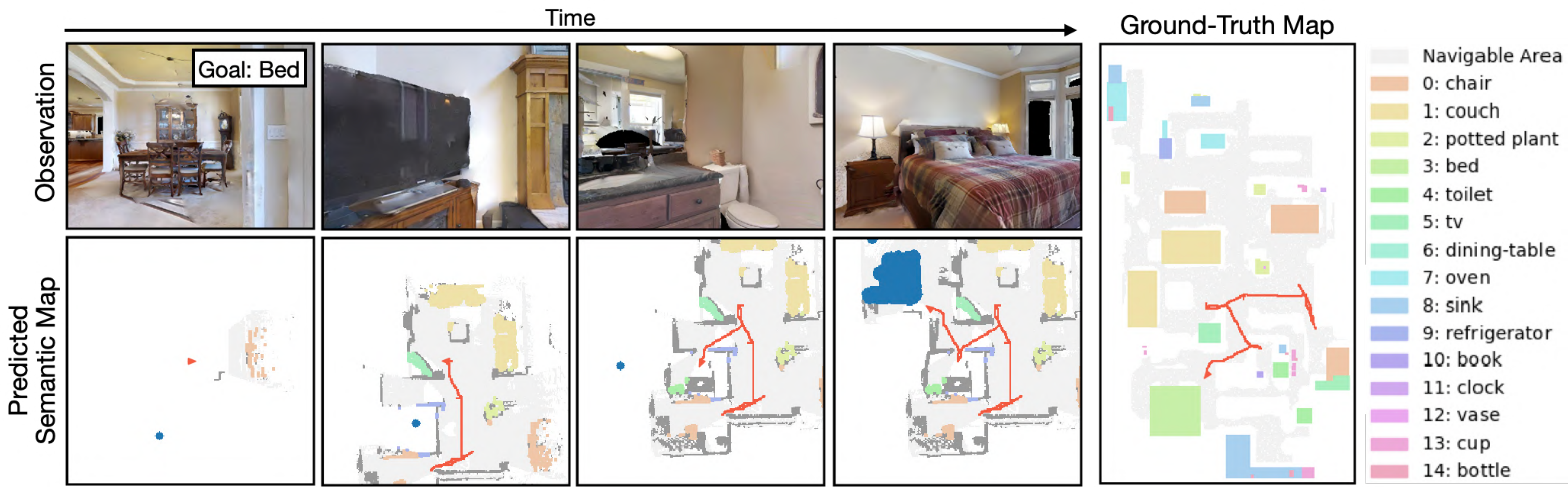}
\caption{\textbf{Example Trajectory.} Figure showing an example trajectory of the \methodabbr{} model in a scene from the Gibson test set. Sample images seen by the agent are shown on the top and the predicted semantic map is shown below. The goal object is `bed'. The long-term goal selected by the Goal-driven Semantic Policy is shown in blue. The ground-truth map (not visible to the agent) with the agent trajectory is shown on the right for reference.}
\label{fig_sem_exp:habitat_episode}
\end{figure*}

For the object goal, we use object categories which are common between Gibson, MP3D, and MS-COCO~\cite{lin2014microsoft} datasets. This leads to set of 6 object goal categories: `chair', `couch', `potted plant', `bed', `toilet' and `tv'. We use a Mask-RCNN~\cite{mask_rcnn} using Feature Pyramid Networks~\cite{lin2017feature} with ResNet50~\cite{he2016deep} backbone pretrained on MS-COCO for object detection and instance segmentation. Although we use 6 categories for object goals, we build a semantic map with 15 categories (shown on the right in Figure~\ref{fig_sem_exp:habitat_episode}) to encode more information for learning semantic priors.\\

\noindent\textbf{Architecture and Hyperparameter details.} We use PyTorch~\citep{paszke2017automatic} for implementing and training our model. The denoising network in the Semantic Mapping module is a 5-layer fully convolutional network. We freeze the Mask RCNN weights in the Semantic Mapping module (except for results on Habitat Challenge in Section~\ref{sec_sem_exp:challenge}) as Matterport does not contain labels for all 15 categories in our semantic map. We train the denoising network with the map-based loss on all 15 categories for Gibson frames and only 6 categories on MP3D frames. The Goal-driven Semantic Policy is a 5 layer convolutional network followed by 3 fully connected layers. In addition to the semantic map, we also pass the agent orientation and goal object category index as separate inputs to this Policy. They are processed by separate Embedding layers and added as an input to the fully-connected layers. 

We train both the modules with 86 parallel threads, with each thread using one scene in the training set. We maintain a FIFO memory of size 500000 for training the Semantic Mapping module. After one step in each thread, we perform 10 updates to the Semantic Mapping module with a batch size of 64. We use Adam optimizer with a learning rate of $0.0001$. We use binary cross-entropy loss for semantic map prediction. The Goal-driven Policy samples a new goal every $u=25$ timesteps. For training this policy, we use Proximal Policy Optimization (PPO)~\citep{schulman2017proximal} with a time horizon of 20 steps, 36 mini-batches, and 4 epochs in each PPO update. Our PPO implementation is based on~\cite{pytorchrl}. The reward for the policy is the decrease in distance to the nearest goal object. We use Adam optimizer with a learning rate of $0.000025$, a discount factor of $\gamma = 0.99$, an entropy coefficient of $0.001$, value loss coefficient of $0.5$ for training the Goal-driven Policy.\\

\noindent\textbf{Metrics.} We use 3 metrics for comparing all the methods:
\textbf{Success.} Ratio of episodes where the method was successful.
\textbf{SPL.} Success weighted by Path Length as proposed by~\cite{anderson2018evaluation}. This metric measures the efficiency of reaching the goal in addition to the success rate.
\textbf{DTS:} Distance to Success. This is the distance of the agent from the success threshold boundary when the episode ends. This is computed as follows:
$$ DTS = max(||x_T - G||_2 - d_s, 0) $$
where $|| x_T - G ||_2$ is the L2 distance of the agent from the goal location at the end of the episode, $d_s$ is the success threshold.

\vspace{-6pt}
\subsection{Baselines.}
\vspace{-8pt}
We use two end-to-end Reinforcement Learning (RL) methods as baselines:\\\vspace{-8pt}

\noindent\textbf{RGBD + RL:} A vanilla recurrent RL Policy initialized with ResNet18~\citep{he2016deep} backbone followed by a GRU adapted from ~\Mycite{savva2019habitat}. Agent pose and goal object category are passed through an embedding layer and append to the recurrent layer input.\\\vspace{-8pt}

\noindent\textbf{RGBD + Semantics + RL~\cite{mousavian2019visual}:} This baseline is adapted from~\Mycite{mousavian2019visual} who pass semantic segmentation and object detection predictions along with RGBD input to a recurrent RL policy. We use a pretrained Mask RCNN identical to the one used in the proposed model for semantic segmentation and object detection in this baseline. RGBD observations are encoded with a ResNet18 backbone visual encoder, and agent pose and goal object are encoded usinng embedding layers as described above. \\\vspace{-8pt}

Both the RL based baselines are trained with Proximal Policy Optimization~\cite{schulman2017proximal} using a dense reward of distance reduced to the nearest goal object. We design two more baselines based on goal-agnostic exploration methods combined with heuristic-based local goal-driven policy.\\\vspace{-8pt}

\noindent\textbf{Classical Mapping + FBE~\cite{yamauchi1997frontier}:} This baseline use classical robotics pipeline for mapping followed by classical frontier-based exploration (FBE)~\cite{yamauchi1997frontier} algorithm. We use a heuristic-based local policy using a pretrained Mask-RCNN. Whenever the Mask RCNN detects the goal object category, the local policy tries to go towards the object using an analytical planner. \\\vspace{-8pt}

\noindent\textbf{Active Neural SLAM~\cite{ans}:} In this baseline, we use an exploration policy trained to maximize coverage from~\cite{ans}, followed by the heuristic-based local policy as described above.

\setlength{\tabcolsep}{4.1pt}
\begin{table}[]
\centering
\begin{tabular}{@{}llccccccc@{}}
\toprule
                      &           & \multicolumn{3}{c}{Gibson}                       &           & \multicolumn{3}{c}{MP3D}                         \\
Method                &           & SPL            & Success           & DTS (m)           &           & SPL            & Success           & DTS (m)           \\ \midrule
Random                &           & 0.004          & 0.004          & 3.893          &           & 0.005          & 0.005          & 8.048          \\
RGBD + RL             &           & 0.027          & 0.082          & 3.310          &           & 0.017          & 0.037          & 7.654          \\
RGBD + Semantics + RL &           & 0.049          & 0.159          & 3.203          &           & 0.015          & 0.031          & 7.612          \\
Classical Map + FBE   &           & 0.124          & 0.403          & 2.432          &           & 0.117          & 0.311          & 7.102          \\
Active Neural SLAM   &           & 0.145          & 0.446          & 2.275          &           & 0.119          & 0.321          & 7.056          \\
\methodabbr{}       &  & \textbf{0.199} & \textbf{0.544} & \textbf{1.723} &  & \textbf{0.144} & \textbf{0.360} & \textbf{6.733} \\ \bottomrule
\end{tabular}
\caption{\small \textbf{Results.} Performance of \methodabbr{} as compared to the baselines on the Gibson and MP3D datasets across three different metrics: Success rate, SPL (Success weighted by Path Length) and DTS (Distance To Goal in meters). See text for more details.}
\label{tab_sem_exp:results}
\end{table}

\vspace{-3pt}
\section{Results}
\vspace{-7pt}
We train all the baselines and the proposed model for 10 million frames and evaluate them on the Gibson and MP3D scenes in our test set separately. We run 200 evaluations episode per scene, leading to a total of 1000 episodes in Gibson (5 scenes) and 2000 episodes in MP3D (10 scenes, 1 scene did not contain any object of the 6 possible categories). Figure~\ref{fig_sem_exp:habitat_episode} visualizes an exmaple trajectory using the proposed \methodabbr{} showing the agent observations and predicted semantic map\footnote{See demo videos at \url{https://devendrachaplot.github.io/projects/semantic-exploration}}. The quantitative results are shown in Table~\ref{tab_sem_exp:results}. \methodabbr{} outperforms all the baselines by a considerable margin consistently across both the datasets (achieving a success rate 54.4\%/36.0\% on Gibson/MP3D vs 44.6\%/32.1\% for the Active Neural SLAM baseline) . The absolute numbers are higher on the Gibson set, as the scenes are comparatively smaller. The Distance to Success (DTS) threshold for Random in Table~\ref{tab_sem_exp:results} indicates the difficulty of the dataset. Interestingly, the baseline combining classical exploration with pretrained object detectors outperforms the end-to-end RL baselines. We observed that the training performance of the RL-based baselines was much higher indicating that they memorize the object locations and appearance in the training scenes and generalize poorly. The increase in performance of \methodabbr{} over the Active Neural SLAM baseline shows the importance of incorporating semantics and the goal object in exploration.

\setlength{\tabcolsep}{22pt}
\begin{table}[]
\centering
\begin{tabular}{@{}lllll@{}}
\toprule
Method                                  &  & SPL   & Success & DTS (m) \\ \midrule
\methodabbr{} w.o. Semantic Map         &  & 0.165 & 0.488   & 2.084   \\
\methodabbr{} w.o. Goal Policy          &  & 0.148 & 0.450   & 2.315   \\ 
\methodabbr{}                           &  & 0.199 & 0.544   & 1.723   \\
\methodabbr{} w. GT SemSeg              &  & 0.457 & 0.731   & 1.089   \\ \bottomrule
\end{tabular}
\caption{\small \textbf{Ablations and Error Analysis.} Table showing comparison of the proposed model, \methodabbr{}, with 2 ablations and with Ground Truth semantic segmentation on the Gibson dataset.}
\label{tab_sem_exp:ablations}
\end{table}

\begin{figure*}[t]
\centering
\includegraphics[width=0.97\linewidth,height=\textheight,keepaspectratio]{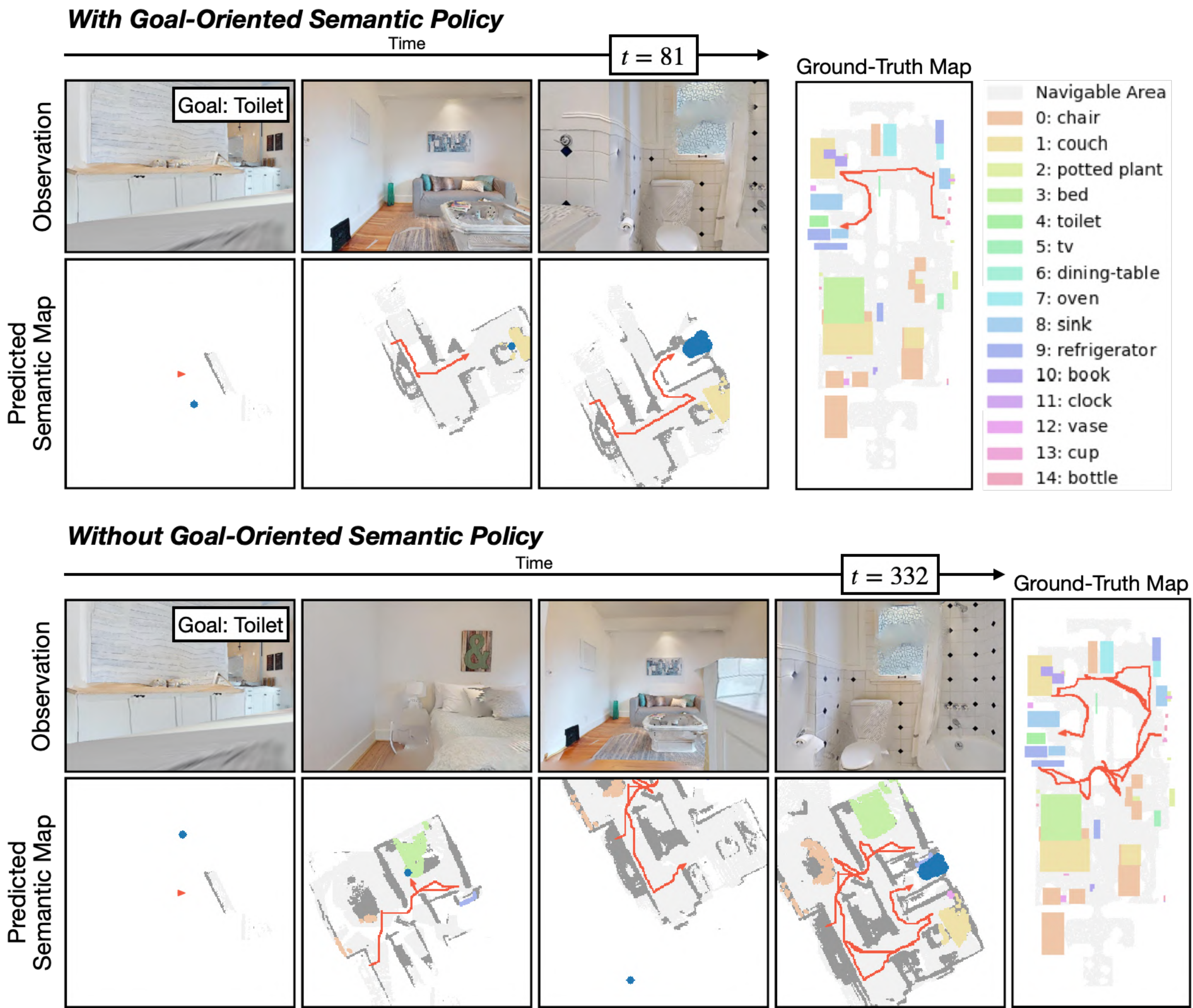}
\caption{Figure showing an example comparing the proposed model with (\textbf{top}) and without (\textbf{bottom}) Goal-Oriented Semantic Policy. Starting at the same location with the same goal object of `toilet', the proposed model with Goal-Oriented Policy can find the target object much faster than without Goal-Oriented Exploration.}
\label{fig_sem_exp:ablation}
\vspace{-7pt}
\end{figure*}

\subsection{Ablations and Error Analysis}
\vspace{-7pt}
To understand the importance of both the modules in \methodabbr{}, we consider two ablations:\\{\vspace{-8pt}

\noindent\textbf{\methodabbr{} w.o. Semantic Map.} We replace the Semantic Map with the Obstacle-only Map. As opposed to the Active Neural SLAM baseline, the Goal-oriented Policy is still trained with distance reduced to the nearest object as the reward.\\{\vspace{-8pt}

\noindent\textbf{\methodabbr{} w.o. Goal Policy.} We replace the Goal-driven policy with a goal-agnostic policy trained to maximize exploration coverage as in~\cite{ans}, but still trained with the semantic map as input.\\{\vspace{-8pt}

The results in the top part of Table~\ref{tab_sem_exp:ablations} show the performance of these ablations. The performance of \methodabbr{} without the Goal-oriented policy is comparable to the Active Neural SLAM baseline, indicating that the Goal-oriented policy learns semantic priors for better exploration leading to more efficient navigation. Figure~\ref{fig_sem_exp:ablation} shows an qualitative example indicating the importance of the Goal-oriented Policy. The performance of \methodabbr{} without the Semantic Map also drops, but it is higher than the ablation without Goal Policy. This indicates that it is possible to learn some semantic priors with just the obstacle map without semantic channels.

The performance of the proposed model is still far from perfect. We would like to understand the error modes for future improvements. We observed two main sources of errors, semantic segmentation inaccuracies and inability to find the goal object. In order to quantify the effect of both the error modes, we evaluate the proposed model with ground truth semantic segmentation (see \textbf{\methodabbr{} w. GT SemSeg} in Table~\ref{tab_sem_exp:ablations}) using the `Semantic' sensor in Habitat simulator. This leads to a success rate of 73.1\% vs 54.4\%, which means around 19\% performance can be improved with better semantic segmentation. The rest of the ~27\% episodes are mostly cases where the goal object is not found, which can be improved with better semantic exploration.

\subsection{Result on Habitat Challenge}
\label{sec_sem_exp:challenge}
\vspace{-7pt}
\methodabbr{} also won the CVPR 2020 Habitat ObjectNav Challenge. The challenge task setup is identical to ours except the goal object categories. The challenge uses 21 object categories for the goal object. As these object categories are not overlapping with the COCO categories, we use DeepLabv3~\cite{chen2017rethinking} model for semantic segmentation. We predict the semantic segmentation and the semantic map with all 40 categories in the MP3D dataset including the 21 goal categories. We fine-tune the DeepLabv3 segmentation model by retraining the final layer to predict semantic segmentation for all 40 categories. This segmentation model is also trained with the map-based loss in addition to the first-person segmentation loss. The performance of the top 5 entries to the challenge are shown in Table~\ref{tab_sem_exp:challenge}. The proposed approach outperforms all the other entries with a success rate of 25.3\% as compared to 18.8\% for the second place entry. 

\setlength{\tabcolsep}{15pt}
\begin{table}[]
\centering
\begin{tabular}{@{}llccc@{}}
\toprule
Team Name                        &  & SPL   & Success & Dist To Goal (m) \\ \midrule
	\textbf{SemExp}                  &  & \textbf{0.102} & \textbf{0.253}   & \textbf{6.328}            \\
SRCB-robot-sudoer                &  & 0.099 & 0.188   & 6.908            \\
Active Exploration (Pre-explore) &  & 0.046 & 0.126   & 7.336            \\
Black Sheep                      &  & 0.028 & 0.101   & 7.033            \\
Blue Ox                          &  & 0.021 & 0.069   & 7.233            \\ \bottomrule
\end{tabular}
\caption{\small \textbf{Results on CVPR 2020 Habitat ObjectNav Challenge.} Table showing the performance of top 5 entries on the Test-Challenge dataset. Our submission based on the \methodabbr{} model won the challenge.}
\label{tab_sem_exp:challenge}
\end{table}

\begin{figure*}[t!]
\centering
\includegraphics[width=0.99\linewidth,height=\textheight,keepaspectratio]{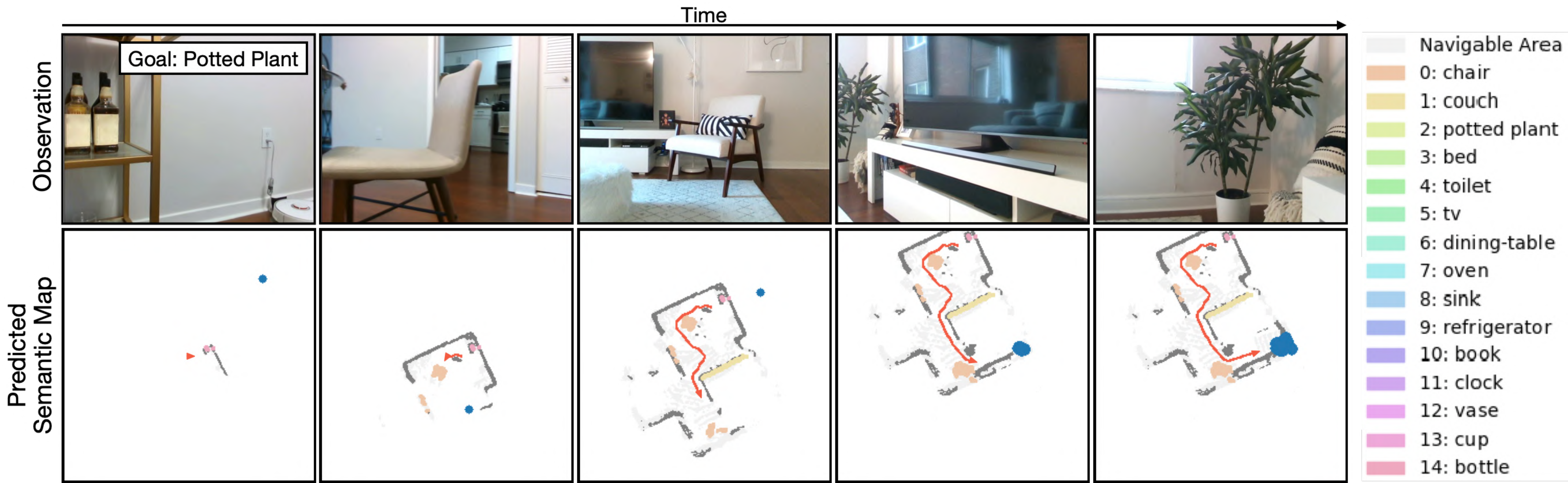}
\caption{\small \textbf{Real-world Transfer.} Figure showing an example trajectory of the \methodabbr{} model transferred to the real-world for the object goal `potted plant'. Sample images seen by the robot are shown on the top and the predicted semantic map is shown below. The long-term goal selected by the Goal-driven Policy is shown in blue.}
\label{fig_sem_exp:real_world_living}
\end{figure*}

\vspace{-4pt}
\subsection{Real World Transfer}
\vspace{-7pt}
We used the Locobot hardware platform and PyRobot API~\citep{pyrobot2019} to deploy the trained policy in the real world. 
In Figure ~\ref{fig_sem_exp:real_world_living} we show an episode of the robot when it was provided `potted plant' as the object goal. The long-term goals sampled by the Goal-driven policy (shown by blue circles on the map) are often towards spaces where there are high chances of finding a potted plant. This indicates that it is learning to exploit the structure in the semantic map to some extent. Out of 20 trials in the real-world, our method succeeded in 13 episodes leading to a success rate of 65\%. 
End-to-end learning-based policies failed consistently in the real-world due to the visual domain gap between simulation environments and the real-world. Our model performs well in the real-world as (1) it is able to leverage Mask RCNN which is trained on real-world data and (2) the other learned modules (map denoising and the goal policy) work on top-down maps which are domain-agnostic. Our trials in the real-world also indicate that perfect pose and depth are not critical to the success of our model as it can be successfully transferred to the real-world where pose and depth are noisy. 

\section{Discussion}
\vspace{-7pt}
In this chapter, we presented a semantically-aware exploration model to tackle the object goal navigation task in large realistic environments. The proposed model makes two major improvements over prior methods, incorporating semantics in explicit episodic memory and learning goal-oriented semantic exploration policies. Our method achieves state-of-the-art performance on the object goal navigation task and won the CVPR2020 Habitat ObjectNav challenge. Ablation studies show that the proposed model learns semantic priors which lead to more efficient goal-driven navigation. Domain-agnostic module design led to successful transfer of our model to the real-world. We also analyze the error modes for our model and quantify the scope for improvement along two important dimensions (semantic mapping and goal-oriented exploration) in the future work. The proposed model can also be extended to tackle a sequence of object goals by utilizing the episodic map for more efficient navigation for subsequent goals.

The semantic mapping learnt by the SemExp model can also be used for active visual learning, where an embodied agent actively explores the environment to collect observations to learn and improve perception models. In the next chapter, we explore a self-supervised approach for training an exploration policy for improving an object detection model.

\chapter{Semantic Curiosity}
\label{chap:semcur}
\blfootnote{Webpage: \url{https://devendrachaplot.github.io/projects/SemanticCuriosity}}

Imagine an agent whose goal is to learn how to detect and categorize objects. How should the agent learn this task? In the case of humans (especially babies), learning is quite interactive in nature. We have the knowledge of what we know and what we don't, and we use that knowledge to guide our future experiences/supervision. Compare this to how current algorithms learn -- we create datasets of random images from the internet and label them, followed by model learning. The model has no control over what data and what supervision it gets -- it is resigned to the static biased dataset of internet images. Why does current learning look quite different from how humans learn? During the early 2000s, as data-driven approaches started to gain acceptance, the computer vision community struggled with comparisons and knowing which approaches work and which don't. As a consequence, the community introduced several benchmarks from BSDS~\cite{MartinFTM01} to VOC~\cite{pascal-voc-2007}. However, a negative side effect of these benchmarks was the use of static training and test datasets. While the pioneering works in computer vision focused on active vision and interactive learning, most of the work in the last two decades focuses on static internet vision. But as things start to work on the model side, we believe it is critical to look at the big picture again and return our focus to an embodied and interactive learning setup.

\begin{figure}[t]
    \centering
    \includegraphics[width=0.75\linewidth]{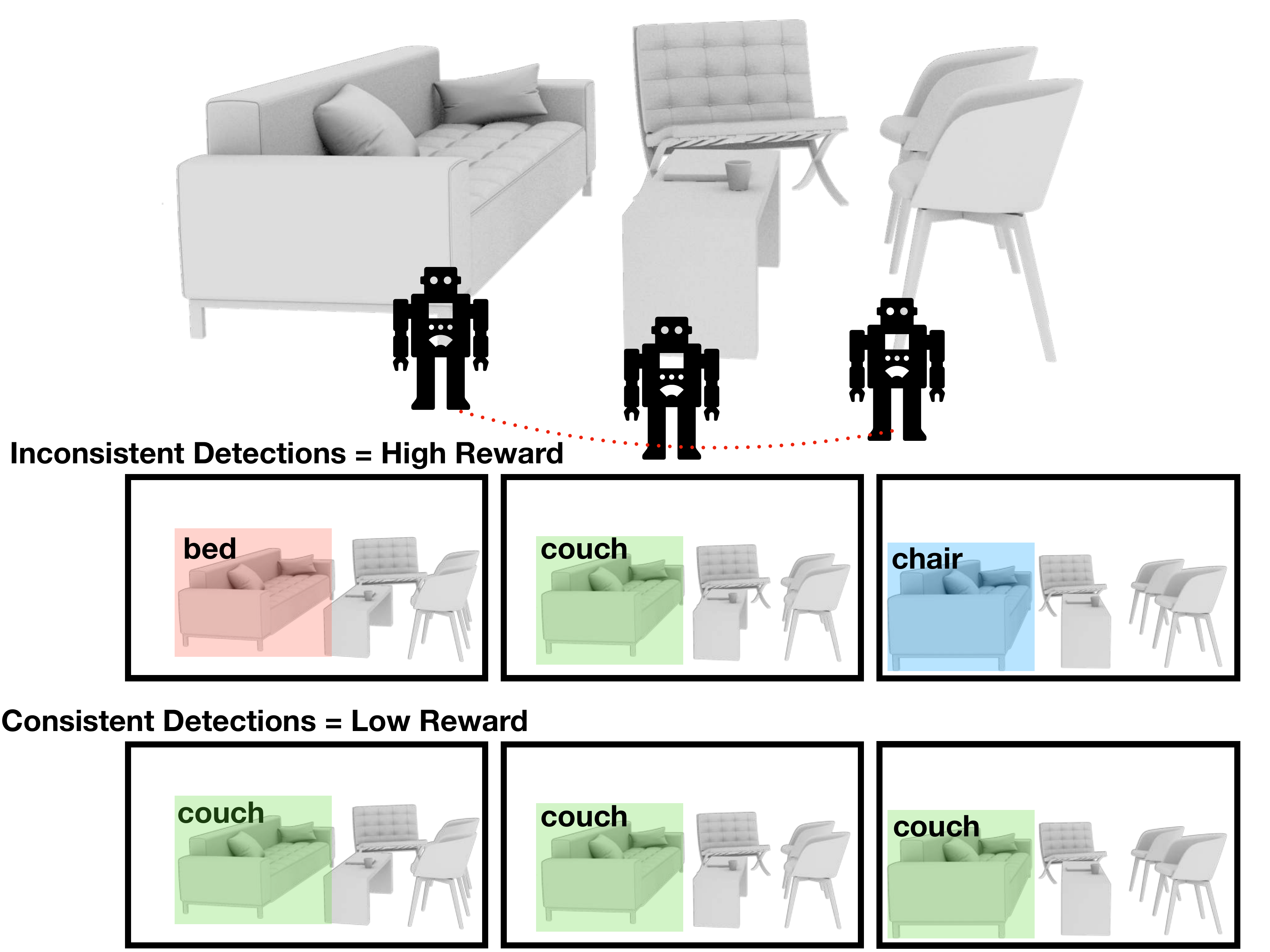}
    \caption{\textbf{Semantic Curiosity}: We propose semantic curiosity to learn exploration for training object detectors. Our semantically curious policy attempts to take actions such that the object detector will produce inconsistent outputs.}
    \label{fig_sem_cur:my_label}
\end{figure}

In an embodied interactive learning setup, an agent has to perform actions such that observations generated from these actions can be useful for learning to perform the semantic task. Several core research questions need to be answered: (a) what is the policy of exploration that generates these observations? (b) what should be labeled in these observations - one object, one frame, or the whole trajectory? (c) and finally, how do we get these labels? In this chapter, we focus on the first task -- what should the exploration policy be to generate observations which can be useful in training an object detector? Instead of using labels, we focus on learning these trajectories in an unsupervised/self-supervised manner. Once the policy has been learned, we use the policy in novel (previously unseen) scenes to perform actions. As observations are generated, we assume that an oracle will densely label all the objects of interest in the trajectories. 

So what are the characteristics of a good exploration policy for visual learning, and how do we learn it? A good semantic exploration policy is one which generates observations of objects and not free-space or the wall/ceiling. But not only should the observations be objects, but a good exploration policy should also observe many unique objects. Finally, a good exploration policy will move to parts of the observation space where the current object detection model fails or does not work. Given these characteristics, how should we define a reward function that could be used to learn this exploration policy? Note, as one of the primary requirements, we assume the policy is learned in a self-supervised manner -- that is, we do not have ground-truth objects labeled which can help us figure out where the detections work or fail. 

Inspired by recent work in intrinsic motivation and curiosity for training policies without external rewards~\cite{pathak2017curiosity,pathak19disagreement}, we propose a new intrinsic reward called semantic curiosity that can be used for the exploration and training of semantic object detectors. In the standard curiosity reward, a policy is rewarded if the predicted future observation does not match the true future observation. The loss is generally formulated in the pixel-based feature space. A corresponding reward function for semantic exploration would be to compare semantic predictions with the current model and then confirm with ground-truth labels -- however, this requires external labels (and hence is not self-supervised anymore). Instead, we formulate semantic curiosity based on the meta-supervisory signal of consistency in semantic prediction -- that is, if our model truly understands the object, it should predict the same label for the object even as we move around and change viewpoints. Therefore, we exploit consistency in label prediction to reward our policies. Our semantic curiosity rewards trajectories which lead to inconsistent labeling behavior of the same object by the semantic classifier. Our experiments indicate that training an exploration policy via semantic curiosity generalizes to novel scenes and helps train an object detector which outperforms baselines trained with other possible alternatives such as random exploration, pixel curiosity, and free space/map curiosity. We also perform a large set of experiments to understand the behavior of a policy trained with semantic curiosity.

We study the problem of how to sample training data in embodied contexts. This is related to active learning (picking what sample to label), active perception (how to move around to gain more information), intrinsic motivation (picking what parts of the environment to explore). Learning in embodied contexts can also leverage spatio-temporal consistency. We survey these research areas below. 

\textbf{Active Perception.} Active perception~\cite{bajcsy1988active} refers to the problem of actively moving the sensors around at \textit{test time} to improve performance on the task by gaining more information about the environment. This has been instantiated in the context of object detection~\cite{ammirato2017dataset}, amodal object detection~\cite{yang2019embodied}, scene completion~\cite{jayaraman2018learning}, and localization~\cite{chaplot2018active, fox1998active}. We consider the problem in a different setting and study how to efficiently move around to best \textit{learn} a model. Furthermore, our approach to learn this movement policy is self-supervised and does not rely on end-task rewards, which were used in~\cite{jayaraman2018learning, chaplot2018active, yang2019embodied}. 

\textbf{Active Learning.} Our problem is more related to that of active learning~\cite{settles2009active}, where an agent actively acquires labels for unlabeled data to improve its model at the fastest rate~\cite{sener2017active, gal2017deep, yoo2019learning}. This has been used in a number of applications such as medical image analysis~\cite{kuo2018cost}, training object detectors~\cite{vijayanarasimhan2014large, yang2018visual}, video segmentation~\cite{fathi2011combining}, and visual question answering~\cite{misra2017lba}. Most works tackle the setting in which the unlabeled data has already been collected. In contrast, we study learning a policy for efficiently acquiring effective unlabeled data, which is complementary to such active learning efforts.

\textbf{Intrinsic Rewards.}
Our work is also related to work on exploration in reinforcement learning~\cite{schmidhuber1991possibility, sutton1998rli, auer2002using, jaksch2010near}. The goal of these works is to effectively explore a Markov Decision Process to find high reward paths. A number of works formulate this as a problem of maximizing an intrinsic reward function which is designed to incentivize the agent to seek previously unseen~\cite{eysenbach2018diversity} or poorly understood~\cite{pathak2017curiosity} parts of the environment. This is similar to our work, as we also seek poorly understood parts of the environment. However, we measure this understanding via multi-view consistency in semantics. This is in a departure from existing works that measure it in 2D image space~\cite{pathak2017curiosity}, or consistency among multiple models~\cite{pathak19disagreement}. Furthermore, our focus is not effective exhaustive exploration, but exploration for the purpose of improving semantic models.

\textbf{Spatio-Temporal smoothing at test time.}
A number of papers use spatio-temporal consistency at test time for better and more consistent predictions~\cite{chandra2018deep, gadde2017semantic}. Much like the distinction from active perception, our focus is using it to generate better data at train time.

\textbf{Spatio-temporal consistency as training signal.} 
Labels have been propagated in videos to simplify annotations~\cite{vatic}, improve prediction performance given limited data~\cite{badrinarayanan2010label, bengio200611}, as well as collect images~\cite{chen2013neil}. 
This line of work leverages spatio-temporal consistency to propagate labels for more efficient labeling. 
Researchers have also used multi-view consistency to learn about 3D shape from weak supervision~\cite{drcTulsiani19}. 
We instead leverage spatio-temporal consistency as a cue to identify what the model does not know. \cite{siddiqui2019viewal} is more directly related, but we tackle the problem in an embodied context and study how to navigate to gather the data, rather than analyzing passive datasets for what labels to acquire.

\textbf{Visual Navigation and Exploration.} Prior work on visual navigation can broadly be categorized into two classes based on whether the location of the goal is known or unknown. Navigation scenarios, where the location of the goal is known, include the most common \textit{pointgoal} task where the coordinate to the goal is given~\cite{gupta2017cognitive, savva2017minos}. Another example of a task in this category is vision and language navigation~\cite{anderson2018vision} where the path to the goal is described in natural language. Tasks in this category do not require exhaustive exploration of the environment as the location of the goal is known explicitly (coordinates) or implicitly (path). 

Navigation scenarios, where the location of the goal is not known, include a wide variety of tasks. These include navigating to a fixed set of objects~\cite{lample2016playing, dosovitskiy2016learning, wu2016training, chaplot2017arnold, mirowski2016learning, gupta2017cognitive}, navigating to an object specified by language~\cite{hermann2017grounded, chaplot2017gated} or by an image~\cite{zhu2017target, chaplot2020neural}, and navigating to a set of objects in order to answer a question~\cite{das2017embodied, gordon2018iqa}. Tasks in this second category essentially involve efficiently and exhaustively exploring the environment to search the desired object. However, most of the above approaches overlook the exploration problem by spawning the target a few steps away from the goal and instead focus on other challenges. For example, models for playing FPS games~\cite{lample2016playing, dosovitskiy2016learning, wu2016training, chaplot2017arnold} show that end-to-end RL policies are effective at reactive navigation and short-term planning such as avoiding obstacles and picking positive reward objects as they randomly appear in the environment. Other works show that learned policies are effective at tackling challenges such as perception (in recognizing the visual goal)~\cite{zhu2017target, chaplot2020neural}, grounding (in understanding the goal described by language)~\cite{hermann2017grounded, chaplot2017gated} or reasoning (about visual properties of the target object)~\cite{das2017embodied, gordon2018iqa}. While end-to-end reinforcement learning is shown to be effective in addressing these challenges, they are ineffective at exhaustive exploration and long-term planning in a large environment as the exploration search space increases exponentially as the distance to the goal increases. 

Some very recent works explicitly tackle the problem of exploration by training end-to-end RL policies maximizing the explored area~\cite{ans, chen2018learning, fang2019scene}. The difference between these approaches and our method is twofold: first, we train semantically-aware exploration policies as compared spatial coverage maximization in some prior works~\cite{ans, chen2018learning}, and second, we train our policy in an unsupervised fashion, without requiring any ground truth labels from the simulator as compared to prior works trained using rewards based on ground-truth labels~\cite{fang2019scene}.

\section{Overview}
Our goal is to learn an exploration policy such that if we use this policy to generate trajectories in a novel scene (and hence observations) and train the object detector from the trajectory data, it would provide a robust, high-performance detector. In literature, most approaches use on-policy exploration; that is, they use the external reward to train the policy itself. However, training an action policy to sample training data for object recognition requires labeling objects. Specifically, these approaches would use the current detector to predict objects and compare them to the ground-truth; they reward the policy if the predictions do not match the ground-truth (the policy is being rewarded to explore regions where the current object detection model fails). However, training such a policy via semantic supervision and external rewards would have a huge bottleneck of supervision. Given that our RL policies require millions of samples (in our case, we train using 10M samples), using ground-truth supervision is clearly not the way. What we need is an intrinsic motivation reward that can help train a policy which can help sample training data for object detectors.

We propose a semantic curiosity formulation. Our work is inspired by a plethora of efforts in active learning~\cite{settles2009active} and recent work on intrinsic reward using disagreement~\cite{pathak19disagreement}. The core idea is simple -- a good object detector has not only high mAP performance but is also consistent in predictions. That is, the detector should predict the same label for different views of the same object. We use this meta-signal of consistency to train our action policy by rewarding trajectories that expose inconsistencies in an object detector. We measure inconsistencies by measuring temporal entropy of prediction -- that is, if an object is labeled with different classes as the viewpoint changes, it will have high temporal entropy. The trajectories with high temporal entropy are then labeled via an oracle and used as the data to retrain the object detector (See Figure~\ref{fig_sem_cur:overview}). 

\begin{figure*}[t]
    \centering
    \includegraphics[width=1\linewidth]{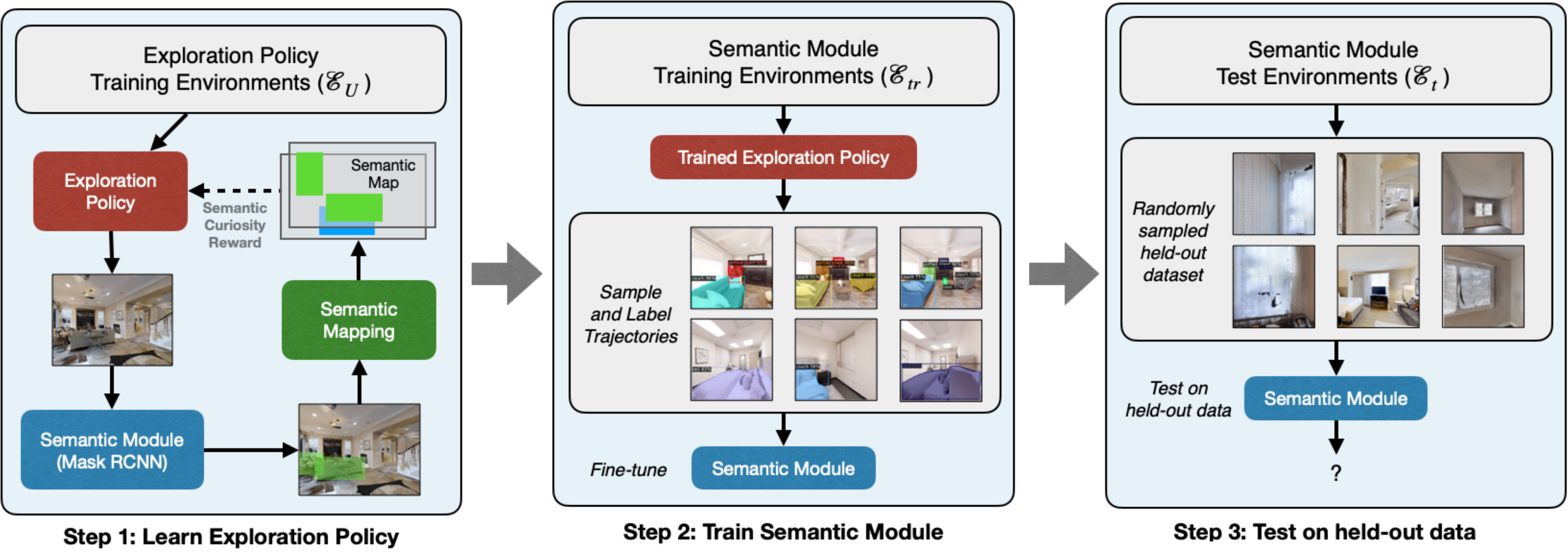}
    \caption{\textbf{Embodied Active Visual Learning}: We use semantic curiosity to learn an exploration policy on $\mathcal{E}_U$ scenes. The exploration policy is learned by projecting segmentation masks on the top-down view to create semantic maps. The entropy of semantic map defines the inconsistency of the object detection module. The learned exploration policy is then used to generate training data for the  object detection/segmentation module. The labeled data is then used to finetune and evaluate the object detection/segmentation.}
    \label{fig_sem_cur:overview}
\end{figure*}

\section{Methodology}
Consider an agent $\mathcal{A}$ which can perform actions in environments $\mathcal{E}$. The agent has an exploration policy $a_t = \pi(x_t,\theta)$ that predicts the action that the agent must take for current observation $x_t$. $\theta$ represents the parameters of the policy that have to be learned. The agent also has an object detection model $\mathcal{O}$ which takes as input an image (the current observation) and generates a set of bounding boxes along with their categories and confidence scores. 

The goal is to learn an exploration policy $\pi$, which is used to sample $N$ trajectories $\tau_1...\tau_N$ in a set of novel environments (and get them semantically labeled). When used to train an object detector, this labeled data would yield a high-performance object detector. In our setup, we divide the environments into three non-overlapping sets $(\mathcal{E}_U, \mathcal{E}_{tr}, \mathcal{E}_t)$ -- the first set is the set of environments where the agent will learn the exploration policy $\pi$, the second set is the object detection training environments where we use $\pi$ to sample trajectories and label them, and the third set is the test environment where we sample random images and test the performance of the object detector on those images.

\begin{figure*}
    \centering
    \includegraphics[width=1\linewidth]{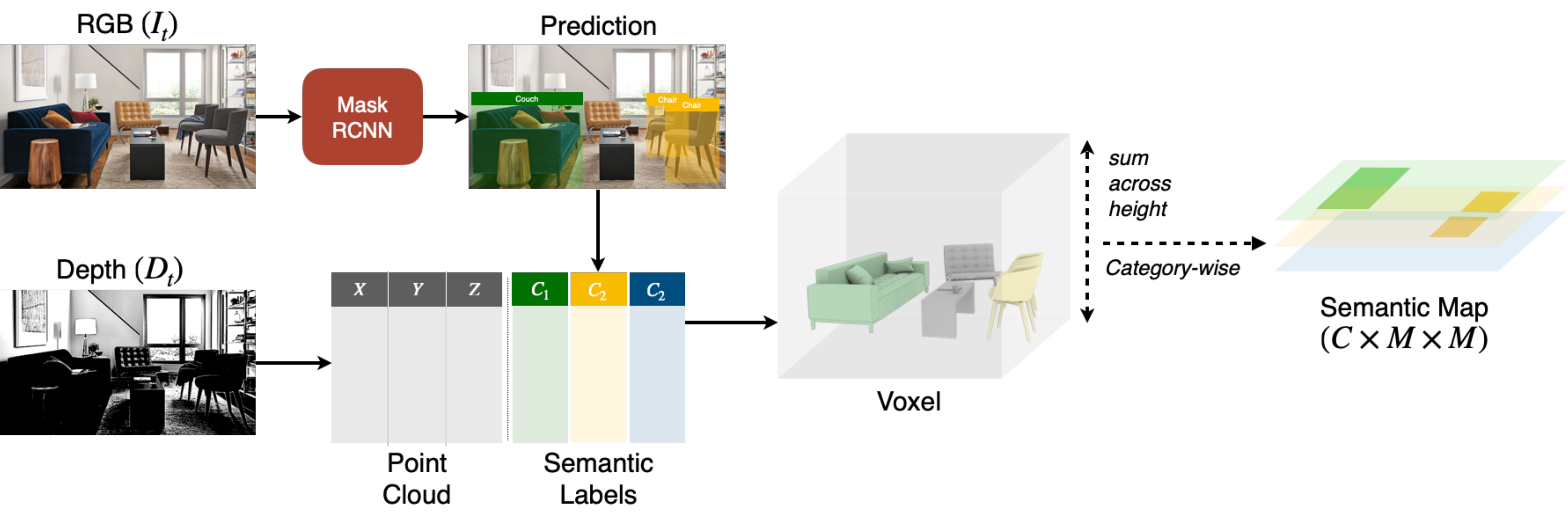}
    \caption{\textbf{Semantic Mapping.} The Semantic Mapping module takes in a sequence of RGB ($I_t$) and Depth ($D_t$) images and produces a top-down Semantic Map.}
    \label{fig_sem_cur:semantic mapping}
\end{figure*}

\subsection{Semantic Curiosity Policy}
\label{sec_sem_cur:semcur}
We define semantic curiosity as the temporal inconsistency in object detection and segmentation predictions from the current model. We use a Mask RCNN to obtain the object predictions. In order to associate the predictions across frames in a trajectory, we use a semantic mapping module as described below.\\

\noindent\textbf{Semantic Mapping.} The Semantic Mapping module takes in a sequence of RGB ($I_t$) and Depth ($D_t$) images and produces a top-down semantic map ($M^{Sem}_t$) represented by a 3-dimensional tensor $C \times M \times M$, where $M$ is the length of the square top-down map, and $C$ is the number of semantic categories. Each element $(c, i, j)$ in this semantic map is $1$ if the Mask RCNN predicted the object category $c$ at coordinates $(i,j)$ on the map in any frame during the whole trajectory and $0$ otherwise. Figure~\ref{fig_sem_cur:semantic mapping} shows how the semantic map is generated for a single frame. The input RGB frame ($I_t$) is passed through a current Mask RCNN to obtain object segmentation predictions, while the Depth frame is used to calculate the point cloud. Each point in the point cloud is associated with its semantic labels based on Mask RCNN predictions. Note that these are not ground-truth labels, as each pixel is assigned the category of the highest-confidence Mask RCNN segmentation prediction on the corresponding pixel. The point cloud with the associated semantic labels is projected to a 3-dimensional voxel map using geometric computations. The voxel representation is converted to a top-down map by max-pooling the values across the height. The resulting 2D map is converted to a 3-dimensional Semantic Map, such that each channel represents an object category.

The above gives a first-person egocentric projection of the semantic map at each time-step. The egocentric projections at each time step are used to compute a geocentric map over time using a spatial transformation technique similar to Chaplot et al.~\cite{ans}. The egocentric projections are converted to the geocentric projections by doing a spatial transformation based on the agent pose. The semantic map at each time step is computed by pooling the semantic map at the previous timestep with the current geocentric prediction. Please refer to~\cite{ans} for more details on these transformations. \\

\noindent\textbf{Semantic Curiosity Reward.} The semantic map allows us to associate the object predictions across different frames as the agent is moving. We define the semantic curiosity reward based on the temporal inconsistency of the object predictions. If an object is predicted to have different labels across different frames, multiple channels in the semantic map at the coordinates corresponding to the object will have $1$s. Such inconsistencies are beneficial for visual learning in downstream tasks, and hence, favorable for the semantic curiosity policy. Thus, we define the cumulative semantic curiosity reward to be proportional to the sum of all the elements in the semantic map. Consequently, the semantic curiosity reward per step is just the increase in the sum of all elements in the semantic map as compared to the previous time step:
$$
r_{SC} = \lambda_{SC} \Sigma_{(c,i,j) \in (C,M,M)} (M^{Sem}_t[c,i,j] - M^{Sem}_{t-1}[c,i,j])
$$
where $\lambda_{SC}$ is the semantic curiosity reward coefficient. Summation over the channels encourages exploring frames with temporally inconsistent predictions. Summation across the coordinates encourages exploring as many objects with temporally inconsistent predictions as possible. 

The proposed Semantic Curiosity Policy is trained using reinforcement learning to maximize the cumulative semantic curiosity reward. Note that although the depth image and agent pose are used to compute the semantic reward, we train the policy only on RGB images.

\section{Experimental Setup}
We use the Habitat simulator~\cite{savva2019habitat} with three different datasets for our experiments: Gibson~\cite{gibsonenv}, Matterport~\cite{Matterport3D} and Replica~\cite{replica19arxiv}.  While the RGB images used in our experiments are visually realistic as they are based on real-world reconstructions, we note that the agent pose and depth images in the simulator are noise-free unlike the real-world. Prior work has shown that both depth and agent pose can be estimated effectively from RGB images under noisy odometry~\cite{ans}. In this chapter, we assume access to perfect agent pose and depth images, as these challenges are orthogonal to the focus of this chapter. Furthermore, these assumptions are only required in the unsupervised pre-training phase for calculating the semantic curiosity reward and not at inference time when our trained semantic-curiosity policy (based only on RGB images) is used to gather exploration trajectories for training the object detector.\\

In a perfectly interactive learning setup, the current model's uncertainty will be used to sample a trajectory in a new scene, followed by labeling and updating the learned visual model (Mask-RCNN). However, due to the complexity of this online training mechanism, we show results on batch training. We use a pre-trained COCO Mask-RCNN as an initial model and train the exploration policy on that model. Once the exploration policy is trained, we collect trajectories in the training environments and then obtain the labels on these trajectories. The labeled examples are then used to fine-tune the Mask-RCNN detector.

\subsection{Implementation details}
\noindent {\bf Exploration Policy:} We train our semantic curiosity policy on the Gibson dataset and test it on the Matterport and Replica datasets. We train the policy on the set of 72 training scenes in the Gibson dataset specified by Savva et al.~\cite{savva2019habitat}. Our policy is trained with reinforcement learning using Proximal Policy Optimization~\cite{schulman2017proximal}. The policy architecture consists of convolutional layers of a pre-trained ResNet18 visual encoder, followed by two fully connected layers and a GRU layer leading to action distribution as well as value prediction. We use 72 parallel threads (one for each scene) with a time horizon on 100 steps and 36 mini batches per PPO epoch. The curiosity reward coefficient is set to $\lambda_{SC} = 2.5\times10^{-3}$. We use an entropy coefficient of 0.001, the value loss coefficient of 0.5. We use Adam optimizer with a learning rate of $1\times10^{-5}$.  The maximum episode length during training is 500 steps.

\noindent {\bf Fine-tuned Object Detector:} We consider 5 classes of objects, chosen because they overlap with the COCO dataset~\cite{lin2014microsoft} and correspond to objects commonly seen in an indoor scene: chair, bed, toilet, couch, and potted plant. To start, we pre-train a Faster-RCNN model~\cite{ren2015faster} with FPN~\cite{lin2017feature} using ResNet-50 as the backbone on the COCO dataset labeled with these 5 overlapping categories. We then fine-tuned our models on the trajectories collected by the exploration policies for 90000 iterations using a batch size of 12 and a learning rate of 0.001, with annealing by a factor of 0.1 at iterations 60000 and 80000. We use the Detectron2 codebase~\cite{wu2019detectron2} and set all other hyperparameters to their defaults in this codebase. We compute the AP50 score (i.e., average precision using an IoU threshold of 50) on the validation set every 5000 iterations. 

\begin{figure}[t]
    \centering
    \includegraphics[width=0.9\linewidth]{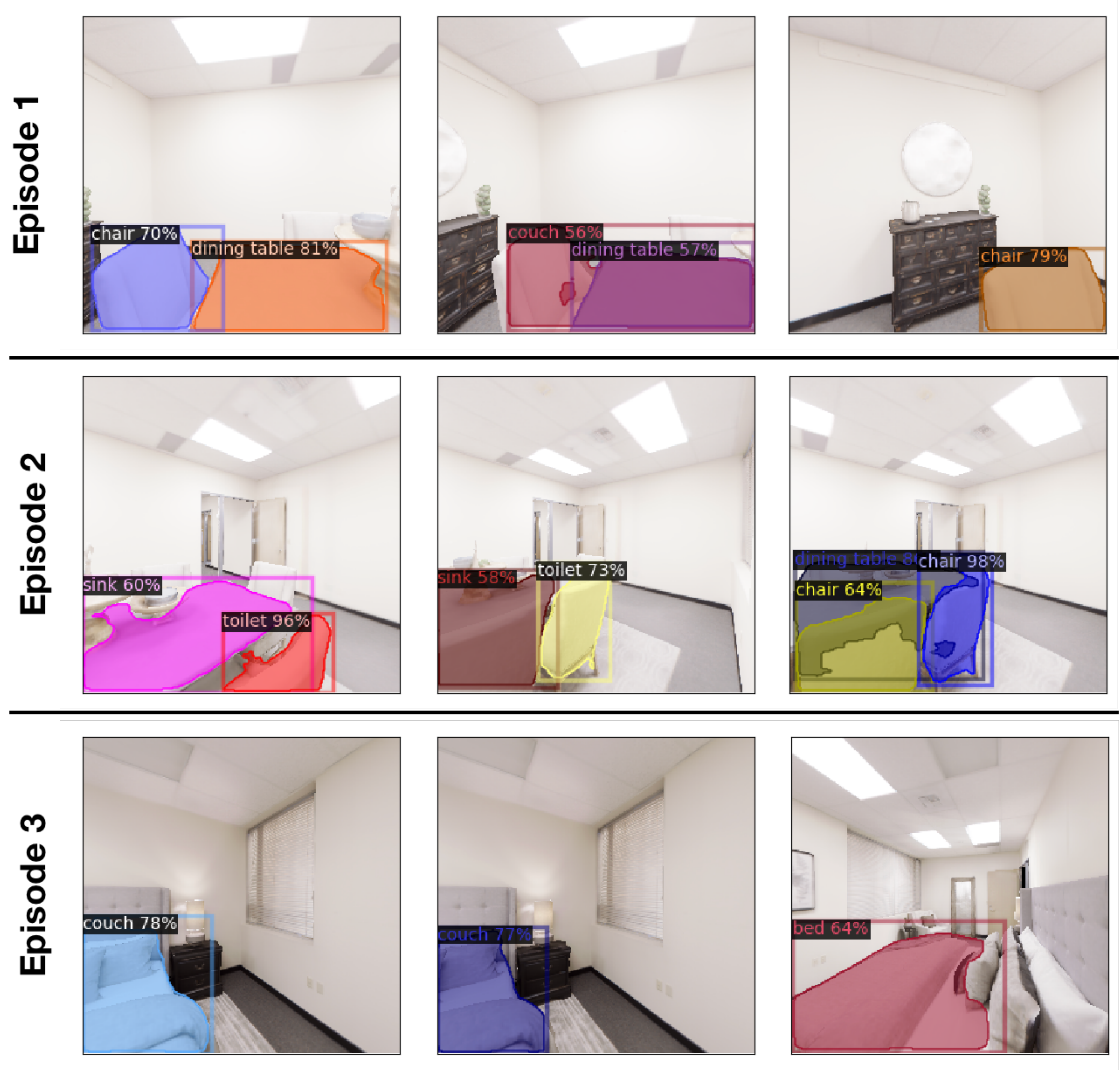}
    \caption{\textbf{Temporal Inconsistency Examples.} Figure showing example trajectories sampled from the semantic curiosity exploration policy. We highlight the segmentation/detection inconsistencies of Mask RCNN. By obtaining labels for these images, the Mask RCNN pipeline improves the detection performance significantly.}
  
    \label{fig_sem_cur:inconsistent}
\end{figure}

\subsection{Baselines}
We use a range of baselines to gather exploration trajectories and compare them to the proposed Semantic Curiosity policy:

\begin{itemize}
\item \textbf{Random.} A baseline sampling actions randomly.
\item \textbf{Prediction Error Curiosity.} This baseline is based on Pathak et al.~\cite{pathak2017curiosity}, which trains an RL policy to maximize error in a forward prediction model.
\item \textbf{Object Exploration.} Object Exploration is a naive baseline where an RL policy is trained to maximize the number of pre-trained Mask R-CNN detections. The limitation of simply maximizing the number of detections is that the policy can learn to look at frames with more objects but might not learn to look at different objects across frames or objects with low confidence.
\item \textbf{Coverage Exploration.} This baseline is based on Chen et al.~\cite{chen2018learning}, where an RL policy is trained to maximize the total explored area.
\item \textbf{Active Neural SLAM.} This baseline is based on Chaplot et al.~\cite{ans} and uses a modular and hierarchical system to maximize the total explored area.
\end{itemize}

After training the proposed policy and the baselines in the Gibson domain, we use them directly (without fine-tuning) in the Matterport and Replica domains. We sample trajectories using each exploration policy, using the images and ground-truth labels to train an object detection model. 

\setlength{\tabcolsep}{10.0pt}
\begin{table}[]
\centering
\begin{tabular}{@{}llccc@{}}
\toprule
Method Name               &  & \begin{tabular}[c]{@{}c@{}}Semantic Curiosity\\ Reward\end{tabular} & \begin{tabular}[c]{@{}c@{}}Explored \\ Area\end{tabular} & \begin{tabular}[c]{@{}c@{}}Number of Object \\ Detections\end{tabular} \\ \midrule
Random                    &  & 1.631                                                              & 4.794                                                    & 82.83                                                          \\
Curiosity                 &  & 2.891                                                              & 6.781                                                    & 112.24                                                         \\
Object exploration reward &  & 2.168                                                              & 6.082                                                    & 382.27                                                         \\
Coverage Exploration      &  & 3.287                                                              & 10.025                                                   & 203.73                                                         \\
Active Neural SLAM        &  & 3.589                                                              & 11.527                                                   & 231.86                                                         \\
Semantic Curiosity        &  & 4.378                                                              & 9.726                                                    & 291.78                                                         \\ \bottomrule
\end{tabular}
\caption{\textbf{Analysis.} Comparing the proposed Semantic Curiosity policy with the baselines along different exploration metrics.}
\label{tab_sem_cur:analysis}
\end{table}

\setlength{\tabcolsep}{8.4pt}
\begin{table*}[t]
\centering
\begin{tabular}{@{}lcccccc@{}}
\toprule
Method Name          & Chair     & Bed     & Toilet    & Couch    & Potted Plant   & Average  \\
\midrule
Random                   & 46.7     & 28.2     & 46.9        & 60.3     & 39.1 & 44.24 \\
Curiosity  &  49.4   &    18.3 &  1.8  &  67.7    & 49.0  &  37.42 \\
Object Exploration   & 54.3   &   24.8  & 5.7    &   76.6   & 49.6 & 42.2         \\
Coverage Exploration &  48.5   & 23.1  & 69.2  & 66.3 & 48.0  & 51.02  \\
Active Neural SLAM   & 51.3     & 20.5     & 49.4      & 59.7    & 45.6 & 45.3  \\
Semantic Curiosity  & 51.6      & 14.6    & 14.2      & 65.2     & 50.4 & 39.2  \\\bottomrule   
\end{tabular}
\caption{\textbf{Quality of object detection on training trajectories.} We also analyze the training trajectories in terms of how well the pre-trained object detection model works on the trajectories. We want the exploration policy to sample hard data where the pre-trained object detector fails. Data on which the pre-trained model already works well would not be useful for fine-tuning. Thus, lower performance is better.}
\label{tab_sem_cur:analysis2}
\end{table*}

\section{Analyzing Learned Exploration Behavior}
Before we measure the quality of the learned exploration policy for the task of detection/segmentation, we first want to analyze the behavior of the learned policy. This will help characterize the quality of data that is gathered by the exploration policy. We will compare the learned exploration policy against the baselines described above. For all the experiments below, we trained our policy on Gibson scenes and collected statistics in 11 Replica scenes. 

Figure~\ref{fig_sem_cur:inconsistent} shows some examples of temporal inconsistencies in trajectories sampled using the semantic curiosity exploration policy. The pre-trained Mask-RCNN detections are also shown in the observation images. Semantic curiosity prefers trajectories with inconsistent detections. In the top row, the chair and couch detector fire on the same object. In the middle row, the chair is misclassified as a toilet and there is inconsistent labeling in the last trajectory. The bed is misclassified as a couch. By selecting these trajectories and obtaining their labels from an oracle, our approach learns to improve the object detection module.

Table~\ref{tab_sem_cur:analysis} shows the behavior of all of the policies on three different metrics. The first metric is the semantic curiosity reward itself which measures uncertain detections in the trajectory data. Since our policy is trained for this reward, it gets the highest score on the sampled trajectories. The second metric is the amount of explored area. Both \cite{chen2018learning} and \cite{ans} optimize this metric and hence perform the best (they cover a lot of area but most of these areas will either not have objects or not enough contradictory overlapping detections). The third metric is the number of objects in the trajectories. The object exploration baseline optimizes for this reward and hence performs the best but it does so without exploring diverse areas or uncertain detections/segmentations. If we look at the three metrics together it is clear that our policy has the right tradeoff -- it explores a lot of area but still focuses on areas where objects can be detected. Not only does it find a large number of object detections, but our policy also prefers inconsistent object detections and segmentations. In Figure~\ref{fig_sem_cur:example_traj}, we show some examples of trajectories seen by the semantic curiosity exploration along with the semantic map. It shows examples of the same object having different object predictions from different viewpoints and also the representation in the semantic map. In Figure~\ref{fig_sem_cur:Comparison}, we show a qualitative comparison of maps and objects explored by the proposed model and all the baselines. Example trajectories in this figure indicate that the semantic curiosity policy explores more unique objects with higher temporal inconsistencies.

\begin{figure*}[t]
    \centering
    \includegraphics[width=0.99\linewidth]{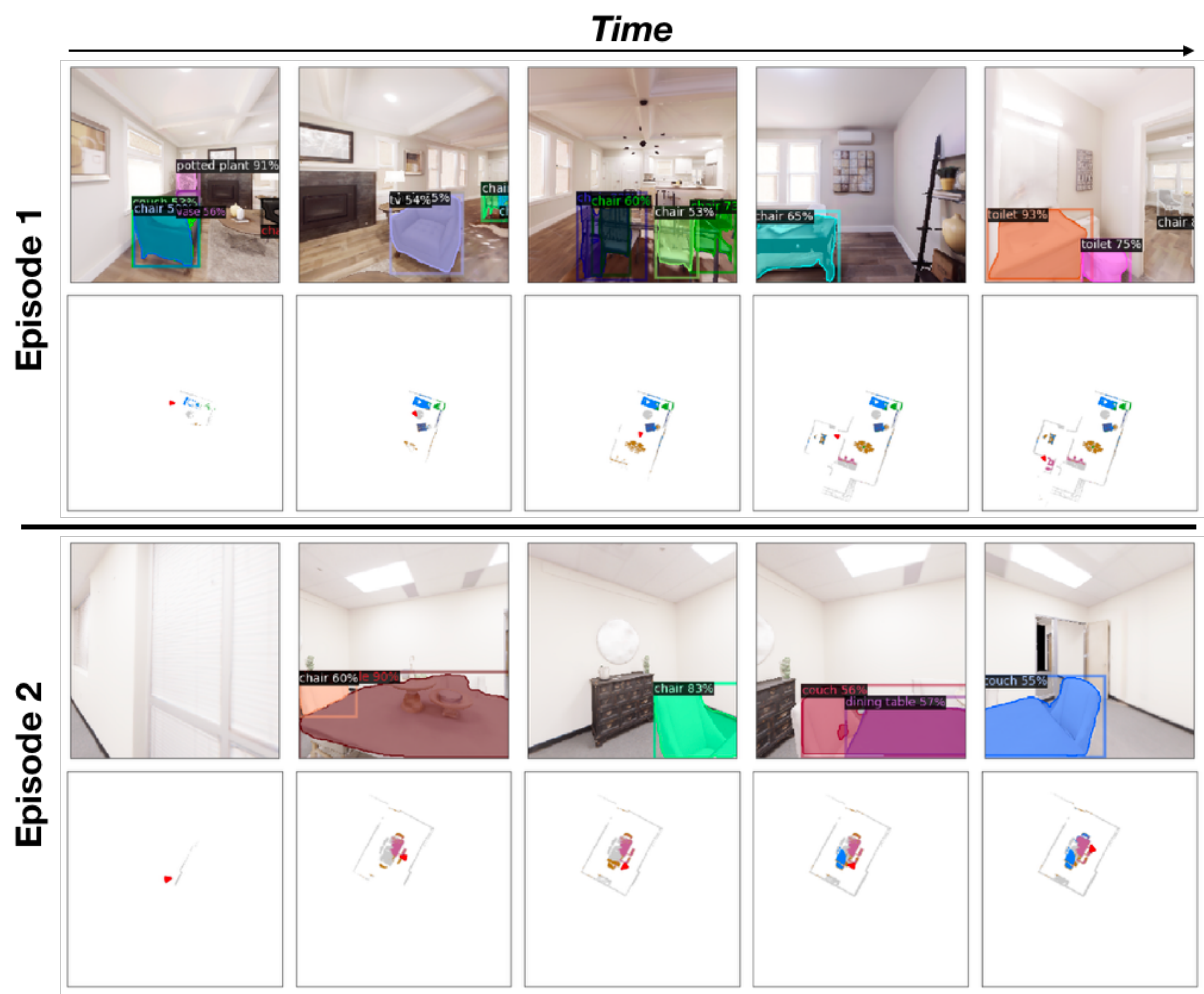}
    \caption{\textbf{Example trajectories.} Figure showing example trajectories sampled from the semantic curiosity exploration policy. In each episode the top row shows the first-person images seen by the agent and the pre-trained Mask R-CNN predictions. The bottom rows show a visualization of the semantic map where colors denote different object categories. Different colors for the same object indicate that the same object is predicted to have different categories from different view points.}
    \label{fig_sem_cur:example_traj}
\end{figure*}

Next, we analyze the trajectories created by different exploration policies during the object detection training stage. Specifically, we want to analyze the kind of data that is sampled by these trajectories. How is the performance of a pre-trained detector on this data? If the pre-trained detector already works well on the sampled trajectories, we would not see much improvement in performance by fine-tuning with this data. In Table~\ref{tab_sem_cur:analysis2}, we show the results of this analysis for these trajectories. As the results indicate, the mAP50 score is low for the data obtained by the semantic curiosity policy.\footnote{Note that curiosity-based policy has the lowest mAP because of outlier toilet category.} As the pre-trained object detector fails more on the data sampled by semantic curiosity, labeling this data would intuitively improve the detection performance. 

\begin{figure}[]
    \centering
    \includegraphics[width=0.99\linewidth]{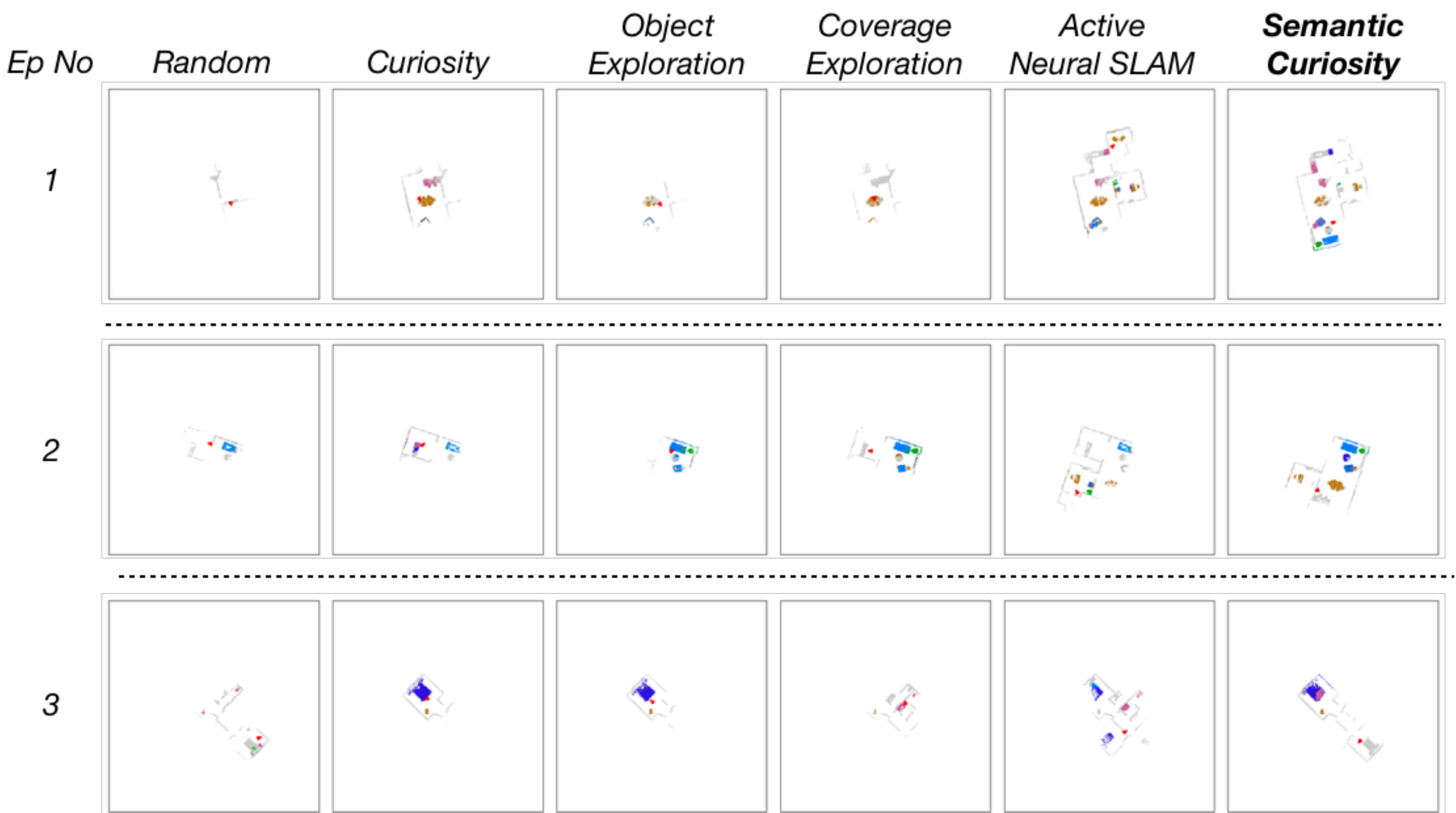}
    \caption{\textbf{Qualitative Comparison.} Figure showing map and objects explored by the proposed Semantic Curiosity policy and the baselines in 3 example episodes. Semantic Curiosity Policy explores more unique objects with higher temporal inconsistency (denoted by different colors for the same object).}
    \label{fig_sem_cur:Comparison}
\end{figure}

\setlength{\tabcolsep}{8.4pt}
\begin{table}[t]
\centering
\begin{tabular}{@{}lcccccc@{}}
\toprule
Method Name          & Chair     & Bed     & Toilet    & Couch    & Potted Plant   & Average  \\
\midrule
PreTrained               & 41.8      & 17.3     & 34.9      & 41.6     & 23.0    & 31.72   \\
Random                   & 51.7      & 17.2     & 43.0        & 45.1     & 30.0  & 37.4    \\
Curiosity  & 48.4     &  18.5   &  42.3     &   44.3   &   32.8 &  37.26 \\
Object Exploration   & 50.3   & 16.4    &   40.0  &   39.7   & 29.9 &  35.26         \\
Coverage Exploration &  50.0    & 19.1    &   38.1    &  42.1   &  33.5 &  36.56      \\
Active Neural SLAM   & 53.1     & 19.5     & 42.0      & 44.5    & 33.4 & 38.5  \\
Semantic Curiosity   & 52.3      & 22.6    & 42.9      & 45.7     & 36.3 & \textbf{39.96}  \\\bottomrule   
\end{tabular}
\caption{\textbf{Object Detection Results.} Object detection results in the Matterport domain using the proposed Semantic Curiosity policy and the baselines. We report AP50 scores on randomly sampled images in the test scenes. Training on data gathered from the semantic curiosity trajectories results in improved object detection scores.}
\label{tab_sem_cur:trained_ap}
\end{table}

\section{Actively Learned Object Detection}
Finally, we evaluate the performance of our semantic curiosity policy for the task of object detection. The semantic curiosity exploration policy is trained on 72 Gibson scenes. The exploration policy is then used to sample data on 50 Matterport scenes. Finally, the learned object detector is tested on 11 Matterport scenes. For each training scene, we sample 5 trajectories of 300 timesteps leading to 75,000 total training images with ground-truth labels. For test scenes, we randomly sample images from test scenes.

In Table~\ref{tab_sem_cur:trained_ap}, we report the top AP50 scores for each method. Our results demonstrate that the proposed semantic curiosity policy obtains higher quality data for performing object detection tasks over alternative methods of exploration. First, we outperform the policy that tries to see maximum coverage area (and hence the most novel images). Second, our approach also outperforms the policy that detects a lot of objects. Finally, apart from outperforming the random policy, visual curiosity~\cite{pathak2017curiosity}, and coverage; we also outperform the highly-tuned approach of ~\cite{ans}. The underlying algorithm is tuned on this data and was the winner of the RGB and RGBD challenge in Habitat.

\section{Discussion}
We argue that we should go from detection/segmentation driven by static datasets to a more embodied active learning setting. In this setting, an agent can move in the scene and create its own datapoints. An oracle labels these datapoints and helps the agent learn a better semantic object detector. This setting is closer to how humans learn to detect and recognize objects. In this chapter, we focus on the exploration policy for sampling images to be labeled. We ask a basic question -- how should an agent explore to learn how to detect objects? Should the agent try to cover as many scenes as possible in the hopes of seeing more diverse examples, or should the agent focus on observing as many objects as possible? 

We propose semantic curiosity as a reward to train the exploration policy. Semantic curiosity encourages trajectories which will lead to inconsistent detection behavior from an object detector. Our experiments indicate that exploration driven by semantic curiosity shows all of the good characteristics of an exploration policy: uncertain/high entropy detections, attention to objects rather than the entire scene and also high coverage for diverse training data. We also show that an object detector trained on trajectories from a semantic curiosity policy leads to the best performance compared to a plethora of baselines. For future work, this chapter is just the first step in embodied active visual learning. It assumes perfect odometry, localization and zero trajectory labeling costs. It also assumes that the trajectories will be labeled -- a topic of interest would be to sample trajectories with which minimal labels can learn the best detector. Finally, the current approach is demonstrated in simulators - it will be interesting to see whether the performance can transfer to real-world robots.

\chapter{Conclusion}
\label{chap:conc}
In this thesis, we presented several methods which make progress towards building intelligent autonomous navigation agents. In chapters~\ref{chap:fps} and \ref{chap:grounding}, we presented end-to-end reinforcement learning methods to tackle \textit{short-term navigation} tasks involving both spatial and semantic understanding. We trained navigation policies in game-like environments which learn fine-grained motor from only raw-pixels and are able to tackle challenges such as obstacle avoidance, semantic perception, language grounding, and reasoning. They are able to learn only from rewards and generalize to unseen environments and unseen language instructions.

While the above method and many other embodied navigational agents trained with end-to-end reinforcement learning are shown to perform well on reactive tasks, where the optimal behavior depends only on the last few observations, they struggle in larger environments requiring long-term memory and planning. In the second part of this thesis, we designed a new class of navigation methods based on modular learning and structured explicit map representations, which leverage the strengths of both classical and end-to-end learning methods. We presented different modular learning models in chapters~\ref{chap:anl} to \ref{chap:semcur}, tackling different challenges involved in long-term spatial and semantic navigation tasks. In this chapter, we summarize different methods presented in this thesis along with their strengths and shortcomings and discuss future directions to address these shortcomings. 

\begin{figure*}[t]
\centering
\includegraphics[width=0.99\linewidth,height=\textheight,keepaspectratio]{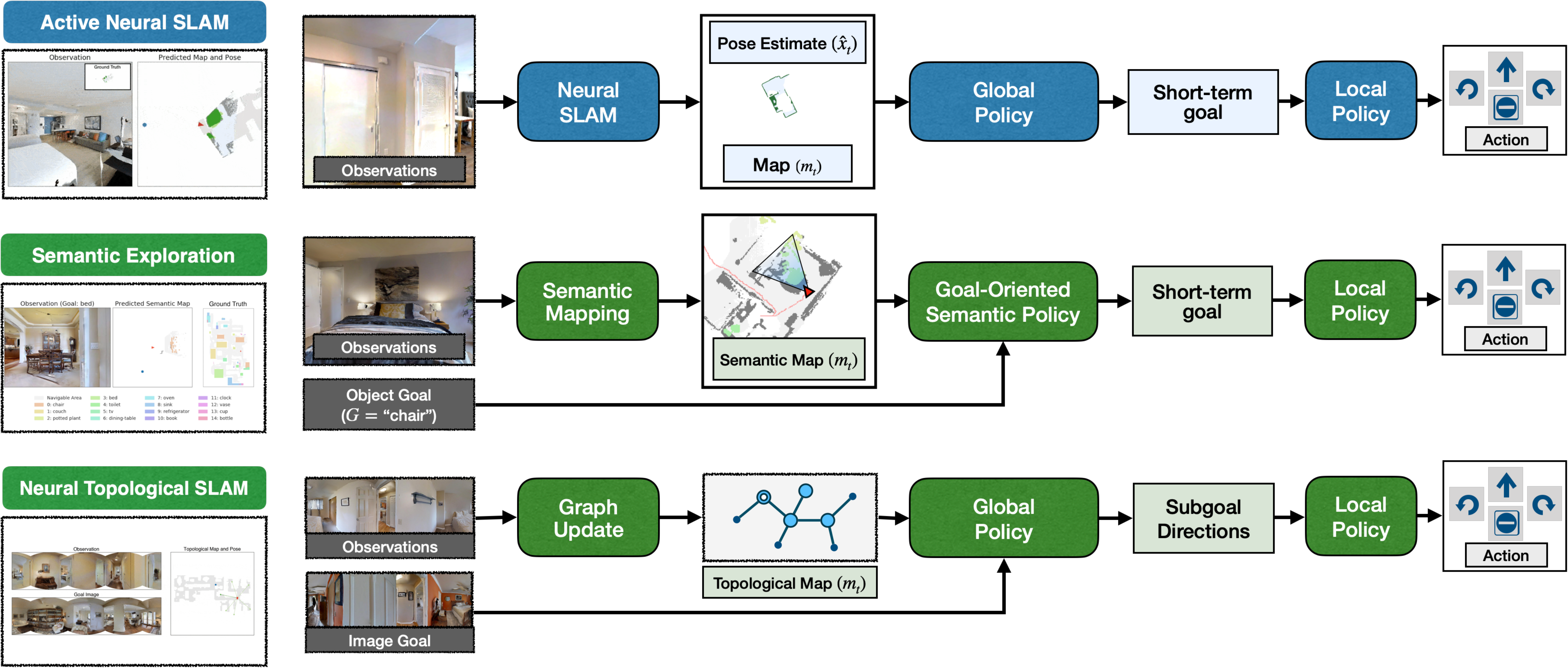}
	\caption{\small \textbf{Modular Learning-based Navigation Methods.} Figure showing the similarities in the design of Active Neural SLAM, Semantic Exploration and Neural Topological SLAM models presented in this thesis.}
\label{conc_fig:mod_nav}
\end{figure*}

\section{Summary}
The modular navigation methods presented in this thesis include Active Neural SLAM (Ch.~\ref{chap:ans}) for the task of exploration and Point-Goal Navigation, Neural Topological SLAM (Ch.~\ref{chap:nts}) for Image-Goal Navigation and Semantic Exploration (Ch.~\ref{chap:semexp}) for Object-Goal Navigation. The design of all these models is very similar as shown in Figure~\ref{conc_fig:mod_nav}, which consists of three modules: the first module is responsible for perception, localization, mapping, and building an episodic memory, the second module is responsible for long-term waypoint selection based on the task and the third module is responsible for low-level control and obstacle avoidance.

We compared each of the above modular learning methods to both end-to-end learning as well classical navigation methods and observed consistent trends. The end-to-end learning-based models performed well at short-term spatial as well as semantic understanding tasks, but as we increased the size of the environment, and increased the distance to the goal location, end-to-end learning methods consistently led to poor performance for all different navigation tasks and goal specifications. This is because implicitly learning about mapping and planning from just rewards is difficult and exploration sample complexity increases exponentially with increase in the distance to the goal.

The classical navigation approaches outperformed end-to-end learning-based approaches at some long-term spatial navigation tasks given access to a depth sensor. The use of explicit maps allows classical methods to be very effective at spatial or geometric understanding of the scene. Explicit maps are also an effective form of long-term memory and can be used for planning to distant goals easily. However, they lack semantic understanding and can not tackle semantic navigation tasks as they build only geometric maps and use a rule-based policy. Also, their scalability is limited as the long-term policy or goal locations need to be manually coded, they typically require pre-computed maps, and they often rely on specialized sensors.

\begin{figure*}[t]
\centering
\includegraphics[width=0.99\linewidth,height=\textheight,keepaspectratio]{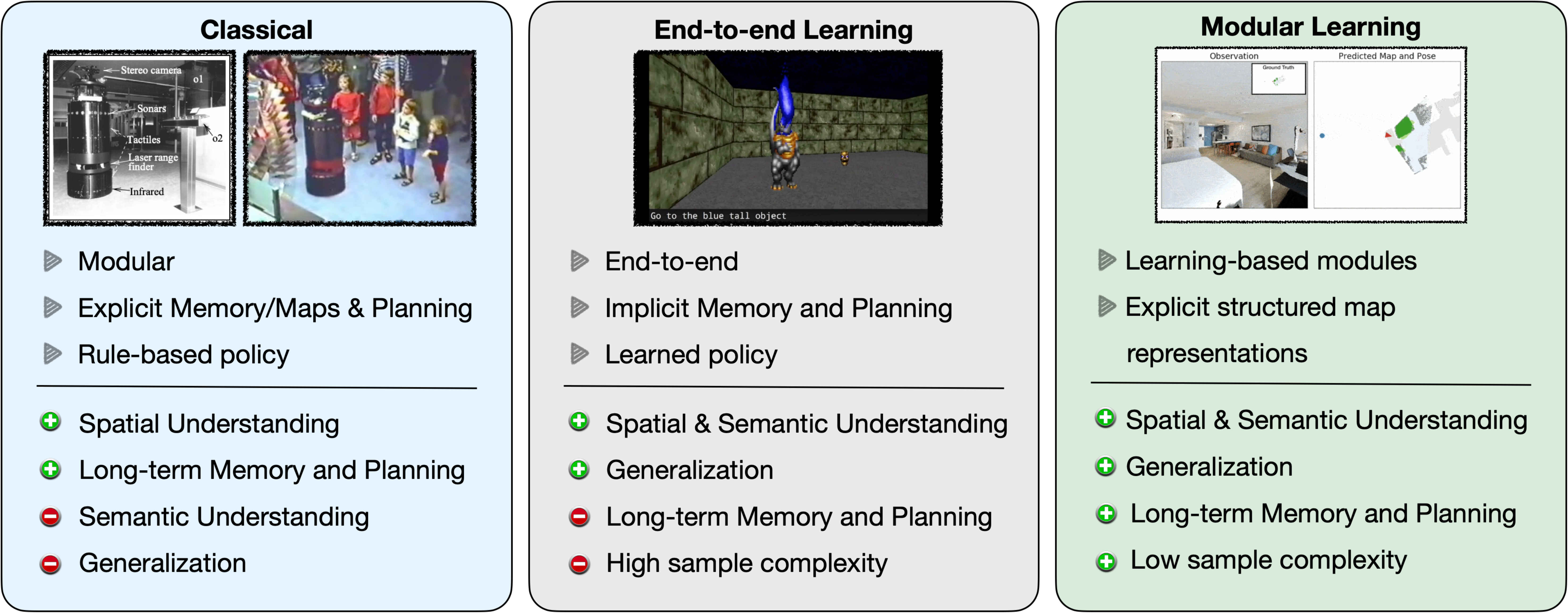}
	\caption{\small \textbf{Advantages of Modular Learning.} Figure summarizing the proerties, advantages and disadvantages of navigation systems based on classical, end-to-end learning and modular learning approaches. Modular Learniing combines the benefits of the other two classes of methods.}
\label{conc_fig:modular_learning_advantages}
\end{figure*} 

Our modular learning methods leverage the strengths of both classical and end-to-end learning methods, using two keys properties: first, they modularize the navigation system but use learning to train each module, and second, they use explicit structured map representations. Explicit maps are an effective form of episodic memory which allow keeping a track of explored and unexplored areas. They also allow us to use analytical planning algorithms like classical approaches which is much more effective than learning to plan implicitly just from rewards. At the same time, the use of learning within each module allows us to learn statistical regularities and semantic priors present in the data and generalize to unseen environments and goals. We saw that modular learning methods consistently outperformed both classical as well as end-to-end learning methods for various long-term navigation tasks such as exploration, image-goal navigation, and object-goal navigation. Figure~\ref{conc_fig:modular_learning_advantages} summarizes the advantages of modular learning approaches over both classical methods as well as end-to-end learning.

\begin{figure*}[t]
\centering
\includegraphics[width=0.99\linewidth,height=\textheight,keepaspectratio]{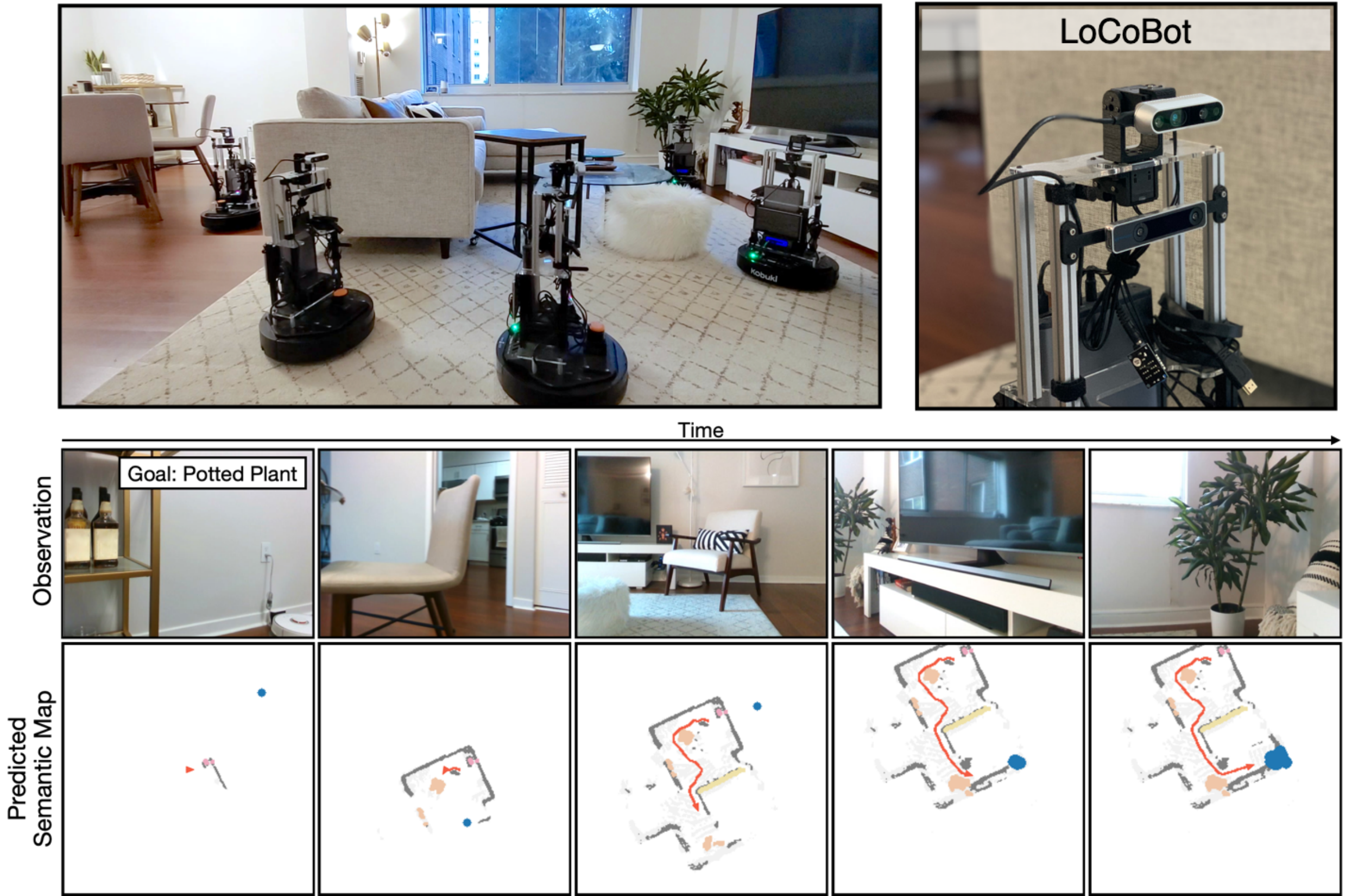}
	\caption{\small \textbf{Real-World Navigation}. Figure showing an example of the Goal-Oriented Semantic Exploration model (Chapter~\ref{chap:semexp}) deployed on a mobile robot and used for Object Goal navigation.}
\label{fig:rw_eg}
\end{figure*}

Another benefit of our modular learning methods is effective \textbf{real-world transfer}. While end-to-end learning methods typically suffer from the physical and visual domain gap between simulation and the real-world, we were able to successfully transfer the Active Neural SLAM and Semantic Exploration models to a mobile robot platform in the real-world and achieve similar performance at both exploration and object-goal navigation (Figure~\ref{fig:rw_eg} shows an example). The effective real-world transfer was a result of two key approaches:

\begin{itemize}
\item \textbf{Modularity:} Due to the modularity, the long-term policy in both Active Neural SLAM and Semantic Exploration models just takes the top-down map as input and does not require first-person observations. The top-down map space is domain-invariant which allows for easy transfer of long-term policies from simulation to the real-world. Furthermore, modularity also allows us to leverage the MaskRCNN semantic segmentation model which is pretrained on real-world data. This allows us to bridge the visual domain gap between simulation and the real-world.
\item \textbf{Realistic noise models:} In chapter~\ref{chap:ans}, we built actuation and motion noise models using data collected in the real-world. We implemented these noise models in simulation and used them for training Pose Estimator modules in our methods. Realistic motion and actuation noise models allow us to bridge the physical domain gap and lead to much better real-world transfer.
\end{itemize}

\section{Future Directions}

Modular learning methods achieved reasonably good performance at spatial or geometric navigation tasks achieving $92\%$ coverage in the Exploration task and $95\%$ success rate in Point-Goal Navigation in our experiments. The majority of the errors were attributed to errors in mapping. The semantic navigation tasks still have ample room for improvement. Our methods achieve around $55-65\%$ success rate at Object-Goal and Image-Goal Navigation tasks. We found the errors were approximately equally attributed to inefficient exploration and ineffective perception. This means that in around $20\%$ of the episodes, the goal location was not found and in another $20\%$ of the episodes, the goal location was found but the agent did not decide to stop at the goal location. Both exploration efficiency and perception performance can be improved in future methods.

A key limitation of all modular learning methods presented is inefficient low-level control. Our methods are efficient in terms of the number of actions but not in terms of wall-clock time. We assumed a discrete action space which leads to slow low-level control when deployed on a physical robot. We also assumed sequential processing of observations, taking physical actions, and sampling next observations. This means the robot needs to stop after taking each action, take the next observation from the sensors, process the observations to compute the action, and then take the next action. Asynchronous observation processing and motion will lead to much more efficient navigation in terms of wall-clock time. This will introduce challenges such as motion blur, lowering model inference time, and tackling the delay between perception and control, which need to be tackled in the future.

Another big avenue of research is the use of modular models for Embodied Active Visual Learning. In biological beings, perception and action is essentially a loop where perception is used to perform actions in the world, and actions are used to better perceive the world. Navigation is just one part of the loop where we are using perception to move in the world, but inversely we can use actions to learn and improve perception. In an embodied active learning setup, an agent has to perform actions such that observations generated from these actions can be useful for learning better perception. We barely scratched the surface of the potential of modular models for active visual learning in Chapter~\ref{chap:semcur}. Several core research questions need to be answered for large-scale applications: (a) what is the policy of exploration that generates these observations? (b) what should be labeled in these observations - one object, one frame, or the whole trajectory? (c) and finally, how do we get these labels? Active Visual Learning can potentially enable machine learning models to overcome the requirement of large-scale annotated datasets and learn more efficiently with fewer labels by being able to look at the world from different viewpoints, just like humans.

\backmatter



\bibliographystyle{plainnat}
\bibliography{references} 

\begin{thebibliography}{222}
\providecommand{\natexlab}[1]{#1}
\providecommand{\url}[1]{\texttt{#1}}
\expandafter\ifx\csname urlstyle\endcsname\relax
  \providecommand{\doi}[1]{doi: #1}\else
  \providecommand{\doi}{doi: \begingroup \urlstyle{rm}\Url}\fi

\bibitem[hab({\natexlab{a}})]{habitatchallenge}
{Habitat Challenge 2019}.
\newblock \url{https://aihabitat.org/challenge/2019/}, {\natexlab{a}}.

\bibitem[hab({\natexlab{b}})]{habitatchallenge2020}
{Habitat Challenge 2020}.
\newblock \url{https://aihabitat.org/challenge/2020/}, {\natexlab{b}}.

\bibitem[Ammirato et~al.(2017)Ammirato, Poirson, Park, Ko{\v{s}}eck{\'a}, and
  Berg]{ammirato2017dataset}
Phil Ammirato, Patrick Poirson, Eunbyung Park, Jana Ko{\v{s}}eck{\'a}, and
  Alexander~C Berg.
\newblock A dataset for developing and benchmarking active vision.
\newblock In \emph{2017 IEEE International Conference on Robotics and
  Automation (ICRA)}, pages 1378--1385. IEEE, 2017.

\bibitem[Anderson et~al.(2018{\natexlab{a}})Anderson, Chang, Chaplot,
  Dosovitskiy, Gupta, Koltun, Kosecka, Malik, Mottaghi, Savva, and
  Zamir]{anderson2018evaluation}
Peter Anderson, Angel Chang, Devendra~Singh Chaplot, Alexey Dosovitskiy,
  Saurabh Gupta, Vladlen Koltun, Jana Kosecka, Jitendra Malik, Roozbeh
  Mottaghi, Manolis Savva, and Amir Zamir.
\newblock On evaluation of embodied navigation agents.
\newblock \emph{arXiv preprint arXiv:1807.06757}, 2018{\natexlab{a}}.

\bibitem[Anderson et~al.(2018{\natexlab{b}})Anderson, Wu, Teney, Bruce,
  Johnson, S{\"u}nderhauf, Reid, Gould, and van~den Hengel]{anderson2018vision}
Peter Anderson, Qi~Wu, Damien Teney, Jake Bruce, Mark Johnson, Niko
  S{\"u}nderhauf, Ian Reid, Stephen Gould, and Anton van~den Hengel.
\newblock Vision-and-language navigation: Interpreting visually-grounded
  navigation instructions in real environments.
\newblock In \emph{Proceedings of the IEEE Conference on Computer Vision and
  Pattern Recognition}, pages 3674--3683, 2018{\natexlab{b}}.

\bibitem[Armeni et~al.(2019)Armeni, He, Gwak, Zamir, Fischer, Malik, and
  Savarese]{armeni20193d}
Iro Armeni, Zhi-Yang He, JunYoung Gwak, Amir~R Zamir, Martin Fischer, Jitendra
  Malik, and Silvio Savarese.
\newblock 3d scene graph: A structure for unified semantics, 3d space, and
  camera.
\newblock In \emph{Proceedings of the IEEE International Conference on Computer
  Vision}, pages 5664--5673, 2019.

\bibitem[Artzi and Zettlemoyer(2013)]{artzi2013weakly}
Yoav Artzi and Luke Zettlemoyer.
\newblock Weakly supervised learning of semantic parsers for mapping
  instructions to actions.
\newblock \emph{Transactions of the Association for Computational Linguistics},
  1:\penalty0 49--62, 2013.

\bibitem[Auer(2002)]{auer2002using}
Peter Auer.
\newblock Using confidence bounds for exploitation-exploration trade-offs.
\newblock \emph{Journal of Machine Learning Research}, 3\penalty0
  (Nov):\penalty0 397--422, 2002.

\bibitem[Ba et~al.(2014)Ba, Mnih, and Kavukcuoglu]{ba2014multiple}
Jimmy Ba, Volodymyr Mnih, and Koray Kavukcuoglu.
\newblock Multiple object recognition with visual attention.
\newblock \emph{arXiv preprint arXiv:1412.7755}, 2014.

\bibitem[Badrinarayanan et~al.(2010)Badrinarayanan, Galasso, and
  Cipolla]{badrinarayanan2010label}
Vijay Badrinarayanan, Fabio Galasso, and Roberto Cipolla.
\newblock Label propagation in video sequences.
\newblock In \emph{2010 IEEE Computer Society Conference on Computer Vision and
  Pattern Recognition}, pages 3265--3272. IEEE, 2010.

\bibitem[Bagnell(2015)]{bagnell2015invitation}
J~Andrew Bagnell.
\newblock An invitation to imitation.
\newblock Technical report, DTIC Document, 2015.

\bibitem[Bajcsy(1988)]{bajcsy1988active}
Ruzena Bajcsy.
\newblock Active perception.
\newblock \emph{Proceedings of the IEEE}, 76\penalty0 (8):\penalty0 966--1005,
  1988.

\bibitem[Bansal et~al.(2019)Bansal, Tolani, Gupta, Malik, and
  Tomlin]{bansal2019combining}
Somil Bansal, Varun Tolani, Saurabh Gupta, Jitendra Malik, and Claire Tomlin.
\newblock Combining optimal control and learning for visual navigation in novel
  environments.
\newblock \emph{arXiv preprint arXiv:1903.02531}, 2019.

\bibitem[Barto and Mahadevan(2003)]{barto2003recent}
Andrew~G Barto and Sridhar Mahadevan.
\newblock Recent advances in hierarchical reinforcement learning.
\newblock \emph{Discrete event dynamic systems}, 13\penalty0 (1-2):\penalty0
  41--77, 2003.

\bibitem[Batra et~al.(2020)Batra, Gokaslan, Kembhavi, Maksymets, Mottaghi,
  Savva, Toshev, and Wijmans]{batra2020objectnav}
Dhruv Batra, Aaron Gokaslan, Aniruddha Kembhavi, Oleksandr Maksymets, Roozbeh
  Mottaghi, Manolis Savva, Alexander Toshev, and Erik Wijmans.
\newblock Objectnav revisited: On evaluation of embodied agents navigating to
  objects.
\newblock \emph{arXiv preprint arXiv:2006.13171}, 2020.

\bibitem[Beetz et~al.(2011)Beetz, Klank, Kresse, Maldonado, M{\"o}senlechner,
  Pangercic, R{\"u}hr, and Tenorth]{beetz2011robotic}
Michael Beetz, Ulrich Klank, Ingo Kresse, Alexis Maldonado, Lorenz
  M{\"o}senlechner, Dejan Pangercic, Thomas R{\"u}hr, and Moritz Tenorth.
\newblock Robotic roommates making pancakes.
\newblock In \emph{Humanoid Robots (Humanoids), 2011 11th IEEE-RAS
  International Conference on}, pages 529--536. IEEE, 2011.

\bibitem[Bellemare et~al.(2012)Bellemare, Naddaf, Veness, and
  Bowling]{bellemare2012arcade}
Marc~G Bellemare, Yavar Naddaf, Joel Veness, and Michael Bowling.
\newblock The arcade learning environment: An evaluation platform for general
  agents.
\newblock \emph{Journal of Artificial Intelligence Research}, 2012.

\bibitem[Bengio et~al.(2006)Bengio, Delalleau, and Le~Roux]{bengio200611}
Yoshua Bengio, Olivier Delalleau, and Nicolas Le~Roux.
\newblock Label propagation and quadratic criterion.
\newblock In \emph{Semi-Supervised Learning}, 2006.

\bibitem[Bollini et~al.(2013)Bollini, Tellex, Thompson, Roy, and
  Rus]{bollini2013interpreting}
Mario Bollini, Stefanie Tellex, Tyler Thompson, Nicholas Roy, and Daniela Rus.
\newblock Interpreting and executing recipes with a cooking robot.
\newblock In \emph{Experimental Robotics}, pages 481--495. Springer, 2013.

\bibitem[Bonin-Font et~al.(2008)Bonin-Font, Ortiz, and Oliver]{bonin2008visual}
Francisco Bonin-Font, Alberto Ortiz, and Gabriel Oliver.
\newblock Visual navigation for mobile robots: A survey.
\newblock \emph{Journal of intelligent and robotic systems}, 53\penalty0
  (3):\penalty0 263--296, 2008.

\bibitem[Borenstein et~al.(1996)Borenstein, Everett, and
  Feng]{Borenstein:1996:NMR:546829}
Johann Borenstein, H.~R. Everett, and Liqiang Feng.
\newblock \emph{Navigating Mobile Robots: Systems and Techniques}.
\newblock A. K. Peters, Ltd., Natick, MA, USA, 1996.
\newblock ISBN 1568810660.

\bibitem[Bowman et~al.(2017)Bowman, Atanasov, Daniilidis, and
  Pappas]{bowman2017probabilistic}
Sean~L Bowman, Nikolay Atanasov, Kostas Daniilidis, and George~J Pappas.
\newblock Probabilistic data association for semantic slam.
\newblock In \emph{2017 IEEE International Conference on Robotics and
  Automation (ICRA)}, pages 1722--1729. IEEE, 2017.

\bibitem[Bruce et~al.(2018)Bruce, Sunderhauf, Mirowski, Hadsell, and
  Milford]{bruce2018deployable}
Jake Bruce, Niko Sunderhauf, Piotr Mirowski, Raia Hadsell, and Michael Milford.
\newblock Learning deployable navigation policies at kilometer scale from a
  single traversal.
\newblock In Aude Billard, Anca Dragan, Jan Peters, and Jun Morimoto, editors,
  \emph{Proceedings of The 2nd Conference on Robot Learning}, volume~87 of
  \emph{Proceedings of Machine Learning Research}, pages 346--361. PMLR, 29--31
  Oct 2018.
\newblock URL \url{http://proceedings.mlr.press/v87/bruce18a.html}.

\bibitem[Burgard et~al.(1996)Burgard, Fox, Hennig, and
  Schmidt]{burgard1996estimating}
Wolfram Burgard, Dieter Fox, Daniel Hennig, and Timo Schmidt.
\newblock Estimating the absolute position of a mobile robot using position
  probability grids.
\newblock In \emph{Proceedings of the national conference on artificial
  intelligence}, pages 896--901, 1996.

\bibitem[Burgard et~al.(1998)Burgard, Derr, Fox, and
  Cremers]{burgard1998integrating}
Wolfram Burgard, Andrcas Derr, Dieter Fox, and Armin~B Cremers.
\newblock Integrating global position estimation and position tracking for
  mobile robots: the dynamic markov localization approach.
\newblock In \emph{Intelligent Robots and Systems, 1998. Proceedings., 1998
  IEEE/RSJ International Conference on}, volume~2, pages 730--735. IEEE, 1998.

\bibitem[Canny(1988)]{canny1988complexity}
John Canny.
\newblock \emph{The complexity of robot motion planning}.
\newblock MIT press, 1988.

\bibitem[Carlone et~al.(2014)Carlone, Du, Ng, Bona, and
  Indri]{carlone2014active}
Luca Carlone, Jingjing Du, Miguel~Kaouk Ng, Basilio Bona, and Marina Indri.
\newblock Active slam and exploration with particle filters using
  kullback-leibler divergence.
\newblock \emph{Journal of Intelligent \& Robotic Systems}, 75\penalty0
  (2):\penalty0 291--311, 2014.

\bibitem[Chandra et~al.(2018)Chandra, Couprie, and Kokkinos]{chandra2018deep}
Siddhartha Chandra, Camille Couprie, and Iasonas Kokkinos.
\newblock Deep spatio-temporal random fields for efficient video segmentation.
\newblock In \emph{Proceedings of the IEEE Conference on Computer Vision and
  Pattern Recognition}, pages 8915--8924, 2018.

\bibitem[Chang et~al.(2017)Chang, Dai, Funkhouser, Halber, Niessner, Savva,
  Song, Zeng, and Zhang]{Matterport3D}
Angel Chang, Angela Dai, Thomas Funkhouser, Maciej Halber, Matthias Niessner,
  Manolis Savva, Shuran Song, Andy Zeng, and Yinda Zhang.
\newblock Matterport3d: Learning from rgb-d data in indoor environments.
\newblock \emph{International Conference on 3D Vision (3DV)}, 2017.

\bibitem[Chao et~al.(2011)Chao, Cakmak, and Thomaz]{chao2011towards}
Crystal Chao, Maya Cakmak, and Andrea~L Thomaz.
\newblock Towards grounding concepts for transfer in goal learning from
  demonstration.
\newblock In \emph{Development and Learning (ICDL), 2011 IEEE International
  Conference on}, volume~2, pages 1--6. IEEE, 2011.

\bibitem[Chaplot and Lample(2017)]{chaplot2017arnold}
Devendra~Singh Chaplot and Guillaume Lample.
\newblock Arnold: An autonomous agent to play fps games.
\newblock In \emph{Thirty-First AAAI Conference on Artificial Intelligence},
  2017.

\bibitem[Chaplot et~al.(2016)Chaplot, Lample, Sathyendra, and
  Salakhutdinov]{chaplottransfer}
Devendra~Singh Chaplot, Guillaume Lample, Kanthashree~Mysore Sathyendra, and
  Ruslan Salakhutdinov.
\newblock Transfer deep reinforcement learning in 3d environments: An empirical
  study.
\newblock In \emph{NIPS Deep Reinforcemente Leaning Workshop}, 2016.

\bibitem[Chaplot et~al.(2018{\natexlab{a}})Chaplot, Parisotto, and
  Salakhutdinov]{chaplot2018active}
Devendra~Singh Chaplot, Emilio Parisotto, and Ruslan Salakhutdinov.
\newblock Active neural localization.
\newblock \emph{ICLR}, 2018{\natexlab{a}}.

\bibitem[Chaplot et~al.(2018{\natexlab{b}})Chaplot, Sathyendra, Pasumarthi,
  Rajagopal, and Salakhutdinov]{chaplot2017gated}
Devendra~Singh Chaplot, Kanthashree~Mysore Sathyendra, Rama~Kumar Pasumarthi,
  Dheeraj Rajagopal, and Ruslan Salakhutdinov.
\newblock Gated-attention architectures for task-oriented language grounding.
\newblock In \emph{AAAI}, 2018{\natexlab{b}}.

\bibitem[Chaplot et~al.(2020{\natexlab{a}})Chaplot, Gandhi, Gupta, Gupta, and
  Salakhutdinov]{ans}
Devendra~Singh Chaplot, Dhiraj Gandhi, Saurabh Gupta, Abhinav Gupta, and Ruslan
  Salakhutdinov.
\newblock Learning to explore using active neural slam.
\newblock In \emph{ICLR}, 2020{\natexlab{a}}.

\bibitem[Chaplot et~al.(2020{\natexlab{b}})Chaplot, Jiang, Gupta, and
  Gupta]{chaplot2020semantic}
Devendra~Singh Chaplot, Helen Jiang, Saurabh Gupta, and Abhinav Gupta.
\newblock Semantic curiosity for active visual learning.
\newblock In \emph{ECCV}, 2020{\natexlab{b}}.

\bibitem[Chaplot et~al.(2020{\natexlab{c}})Chaplot, Salakhutdinov, Gupta, and
  Gupta]{chaplot2020neural}
Devendra~Singh Chaplot, Ruslan Salakhutdinov, Abhinav Gupta, and Saurabh Gupta.
\newblock Neural topological slam for visual navigation.
\newblock In \emph{CVPR}, 2020{\natexlab{c}}.

\bibitem[Chen and Mooney(2011)]{chen2011learning}
David~L Chen and Raymond~J Mooney.
\newblock Learning to interpret natural language navigation instructions from
  observations.
\newblock In \emph{AAAI}, volume~2, pages 1--2, 2011.

\bibitem[Chen et~al.(2019{\natexlab{a}})Chen, de~Vicente, Sepulveda, Xia, Soto,
  V{\'a}zquez, and Savarese]{chen2019behavioral}
Kevin Chen, Juan~Pablo de~Vicente, Gabriel Sepulveda, Fei Xia, Alvaro Soto,
  Marynel V{\'a}zquez, and Silvio Savarese.
\newblock A behavioral approach to visual navigation with graph localization
  networks.
\newblock In \emph{Robotics: Science and Systems}, 2019{\natexlab{a}}.

\bibitem[Chen et~al.(2017)Chen, Papandreou, Schroff, and
  Adam]{chen2017rethinking}
Liang-Chieh Chen, George Papandreou, Florian Schroff, and Hartwig Adam.
\newblock Rethinking atrous convolution for semantic image segmentation.
\newblock \emph{arXiv preprint arXiv:1706.05587}, 2017.

\bibitem[Chen et~al.(2019{\natexlab{b}})Chen, Gupta, and
  Gupta]{chen2018learning}
Tao Chen, Saurabh Gupta, and Abhinav Gupta.
\newblock Learning exploration policies for navigation.
\newblock In \emph{International Conference on Learning Representations},
  2019{\natexlab{b}}.
\newblock URL \url{https://openreview.net/forum?id=SyMWn05F7}.

\bibitem[Chen et~al.(2013)Chen, Shrivastava, and Gupta]{chen2013neil}
Xinlei Chen, Abhinav Shrivastava, and Abhinav Gupta.
\newblock Neil: Extracting visual knowledge from web data.
\newblock In \emph{Proceedings of the IEEE International Conference on Computer
  Vision}, pages 1409--1416, 2013.

\bibitem[Cho et~al.(2014)Cho, Van~Merri{\"e}nboer, Bahdanau, and
  Bengio]{cho2014properties}
Kyunghyun Cho, Bart Van~Merri{\"e}nboer, Dzmitry Bahdanau, and Yoshua Bengio.
\newblock On the properties of neural machine translation: Encoder-decoder
  approaches.
\newblock \emph{Eighth Workshop on Syntax, Semantics and Structure in
  Statistical Translation}, 2014.

\bibitem[Choset and Nagatani(2001)]{choset2001topological}
Howie Choset and Keiji Nagatani.
\newblock Topological simultaneous localization and mapping (slam): toward
  exact localization without explicit localization.
\newblock \emph{IEEE Transactions on robotics and automation}, 17\penalty0
  (2):\penalty0 125--137, 2001.

\bibitem[Chu et~al.(2013)Chu, McMahon, Riano, McDonald, He, Perez-Tejada,
  Arrigo, Fitter, Nappo, and Darrell]{chu2013using}
Vivian Chu, Ian McMahon, Lorenzo Riano, Craig~G McDonald, Qin He,
  Jorge~Martinez Perez-Tejada, Michael Arrigo, Naomi Fitter, John~C Nappo, and
  Trevor Darrell.
\newblock Using robotic exploratory procedures to learn the meaning of haptic
  adjectives.
\newblock In \emph{Robotics and Automation (ICRA)}, pages 3048--3055, 2013.

\bibitem[Chung et~al.(2014)Chung, Gulcehre, Cho, and
  Bengio]{chung2014empirical}
Junyoung Chung, Caglar Gulcehre, KyungHyun Cho, and Yoshua Bengio.
\newblock Empirical evaluation of gated recurrent neural networks on sequence
  modeling.
\newblock \emph{arXiv preprint arXiv:1412.3555}, 2014.

\bibitem[Chunjie et~al.(2017)Chunjie, Qiang, et~al.]{chunjie2017cosine}
Luo Chunjie, Yang Qiang, et~al.
\newblock Cosine normalization: Using cosine similarity instead of dot product
  in neural networks.
\newblock \emph{arXiv preprint arXiv:1702.05870}, 2017.

\bibitem[Clark et~al.(2017{\natexlab{a}})Clark, Wang, Markham, Trigoni, and
  Wen]{clark2017vidloc}
Ronald Clark, Sen Wang, Andrew Markham, Niki Trigoni, and Hongkai Wen.
\newblock Vidloc: 6-dof video-clip relocalization.
\newblock \emph{arXiv preprint arXiv:1702.06521}, 2017{\natexlab{a}}.

\bibitem[Clark et~al.(2017{\natexlab{b}})Clark, Wang, Wen, Markham, and
  Trigoni]{clark2017vinet}
Ronald Clark, Sen Wang, Hongkai Wen, Andrew Markham, and Niki Trigoni.
\newblock Vinet: Visual-inertial odometry as a sequence-to-sequence learning
  problem.
\newblock In \emph{AAAI}, pages 3995--4001, 2017{\natexlab{b}}.

\bibitem[Cox and Leonard(1994)]{cox1994modeling}
Ingemar~J Cox and John~J Leonard.
\newblock Modeling a dynamic environment using a bayesian multiple hypothesis
  approach.
\newblock \emph{Artificial Intelligence}, 66\penalty0 (2):\penalty0 311--344,
  1994.

\bibitem[Cox and Wilfong(1990)]{Cox:1990:ARV:93002}
Ingemar~J. Cox and Gordon~T. Wilfong, editors.
\newblock \emph{Autonomous Robot Vehicles}.
\newblock Springer-Verlag New York, Inc., New York, NY, USA, 1990.
\newblock ISBN 0-387-97240-4.

\bibitem[Das et~al.(2018{\natexlab{a}})Das, Datta, Gkioxari, Lee, Parikh, and
  Batra]{das2017embodied}
Abhishek Das, Samyak Datta, Georgia Gkioxari, Stefan Lee, Devi Parikh, and
  Dhruv Batra.
\newblock Embodied question answering.
\newblock In \emph{CVPR}, 2018{\natexlab{a}}.

\bibitem[Das et~al.(2018{\natexlab{b}})Das, Gkioxari, Lee, Parikh, and
  Batra]{das2018neural}
Abhishek Das, Georgia Gkioxari, Stefan Lee, Devi Parikh, and Dhruv Batra.
\newblock Neural modular control for embodied question answering.
\newblock In \emph{Conference on Robot Learning}, pages 53--62,
  2018{\natexlab{b}}.

\bibitem[Dayan and Hinton(1993)]{dayan1993feudal}
Peter Dayan and Geoffrey~E Hinton.
\newblock Feudal reinforcement learning.
\newblock In \emph{Advances in neural information processing systems}, pages
  271--278, 1993.

\bibitem[Dellaert et~al.(1999)Dellaert, Fox, Burgard, and
  Thrun]{dellaert1999monte}
Frank Dellaert, Dieter Fox, Wolfram Burgard, and Sebastian Thrun.
\newblock Monte carlo localization for mobile robots.
\newblock In \emph{ICRA}, volume~2, pages 1322--1328, 1999.

\bibitem[Dhingra et~al.(2017)Dhingra, Liu, Yang, Cohen, and
  Salakhutdinov]{dhingra2016gated}
Bhuwan Dhingra, Hanxiao Liu, Zhilin Yang, William Cohen, and Ruslan
  Salakhutdinov.
\newblock Gated-attention readers for text comprehension.
\newblock \emph{ACL}, 2017.

\bibitem[Dornhege and Kleiner(2013)]{dornhege2013frontier}
Christian Dornhege and Alexander Kleiner.
\newblock A frontier-void-based approach for autonomous exploration in 3d.
\newblock \emph{Advanced Robotics}, 27\penalty0 (6):\penalty0 459--468, 2013.

\bibitem[Dosovitskiy and Koltun(2017)]{dosovitskiy2016learning}
Alexey Dosovitskiy and Vladlen Koltun.
\newblock Learning to act by predicting the future.
\newblock In \emph{ICLR}, 2017.

\bibitem[Elfes(1989)]{elfes1989using}
Alberto Elfes.
\newblock Using occupancy grids for mobile robot perception and navigation.
\newblock \emph{Computer}, 22\penalty0 (6):\penalty0 46--57, 1989.

\bibitem[Everingham et~al.()Everingham, Van~Gool, Williams, Winn, and
  Zisserman]{pascal-voc-2007}
M.~Everingham, L.~Van~Gool, C.~K.~I. Williams, J.~Winn, and A.~Zisserman.
\newblock The {PASCAL} {V}isual {O}bject {C}lasses {C}hallenge 2007 {(VOC2007)}
  {R}esults.
\newblock
  http://www.pascal-network.org/challenges/VOC/voc2007/workshop/index.html.

\bibitem[Eysenbach et~al.(2019{\natexlab{a}})Eysenbach, Gupta, Ibarz, and
  Levine]{eysenbach2018diversity}
Benjamin Eysenbach, Abhishek Gupta, Julian Ibarz, and Sergey Levine.
\newblock Diversity is all you need: Learning skills without a reward function.
\newblock In \emph{International Conference on Learning Representations},
  2019{\natexlab{a}}.
\newblock URL \url{https://openreview.net/forum?id=SJx63jRqFm}.

\bibitem[Eysenbach et~al.(2019{\natexlab{b}})Eysenbach, Salakhutdinov, and
  Levine]{eysenbach2019search}
Benjamin Eysenbach, Ruslan Salakhutdinov, and Sergey Levine.
\newblock Search on the replay buffer: Bridging planning and reinforcement
  learning.
\newblock \emph{arXiv preprint arXiv:1906.05253}, 2019{\natexlab{b}}.

\bibitem[Fang et~al.(2019{\natexlab{a}})Fang, Toshev, Fei-Fei, and
  Savarese]{fang2019scene}
Kuan Fang, Alexander Toshev, Li~Fei-Fei, and Silvio Savarese.
\newblock Scene memory transformer for embodied agents in long-horizon tasks.
\newblock In \emph{CVPR}, 2019{\natexlab{a}}.

\bibitem[Fang et~al.(2019{\natexlab{b}})Fang, Toshev, Fei-Fei, and
  Savarese]{fang2019smt}
Kuan Fang, Alexander Toshev, Li~Fei-Fei, and Silvio Savarese.
\newblock Scene memory transformer for embodied agents in long-horizon tasks.
\newblock In \emph{CVPR}, 2019{\natexlab{b}}.

\bibitem[Fasola and Mataric(2013)]{fasola2013using}
Juan Fasola and Maja~J Mataric.
\newblock Using semantic fields to model dynamic spatial relations in a robot
  architecture for natural language instruction of service robots.
\newblock In \emph{Intelligent Robots and Systems (IROS)}, pages 143--150.
  IEEE, 2013.

\bibitem[Fathi et~al.(2011)Fathi, Balcan, Ren, and Rehg]{fathi2011combining}
Alireza Fathi, Maria~Florina Balcan, Xiaofeng Ren, and James~M Rehg.
\newblock Combining self training and active learning for video segmentation.
\newblock In \emph{Proceedings of the British Machine Vision Conference}.
  Georgia Institute of Technology, 2011.

\bibitem[Foerster et~al.(2016)Foerster, Assael, de~Freitas, and
  Whiteson]{foerster2016learning}
Jakob~N Foerster, Yannis~M Assael, Nando de~Freitas, and Shimon Whiteson.
\newblock Learning to communicate to solve riddles with deep distributed
  recurrent q-networks.
\newblock \emph{arXiv preprint arXiv:1602.02672}, 2016.

\bibitem[Fox(1998)]{fox1998markov}
Dieter Fox.
\newblock \emph{Markov localization-a probabilistic framework for mobile robot
  localization and navigation.}
\newblock PhD thesis, Universit{\"a}t Bonn, 1998.

\bibitem[Fox et~al.(1998)Fox, Burgard, and Thrun]{fox1998active}
Dieter Fox, Wolfram Burgard, and Sebastian Thrun.
\newblock Active markov localization for mobile robots.
\newblock \emph{Robotics and Autonomous Systems}, 25\penalty0 (3-4):\penalty0
  195--207, 1998.

\bibitem[Fox et~al.(2003)Fox, Hightower, Liao, Schulz, and
  Borriello]{fox2003bayesian}
V~Fox, Jeffrey Hightower, Lin Liao, Dirk Schulz, and Gaetano Borriello.
\newblock Bayesian filtering for location estimation.
\newblock \emph{IEEE pervasive computing}, 2\penalty0 (3):\penalty0 24--33,
  2003.

\bibitem[Fuentes-Pacheco et~al.(2015)Fuentes-Pacheco, Ruiz-Ascencio, and
  Rend\'{o}n-Mancha]{slam-survey:2015}
J.~Fuentes-Pacheco, J.~Ruiz-Ascencio, and J.~M. Rend\'{o}n-Mancha.
\newblock Visual simultaneous localization and mapping: a survey.
\newblock \emph{Artificial Intelligence Review}, 2015.

\bibitem[Gadde et~al.(2017)Gadde, Jampani, and Gehler]{gadde2017semantic}
Raghudeep Gadde, Varun Jampani, and Peter~V Gehler.
\newblock Semantic video cnns through representation warping.
\newblock In \emph{Proceedings of the IEEE International Conference on Computer
  Vision}, pages 4453--4462, 2017.

\bibitem[Gal et~al.(2017)Gal, Islam, and Ghahramani]{gal2017deep}
Yarin Gal, Riashat Islam, and Zoubin Ghahramani.
\newblock Deep bayesian active learning with image data.
\newblock In \emph{Proceedings of the 34th International Conference on Machine
  Learning-Volume 70}, pages 1183--1192. JMLR. org, 2017.

\bibitem[Gandhi et~al.(2017)Gandhi, Pinto, and Gupta]{gandhi2017learning}
Dhiraj Gandhi, Lerrel Pinto, and Abhinav Gupta.
\newblock Learning to fly by crashing.
\newblock In \emph{2017 IEEE/RSJ International Conference on Intelligent Robots
  and Systems (IROS)}, pages 3948--3955. IEEE, 2017.

\bibitem[Gaskett et~al.(1999)Gaskett, Wettergreen, and Zelinsky]{gaskett1999q}
Chris Gaskett, David Wettergreen, and Alexander Zelinsky.
\newblock Q-learning in continuous state and action spaces.
\newblock In \emph{Australasian Joint Conference on Artificial Intelligence},
  pages 417--428. Springer, 1999.

\bibitem[Gillner and Mallot(1998)]{gillner1998navigation}
Sabine Gillner and Hanspeter~A Mallot.
\newblock Navigation and acquisition of spatial knowledge in a virtual maze.
\newblock \emph{Journal of cognitive neuroscience}, 10\penalty0 (4):\penalty0
  445--463, 1998.

\bibitem[Godard et~al.(2017)Godard, Mac~Aodha, and
  Brostow]{godard2017unsupervised}
Cl{\'e}ment Godard, Oisin Mac~Aodha, and Gabriel~J Brostow.
\newblock Unsupervised monocular depth estimation with left-right consistency.
\newblock In \emph{Proceedings of the IEEE Conference on Computer Vision and
  Pattern Recognition}, pages 270--279, 2017.

\bibitem[Gordon et~al.(2018)Gordon, Kembhavi, Rastegari, Redmon, Fox, and
  Farhadi]{gordon2018iqa}
Daniel Gordon, Aniruddha Kembhavi, Mohammad Rastegari, Joseph Redmon, Dieter
  Fox, and Ali Farhadi.
\newblock Iqa: Visual question answering in interactive environments.
\newblock In \emph{Proceedings of the IEEE Conference on Computer Vision and
  Pattern Recognition}, pages 4089--4098, 2018.

\bibitem[Guadarrama et~al.(2013)Guadarrama, Riano, Golland, Go, Jia, Klein,
  Abbeel, Darrell, et~al.]{guadarrama2013grounding}
Sergio Guadarrama, Lorenzo Riano, Dave Golland, Daniel Go, Yangqing Jia, Dan
  Klein, Pieter Abbeel, Trevor Darrell, et~al.
\newblock Grounding spatial relations for human-robot interaction.
\newblock In \emph{IROS}, pages 1640--1647. IEEE, 2013.

\bibitem[Guadarrama et~al.(2014)Guadarrama, Rodner, Saenko, Zhang, Farrell,
  Donahue, and Darrell]{guadarrama2014open}
Sergio Guadarrama, Erik Rodner, Kate Saenko, Ning Zhang, Ryan Farrell, Jeff
  Donahue, and Trevor Darrell.
\newblock Open-vocabulary object retrieval.
\newblock In \emph{Robotics: science and systems}, volume~2, page~6. Citeseer,
  2014.

\bibitem[Gupta et~al.(2017)Gupta, Davidson, Levine, Sukthankar, and
  Malik]{gupta2017cognitive}
Saurabh Gupta, James Davidson, Sergey Levine, Rahul Sukthankar, and Jitendra
  Malik.
\newblock Cognitive mapping and planning for visual navigation.
\newblock In \emph{Proceedings of the IEEE Conference on Computer Vision and
  Pattern Recognition}, pages 2616--2625, 2017.

\bibitem[Hartley and Zisserman(2003)]{hartley2003multiple}
Richard Hartley and Andrew Zisserman.
\newblock \emph{Multiple view geometry in computer vision}.
\newblock Cambridge university press, 2003.

\bibitem[Hausknecht and Stone(2015)]{hausknecht2015deep}
Matthew Hausknecht and Peter Stone.
\newblock Deep recurrent q-learning for partially observable mdps.
\newblock \emph{arXiv preprint arXiv:1507.06527}, 2015.

\bibitem[{He} et~al.(2017){He}, {Gkioxari}, {Dollár}, and
  {Girshick}]{mask_rcnn}
K.~{He}, G.~{Gkioxari}, P.~{Dollár}, and R.~{Girshick}.
\newblock Mask r-cnn.
\newblock In \emph{2017 IEEE International Conference on Computer Vision
  (ICCV)}, pages 2980--2988, 2017.

\bibitem[He et~al.(2016)He, Zhang, Ren, and Sun]{he2016deep}
Kaiming He, Xiangyu Zhang, Shaoqing Ren, and Jian Sun.
\newblock Deep residual learning for image recognition.
\newblock In \emph{Proceedings of the IEEE conference on computer vision and
  pattern recognition}, pages 770--778, 2016.

\bibitem[Heess et~al.(2015)Heess, Wayne, Silver, Lillicrap, Erez, and
  Tassa]{heess2015learning}
Nicolas Heess, Gregory Wayne, David Silver, Tim Lillicrap, Tom Erez, and Yuval
  Tassa.
\newblock Learning continuous control policies by stochastic value gradients.
\newblock In \emph{Advances in Neural Information Processing Systems}, pages
  2944--2952, 2015.

\bibitem[Henriques and Vedaldi(2018)]{henriques2018mapnet}
Joao~F Henriques and Andrea Vedaldi.
\newblock Mapnet: An allocentric spatial memory for mapping environments.
\newblock In \emph{proceedings of the IEEE Conference on Computer Vision and
  Pattern Recognition}, pages 8476--8484, 2018.

\bibitem[Henry et~al.(2014)Henry, Krainin, Herbst, Ren, and Fox]{henry2014rgb}
Peter Henry, Michael Krainin, Evan Herbst, Xiaofeng Ren, and Dieter Fox.
\newblock Rgb-d mapping: Using depth cameras for dense 3d modeling of indoor
  environments.
\newblock In \emph{Experimental robotics}, pages 477--491. Springer, 2014.

\bibitem[Hermann et~al.(2017)Hermann, Hill, Green, Wang, Faulkner, Soyer,
  Szepesvari, Czarnecki, Jaderberg, Teplyashin, et~al.]{hermann2017grounded}
Karl~Moritz Hermann, Felix Hill, Simon Green, Fumin Wang, Ryan Faulkner, Hubert
  Soyer, David Szepesvari, Wojtek Czarnecki, Max Jaderberg, Denis Teplyashin,
  et~al.
\newblock Grounded language learning in a simulated 3d world.
\newblock \emph{arXiv preprint arXiv:1706.06551}, 2017.

\bibitem[Hochreiter and Schmidhuber(1997)]{hochreiter1997long}
Sepp Hochreiter and J{\"u}rgen Schmidhuber.
\newblock Long short-term memory.
\newblock \emph{Neural computation}, 9\penalty0 (8):\penalty0 1735--1780, 1997.

\bibitem[Holz et~al.(2010)Holz, Basilico, Amigoni, and
  Behnke]{holz2010evaluating}
Dirk Holz, Nicola Basilico, Francesco Amigoni, and Sven Behnke.
\newblock Evaluating the efficiency of frontier-based exploration strategies.
\newblock In \emph{ISR 2010 (41st International Symposium on Robotics) and
  ROBOTIK 2010 (6th German Conference on Robotics)}, pages 1--8. VDE, 2010.

\bibitem[Horn(1990)]{horn1990hadamard}
Roger~A Horn.
\newblock The hadamard product.
\newblock In \emph{Proc. Symp. Appl. Math}, volume~40, pages 87--169, 1990.

\bibitem[Hornung et~al.(2013)Hornung, Wurm, Bennewitz, Stachniss, and
  Burgard]{hornung13auro}
Armin Hornung, Kai~M. Wurm, Maren Bennewitz, Cyrill Stachniss, and Wolfram
  Burgard.
\newblock {OctoMap}: An efficient probabilistic {3D} mapping framework based on
  octrees.
\newblock \emph{Autonomous Robots}, 2013.
\newblock \doi{10.1007/s10514-012-9321-0}.
\newblock URL \url{http://octomap.github.com}.

\bibitem[Huang et~al.(2013)Huang, He, Gao, Deng, Acero, and
  Heck]{huang2013learning}
Po-Sen Huang, Xiaodong He, Jianfeng Gao, Li~Deng, Alex Acero, and Larry Heck.
\newblock Learning deep structured semantic models for web search using
  clickthrough data.
\newblock In \emph{Proceedings of the 22nd ACM international conference on
  Conference on information \& knowledge management}, pages 2333--2338. ACM,
  2013.

\bibitem[Huber(1964)]{huber1964robust}
Peter~J Huber.
\newblock Robust estimation of a location parameter.
\newblock \emph{The Annals of Mathematical Statistics}, 35\penalty0
  (1):\penalty0 73--101, 1964.

\bibitem[Izadi et~al.(2011)Izadi, Kim, Hilliges, Molyneaux, Newcombe, Kohli,
  Shotton, Hodges, Freeman, Davison, and Fitzgibbon]{izadiUIST11}
Shahram Izadi, David Kim, Otmar Hilliges, David Molyneaux, Richard Newcombe,
  Pushmeet Kohli, Jamie Shotton, Steve Hodges, Dustin Freeman, Andrew Davison,
  and Andrew Fitzgibbon.
\newblock {KinectFusion}: real-time {3D} reconstruction and interaction using a
  moving depth camera.
\newblock \emph{UIST}, 2011.

\bibitem[Jaderberg et~al.(2015)Jaderberg, Simonyan, Zisserman,
  et~al.]{jaderberg2015spatial}
Max Jaderberg, Karen Simonyan, Andrew Zisserman, et~al.
\newblock Spatial transformer networks.
\newblock In \emph{Advances in neural information processing systems}, pages
  2017--2025, 2015.

\bibitem[Jaksch et~al.(2010)Jaksch, Ortner, and Auer]{jaksch2010near}
Thomas Jaksch, Ronald Ortner, and Peter Auer.
\newblock Near-optimal regret bounds for reinforcement learning.
\newblock \emph{Journal of Machine Learning Research}, 11\penalty0
  (Apr):\penalty0 1563--1600, 2010.

\bibitem[Jayaraman and Grauman(2018)]{jayaraman2018learning}
Dinesh Jayaraman and Kristen Grauman.
\newblock Learning to look around: Intelligently exploring unseen environments
  for unknown tasks.
\newblock In \emph{Proceedings of the IEEE Conference on Computer Vision and
  Pattern Recognition}, pages 1238--1247, 2018.

\bibitem[Jensfelt and Kristensen(2001)]{jensfelt2001active}
Patric Jensfelt and Steen Kristensen.
\newblock Active global localization for a mobile robot using multiple
  hypothesis tracking.
\newblock \emph{IEEE Transactions on Robotics and Automation}, 17\penalty0
  (5):\penalty0 748--760, 2001.

\bibitem[Kalman et~al.(1960)]{kalman1960new}
Rudolph~Emil Kalman et~al.
\newblock A new approach to linear filtering and prediction problems.
\newblock \emph{Journal of basic Engineering}, 82\penalty0 (1):\penalty0
  35--45, 1960.

\bibitem[Karkus et~al.(2017)Karkus, Hsu, and Lee]{qmdpnet17}
Peter Karkus, David Hsu, and Wee~Sun Lee.
\newblock Qmdp-net: Deep learning for planning under partial observability.
\newblock \emph{CoRR}, abs/1703.06692, 2017.
\newblock URL \url{https://arxiv.org/abs/1703.06692}.

\bibitem[Kaufmann et~al.(2019)Kaufmann, Gehrig, Foehn, Ranftl, Dosovitskiy,
  Koltun, and Scaramuzza]{kaufmann2019beauty}
Elia Kaufmann, Mathias Gehrig, Philipp Foehn, Ren{\'e} Ranftl, Alexey
  Dosovitskiy, Vladlen Koltun, and Davide Scaramuzza.
\newblock Beauty and the beast: Optimal methods meet learning for drone racing.
\newblock In \emph{2019 International Conference on Robotics and Automation
  (ICRA)}, pages 690--696. IEEE, 2019.

\bibitem[Kavraki et~al.(1996)Kavraki, Svestka, Latombe, and
  Overmars]{kavraki1996probabilistic}
Lydia~E Kavraki, Petr Svestka, J-C Latombe, and Mark~H Overmars.
\newblock Probabilistic roadmaps for path planning in high-dimensional
  configuration spaces.
\newblock \emph{RA}, 1996.

\bibitem[Kearns and Singh(2002)]{kearns2002near}
Michael Kearns and Satinder Singh.
\newblock Near-optimal reinforcement learning in polynomial time.
\newblock \emph{Machine learning}, 49\penalty0 (2-3):\penalty0 209--232, 2002.

\bibitem[Kempka et~al.(2016)Kempka, Wydmuch, Runc, Toczek, and
  Ja{\'s}kowski]{kempka2016vizdoom}
Micha{\l} Kempka, Marek Wydmuch, Grzegorz Runc, Jakub Toczek, and Wojciech
  Ja{\'s}kowski.
\newblock Vizdoom: A doom-based ai research platform for visual reinforcement
  learning.
\newblock \emph{arXiv preprint arXiv:1605.02097}, 2016.

\bibitem[Kendall et~al.(2015)Kendall, Grimes, and Cipolla]{kendall2015posenet}
Alex Kendall, Matthew Grimes, and Roberto Cipolla.
\newblock Posenet: A convolutional network for real-time 6-dof camera
  relocalization.
\newblock In \emph{Proceedings of the IEEE international conference on computer
  vision}, pages 2938--2946, 2015.

\bibitem[Khan et~al.(2017)Khan, Zhang, Atanasov, Karydis, Lee, and
  Kumar]{khan2017end}
Arbaaz Khan, Clark Zhang, Nikolay Atanasov, Konstantinos Karydis, Daniel~D Lee,
  and Vijay Kumar.
\newblock End-to-end navigation in unknown environments using neural networks.
\newblock \emph{arXiv preprint arXiv:1707.07385}, 2017.

\bibitem[Kingma and Ba(2014)]{kingma2014adam}
Diederik~P Kingma and Jimmy Ba.
\newblock Adam: A method for stochastic optimization.
\newblock \emph{arXiv preprint arXiv:1412.6980}, 2014.

\bibitem[Koch et~al.(2015)Koch, Zemel, and Salakhutdinov]{koch2015siamese}
Gregory Koch, Richard Zemel, and Ruslan Salakhutdinov.
\newblock Siamese neural networks for one-shot image recognition.
\newblock In \emph{ICML Deep Learning Workshop}, volume~2, 2015.

\bibitem[Kollar and Roy(2008)]{kollar2008trajectory}
Thomas Kollar and Nicholas Roy.
\newblock Trajectory optimization using reinforcement learning for map
  exploration.
\newblock \emph{The International Journal of Robotics Research}, 27\penalty0
  (2):\penalty0 175--196, 2008.

\bibitem[Kostrikov(2018)]{pytorchrl}
Ilya Kostrikov.
\newblock Pytorch implementations of reinforcement learning algorithms.
\newblock \url{https://github.com/ikostrikov/pytorch-a2c-ppo-acktr-gail}, 2018.

\bibitem[Koutn{\'\i}k et~al.(2013)Koutn{\'\i}k, Cuccu, Schmidhuber, and
  Gomez]{koutnik2013evolving}
Jan Koutn{\'\i}k, Giuseppe Cuccu, J{\"u}rgen Schmidhuber, and Faustino Gomez.
\newblock Evolving large-scale neural networks for vision-based reinforcement
  learning.
\newblock In \emph{Proceedings of the 15th annual conference on Genetic and
  evolutionary computation}, pages 1061--1068. ACM, 2013.

\bibitem[Kruskal(1956)]{kruskal1956shortest}
Joseph~B Kruskal.
\newblock On the shortest spanning subtree of a graph and the traveling
  salesman problem.
\newblock \emph{Proceedings of the American Mathematical society}, 7\penalty0
  (1):\penalty0 48--50, 1956.

\bibitem[Kuipers and Byun(1991)]{kuipers1991robot}
Benjamin Kuipers and Yung-Tai Byun.
\newblock A robot exploration and mapping strategy based on a semantic
  hierarchy of spatial representations.
\newblock \emph{Robotics and autonomous systems}, 8\penalty0 (1-2):\penalty0
  47--63, 1991.

\bibitem[Kulick et~al.(2013)Kulick, Toussaint, Lang, and
  Lopes]{kulick2013active}
Johannes Kulick, Marc Toussaint, Tobias Lang, and Manuel Lopes.
\newblock Active learning for teaching a robot grounded relational symbols.
\newblock In \emph{IJCAI}, 2013.

\bibitem[Kulkarni et~al.(2016)Kulkarni, Saeedi, Gautam, and
  Gershman]{kulkarni2016deep}
Tejas~D Kulkarni, Ardavan Saeedi, Simanta Gautam, and Samuel~J Gershman.
\newblock Deep successor reinforcement learning.
\newblock \emph{arXiv preprint arXiv:1606.02396}, 2016.

\bibitem[Kumar et~al.(2018)Kumar, Gupta, Fouhey, Levine, and
  Malik]{kumar2018visual}
Ashish Kumar, Saurabh Gupta, David Fouhey, Sergey Levine, and Jitendra Malik.
\newblock Visual memory for robust path following.
\newblock In \emph{Advances in Neural Information Processing Systems}, 2018.

\bibitem[Kuo et~al.(2018)Kuo, H{\"a}ne, Yuh, Mukherjee, and Malik]{kuo2018cost}
Weicheng Kuo, Christian H{\"a}ne, Esther Yuh, Pratik Mukherjee, and Jitendra
  Malik.
\newblock Cost-sensitive active learning for intracranial hemorrhage detection.
\newblock In \emph{International Conference on Medical Image Computing and
  Computer-Assisted Intervention}, pages 715--723. Springer, 2018.

\bibitem[Lample and Chaplot(2017)]{lample2016playing}
Guillaume Lample and Devendra~Singh Chaplot.
\newblock Playing {FPS} games with deep reinforcement learning.
\newblock In \emph{Thirty-First AAAI Conference on Artificial Intelligence},
  2017.

\bibitem[Lavalle and Kuffner~Jr(2000)]{lavalle2000rapidly}
Steven~M Lavalle and James~J Kuffner~Jr.
\newblock Rapidly-exploring random trees: Progress and prospects.
\newblock In \emph{Algorithmic and Computational Robotics: New Directions},
  2000.

\bibitem[LeCun et~al.(1995)LeCun, Bengio, et~al.]{lecun1995convolutional}
Yann LeCun, Yoshua Bengio, et~al.
\newblock Convolutional networks for images, speech, and time series.
\newblock \emph{The handbook of brain theory and neural networks},
  3361\penalty0 (10):\penalty0 1995, 1995.

\bibitem[Lee et~al.(2017)Lee, Levine, and Abbeel]{lee2017learning}
Alex~X Lee, Sergey Levine, and Pieter Abbeel.
\newblock Learning visual servoing with deep features and fitted q-iteration.
\newblock \emph{arXiv preprint arXiv:1703.11000}, 2017.

\bibitem[Lee et~al.(2018)Lee, Parisotto, Chaplot, Xing, and
  Salakhutdinov]{lee2018gated}
Lisa Lee, Emilio Parisotto, Devendra~Singh Chaplot, Eric Xing, and Ruslan
  Salakhutdinov.
\newblock Gated path planning networks.
\newblock \emph{arXiv preprint arXiv:1806.06408}, 2018.

\bibitem[Lemaignan et~al.(2012)Lemaignan, Ros, Sisbot, Alami, and
  Beetz]{lemaignan2012grounding}
S{\'e}verin Lemaignan, Raquel Ros, E~Akin Sisbot, Rachid Alami, and Michael
  Beetz.
\newblock Grounding the interaction: Anchoring situated discourse in everyday
  human-robot interaction.
\newblock \emph{International Journal of Social Robotics}, 4\penalty0
  (2):\penalty0 181--199, 2012.

\bibitem[Levine et~al.(2016)Levine, Finn, Darrell, and Abbeel]{levine2016end}
Sergey Levine, Chelsea Finn, Trevor Darrell, and Pieter Abbeel.
\newblock End-to-end training of deep visuomotor policies.
\newblock \emph{Journal of Machine Learning Research}, 17\penalty0
  (39):\penalty0 1--40, 2016.

\bibitem[Lin(1993)]{lin1993reinforcement}
Long-Ji Lin.
\newblock Reinforcement learning for robots using neural networks.
\newblock Technical report, DTIC Document, 1993.

\bibitem[Lin et~al.(2014)Lin, Maire, Belongie, Hays, Perona, Ramanan,
  Doll{\'a}r, and Zitnick]{lin2014microsoft}
Tsung-Yi Lin, Michael Maire, Serge Belongie, James Hays, Pietro Perona, Deva
  Ramanan, Piotr Doll{\'a}r, and C~Lawrence Zitnick.
\newblock Microsoft coco: Common objects in context.
\newblock In \emph{ECCV}, 2014.

\bibitem[Lin et~al.(2017)Lin, Doll{\'a}r, Girshick, He, Hariharan, and
  Belongie]{lin2017feature}
Tsung-Yi Lin, Piotr Doll{\'a}r, Ross Girshick, Kaiming He, Bharath Hariharan,
  and Serge Belongie.
\newblock Feature pyramid networks for object detection.
\newblock In \emph{Proceedings of the IEEE conference on computer vision and
  pattern recognition}, pages 2117--2125, 2017.

\bibitem[Ma et~al.(2017)Ma, St{\"u}ckler, Kerl, and Cremers]{ma2017multi}
Lingni Ma, J{\"o}rg St{\"u}ckler, Christian Kerl, and Daniel Cremers.
\newblock Multi-view deep learning for consistent semantic mapping with rgb-d
  cameras.
\newblock In \emph{2017 IEEE/RSJ International Conference on Intelligent Robots
  and Systems (IROS)}, pages 598--605. IEEE, 2017.

\bibitem[Martin et~al.(2001)Martin, Fowlkes, Tal, and Malik]{MartinFTM01}
D.~Martin, C.~Fowlkes, D.~Tal, and J.~Malik.
\newblock A database of human segmented natural images and its application to
  evaluating segmentation algorithms and measuring ecological statistics.
\newblock In \emph{Proc. 8th Int'l Conf. Computer Vision}, volume~2, pages
  416--423, July 2001.

\bibitem[Martinez-Cantin et~al.(2009)Martinez-Cantin, de~Freitas, Brochu,
  Castellanos, and Doucet]{martinez2009bayesian}
Ruben Martinez-Cantin, Nando de~Freitas, Eric Brochu, Jos{\'e} Castellanos, and
  Arnaud Doucet.
\newblock A bayesian exploration-exploitation approach for optimal online
  sensing and planning with a visually guided mobile robot.
\newblock \emph{Autonomous Robots}, 27\penalty0 (2):\penalty0 93--103, 2009.

\bibitem[McPartland and Gallagher(2008)]{mcpartland2008learning}
Michelle McPartland and Marcus Gallagher.
\newblock Learning to be a bot: Reinforcement learning in shooter games.
\newblock In \emph{AIIDE}, 2008.

\bibitem[Mei et~al.(2015)Mei, Bansal, and Walter]{mei2015listen}
Hongyuan Mei, Mohit Bansal, and Matthew~R Walter.
\newblock Listen, attend, and walk: Neural mapping of navigational instructions
  to action sequences.
\newblock \emph{arXiv preprint arXiv:1506.04089}, 2015.

\bibitem[Mei et~al.(2016)Mei, Bansal, and Walter]{mei2016listen}
Hongyuan Mei, Mohit Bansal, and Matthew~R Walter.
\newblock Listen, attend, and walk: Neural mapping of navigational instructions
  to action sequences.
\newblock In \emph{AAAI}, volume~1, page~2, 2016.

\bibitem[Meng and Kak(1993)]{meng1993mobile}
Min Meng and Avinash~C Kak.
\newblock Mobile robot navigation using neural networks and nonmetrical
  environmental models.
\newblock \emph{IEEE Control Systems Magazine}, 13\penalty0 (5):\penalty0
  30--39, 1993.

\bibitem[Mirowski et~al.(2016)Mirowski, Pascanu, Viola, Soyer, Ballard, Banino,
  Denil, Goroshin, Sifre, Kavukcuoglu, et~al.]{mirowski2016learning}
Piotr Mirowski, Razvan Pascanu, Fabio Viola, Hubert Soyer, Andrew~J Ballard,
  Andrea Banino, Misha Denil, Ross Goroshin, Laurent Sifre, Koray Kavukcuoglu,
  et~al.
\newblock Learning to navigate in complex environments.
\newblock \emph{arXiv preprint arXiv:1611.03673}, 2016.

\bibitem[Mirowski et~al.(2018)Mirowski, Grimes, Malinowski, Hermann, Anderson,
  Teplyashin, Simonyan, Kavukcuoglu, Zisserman, and
  Hadsell]{mirowski2018learning}
Piotr Mirowski, Matthew~Koichi Grimes, Mateusz Malinowski, Karl~Moritz Hermann,
  Keith Anderson, Denis Teplyashin, Karen Simonyan, Koray Kavukcuoglu, Andrew
  Zisserman, and Raia Hadsell.
\newblock Learning to navigate in cities without a map.
\newblock In \emph{Neural Information Processing Systems (NeurIPS)}, 2018.

\bibitem[Misra et~al.(2017)Misra, Langford, and Artzi]{misra2017mapping}
Dipendra~K Misra, John Langford, and Yoav Artzi.
\newblock Mapping instructions and visual observations to actions with
  reinforcement learning.
\newblock \emph{arXiv preprint arXiv:1704.08795}, 2017.

\bibitem[Misra et~al.(2018)Misra, Girshick, Fergus, Hebert, Gupta, and van~der
  Maaten]{misra2017lba}
Ishan Misra, Ross Girshick, Rob Fergus, Martial Hebert, Abhinav Gupta, and
  Laurens van~der Maaten.
\newblock {Learning by Asking Questions}.
\newblock In \emph{{CVPR}}, 2018.

\bibitem[Mnih et~al.(2013)Mnih, Kavukcuoglu, Silver, Graves, Antonoglou,
  Wierstra, and Riedmiller]{mnih2013playing}
Volodymyr Mnih, Koray Kavukcuoglu, David Silver, Alex Graves, Ioannis
  Antonoglou, Daan Wierstra, and Martin Riedmiller.
\newblock Playing atari with deep reinforcement learning.
\newblock \emph{arXiv preprint arXiv:1312.5602}, 2013.

\bibitem[Mnih et~al.(2016)Mnih, Badia, Mirza, Graves, Lillicrap, Harley,
  Silver, and Kavukcuoglu]{mnih2016asynchronous}
Volodymyr Mnih, Adria~Puigdomenech Badia, Mehdi Mirza, Alex Graves, Timothy
  Lillicrap, Tim Harley, David Silver, and Koray Kavukcuoglu.
\newblock Asynchronous methods for deep reinforcement learning.
\newblock In \emph{ICML}, 2016.

\bibitem[Mousavian et~al.(2019)Mousavian, Toshev, Fi{\v{s}}er,
  Ko{\v{s}}eck{\'a}, Wahid, and Davidson]{mousavian2019visual}
Arsalan Mousavian, Alexander Toshev, Marek Fi{\v{s}}er, Jana Ko{\v{s}}eck{\'a},
  Ayzaan Wahid, and James Davidson.
\newblock Visual representations for semantic target driven navigation.
\newblock In \emph{2019 International Conference on Robotics and Automation
  (ICRA)}, pages 8846--8852. IEEE, 2019.

\bibitem[Mur-Artal and Tard{\'o}s(2017)]{mur2017orb}
Raul Mur-Artal and Juan~D Tard{\'o}s.
\newblock Orb-slam2: An open-source slam system for monocular, stereo, and
  rgb-d cameras.
\newblock \emph{IEEE Transactions on Robotics}, 33\penalty0 (5):\penalty0
  1255--1262, 2017.

\bibitem[Mur-Artal et~al.(2015)Mur-Artal, Montiel, and Tardos]{mur2015orb}
Raul Mur-Artal, Jose Maria~Martinez Montiel, and Juan~D Tardos.
\newblock Orb-slam: a versatile and accurate monocular slam system.
\newblock \emph{IEEE transactions on robotics}, 31\penalty0 (5):\penalty0
  1147--1163, 2015.

\bibitem[Murali et~al.(2019)Murali, Chen, Alwala, Gandhi, Pinto, Gupta, and
  Gupta]{pyrobot2019}
Adithyavairavan Murali, Tao Chen, Kalyan~Vasudev Alwala, Dhiraj Gandhi, Lerrel
  Pinto, Saurabh Gupta, and Abhinav Gupta.
\newblock Pyrobot: An open-source robotics framework for research and
  benchmarking.
\newblock \emph{arXiv preprint arXiv:1906.08236}, 2019.

\bibitem[Nair and Hinton(2010)]{nair2010rectified}
Vinod Nair and Geoffrey~E Hinton.
\newblock Rectified linear units improve restricted boltzmann machines.
\newblock In \emph{Proceedings of the 27th international conference on machine
  learning (ICML-10)}, pages 807--814, 2010.

\bibitem[Newcombe et~al.(2011{\natexlab{a}})Newcombe, Izadi, Hilliges,
  Molyneaux, Kim, Davison, Kohli, Shotton, Hodges, and
  Fitzgibbon]{newcombe2011kinectfusion}
Richard~A Newcombe, Shahram Izadi, Otmar Hilliges, David Molyneaux, David Kim,
  Andrew~J Davison, Pushmeet Kohli, Jamie Shotton, Steve Hodges, and Andrew~W
  Fitzgibbon.
\newblock Kinectfusion: Real-time dense surface mapping and tracking.
\newblock In \emph{ISMAR}, volume~11, pages 127--136, 2011{\natexlab{a}}.

\bibitem[Newcombe et~al.(2011{\natexlab{b}})Newcombe, Lovegrove, and
  Davison]{newcombe2011dtam}
Richard~A Newcombe, Steven~J Lovegrove, and Andrew~J Davison.
\newblock Dtam: Dense tracking and mapping in real-time.
\newblock In \emph{2011 international conference on computer vision}, pages
  2320--2327. IEEE, 2011{\natexlab{b}}.

\bibitem[Ng(2003)]{ng2003shaping}
Andrew~Y Ng.
\newblock \emph{Shaping and policy search in reinforcement learning}.
\newblock PhD thesis, University of California, Berkeley, 2003.

\bibitem[Nist{\'e}r et~al.(2006)Nist{\'e}r, Naroditsky, and
  Bergen]{nister2006visual}
David Nist{\'e}r, Oleg Naroditsky, and James Bergen.
\newblock Visual odometry for ground vehicle applications.
\newblock \emph{Journal of Field Robotics}, 23\penalty0 (1):\penalty0 3--20,
  2006.

\bibitem[Oh et~al.(2016)Oh, Chockalingam, Singh, and Lee]{oh2016control}
Junhyuk Oh, Valliappa Chockalingam, Satinder Singh, and Honglak Lee.
\newblock Control of memory, active perception, and action in minecraft.
\newblock In \emph{ICML}, 2016.

\bibitem[Parisotto and Salakhutdinov(2018)]{parisotto2017neural}
Emilio Parisotto and Ruslan Salakhutdinov.
\newblock Neural map: Structured memory for deep reinforcement learning.
\newblock In \emph{ICLR}, 2018.

\bibitem[Paszke et~al.(2017)Paszke, Gross, Chintala, Chanan, Yang, DeVito, Lin,
  Desmaison, Antiga, and Lerer]{paszke2017automatic}
Adam Paszke, Sam Gross, Soumith Chintala, Gregory Chanan, Edward Yang, Zachary
  DeVito, Zeming Lin, Alban Desmaison, Luca Antiga, and Adam Lerer.
\newblock Automatic differentiation in pytorch.
\newblock 2017.

\bibitem[Pathak et~al.(2017{\natexlab{a}})Pathak, Agrawal, Efros, and
  Darrell]{pathak2017curiosity}
Deepak Pathak, Pulkit Agrawal, Alexei~A Efros, and Trevor Darrell.
\newblock Curiosity-driven exploration by self-supervised prediction.
\newblock In \emph{International Conference on Machine Learning (ICML)},
  2017{\natexlab{a}}.

\bibitem[Pathak et~al.(2017{\natexlab{b}})Pathak, Agrawal, Efros, and
  Darrell]{pathakICMl17curiosity}
Deepak Pathak, Pulkit Agrawal, Alexei~A. Efros, and Trevor Darrell.
\newblock Curiosity-driven exploration by self-supervised prediction.
\newblock In \emph{ICML}, 2017{\natexlab{b}}.

\bibitem[Pathak et~al.(2019)Pathak, Gandhi, and Gupta]{pathak19disagreement}
Deepak Pathak, Dhiraj Gandhi, and Abhinav Gupta.
\newblock Self-supervised exploration via disagreement.
\newblock In \emph{ICML}, 2019.

\bibitem[Pronobis and Jensfelt(2012)]{pronobis2012large}
Andrzej Pronobis and Patric Jensfelt.
\newblock Large-scale semantic mapping and reasoning with heterogeneous
  modalities.
\newblock In \emph{2012 IEEE International Conference on Robotics and
  Automation}, pages 3515--3522. IEEE, 2012.

\bibitem[Quigley et~al.(2009)Quigley, Gerkey, Conley, Faust, Foote, Leibs,
  Berger, Wheeler, and Ng]{Quigley09}
Morgan Quigley, Brian Gerkey, Ken Conley, Josh Faust, Tully Foote, Jeremy
  Leibs, Eric Berger, Rob Wheeler, and Andrew Ng.
\newblock Ros: an open-source robot operating system.
\newblock In \emph{Proc. of the IEEE Intl. Conf. on Robotics and Automation
  (ICRA) Workshop on Open Source Robotics}, Kobe, Japan, May 2009.

\bibitem[Ren et~al.(2015)Ren, He, Girshick, and Sun]{ren2015faster}
Shaoqing Ren, Kaiming He, Ross Girshick, and Jian Sun.
\newblock Faster r-cnn: Towards real-time object detection with region proposal
  networks.
\newblock In \emph{Advances in neural information processing systems}, pages
  91--99, 2015.

\bibitem[Ross et~al.(2011)Ross, Gordon, and Bagnell]{ross2011reduction}
St{\'e}phane Ross, Geoffrey Gordon, and Drew Bagnell.
\newblock A reduction of imitation learning and structured prediction to
  no-regret online learning.
\newblock In \emph{Proceedings of the fourteenth international conference on
  artificial intelligence and statistics}, pages 627--635, 2011.

\bibitem[Roumeliotis and Bekey(2000)]{Roumeliotis00bayesianestimation}
Stergios~I. Roumeliotis and George~A. Bekey.
\newblock Bayesian estimation and kalman filtering: A unified framework for
  mobile robot localization, 2000.

\bibitem[Russell and Norvig(1995)]{russell1995modern}
Stuart Russell and Peter Norvig.
\newblock A modern approach.
\newblock \emph{Artificial Intelligence. Prentice-Hall, Egnlewood Cliffs},
  25:\penalty0 27, 1995.

\bibitem[Sadeghi and Levine(2016)]{sadeghi2016cad2rl}
Fereshteh Sadeghi and Sergey Levine.
\newblock Cad2rl: Real single-image flight without a single real image.
\newblock \emph{arXiv preprint arXiv:1611.04201}, 2016.

\bibitem[Savinov et~al.(2018{\natexlab{a}})Savinov, Dosovitskiy, and
  Koltun]{savinov2018semi}
Nikolay Savinov, Alexey Dosovitskiy, and Vladlen Koltun.
\newblock Semi-parametric topological memory for navigation.
\newblock In \emph{International Conference on Learning Representations
  ({ICLR})}, 2018{\natexlab{a}}.

\bibitem[Savinov et~al.(2018{\natexlab{b}})Savinov, Raichuk, Marinier, Vincent,
  Pollefeys, Lillicrap, and Gelly]{savinov2018episodic}
Nikolay Savinov, Anton Raichuk, Rapha{\"e}l Marinier, Damien Vincent, Marc
  Pollefeys, Timothy Lillicrap, and Sylvain Gelly.
\newblock Episodic curiosity through reachability.
\newblock \emph{arXiv preprint arXiv:1810.02274}, 2018{\natexlab{b}}.

\bibitem[Savva et~al.(2017)Savva, Chang, Dosovitskiy, Funkhouser, and
  Koltun]{savva2017minos}
Manolis Savva, Angel~X. Chang, Alexey Dosovitskiy, Thomas Funkhouser, and
  Vladlen Koltun.
\newblock {MINOS}: Multimodal indoor simulator for navigation in complex
  environments.
\newblock \emph{arXiv:1712.03931}, 2017.

\bibitem[Savva et~al.(2019)Savva, Kadian, Maksymets, Zhao, Wijmans, Jain,
  Straub, Liu, Koltun, Malik, et~al.]{savva2019habitat}
Manolis Savva, Abhishek Kadian, Oleksandr Maksymets, Yili Zhao, Erik Wijmans,
  Bhavana Jain, Julian Straub, Jia Liu, Vladlen Koltun, Jitendra Malik, et~al.
\newblock Habitat: A platform for embodied ai research.
\newblock In \emph{ICCV}, 2019.

\bibitem[Schaul et~al.(2015)Schaul, Quan, Antonoglou, and
  Silver]{schaul2015prioritized}
Tom Schaul, John Quan, Ioannis Antonoglou, and David Silver.
\newblock Prioritized experience replay.
\newblock \emph{arXiv preprint arXiv:1511.05952}, 2015.

\bibitem[Schmidhuber(1991)]{schmidhuber1991possibility}
J{\"u}rgen Schmidhuber.
\newblock A possibility for implementing curiosity and boredom in
  model-building neural controllers.
\newblock In \emph{Proc. of the international conference on simulation of
  adaptive behavior: From animals to animats}, pages 222--227, 1991.

\bibitem[Schulman et~al.(2015)Schulman, Moritz, Levine, Jordan, and
  Abbeel]{schulman2015high}
John Schulman, Philipp Moritz, Sergey Levine, Michael Jordan, and Pieter
  Abbeel.
\newblock High-dimensional continuous control using generalized advantage
  estimation.
\newblock \emph{arXiv preprint arXiv:1506.02438}, 2015.

\bibitem[Schulman et~al.(2017)Schulman, Wolski, Dhariwal, Radford, and
  Klimov]{schulman2017proximal}
John Schulman, Filip Wolski, Prafulla Dhariwal, Alec Radford, and Oleg Klimov.
\newblock Proximal policy optimization algorithms.
\newblock \emph{arXiv preprint arXiv:1707.06347}, 2017.

\bibitem[Sener and Savarese(2018)]{sener2017active}
Ozan Sener and Silvio Savarese.
\newblock Active learning for convolutional neural networks: A core-set
  approach.
\newblock In \emph{International Conference on Learning Representations}, 2018.
\newblock URL \url{https://openreview.net/forum?id=H1aIuk-RW}.

\bibitem[Sethian(1996)]{sethian1996fast}
James~A Sethian.
\newblock A fast marching level set method for monotonically advancing fronts.
\newblock \emph{Proceedings of the National Academy of Sciences}, 93\penalty0
  (4):\penalty0 1591--1595, 1996.

\bibitem[Settles(2009)]{settles2009active}
Burr Settles.
\newblock Active learning literature survey.
\newblock Technical report, University of Wisconsin-Madison Department of
  Computer Sciences, 2009.

\bibitem[Shah et~al.(2017)Shah, Dey, Lovett, and Kapoor]{airsim2017fsr}
Shital Shah, Debadeepta Dey, Chris Lovett, and Ashish Kapoor.
\newblock Airsim: High-fidelity visual and physical simulation for autonomous
  vehicles.
\newblock In \emph{Field and Service Robotics}, 2017.
\newblock URL \url{https://arxiv.org/abs/1705.05065}.

\bibitem[Siddiqui et~al.(2020)Siddiqui, Valentin, and
  Nie{\ss}ner]{siddiqui2019viewal}
Yawar Siddiqui, Julien Valentin, and Matthias Nie{\ss}ner.
\newblock Viewal: Active learning with viewpoint entropy for semantic
  segmentation.
\newblock In \emph{Proceedings of the IEEE/CVF Conference on Computer Vision
  and Pattern Recognition}, pages 9433--9443, 2020.

\bibitem[Silver et~al.(2016)Silver, Huang, Maddison, Guez, Sifre, Van
  Den~Driessche, Schrittwieser, Antonoglou, Panneershelvam, Lanctot,
  et~al.]{silver2016mastering}
David Silver, Aja Huang, Chris~J Maddison, Arthur Guez, Laurent Sifre, George
  Van Den~Driessche, Julian Schrittwieser, Ioannis Antonoglou, Veda
  Panneershelvam, Marc Lanctot, et~al.
\newblock Mastering the game of go with deep neural networks and tree search.
\newblock \emph{Nature}, 529\penalty0 (7587):\penalty0 484--489, 2016.

\bibitem[Smith et~al.(1990)Smith, Self, and Cheeseman]{smith1990estimating}
Randall Smith, Matthew Self, and Peter Cheeseman.
\newblock Estimating uncertain spatial relationships in robotics.
\newblock In \emph{Autonomous robot vehicles}, pages 167--193. Springer, 1990.

\bibitem[Snavely et~al.(2008)Snavely, Seitz, and Szeliski]{snavely2008modeling}
Noah Snavely, Steven~M Seitz, and Richard Szeliski.
\newblock Modeling the world from internet photo collections.
\newblock \emph{International journal of computer vision}, 80\penalty0
  (2):\penalty0 189--210, 2008.

\bibitem[Song et~al.(2017)Song, Yu, Zeng, Chang, Savva, and
  Funkhouser]{song2016ssc}
Shuran Song, Fisher Yu, Andy Zeng, Angel~X Chang, Manolis Savva, and Thomas
  Funkhouser.
\newblock Semantic scene completion from a single depth image.
\newblock In \emph{CVPR}, 2017.

\bibitem[Stachniss et~al.(2005)Stachniss, Grisetti, and
  Burgard]{stachniss2005information}
Cyrill Stachniss, Giorgio Grisetti, and Wolfram Burgard.
\newblock Information gain-based exploration using rao-blackwellized particle
  filters.
\newblock In \emph{Robotics: Science and Systems}, volume~2, pages 65--72,
  2005.

\bibitem[Straub et~al.(2019)Straub, Whelan, Ma, Chen, Wijmans, Green, Engel,
  Mur-Artal, Ren, Verma, Clarkson, Yan, Budge, Yan, Pan, Yon, Zou, Leon,
  Carter, Briales, Gillingham, Mueggler, Pesqueira, Savva, Batra, Strasdat,
  Nardi, Goesele, Lovegrove, and Newcombe]{replica19arxiv}
Julian Straub, Thomas Whelan, Lingni Ma, Yufan Chen, Erik Wijmans, Simon Green,
  Jakob~J. Engel, Raul Mur-Artal, Carl Ren, Shobhit Verma, Anton Clarkson,
  Mingfei Yan, Brian Budge, Yajie Yan, Xiaqing Pan, June Yon, Yuyang Zou,
  Kimberly Leon, Nigel Carter, Jesus Briales, Tyler Gillingham, Elias Mueggler,
  Luis Pesqueira, Manolis Savva, Dhruv Batra, Hauke~M. Strasdat, Renzo~De
  Nardi, Michael Goesele, Steven Lovegrove, and Richard Newcombe.
\newblock The {R}eplica dataset: A digital replica of indoor spaces.
\newblock \emph{arXiv preprint arXiv:1906.05797}, 2019.

\bibitem[Sutton and Barto(1998)]{sutton1998rli}
Richard~S. Sutton and Andrew~G. Barto.
\newblock \emph{Reinforcement Learning: An Introduction}.
\newblock MIT Press, 1998.
\newblock URL \url{http://www.cs.ualberta.ca/~sutton/book/the-book.html}.

\bibitem[Sutton and Barto(2018)]{sutton2018reinforcement}
Richard~S Sutton and Andrew~G Barto.
\newblock \emph{Reinforcement learning: An introduction}.
\newblock MIT press, 2018.

\bibitem[Sutton et~al.(1999)Sutton, Precup, and Singh]{sutton1999between}
Richard~S Sutton, Doina Precup, and Satinder Singh.
\newblock Between mdps and semi-mdps: A framework for temporal abstraction in
  reinforcement learning.
\newblock \emph{Artificial intelligence}, 112\penalty0 (1-2):\penalty0
  181--211, 1999.

\bibitem[Tamar et~al.(2016)Tamar, Wu, Thomas, Levine, and
  Abbeel]{tamar2016value}
Aviv Tamar, Yi~Wu, Garrett Thomas, Sergey Levine, and Pieter Abbeel.
\newblock Value iteration networks.
\newblock In \emph{Advances in Neural Information Processing Systems}, pages
  2154--2162, 2016.

\bibitem[Tastan and Sukthankar(2011)]{tastan2011learning}
Bulent Tastan and Gita~Reese Sukthankar.
\newblock Learning policies for first person shooter games using inverse
  reinforcement learning.
\newblock In \emph{AIIDE}. Citeseer, 2011.

\bibitem[Tellex et~al.(2011)Tellex, Kollar, Dickerson, Walter, Banerjee,
  Teller, and Roy]{tellex2011understanding}
Stefanie~A Tellex, Thomas~Fleming Kollar, Steven~R Dickerson, Matthew~R Walter,
  Ashis Banerjee, Seth Teller, and Nicholas Roy.
\newblock Understanding natural language commands for robotic navigation and
  mobile manipulation.
\newblock 2011.

\bibitem[Thrun et~al.(1998)Thrun, Gutmann, Fox, Burgard, Kuipers,
  et~al.]{thrun1998integrating}
Sebastian Thrun, Jens-Steffen Gutmann, Dieter Fox, Wolfram Burgard, Benjamin
  Kuipers, et~al.
\newblock Integrating topological and metric maps for mobile robot navigation:
  A statistical approach.
\newblock In \emph{AAAI/IAAI}, pages 989--995, 1998.

\bibitem[Thrun et~al.(2001)Thrun, Fox, Burgard, and Dellaert]{thrun2001robust}
Sebastian Thrun, Dieter Fox, Wolfram Burgard, and Frank Dellaert.
\newblock Robust monte carlo localization for mobile robots.
\newblock \emph{Artificial intelligence}, 128\penalty0 (1-2):\penalty0 99--141,
  2001.

\bibitem[Thrun et~al.(2005)Thrun, Burgard, and Fox]{thrun2005probabilistic}
Sebastian Thrun, Wolfram Burgard, and Dieter Fox.
\newblock \emph{Probabilistic robotics}.
\newblock MIT press, 2005.

\bibitem[Tieleman and Hinton(2012)]{tieleman2012lecture}
Tijmen Tieleman and Geoffrey Hinton.
\newblock Lecture 6.5-rmsprop: Divide the gradient by a running average of its
  recent magnitude.
\newblock \emph{COURSERA: Neural networks for machine learning}, 4\penalty0
  (2), 2012.

\bibitem[Tobin et~al.(2017)Tobin, Fong, Ray, Schneider, Zaremba, and
  Abbeel]{tobin2017domain}
Josh Tobin, Rachel Fong, Alex Ray, Jonas Schneider, Wojciech Zaremba, and
  Pieter Abbeel.
\newblock Domain randomization for transferring deep neural networks from
  simulation to the real world.
\newblock In \emph{Intelligent Robots and Systems (IROS), 2017 IEEE/RSJ
  International Conference on}, pages 23--30. IEEE, 2017.

\bibitem[Tomatis et~al.(2001)Tomatis, Nourbakhsh, and
  Siegwart]{tomatis2001combining}
Nicola Tomatis, Illah Nourbakhsh, and Roland Siegwart.
\newblock Combining topological and metric: A natural integration for
  simultaneous localization and map building.
\newblock In \emph{Proceedings of the Fourth European Workshop on Advanced
  Mobile Robots (Eurobot)}. ETH-Z{\"u}rich, 2001.

\bibitem[Tulsiani et~al.(2019)Tulsiani, Zhou, Efros, and Malik]{drcTulsiani19}
Shubham Tulsiani, Tinghui Zhou, Alexei~A. Efros, and Jitendra Malik.
\newblock Multi-view supervision for single-view reconstruction via
  differentiable ray consistency.
\newblock \emph{TPAMI}, 2019.

\bibitem[Van~Hasselt et~al.(2015)Van~Hasselt, Guez, and Silver]{van2015deep}
Hado Van~Hasselt, Arthur Guez, and David Silver.
\newblock Deep reinforcement learning with double q-learning.
\newblock \emph{CoRR, abs/1509.06461}, 2015.

\bibitem[Vijayanarasimhan and Grauman(2014)]{vijayanarasimhan2014large}
Sudheendra Vijayanarasimhan and Kristen Grauman.
\newblock Large-scale live active learning: Training object detectors with
  crawled data and crowds.
\newblock \emph{International Journal of Computer Vision}, 108\penalty0
  (1-2):\penalty0 97--114, 2014.

\bibitem[Vondrick et~al.(2013)Vondrick, Patterson, and Ramanan]{vatic}
Carl Vondrick, Donald Patterson, and Deva Ramanan.
\newblock Efficiently scaling up crowdsourced video annotation.
\newblock \emph{International Journal of Computer Vision}, pages 1--21, 2013.
\newblock ISSN 0920-5691.
\newblock URL \url{http://dx.doi.org/10.1007/s11263-012-0564-1}.
\newblock 10.1007/s11263-012-0564-1.

\bibitem[Walter et~al.(2013)Walter, Hemachandra, Homberg, Tellex, and
  Teller]{walter2013learning}
Matthew~R Walter, Sachithra Hemachandra, Bianca Homberg, Stefanie Tellex, and
  Seth Teller.
\newblock Learning semantic maps from natural language descriptions.
\newblock Robotics: Science and Systems, 2013.

\bibitem[Wang and Spelke(2002)]{wang2002human}
Ranxiao~Frances Wang and Elizabeth~S Spelke.
\newblock Human spatial representation: Insights from animals.
\newblock \emph{Trends in cognitive sciences}, 6\penalty0 (9):\penalty0
  376--382, 2002.

\bibitem[Wang et~al.(2017)Wang, Clark, Wen, and Trigoni]{wang2017deepvo}
Sen Wang, Ronald Clark, Hongkai Wen, and Niki Trigoni.
\newblock Deepvo: Towards end-to-end visual odometry with deep recurrent
  convolutional neural networks.
\newblock In \emph{Robotics and Automation (ICRA), 2017 IEEE International
  Conference on}, pages 2043--2050. IEEE, 2017.

\bibitem[Wang et~al.(2015)Wang, de~Freitas, and Lanctot]{wang2015dueling}
Ziyu Wang, Nando de~Freitas, and Marc Lanctot.
\newblock Dueling network architectures for deep reinforcement learning.
\newblock \emph{arXiv preprint arXiv:1511.06581}, 2015.

\bibitem[Wijmans et~al.(2020)Wijmans, Kadian, Morcos, Lee, Essa, Parikh, Savva,
  and Batra]{wijmans2019decentralized}
Erik Wijmans, Abhishek Kadian, Ari Morcos, Stefan Lee, Irfan Essa, Devi Parikh,
  Manolis Savva, and Dhruv Batra.
\newblock Decentralized distributed ppo: Solving pointgoal navigation.
\newblock In \emph{ICLR}, 2020.

\bibitem[Wortsman et~al.(2019)Wortsman, Ehsani, Rastegari, Farhadi, and
  Mottaghi]{wortsman2019learning}
Mitchell Wortsman, Kiana Ehsani, Mohammad Rastegari, Ali Farhadi, and Roozbeh
  Mottaghi.
\newblock Learning to learn how to learn: Self-adaptive visual navigation using
  meta-learning.
\newblock In \emph{Proceedings of the IEEE Conference on Computer Vision and
  Pattern Recognition}, pages 6750--6759, 2019.

\bibitem[Wu et~al.(2018)Wu, Wu, Tamar, Russell, Gkioxari, and
  Tian]{wu2018learning}
Yi~Wu, Yuxin Wu, Aviv Tamar, Stuart Russell, Georgia Gkioxari, and Yuandong
  Tian.
\newblock Learning and planning with a semantic model.
\newblock \emph{arXiv preprint arXiv:1809.10842}, 2018.

\bibitem[Wu et~al.(2019{\natexlab{a}})Wu, Wu, Tamar, Russell, Gkioxari, and
  Tian]{wu2019bayesian}
Yi~Wu, Yuxin Wu, Aviv Tamar, Stuart Russell, Georgia Gkioxari, and Yuandong
  Tian.
\newblock Bayesian relational memory for semantic visual navigation.
\newblock In \emph{CVPR}, pages 2769--2779, 2019{\natexlab{a}}.

\bibitem[Wu et~al.(2016)Wu, Zhang, Zhang, Bengio, and Salakhutdinov]{Wu2016}
Yuhuai Wu, Saizheng Zhang, Ying Zhang, Yoshua Bengio, and Ruslan Salakhutdinov.
\newblock On multiplicative integration with recurrent neural networks.
\newblock In \emph{NIPS}, 2016.

\bibitem[Wu and Tian(2017)]{wu2016training}
Yuxin Wu and Yuandong Tian.
\newblock Training agent for first-person shooter game with actor-critic
  curriculum learning.
\newblock In \emph{ICLR}, 2017.

\bibitem[Wu et~al.(2019{\natexlab{b}})Wu, Kirillov, Massa, Lo, and
  Girshick]{wu2019detectron2}
Yuxin Wu, Alexander Kirillov, Francisco Massa, Wan-Yen Lo, and Ross Girshick.
\newblock Detectron2.
\newblock \url{https://github.com/facebookresearch/detectron2},
  2019{\natexlab{b}}.

\bibitem[Xia et~al.(2018{\natexlab{a}})Xia, R.~Zamir, He, Sax, Malik, and
  Savarese]{gibsonenv}
Fei Xia, Amir R.~Zamir, Zhi-Yang He, Alexander Sax, Jitendra Malik, and Silvio
  Savarese.
\newblock Gibson {Env}: real-world perception for embodied agents.
\newblock In \emph{Computer Vision and Pattern Recognition (CVPR), 2018 IEEE
  Conference on}. IEEE, 2018{\natexlab{a}}.

\bibitem[Xia et~al.(2018{\natexlab{b}})Xia, R.~Zamir, He, Sax, Malik, and
  Savarese]{xiazamirhe2018gibsonenv}
Fei Xia, Amir R.~Zamir, Zhi-Yang He, Alexander Sax, Jitendra Malik, and Silvio
  Savarese.
\newblock Gibson {Env}: real-world perception for embodied agents.
\newblock In \emph{Computer Vision and Pattern Recognition (CVPR), 2018 IEEE
  Conference on}. IEEE, 2018{\natexlab{b}}.

\bibitem[Xu et~al.(2017)Xu, Zheng, Yan, Yan, Zhang, Niessner, Deussen,
  Cohen-Or, and Huang]{xu2017autonomous}
Kai Xu, Lintao Zheng, Zihao Yan, Guohang Yan, Eugene Zhang, Matthias Niessner,
  Oliver Deussen, Daniel Cohen-Or, and Hui Huang.
\newblock Autonomous reconstruction of unknown indoor scenes guided by
  time-varying tensor fields.
\newblock \emph{ACM Transactions on Graphics (TOG)}, 36\penalty0 (6):\penalty0
  202, 2017.

\bibitem[Yamauchi(1997)]{yamauchi1997frontier}
Brian Yamauchi.
\newblock A frontier-based approach for autonomous exploration.
\newblock In \emph{cira}, volume~97, page 146, 1997.

\bibitem[Yang et~al.(2019)Yang, Ren, Xu, Chen, Crandall, Parikh, and
  Batra]{yang2019embodied}
Jianwei Yang, Zhile Ren, Mingze Xu, Xinlei Chen, David~J Crandall, Devi Parikh,
  and Dhruv Batra.
\newblock Embodied amodal recognition: Learning to move to perceive objects.
\newblock In \emph{Proceedings of the IEEE International Conference on Computer
  Vision}, pages 2040--2050, 2019.

\bibitem[Yang et~al.(2018)Yang, Wang, Farhadi, Gupta, and
  Mottaghi]{yang2018visual}
Wei Yang, Xiaolong Wang, Ali Farhadi, Abhinav Gupta, and Roozbeh Mottaghi.
\newblock Visual semantic navigation using scene priors.
\newblock \emph{arXiv preprint arXiv:1810.06543}, 2018.

\bibitem[Yoo and Kweon(2019)]{yoo2019learning}
Donggeun Yoo and In~So Kweon.
\newblock Learning loss for active learning.
\newblock In \emph{Proceedings of the IEEE Conference on Computer Vision and
  Pattern Recognition}, pages 93--102, 2019.

\bibitem[Yu et~al.(2017)Yu, Zhang, and Xu]{yu2017deep}
Haonan Yu, Haichao Zhang, and Wei Xu.
\newblock A deep compositional framework for human-like language acquisition in
  virtual environment.
\newblock \emph{arXiv preprint arXiv:1703.09831}, 2017.

\bibitem[Zhang et~al.(2018{\natexlab{a}})Zhang, Lerer, Sukhbaatar, Fergus, and
  Szlam]{zhang2018composable}
Amy Zhang, Adam Lerer, Sainbayar Sukhbaatar, Rob Fergus, and Arthur Szlam.
\newblock Composable planning with attributes.
\newblock \emph{arXiv preprint arXiv:1803.00512}, 2018{\natexlab{a}}.

\bibitem[Zhang et~al.(2017)Zhang, Tai, Boedecker, Burgard, and
  Liu]{zhang2017neural}
Jingwei Zhang, Lei Tai, Joschka Boedecker, Wolfram Burgard, and Ming Liu.
\newblock Neural slam: Learning to explore with external memory.
\newblock \emph{arXiv preprint arXiv:1706.09520}, 2017.

\bibitem[Zhang et~al.(2018{\natexlab{b}})Zhang, Wei, Shen, Wei, Zhu, and
  Song]{zhang2018semantic}
Liang Zhang, Leqi Wei, Peiyi Shen, Wei Wei, Guangming Zhu, and Juan Song.
\newblock Semantic slam based on object detection and improved octomap.
\newblock \emph{IEEE Access}, 6:\penalty0 75545--75559, 2018{\natexlab{b}}.

\bibitem[Zhu et~al.(2017)Zhu, Mottaghi, Kolve, Lim, Gupta, Fei-Fei, and
  Farhadi]{zhu2017target}
Yuke Zhu, Roozbeh Mottaghi, Eric Kolve, Joseph~J Lim, Abhinav Gupta,
  Li~Fei-Fei, and Ali Farhadi.
\newblock Target-driven visual navigation in indoor scenes using deep
  reinforcement learning.
\newblock In \emph{2017 IEEE international conference on robotics and
  automation (ICRA)}, pages 3357--3364. IEEE, 2017.

\end{thebibliography}

\end{document}


\maketitle

\appendix


\section{Dataset details}

\textbf{Dataset splits.} We use standard train and val splits provided in the Gibson and MP3D datasets as our train and test sets in our experiments, as test scenes in the Habitat simulator format are held-out for an online evaluation server. The list of scenes used in train and test splits for both the datasets are shown in Table~\ref{tab:splits}. For the MP3D dataset, we use all 72 scenes (61 train + 11 test), whereas for the Gibson dataset, we use the subset of 30 scenes (25 train + 5 test) which have ground-truth semantic segmentation annotations available.

\textbf{Object categories}
The 15 object categories used in the semantic map were:
chair,
couch,
potted plant,
bed,
toilet,
tv,
dining-table,
oven,
sink,
refrigerator,
book,
clock,
vase,
cup,
bottle.
Out of these, the first 6 (chair,
couch,
potted plant,
bed,
toilet,
tv) are categories common between Gibson, MP3D and MS-COCO datasets are used for object goals.

\setlength{\tabcolsep}{6pt}
\begin{table}[h!]

\caption{\textbf{Dataset Splits.} Table showing the list of scenes used in the training and test splits for both the Gibson and Matterport3D datasets.}
\vspace{-5pt}
\label{tab:splits}
\centering
{\scriptsize
\begin{tabular}{@{}llclllllc@{}}

\toprule
Dataset &  & \multicolumn{5}{c}{Train split} &  & Test split  \\ \midrule
Gibson  &  & \begin{tabular}[c]{@{}c@{}}Allensville\\ Beechwood\\ Benevolence\\ Coffeen\\ Cosmos\end{tabular}                                                                                                                                                                      & \multicolumn{1}{c}{\begin{tabular}[c]{@{}c@{}}Forkland\\ Hanson\\ Hiteman\\ Klickitat\\ Lakeville\end{tabular}}                                                                                                                                   & \multicolumn{1}{c}{\begin{tabular}[c]{@{}c@{}}Leonardo\\ Lindenwood\\ Marstons\\ Merom\\ Mifflinburg\end{tabular}}                                                                                                                                                      & \multicolumn{1}{c}{\begin{tabular}[c]{@{}c@{}}Newfields\\ Onaga\\ Pinesdale\\ Pomaria\\ Ranchester\end{tabular}}                                                                                                                                  & \multicolumn{1}{c}{\begin{tabular}[c]{@{}c@{}}Shelbyville\\ Stockman\\ Tolstoy\\ Wainscott\\ Woodbine\end{tabular}}                                                                                                                               &  & \begin{tabular}[c]{@{}c@{}}Collierville\\ Corozal\\ Darden\\ Markleeville\\ Wiconisco\end{tabular}                                                                                                                                                                                                          \\\midrule
MP3D    &  & \multicolumn{1}{c}{\begin{tabular}[c]{@{}c@{}}qoiz87JEwZ2\\ r1Q1Z4BcV1o\\ r47D5H71a5s\\ rPc6DW4iMge\\ s8pcmisQ38h\\ S9hNv5qa7GM\\ sKLMLpTHeUy\\ SN83YJsR3w2\\ sT4fr6TAbpF\\ ULsKaCPVFJR\\ uNb9QFRL6hY\\ ur6pFq6Qu1A\end{tabular}} & \begin{tabular}[c]{@{}l@{}}ac26ZMwG7aT\\ B6ByNegPMKs\\ b8cTxDM8gDG\\ cV4RVeZvu5T\\ D7G3Y4RVNrH\\ D7N2EKCX4Sj\\ dhjEzFoUFzH\\ E9uDoFAP3SH\\ e9zR4mvMWw7\\ EDJbREhghzL\\ GdvgFV5R1Z5\\ gTV8FGcVJC9\end{tabular} & \begin{tabular}[c]{@{}l@{}}gZ6f7yhEvPG\\ HxpKQynjfin\\ i5noydFURQK\\ JeFG25nYj2p\\ JF19kD82Mey\\ jh4fc5c5qoQ\\ JmbYfDe2QKZ\\ kEZ7cmS4wCh\\ mJXqzFtmKg4\\ p5wJjkQkbXX\\ Pm6F8kyY3z2\\ pRbA3pwrgk9\\ PuKPg4mmafe\end{tabular} & \begin{tabular}[c]{@{}l@{}}PX4nDJXEHrG\\ qoiz87JEwZ2\\ r1Q1Z4BcV1o\\ r47D5H71a5s\\ rPc6DW4iMge\\ s8pcmisQ38h\\ S9hNv5qa7GM\\ sKLMLpTHeUy\\ SN83YJsR3w2\\ sT4fr6TAbpF\\ ULsKaCPVFJR\\ uNb9QFRL6hY\end{tabular} & \begin{tabular}[c]{@{}l@{}}ur6pFq6Qu1A\\ Uxmj2M2itWa\\ V2XKFyX4ASd\\ VFuaQ6m2Qom\\ VLzqgDo317F\\ VVfe2KiqLaN\\ Vvot9Ly1tCj\\ vyrNrziPKCB\\ VzqfbhrpDEA\\ XcA2TqTSSAj\\ YmJkqBEsHnH\\ ZMojNkEp431\end{tabular} &  & \multicolumn{1}{c}{\begin{tabular}[c]{@{}c@{}}2azQ1b91cZZ\\ 8194nk5LbLH\\ EU6Fwq7SyZv\\ oLBMNvg9in8\\ pLe4wQe7qrG\\ QUCTc6BB5sX\\ TbHJrupSAjP\\ X7HyMhZNoso\\ x8F5xyUWy9e\\ Z6MFQCViBuw\\ zsNo4HB9uLZ\end{tabular}} \\ \bottomrule
\end{tabular}
}
\end{table}
